\documentclass[lettersize,journal]{IEEEtran}
\usepackage{amsmath,amsfonts}
\usepackage{array}
\usepackage[caption=false,font=scriptsize,labelfont=sf,textfont=sf]{subfig}
\usepackage{textcomp}
\usepackage{stfloats}
\usepackage{url}
\usepackage{verbatim}
\usepackage{graphicx}
\usepackage{cite}
\usepackage[T1]{fontenc}
\usepackage{pifont}
\usepackage{arydshln}
\usepackage{multirow}  
\usepackage[normalem]{ulem} 
\usepackage{color} 
\usepackage{mathabx} 
\useunder{\uline}{\ul}{}
\usepackage[ruled,linesnumbered]{algorithm2e} 
\usepackage[colorlinks,
            linkcolor=red,  
            anchorcolor=blue,
            citecolor=green]{hyperref}

\begin{document}

\title{
  Evolutionary Multitasking with Solution Space Cutting for Point Cloud Registration
  }
\author{
Yue~Wu,~\IEEEmembership{Member,~IEEE,}
Peiran~Gong,
Maoguo~Gong,~\IEEEmembership{Senior Member,~IEEE,}
Hangqi~Ding,
Zedong~Tang,
Yibo~Liu,
Wenping~Ma,~\IEEEmembership{Senior Member,~IEEE,}
and~Qiguang~Miao,~\IEEEmembership{Senior Member,~IEEE}
\thanks{Manuscript received XXX; revised XXX; accepted XXX. 
This work is supported by the National Natural Science Foundation of China (62276200, 62036006), the Natural Science Basic Research Plan in Shaanxi Province of China (2022JM-327) and the CAAI-Huawei MINDSPORE Academic Open Fund. 
\textit{(Corresponding author: Maoguo Gong.)}}
\thanks{Yue Wu, Peiran Gong, Hangqi Ding, Yibo Liu and Qiguang Miao are with the School of Computer Science and Technology, Key Laboratory of Collaborative Intelligence Systems, Ministry of Education, Xidian University, Xi'an 710071, China (e-mail: ywu@xidian.edu.cn;  gpr@stu.xidian.edu.cn; hqding@stu.xidian.edu.cn;  yb\_liu@stu.xidian.edu.cn; qgmiao@mail.xidian.edu.cn).}
\thanks{Maoguo Gong and Zedong Tang are with the School of Electronic Engineering, Key Laboratory of Collaborative Intelligence Systems, Ministry of Education, Xidian University, Xi'an 710071, China (e-mail: gong@ieee.org;  omegatangzd@gmail.com).}
\thanks{Wenping Ma is with the Key Laboratory of Intelligent Perception and Image Understanding, Ministry of Education, School of Artificial Intelligence, Xidian University, Xi'an 710071, China (e-mail: wpma@mail.xidian.edu.cn).}
}
\markboth{IEEE TRANSACTIONS ON EMERGING TOPICS IN COMPUTATIONAL INTELLIGENCE}
{Shell \MakeLowercase{\textit{et al.}}: A Sample Article Using IEEEtran.cls for IEEE Journals}

\IEEEpubid{
  \begin{minipage}{\textwidth}\ \\[30pt] \centering
    \copyright 2023 IEEE. Personal use of this material is permitted.
    Permission from IEEE must be obtained for all other uses, in any current or future media, including reprinting/republishing this material for advertising or promotional purposes, creating new collective works, for resale or redistribution to servers or lists, or reuse of any copyrighted component of this work in other works.
  \end{minipage}
}

\maketitle

\begin{abstract}
Point cloud registration (PCR) is a popular research topic in computer vision.
Recently, the registration method in an evolutionary way has received continuous attention because of its robustness to the initial pose and flexibility in objective function design. 
However, most evolving registration methods cannot tackle the local optimum well and they have rarely investigated the success ratio, which implies the probability of not falling into local optima and is closely related to the practicality of the algorithm.
Evolutionary multi-task optimization (EMTO) is a widely used paradigm, which can boost exploration capability through knowledge transfer among related tasks.  
Inspired by this concept, this study proposes a novel evolving registration algorithm via EMTO, where the multi-task configuration is based on the idea of solution space cutting.
Concretely, one task searching in cut space assists another task with complex function landscape in escaping from local optima and enhancing successful registration ratio.
To reduce unnecessary computational cost, a sparse-to-dense strategy is proposed. 
In addition, a novel fitness function robust to various overlap rates as well as a problem-specific metric of computational cost is introduced.
Compared with 8 evolving approaches, 4 traditional approaches and 3 deep learning approaches on the object-scale and scene-scale registration datasets, experimental results demonstrate that the proposed method has superior performances in terms of precision and tackling local optima.
\end{abstract}

\begin{IEEEkeywords}
Evolutionary multi-task optimization, point cloud registration, particle swarm optimization.
\end{IEEEkeywords}

\section{Introduction}
\IEEEPARstart{P}{oint} cloud is a collection of points captured from consecutive surfaces of objects through scanners, and it has been one of the most appropriate data formats to describe the 3D world.
Due to occlusions and the scanner's varying views, only a part of object or scene is contained in the captured point cloud. 
Hence, it is necessary to develop the registration approach to recover the complete object or scene according to a sequence of incomplete point clouds.
Naturally, \emph{point cloud registration} (PCR) has been employed in many practical applications, such as 3D reconstruction \cite{huang2021di, chabra2020deep}, simultaneous localization and mapping (SLAM) \cite{rosinol2020kimera} and object recognition \cite{girish2021lottery}.
The aim of registration is to estimate the transformations that properly align multiple point clouds with dissimilar poses or in different coordinate systems. 
Considering both partial overlap and the noise present in measurement, the PCR problem becomes intractable.
This study considers the case of pairwise registration, which means the number of point clouds to be aligned is two.

\IEEEpubidadjcol{ 
Numerous approaches have been presented to address the PCR problems, which can be roughly classified into three categories, i.e., traditional, deep learning and evolving approaches.
\begin{itemize}
  \item {
  \emph{Traditional approaches} \cite{besl1992method, chetverikov2005robust, biber2003normal, zhou2016fast,  li2020gesac, yang2021sac, chen2022sc2, yang2022correspondence} are generally supported by rigorous mathematical theories \cite{huang2021comprehensive} and their results are deterministic for the given initial poses in repeated execution.
  In order to satisfy the mathematical constraints introduced by the numerical optimization methods, such as derivability or convexity, these approaches typically reduce the modeling accuracy. 
  Consequently, such compromises between constraints and modeling accuracy lead to only convergence to local optimum or the loss of registration precision.
  }
  \item {
  \emph{Deep learning approaches} \cite{zeng20173dmatch, deng2018ppfnet, huang2020feature} can register point clouds in a short time, and they are expected to achieve real-time registration \cite{qin2022geometric} with the help of neural network (NN).
  However, registration on unseen datasets or inadequate training significantly degrades their performance.
  Apart from that, these approaches are resource-consuming.
  }
  \item{
  \emph{Evolving approaches} \cite{silva2005precision, li2017differential, zhu2016automatic, wu2022multi} adopt evolutionary computation (EC) as the optimization method.
  Compared to traditional approaches, these approaches need no consideration of the restrictions on the objective function.
  It is worth noting that this allows the researchers to focus more on which objective function can better measure registration error rather than how to solve it.
  Moreover, due to the strong global search capability of EC, evolving approaches are not sensitive to initial pose \cite{silva2005precision}.
  Notably, this does not imply evolving approaches can avoid local optimum, but rather these approaches can handle the situation where the initial pose converges to local optimum for traditional approaches.
  Unlike deep learning approaches, evolving and traditional approaches both have great generalization ability and no requirement of training \cite{huang2021comprehensive}.
  }
\end{itemize}

However, the results searched by evolving approaches are nondeterministic, since the various operations in EC algorithms are full of randomness.
This signifies that not only the accuracy of successful registration fluctuates, but wrong registration may occur.
More terribly, when tackling the PCR problem, most existing evolutionary algorithms tend to fall into local optima, which will be experimentally proved in Section \ref{sec:RD}.
In the context of PCR, obtaining a local optimum means registration failure whose estimated transformation deviates significantly from the ground truth, thus it should be overcome.
Nevertheless, in the field of evolving registration, how to reduce the probability of getting stuck to local optimum, i.e., improve the ratio of successful registration, has rarely been investigated.

\emph{Evolutionary multi-task optimization} (EMTO) \cite{gupta2015multifactorial, ong2016evolutionary, bai2022multitask} is a promising research topic in EC and has been successfully employed in many optimization tasks, such as hyperspectral image unmixing \cite{li2018evolutionary}, feature selection \cite{chen2021evolutionary, chen2020evolutionary} and image classification \cite{bi2021learning}, as well as many other fields \cite{wu2021evolutionary,gao2022multiobjective}. 
Mimicking parallel learning mechanisms of humans, EMTO is capable of coping with multiple tasks at the same time. 
Its effectiveness depends on the selection/generation of multiple tasks and the knowledge transfer mechanism among tasks.
Furthermore, with the transfer of useful information among tasks, EMTO enables the original algorithm to improve its search ability and reduce falling into local optima.
Considering the advantages of EMTO, it is well suited to solve the problems present in evolving registration.
On the other hand, the potential of EMTO in PCR has not been thoroughly explored.
To this end, our method is implemented based on EMTO.

\emph{Particle swarm optimization} (PSO) \cite{kennedy1995particle}, a meta-heuristic algorithm in EC, is chosen as the engine of the proposed evolving approach since it has the advantage of being simple to implement, quick to converge and having few parameters to manually determine. 
PSO has been successfully applied in different optimization problems, such as continuous optimization \cite{du2022network, zhang2020utility}, large-scale optimization \cite{wang2020dynamic} and multi-objective optimization \cite{wu2021adaptive}.
Furthermore, PSO is easy to incorporate into EMTO \cite{feng2017empirical, han2021self, tang2022multi}, which will combine the strengths of both two methods.

\begin{figure}[!t]
  \centering
  \subfloat[Task $\alpha$]{\includegraphics[width=1.7in]{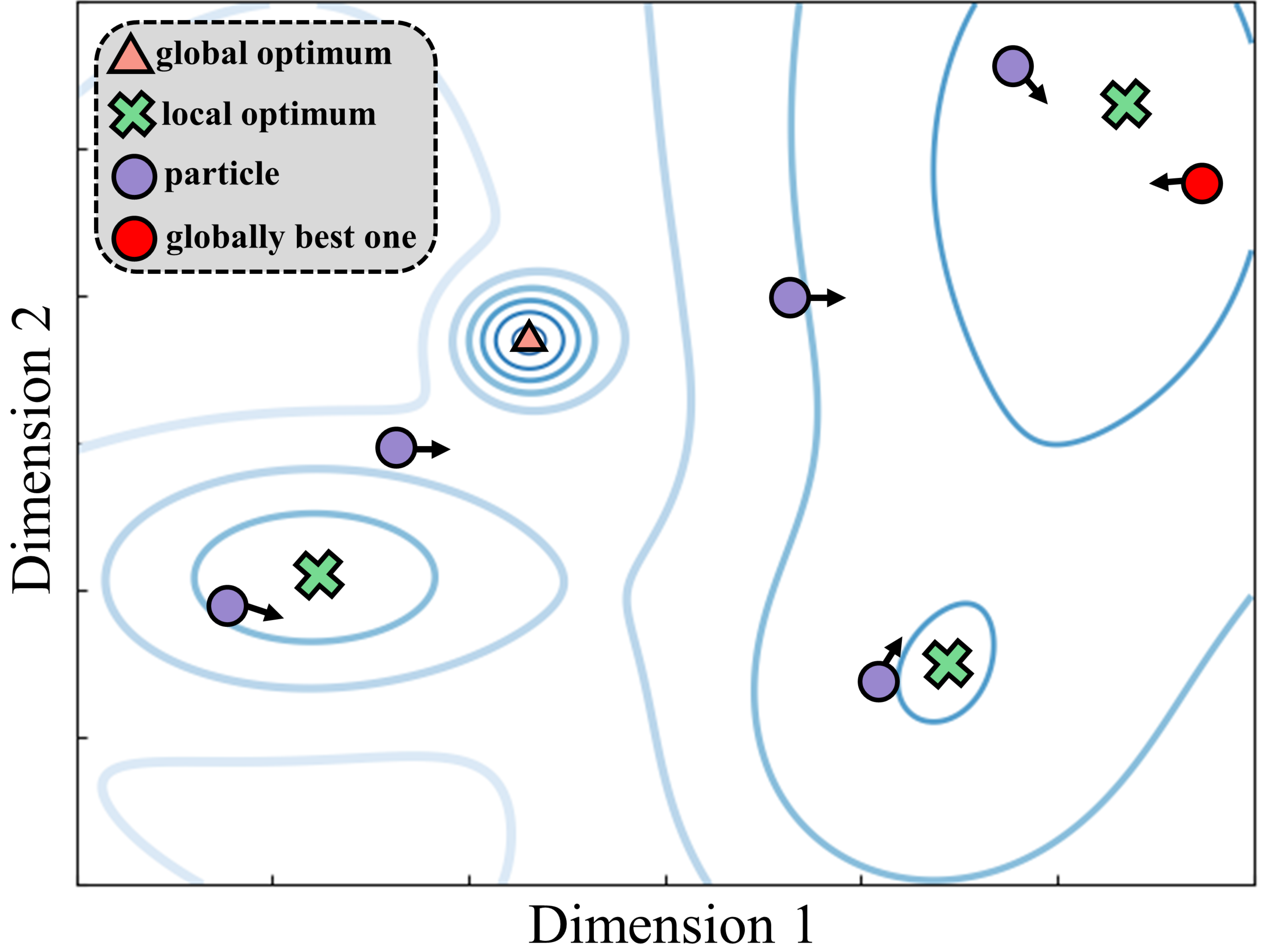}\label{fig:prin_a}}
  \hspace{0.01\linewidth}
  \subfloat[Task $\beta$]{\includegraphics[width=1.7in]{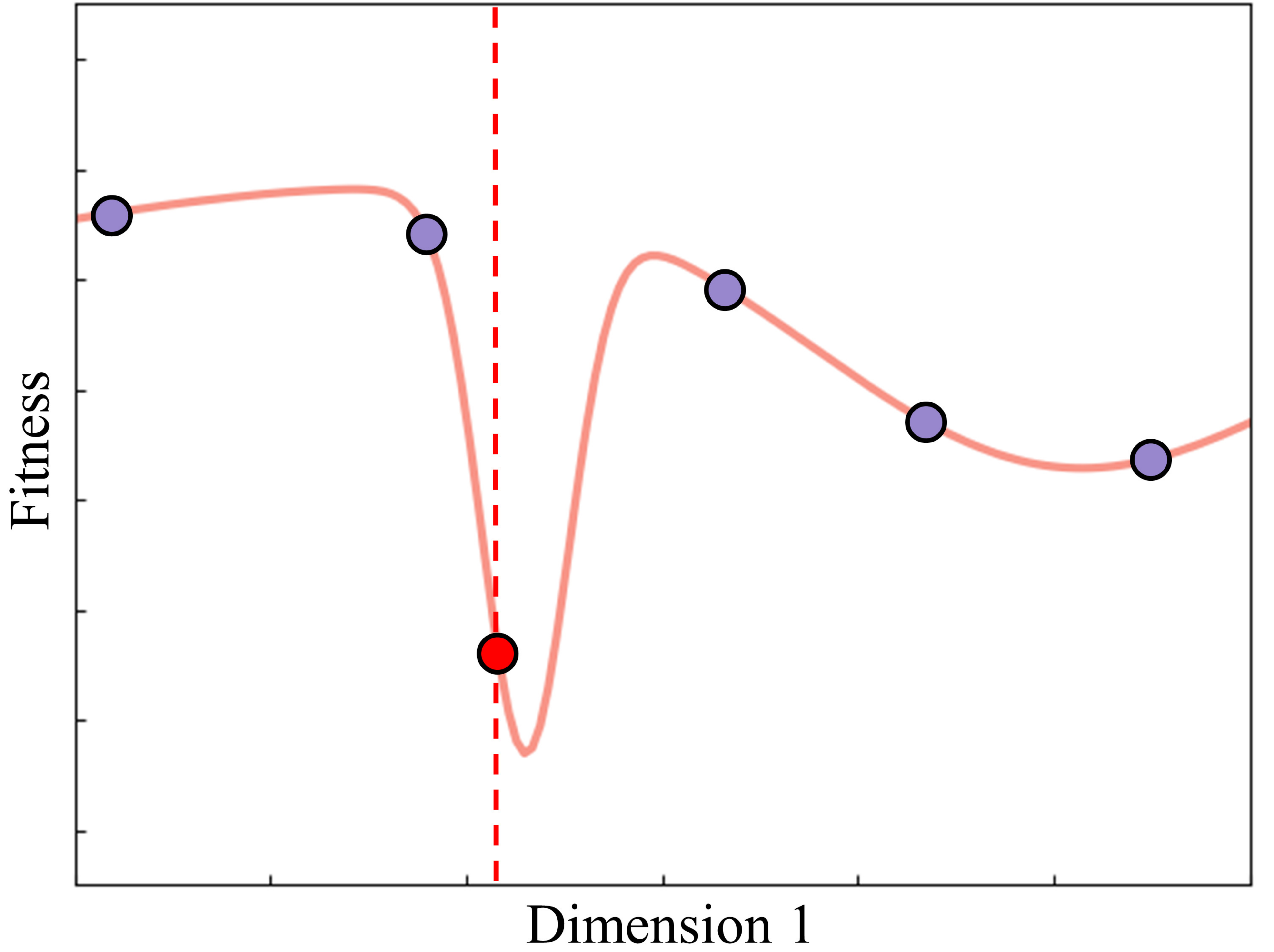}\label{fig:prin_b}}
  \vfill
  \subfloat[After knowledge transfer]{\includegraphics[width=1.7in]{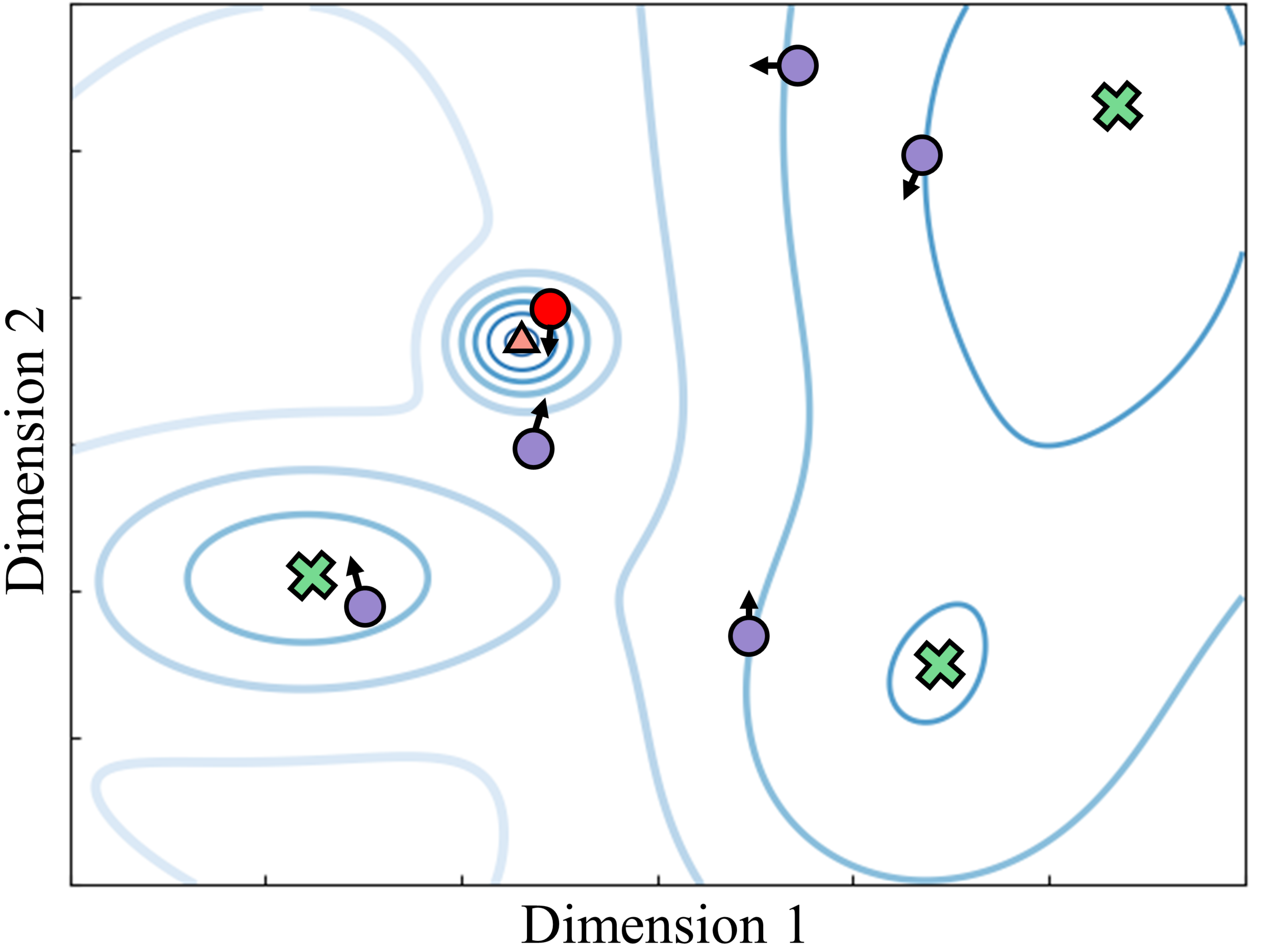}\label{fig:prin_c}}
  \hspace{0.01\linewidth}
  \subfloat[Before knowledge transfer]{\includegraphics[width=1.7in]{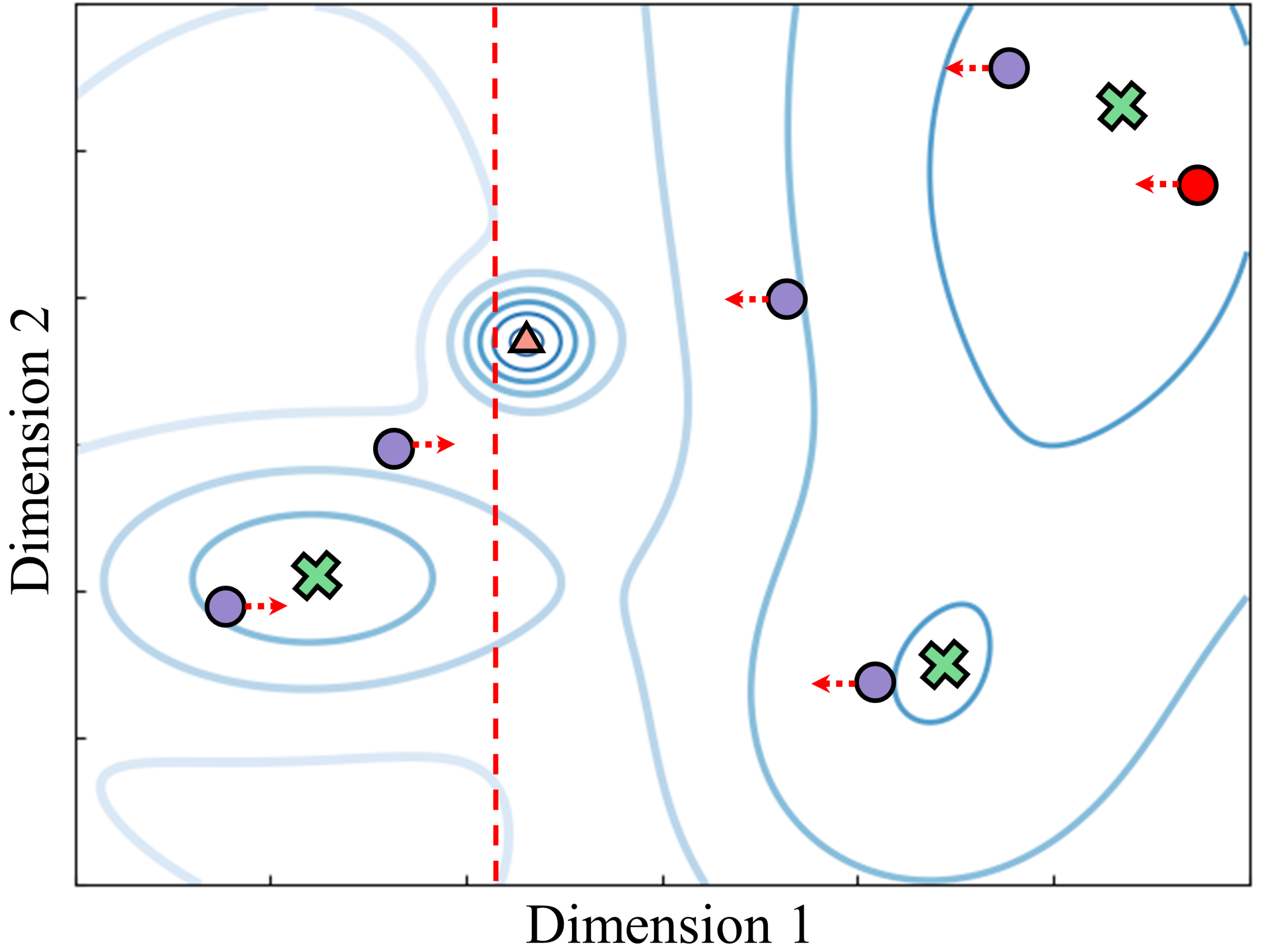}\label{fig:prin_d}}
  \caption{Illustration of the advantage and application of solution space cutting. 
  (b) shows the curve of Task $\beta$, while the other three subfigures display contour plots of Task $\alpha$.
  The dimension of Task $\alpha$ and $\beta$ are 2 and 1, respectively.
  The black and red arrows represent the velocity of the particle and the velocity component from the guidance, respectively.
  The red dotted line represents the guidance from Task $\beta$.
  }
  \label{fig:principle}
  \end{figure}

The idea of \emph{solution space cutting} is used to generate multiple tasks in this article.
Fig. \ref{fig:principle} illustrates the advantage of solution space cutting and its application through the function landscape and population.
Fig. \ref{fig:prin_a} shows the contour of Task $\alpha$, while Fig. \ref{fig:prin_b} displays the curve of Task $\beta$.
Task $\beta$ is generated from Task $\alpha$ by cutting the solution space.
In this paper, cutting solution space is achieved by reducing the dimension of search space.
As shown in these two subfigures, the local optima reduce through cutting the solution space.
Due to smaller solution space and less local optima, Task $\beta$ is more likely to discover the optimal solution.
Based on this observation, we inject the idea of solution space cutting into EMTO.
In the multitasking setting, the optimal solution of Task $\beta$ can be used to provide guidance to Task $\alpha$ in partial dimensions via knowledge transfer.
In Fig. \ref{fig:prin_d}, the red dotted line represents the guidance for Task $\alpha$ in the first dimension.
Then, due to the guidance, the particles in Task $\alpha$ possibly escape local optima and find the optimal solution as shown in Fig. \ref{fig:prin_c}.

Overall, this article aims at further investigating the capability of evolving registration by injecting the ideas of EMTO into algorithm.
The new approach is termed EMTR-SSC, indicating \emph{Evolutionary Multi-Task Registration with Solution Space Cutting} for point cloud data. 
The core idea of EMTR-SSC is to leverage the knowledge learned from the task with reduced solution space to provide guidance towards promising search areas to the original task.
In other words, this method reduces falling into local optima via a novel multi-task configuration and a redesigned knowledge transfer mechanism. 
The performance evaluation of EMTR-SSC is conducted on four object-scale datasets and one scene-scale dataset, and it is compared with 15 approaches, containing 8 evolving ones, 4 traditional ones and 3 deep learning ones, to demonstrate its effectiveness.

The main contributions of this article are given as follows.
\begin{enumerate}
  \item {
  We propose a novel multi-task configuration based on solution space cutting and an evolutionary multi-task registration approach EMTR-SSC for improving the success ratio of evolving point cloud registration.
  Besides, a fitness function, \emph{M-estimator Chamfer Distance}, is designed to further enhance the accuracy and the capability of handling point clouds with various overlap rates.
  }
  \item { 
  A novel \emph{Knowledge Complement} strategy is designed and incorporated into the knowledge transfer mechanism, which enhances the quality of the transferred knowledge.
  In this strategy, the classical numerical optimization method is embedded into evolutionary optimization as a module.
  }
  \item {
  We develop a \emph{Sparse-to-Dense} strategy that switches the point clouds used for evaluation to reduce unnecessary computational cost. 
  Experiments have proven that it can reduce the computational cost by nearly 50\% while minimizing the accuracy loss.
  In addition, a computational cost metric considering the characteristics of the evolving registration is introduced.
  }
  \end{enumerate}
}

The rest of this article is organized as follows. 
Section \ref{sec:Preliminaries} introduces the background and related work about this research topic. 
Section \ref{sec:Methodology} details the proposed method EMTR-SSC. 
Section \ref{sec:ED} describes the experimental design and Section \ref{sec:RD} presents the experimental results and analysis of them. 
Finally, Section \ref{sec:Conclusion} summarizes the work in this article. 

\section{Preliminaries}
\label{sec:Preliminaries}
In this section, we first describe the background knowledge of PCR, EMTO and PSO. 
Then, we review the related work on PCR methods and EMTO methods.
\subsection{Background Knowledge}
\subsubsection{PCR}
Assuming that source point cloud $\mathcal{P}$ and target point cloud $\mathcal{Q}$ are scanned from the same object or scene. 
Let $\mathcal{P} = \{\boldsymbol{p}_i\}_{i=1}^N$ and $\mathcal{Q} = \{\boldsymbol{q}_i\}_{i=1}^M$, where $\boldsymbol{p}_i, \boldsymbol{q}_i \in \mathbb{R}^3$ are 3D coordinates.
$N$ and $M$ are the number of points in the source and target point cloud, respectively.
Since objective functions of PCR have various forms, it is difficult to represent them using a uniform function. 
Hence, an abstract error metric $\mathfrak{E}(\cdot)$ contains the inputs, $\mathcal{P}$ and $\mathcal{Q}$, and outputs, rotation matrix $\boldsymbol{R}$ and translation vector $\boldsymbol{t}$, is employed to represent PCR problems:

\begin{equation}
  \begin{aligned}
    \mathop{\arg\min}_{\boldsymbol{R} \in \mathcal{SO}(3), \boldsymbol{t} \in \mathbb{R}^3} \quad & \mathfrak{E}_{\mathcal{P}, \mathcal{Q}} (\boldsymbol{R}, \boldsymbol{t})  \\
  \end{aligned}
\end{equation}
where $\mathcal{SO}(3)$ represents the 3D special orthogonal group.
For our study, the specific objective function is the fitness function of Task $\alpha$ in Subsection \ref{subsec:fit}.
For evolutionary optimization, the search space of PCR problem is special Euclidean group $\mathcal{SE}(3)$ \footnote{
$\mathcal{SE}(3) = \left\{
\begin{bmatrix} \boldsymbol{R} & \boldsymbol{t} \\
                \boldsymbol{0} & 1              \\  \end{bmatrix}
\in \mathbb{R}^4 | \boldsymbol{R} \in \mathcal{SO}(3), \boldsymbol{t} \in \mathbb{R}^3 
\right\}$
} and its decision vector is rotation and translation parameters. 
The dimension of both rotation and translation parameters is three in our study.
In general, $\mathcal{P}$ and $\mathcal{Q}$ only partially overlap.
Hence, how to design an error evaluation that can cope with partial overlap is the key point.
Besides, choosing an appropriate solving method is also very important for PCR.
\subsubsection{EMTO}
EMTO \cite{gupta2015multifactorial, ong2016evolutionary, bai2022multitask} is a promising and attractive direction for researchers in the field of EC, and its core idea is that multiple tasks share useful information to achieve faster convergence or to help a few tasks escape from local optima.
The framework of EMTO with $N$ tasks can be expressed mathematically as follows:
\begin{equation}
  \begin{aligned}
    \boldsymbol{x}_i^* = \mathop{\arg\min}_{\boldsymbol{x}_i \in \Omega_i} & F_i(\boldsymbol{x}_i), \; i = 1, \ldots, N \\
    \boldsymbol{x}_i & = (x_i^1, \ldots, x_i^{D_i})
  \end{aligned}
\end{equation}
where $\boldsymbol{x}_i^*$ and $F_i(\cdot)$ denote the optimal solution and the fitness function of the $i$-th optimization task, respectively.
$\Omega_i$ represents the solution space of $i$-th task with $D_i$ dimensions.
\subsubsection{PSO}
In PSO \cite{kennedy1995particle}, each particle represents one feasible solution to the optimization problem.
At the $t$-th iteration, the $k$-th particle has two properties, i.e. position and velocity, which both have $D$ dimensions and are denoted by 
$\boldsymbol{x}_k^t = (x_{k1}^t, x_{k2}^t,\dots, x_{kD}^t)$ and $\boldsymbol{v}_k^t = (v_{k1}^t, v_{k2}^t,\dots, v_{kD}^t)$.
During the search process, the velocity update relies on two positions, one is the position with the best fitness found so far for each particle, {\it pbest}, and the other is the best position for the whole swarm, {\it gbest}. 
The velocity and the position update formulas are as follows:
\begin{subequations}
  \begin{align}
    v_{kd}^{t+1} = & \omega v_{kd}^t + c_1  r_1  (pbest_{kd}^t - x_{kd}^t) \notag\\
        & \quad + c_2  r_2  (gbest_d^t - x_{kd}^t), \quad r_1,r_2 \in [0, 1] \label{equ:PSO_v} \\ %
    x_{kd}^{t+1} = & x_{kd}^t + v_{kd}^{t+1} \label{equ:PSO_p}
  \end{align}
  \end{subequations}
where $\omega$ is the inertia weight, $c_1$ and $c_2$ represent the acceleration coefficients, $r_1$ and $r_2$ denote random numbers independently generated by the uniform distribution.

\subsection{Related Work}
\subsubsection{PCR}
The traditional approaches are reviewed first.
Iterative closest point (ICP) \cite{besl1992method}, the milestone in the field of registration, models the PCR as a least squares (LS) problem, and solves it using singular value decomposition (SVD) or quaternions in an iterative manner.
But ICP only handles the case where two point clouds overlap completely, leading to limited performance when it comes to real world registration problems. 
Then trimmed ICP (TrICP) \cite{chetverikov2005robust}, based on the least trimmed squares approach, was designed to tackle outliers caused by partial overlap.
Unlike the above two methods that establish point-to-point correspondences, normal distributions transform (NDT) \cite{biber2003normal} models the distribution of all points by a combination of Gaussian distributions in cells and utilizes Newton's algorithm to optimize.
In general, these approaches only guarantee convergence to local optimum, which causes them to be sensitive to the initial poses.
Therefore, these approaches are commonly performed after the coarse registration using both handcrafted features \cite{rusu2009fast,salti2014shot} and RANSAC \cite{fischler1981random}.
In recent years, feature-based registration methods have received extensive attention again as they can achieve global registration and have less computational cost.
Fast global registration (FGR) \cite{zhou2016fast} evaluates registration error by using a scaled Geman-McClure estimator and then optimizes objective function by the Gauss-Newton method.
Moreover, \cite{li2020gesac,yang2021sac,chen2022sc2,yang2022correspondence} employ geometric constraints to discover the correct ones, the inliers, in all given correspondences obtained by feature.
But these methods are less accurate than local registration methods and rely on the means of removing outliers.

Next are deep learning approaches.
Through multiple 3D convolution operations, 3DMatch \cite{zeng20173dmatch} extracts high-dimensional features from the volumetric patch for aligning point clouds.
PPFNet \cite{deng2018ppfnet} with an N-tuple loss merges global information into local descriptors in order to make descriptors more discriminative and robust.
Unlike extracting features from the sampled points, the feature-metric method \cite{huang2020feature} minimizes the differences of feature maps acquired from whole point clouds to regress transformation parameters.
The disadvantage of this category is that it requires large training data and has poor generalization ability.

Finally, evolving approaches are introduced. 
Silva {\it et al.} \cite{silva2005precision} proposed a hybrid genetic algorithm (GA) for registration, where the fitness function is switched from the one based on mean square error (MSE) to the one using designed surface interpenetration measure (SIM) later in the evolutionary process.
The SIM-based objective function cannot be optimized through numerical optimization methods, and naturally, EC is chosen to solve for the global optimum.
Li {\it et al.} \cite{li2017differential} designed a novel point descriptor for correspondence establishment in evaluation, and a modified differential evolution (DE) is conducted for the search of transformation.
Zhu {\it et al.} \cite{zhu2016automatic} applied the GA to provide the initial pose for fine registration with TrICP, which solves the local convergence problem of ICP-based algorithms.
These aforementioned studies only investigate the performance of evolving single-task for PCR.
And none of them considers the successful registration ratio. 

\begin{figure*}[!t]
  \centering
  \includegraphics[width=6.8in]{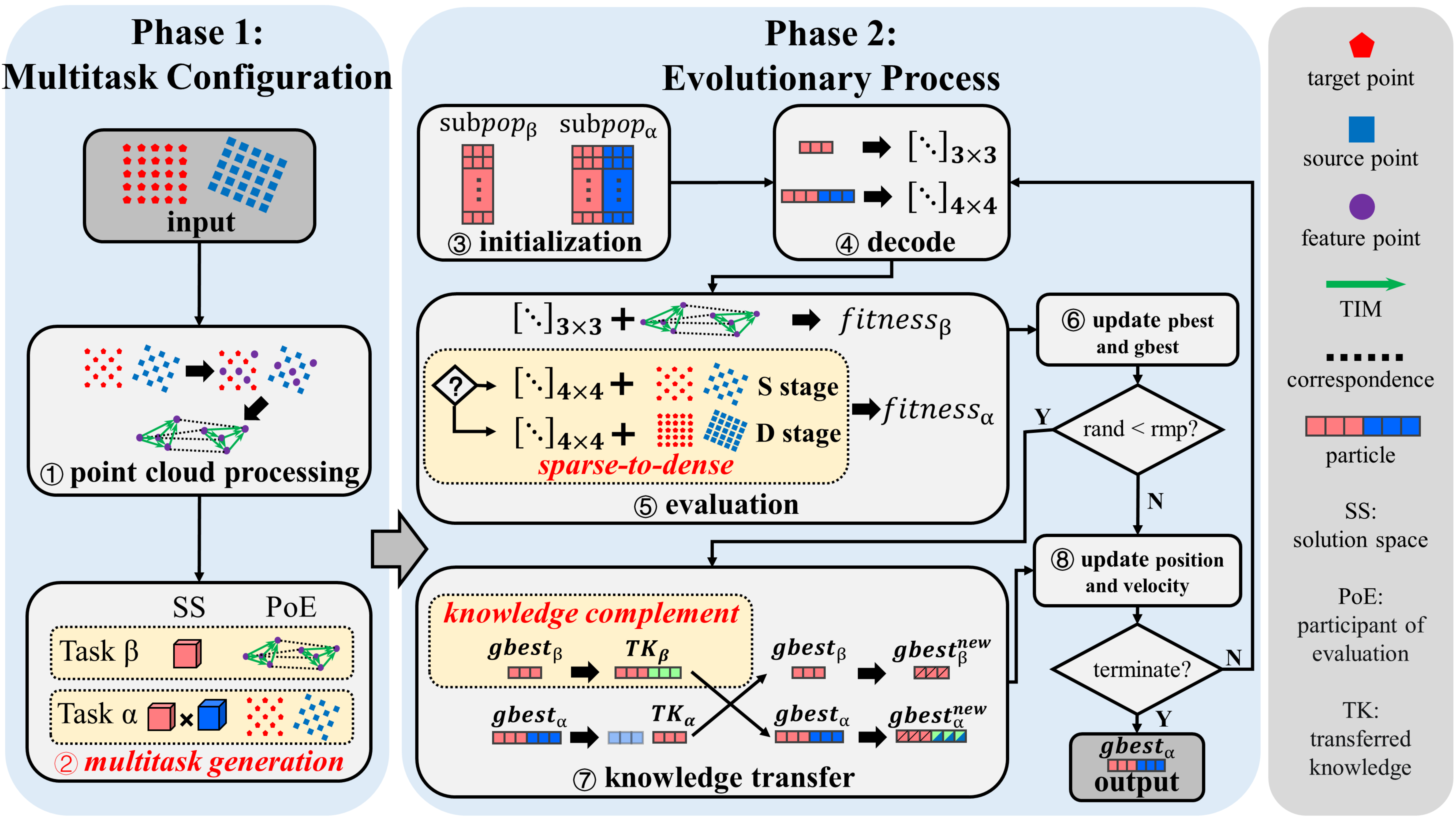}
  \caption{Overview of evolutionary multi-task registration with solution space cutting for point cloud.}
  \label{flowchat}
  \end{figure*}

\subsubsection{EMTO}
In the past few years, the popularity of EMTO research has continued to rise. 
The first product derived from the EMTO paradigm is multifactorial evolutionary algorithm (MFEA) \cite{gupta2015multifactorial}, which mimics the multifactorial inheritance in bio-culture.
MFEA enables individuals to handle the tasks they are good at through skill factors and to transfer beneficial knowledge via crossover which is inherent in GA.
To allay the negative transfer, Bali {\it et al.} \cite{bali2019multifactorial} developed the MFEA-II, the pioneer of intelligent knowledge transfer mechanisms, whose transfer parameters have adaptive and inter-task specific characteristics. 
Considering the restrictive resources, the evolutionary multi-task algorithm with dynamic resource allocating strategy (MTO-DRA) \cite{gong2019evolutionary} dynamically allocates resources according to the requirements of tasks to enhance the efficiency of resource utilization.
Multi-task evolutionary algorithm based on anomaly detection (MTEA-AD) \cite{wang2021solving} integrates algorithm in the machine learning field with EMTO, which realizes the adaptive transference of knowledge and the reduction of negative transfer.
Li {\it et al.} \cite{li2022evolutionary} developed the differential evolution algorithm with an online resource allocation strategy (DEORA), which modifies the knowledge transfer mechanism and appends resource allocation strategy for the special multitasking optimization problem.
Although all the aforementioned EMTO methods have great performance in test functions, they are not suitable for directly optimizing the functions of PCR, whose landscapes are more complex than those of test functions.
The poor performance of these methods concerning the PCR problem will be provided in Section \ref{sec:RD} through experimental results. 

Apart from the above methods, some studies have applied EMTO to PCR.
Multitasking multiview point cloud registration (MTPCR) \cite{wu2022multi} is the first method that combines EMTO with PCR, which accelerates the convergence for the multi-view registration task and achieves highly accurate results by considering global consistency. 
However, due to the difference between multi-view and pairwise registration, this method is not suitable for the pairwise registration problem studied in this paper.
Evolutionary multiform point cloud registration (EMFPCR) \cite{wu2022evolutionary} utilizes two fitness functions with different registration attributes, i.e., robustness and precision, to construct multi-task, which makes the method both robust to noise and highly precise. 
Nevertheless, the investigation in terms of the success ratio is inadequate in this study.

\section{Methodology}
\label{sec:Methodology}
In this section, we introduce the proposed EMTR-SSC approach.
First, the overall approach is presented.
Next, the multi-task configuration and the designed fitness function for each task are detailed.
After that, the knowledge transfer mechanism embedded with the knowledge complement strategy is described.
Finally, the resource-saving strategy, termed sparse-to-dense, is introduced.

\subsection{Overview of EMTR-SSC}
Fig. \ref{flowchat} shows the overview of the proposed EMTR-SSC approach, where our contributions in this article are marked in red font with italic style. 
The input consists of a target point cloud (red pentagon) and a source one (dark blue square), and the output is a 6D parameter including the rotation parameter (rosy) and translation one (blue). 
As can be seen from Fig. \ref{flowchat}, EMTR-SSC consists of two phases. 
The first phase contains the configuration of two tasks, Task $\alpha$ and Task $\beta$.
The solution space and the participant of evaluation are used to represent two tasks in step 2, and the rosy and blue cubes denote the rotation and translation space, respectively. 
Task $\alpha$ searches the whole rigid transformation space, while Task $\beta$ uses the reduced solution space.
They evaluate fitness based on sampled point clouds and translation invariant measurements (TIMs), respectively.
A series of processing is performed before the evolution process to acquire them (step 1).
TIM will be detailed in Subsection \ref{subsec:MC}.

The second phase includes the optimization process of two tasks based on PSO and EMTO.
Most of the steps are inherited from multi-task PSO.
First, two sub-populations corresponding to two tasks with different dimensions are randomly initialized (step 3).
Next, each particle is decoded to obtain one transformation which will be applied to source data (step 4). 
During the evaluation, the transformed source data and the target one are used to calculate registration error by task-specific fitness function for evaluating the particles (step 5).
Step 5 also employs the sparse-to-dense strategy to reduce unnecessary computational cost in Task $\alpha$, where the point clouds in the S and D stages differ in terms of point density.
After the update of {\it gbest} and {\it pbest} (step 6), the algorithm determines whether to perform knowledge transfer.
If knowledge transfer is conducted (step 7), {\it gbest} will be updated again using the shared knowledge.
Especially, the designed knowledge complement strategy completes the incomplete knowledge from Task $\beta$ before transferring.
In step 8, the velocity and position of the particles are updated sequentially.
Finally, if the termination condition is satisfied, the {\it gbest} in subpopulation $\alpha$ is taken as the output of the whole algorithm, otherwise the iteration continues. 
For better presentation of details, the pseudocode of EMTR-SSC is given in Algorithm \ref{alg:framework}.

\begin{algorithm}[!t]
  \caption{Framework of EMTR-SSC}\label{alg:framework}
  \LinesNumbered
  Process inputs and generate two tasks (\ref{subsec:MC})\;
  Initialize two subpopulations\;
  \For{$it\mathrm{\ in\ range (1, \textit{MaxIt})}$}
      {
          Decode and evaluate particles for each subpopulation (\ref{subsec:fit})\;
          Update $pbest$ and $gbest$ for each subpopulation\;
          \If{random number $< rmp$}
          {
              /* knowledge transfer (\ref{subsec:KT}) $(\beta \rightarrow \alpha)$ */ \\
              Decoded {\it gbest} of subpopulation $\beta$ to Euler angles $\theta_1 , \theta_2, \theta_3$ and transform them to rotation matrix $\boldsymbol{R}_{srch}$ using Eq. \eqref{equ:eul2rotm}\;
              \eIf{{\it gbest} of subpopulation $\beta$ updates}
              {
              Estimate $\boldsymbol{t}_{est}$ using Algorithm \ref{alg:trans_est}\;
              }
              {
              Use the previously estimated $\boldsymbol{t}_{est}$\;
              }
              Update crossover parameter $\textit{cr}^{\alpha}$, $\textit{cr}^{\beta}$\;
              Encode $\theta_1 , \theta_2, \theta_3$ and $\boldsymbol{t}_{est}$ to create $\textit{gbest}^{\beta}$\;
              Calculate $\textit{gbest}^{\alpha}_{new}$ using Eq. \eqref{equ:EMTO_v}\;
              /* knowledge transfer $(\alpha \rightarrow \beta)$ */ \\
              Decode {\it gbest} of subpopulation $\alpha$ and extract Euler angles $\theta_1, \theta_2, \theta_3$ to create $\textit{gbest}^{\alpha}$\;
              Calculate $\textit{gbest}^{\beta}_{new}$ using Eq. \eqref{equ:EMTO_v}\;
          }
          \If{$it$ is equal to $\lceil \delta * \textit{MaxIt} \rceil$}
          {
              /* Sparse-to-Dense in Task $\alpha$ (\ref{subsec:S2D}) */ \\
              Switch the fitness function and point clouds for evaluation\;
              Evaluate current positions, $pbest$s and $gbest$\;
              Update $pbest$s and $gbest$\;
          }
          Update velocity and position for each subpopulation\;
      }
  \end{algorithm}

\subsection{Multi-task Configuration}
\label{subsec:MC}
\subsubsection{Task $\alpha$}
High accuracy is the primary concern of Task $\alpha$.
It utilizes point clouds, which contain more detailed information compared with TIMs, to evaluate registration error, and Task $\alpha$ optimizes rotation and translation parameters simultaneously.
Additionally, Task $\alpha$ takes into account the ability to deal with various overlap rates.
To this end, a novel robust fitness function for Task $\alpha$ is designed, which will be detailed in Subsection \ref{subsec:fit}. 
However, due to the complex fitness function and large solution space, Task $\alpha$ suffers from local optima.

\subsubsection{Task $\beta$}
Two keys are under consideration for Task $\beta$.
The first one is the degree of closeness of optimal solutions of Task $\alpha$ and Task $\beta$ in solution space, which can significantly affect the performance of EMTO.
However, if none of the tasks find valuable information, two tasks that meet the above condition cannot help each other via knowledge transfer.
Therefore, the second one is the guarantee of obtaining useful information.
Based on the above two keys, TIM is chosen as the participant of evaluation in Task $\beta$, and the reasons are as follows. 
The optimal transformation of aligning TIMs is the same as that of aligning point clouds, since TIMs are extracted from point clouds.
On the other hand, when leveraging TIM to register, the search space only includes rotation, which reduces the difficulty of finding the optimal solution, i.e., valuable information.

Methods resembling TIM have been employed in recent work \cite{bustos2017guaranteed,yang2020teaser}. 
Since TIM can effectively remove translation parameters from all parameters to be optimized, and it is easy to generate, it is chosen to participate in evaluation in Task $\beta$.
The process of generating TIMs is shown in step 1 of Fig. \ref{flowchat}.
Firstly, the input point clouds are downsampled to make the distribution of points uniform and reduce the computational cost of subsequent operations. 
Secondly, the points with feature vectors are extracted from sampled point clouds, and putative point correspondences are established according to the similarity of feature vectors.
Thirdly, the TIMs are acquired from the feature points and correspondences. 
Given the correct correspondences $(\boldsymbol{p}_j, \boldsymbol{q}_{\varphi(j)}), j \in \{1, \ldots, N\}$ with noise, they obey the following equation:
\begin{equation}
  \label{equ:point_corres}
  \boldsymbol{q}_{\varphi(j)} = \boldsymbol{R}^*\boldsymbol{p}_j + \boldsymbol{t}^* + \epsilon_j
\end{equation}
where $\varphi(\cdot): \{1, \ldots, N\} \to \{1, \ldots, M\}$ is a function for matching two points in different point clouds through feature vectors. 
$\boldsymbol{R}^*$ and $\boldsymbol{t}^*$ denote the correct rotation matrix and translation vector, $\epsilon_j$ represents the measurement noise.
In the same point cloud, every two feature points can generate one TIM, the generative formula of TIM is:
\begin{subequations}
  \begin{align}
  \boldsymbol{q}_{\varphi(k)} - \boldsymbol{q}_{\varphi(j)} 
  & = \boldsymbol{R}^* (\boldsymbol{p}_k - \boldsymbol{p}_j) + \epsilon_k - \epsilon_j \label{equ_sub_a} \\ %
  \vec{\boldsymbol{q}}_{\varphi(jk)} 
  & = \boldsymbol{R}^*\vec{\boldsymbol{p}}_{jk} + {\boldsymbol{\epsilon}}_{jk} \label{equ_sub_b}
  \end{align}
  \end{subequations}

For simplicity, we use $\vec{\boldsymbol{q}}_{\varphi(jk)} = \boldsymbol{q}_{\varphi(k)} - \boldsymbol{q}_{\varphi(j)}$, 
$\vec{\boldsymbol{p}}_{jk} = \boldsymbol{p}_k - \boldsymbol{p}_j$ to represent TIMs, while ${\boldsymbol{\epsilon}}_{jk} = \boldsymbol{\epsilon}_{k}-\boldsymbol{\epsilon}_{j}$ is the measurement noise.
Compared with registration based on the point in Eq. \eqref{equ:point_corres}, Eq. \eqref{equ_sub_b} shows the registration using TIMs needs not to solve translation $\boldsymbol{t}$.
Therefore, Task $\beta$ only searches the rotation space and the idea of solution space cutting is implemented.

\subsection{Fitness Function}
\label{subsec:fit}

\subsubsection{Chamfer Distance with M-estimator in Task $\alpha$}
Chamfer distance is a commonly used registration error metric. 
Based on this, the PCR can be represented by the optimization problem as follows:
\begin{equation}  \label{equ:CD}
  \mathop{\min}_{\boldsymbol{R},\boldsymbol{t}}
  \sum\limits^N_{i=1} \Vert \boldsymbol{R}\boldsymbol{p}_i + \boldsymbol{t} - \boldsymbol{q}_{\phi(i)} \Vert^2 + \sum\limits^M_{i=1}\Vert \boldsymbol{R}\boldsymbol{p}_{\phi(i)} + \boldsymbol{t} - \boldsymbol{q}_i \Vert^2
\end{equation}
where $\phi(\cdot)$ is a function for searching the closest point in the other point cloud.  
Given the $i$-th point in one point cloud, the $\phi(i)$-th point in the other one is its closest point, which is expressed mathematically as follows:
\begin{equation} \label{equ:cloest}
  \begin{aligned}
  & \boldsymbol{q}_{\phi(i)} = \mathop{\arg\min}\limits_{\boldsymbol{q} \in \mathcal{Q} } \Vert \boldsymbol{R}\boldsymbol{p}_i + \boldsymbol{t} - \boldsymbol{q} \Vert, \; \boldsymbol{p}_i \in \mathcal{P}  \\
  & \boldsymbol{p}_{\phi(i)} = \mathop{\arg\min}\limits_{\boldsymbol{p} \in \mathcal{P} } \Vert \boldsymbol{R}\boldsymbol{p} + \boldsymbol{t} - \boldsymbol{q}_i \Vert, \; \boldsymbol{q}_i \in \mathcal{Q} 
  \end{aligned}
\end{equation}
However, chamfer distance works only when the point clouds overlap almost completely.
In practice, due to partial overlap, points outside the overlap region do not have matched points in the other point cloud.
Eq. \eqref{equ:CD} assigns a matched point for every point whether its matched point exists or not, which produces false correspondences, also called outliers.
Generally, one outlier has more influence on the optimization than several inliers, correct correspondences, combined. 
Therefore, by replacing the $L_2$ norm with the robust M-estimator, we develop a new registration error metric, M-estimator Chamfer Distance (MCD).
The fitness function of Task $\alpha$ based on MCD can be written as:
\begin{equation}
  \textit{f}^\alpha=
  \sum\limits^{N}\limits_{i=1}\rho(\Vert \boldsymbol{R}\boldsymbol{p}_i + \boldsymbol{t} - \boldsymbol{q}_{\phi(i)} \Vert) + 
  \sum\limits^{M}\limits_{i=1}\rho(\Vert \boldsymbol{R}\boldsymbol{p}_{\phi(i)} + \boldsymbol{t} - \boldsymbol{q}_i \Vert)
\end{equation}
where $\Vert \boldsymbol{R}\boldsymbol{p}_i + \boldsymbol{t} - \boldsymbol{q}_{\phi(i)} \Vert$ and $\Vert \boldsymbol{R}\boldsymbol{p}_{\phi(i)} + \boldsymbol{t} - \boldsymbol{q}_i \Vert$ are the residuals of the point pair, and they can be represented by $r$.
$\rho(\cdot)$ denotes the M-estimator which is a piecewise function as follows:
\begin{equation}  \label{equ:M-estimator}
  \rho(r) = 
  \begin{cases} 
  1-(1-(\frac{r}{c})^2)^3, & \vert r \vert \leq c \\ 
  1, & \vert r \vert > c 
  \end{cases}
\end{equation}
where $c$ is the threshold to determine whether the point pair is an inlier or not. 
If the residual is greater than the threshold, the point pair is considered an outlier.

\subsubsection{Modified Consensus Maximization in Task $\beta$}
In practice, the inadequate accuracy of extracted feature leads to the wrong matched TIM pair, the outlier, while the correct matched one is termed as the inlier.
Generally, it is assumed the estimated rotation that maximizes the number of inliers is the correct one.
This method is consensus maximization which detects outliers by comparing the residuals of correspondences with a predefined threshold.
For one matched TIM pair $(\vec{\boldsymbol{p}}_{jk}, \vec{\boldsymbol{q}}_{\varphi(jk)})$, the residual is defined as $\Vert \vec{\boldsymbol{q}}_{\varphi(l)} - \boldsymbol{R}\vec{\boldsymbol{p}}_{l} \Vert $, where we use subscript $l$ instead of $jk$ for simplicity. 
$\Vert \cdot \Vert$ represents $L_2$ norm. 
From Eq. \eqref{equ_sub_b}, we can see that threshold is related to noise.
In general, the bound of noise is set to $\tau$.
Due to the subadditivity of norm\footnote{
    $\because \Vert {\boldsymbol{\epsilon}}_{j} \Vert, \Vert {\boldsymbol{\epsilon}}_{k} \Vert \leq \tau
    \quad \therefore \Vert {\boldsymbol{\epsilon}}_{l} \Vert = \Vert {\boldsymbol{\epsilon}}_{jk} \Vert \leq \Vert {\boldsymbol{\epsilon}}_{k} \Vert + \Vert -{\boldsymbol{\epsilon}}_{j} \Vert \leq 2\tau$
  }, we can get $\Vert {\boldsymbol{\epsilon}}_{l} \Vert \leq 2\tau$.
If one TIM pair satisfy $\Vert \vec{\boldsymbol{q}}_{\varphi(l)} - \boldsymbol{R}\vec{\boldsymbol{p}}_{l} \Vert < 2\tau$, 
it is regarded as an inlier under this rotation.
Therefore, the optimization problem can be described as follows:
\begin{equation} 
  \begin{aligned}
    & \mathop{\min}_{\boldsymbol{R} \in \mathcal{SO}(3)}\quad \lfloor \mathcal{C} \backslash \mathcal{L} \rfloor \\
    & s.t.\quad \Vert \vec{\boldsymbol{q}}_{\varphi(l)} - \boldsymbol{R}\vec{\boldsymbol{p}}_l \Vert < 2\tau, \forall l \in \mathcal{L} \subset \mathcal{C}
  \end{aligned}
\end{equation}
where $\lfloor \cdot \rfloor$ denotes the cardinality of set, 
$\mathcal{C}$ and $\mathcal{L}$ are the index set of all TIM pairs and inliers.
There are some plateaus on this function landscape, which makes the population disoriented in the small area.
To specify the optimization direction, the fitness function further considers the sum of the difference between two corresponding TIMs, and it can be expressed as follows:
\begin{equation} \label{equ:fitness1}
  \textit{f}^\beta= \eta \lfloor \mathcal{C} \backslash \mathcal{L} \rfloor + \sum_{l \in \mathcal{L}}{\Vert \vec{\boldsymbol{q}}_{\varphi(l)} - \boldsymbol{R}\vec{\boldsymbol{p}}_l \Vert}
\end{equation}
where $\eta$ is a constant to make two parts of this function at the same order of magnitude.
This is a modified consensus maximization. 
In addition, to further improve the success ratio of Task $\beta$, the threshold is increased to $5\tau$.
The noise bound $\tau$ is usually calculated by the mean point cloud resolution.

Rotation matrix $\boldsymbol{R}$ is an orthonormal matrix\footnote{$\boldsymbol{R}\boldsymbol{R}^T = \boldsymbol{I}$, $\vert\boldsymbol{R}\vert = \pm1$}.
Directly applying $\boldsymbol{R}$ to encode the particle, the updated position is not necessarily a valid rotation matrix.
As a result, this requires many additional operations to make the updated position feasible.
Therefore, {\it Euler angles} are chosen to represent rotation in this article.
Using this representation, each rotation can be represented as three angles $\theta_1, \theta_2$ and $\theta_3$. 
For the sequence of {\it Euler angles} in this study, we adopt the {\it Z-Y-X}, which means the point cloud rotates about the {\it Z} axis by $\theta_1$ first, and then rotates about {\it Y} axis by $\theta_2$, finally rotates about {\it X} axis by $\theta_3$. 
The {\it X, Y, Z} axes are fixed during the rotation process. 
The rotation matrix $\boldsymbol{R}$ can be obtained by these angles via:
\begin{equation}
  \label{equ:eul2rotm}
  \begin{aligned}
  \boldsymbol{R} 
  & = 
  \begin{bmatrix} 
    c_1c_2 & c_1s_2s_3 - s_1c_3 & c_1s_2c_3 + s_1s_3 \\
    s_1c_2 & s_1s_2s_3 + c_1c_3 & s_1s_2c_3 - c_1s_3 \\
    -s_2   & c_2s_3             & c_2c_3             \\ 
  \end{bmatrix}
  \end{aligned}
\end{equation}
where $s$ and $c$ represent {\it sine} and {\it cosine}, e.g. $s_1$ denotes the {\it sine} of $\theta_1$.
In the evolutionary process, we search the position representing {\it Euler angles} and then decode it to a rotation matrix.
The ranges of three angles are $[-\pi, \pi]$, $[-\frac{\pi}{2}, \frac{\pi}{2}]$, $[-\pi, \pi]$, respectively, which constitute rotation search space.

\subsection{Knowledge Transfer}
\label{subsec:KT}
\subsubsection{Transfer Strategy}
The transfer strategy will be introduced from three aspects, the participant of transfer, the controller of transfer and the way of transfer. 

{\it The participant of transfer}: The best particle in each subpopulation is chosen as the participant, since it is most likely to carry information about the optimal solution. 
Some operations need to be performed before transfer. 
The operation for Task $\beta$ is to complement the translation parameters, and for Task $\alpha$, that is to extract the rotation parameters represented by Euler angles. 

{\it The controller of transfer}: Random mating probability ({\it rmp}) is a classical method to control whether to transfer or not \cite{gupta2015multifactorial,li2018evolutionary}.
In each iteration, when the transfer condition (random number $\leq$ {\it rmp}) is met, the useful information discovered in one task is passed on to another task, otherwise, no knowledge transfer occurs.
Hence, a larger {\it rmp} leads to a higher frequency of knowledge transfer, and vice versa.

{\it The way of transfer}: Crossover operation has many advantages for knowledge transfer, such as efficient information sharing, low computational cost, and increased population diversity.
Therefore, the arithmetic crossover is chosen for knowledge transfer, and it can be expressed mathematically as follows:
\begin{equation} \label{equ:EMTO_v}
  gbest_{new}^{\beta/\alpha} = \lambda^{\beta/\alpha} gbest^{\beta/\alpha} + (1-\lambda^{\beta/\alpha}) gbest^{\alpha/\beta}
\end{equation}
where $\lambda^{\beta/\alpha}$ is a linearly varying coefficient.
$\lambda^{\alpha}$ increases linearly during evolution. 
In the early phase, Task $\beta$ can quickly get the rough solution close to the optimal one, and it plays a guiding role in the optimization of Task $\alpha$. 
Since the reachable precision of Task $\beta$ is lower than Task $\alpha$, in the late phase, Task $\beta$ cannot provide sufficient guidance for Task $\alpha$. 
The importance of knowledge transferred from Task $\beta$ needs to be diminished.
On the other hand, $\lambda^{\beta}$ decreases linearly.
This is because the solution of Task $\alpha$ becomes more precise, and it is gradually more helpful for Task $\beta$.

Algorithm \ref{alg:framework} also summarizes the process of knowledge transfer, which is divided into two parts.
The first one is Task $\beta$ transfers knowledge to Task $\alpha$ (from lines 8 to 16).
The second one is Task $\alpha$ shares information with Task $\beta$ (from lines 18 to 19).

\begin{algorithm}[!t]
  \caption{The knowledge complement strategy}\label{alg:trans_est}
  \LinesNumbered
  \KwIn{{\it Source point cloud $\mathcal{P}$}, {\it Target point cloud $\mathcal{Q}$}, {\it Searched rotation matrix} $\boldsymbol{R}_{srch}$, {\it Number of iterations} $MaxIt_{KC}$}
  \KwOut{{\it Estimated translation} $\boldsymbol{t}_{est}$}
  Transform  $\mathcal{P}$ to $T(\mathcal{P})=\{T(\boldsymbol{p}_i)\}_{i=1}^N$ using $\boldsymbol{R}_{srch}$\;
  \For{$k\mathrm{\ in\ range (1, MaxIt_{KC})}$}
      {
      Establish correspondence $(T(\boldsymbol{p}_i), \boldsymbol{q}_{\phi(i)})$ using Eq. \eqref{equ:cloest}\;
      Calculate weight $w_i$ for each correspondence using Eq. \eqref{equ:trans_weight}\;
      Compute weighted centroids ${\bar{\boldsymbol{q}}}_{k} = \frac{ \sum_{i=1}^{n} {w_i \boldsymbol{q}_{\phi(i)}}}{\sum_{i=1}^{n} {w_i}}$, ${\bar{\boldsymbol{p}}}_{k} = \frac{ \sum_{i=1}^{n} w_iT({\boldsymbol{p}}_i)}{\sum_{i=1}^{n} {w_i}}$\; 
      Estimate translation increment $\boldsymbol{t}_{k} = {\bar{\boldsymbol{q}}}_{k} - {\bar{\boldsymbol{p}}}_{k}$\;
      Store $\boldsymbol{t}_{k}$\ and transform $T(\mathcal{P})$ using it;
      }
  Calculate $\boldsymbol{t}_{est}=\sum^{k-1}_{i=1}\boldsymbol{t}_i$\;
  \end{algorithm}

\subsubsection{Knowledge Complement}
Due to the solution space cutting, the solution of Task $\beta$ that only contains rotation parameters is incomplete compared with that of Task $\alpha$.
Incomplete knowledge is helpful for the optimization of Task $\alpha$ but with limited effect, which will be experimentally proved in Section \ref{sec:RD}.
In order to improve the quality of transferred knowledge, the translation parameters need to be complemented.
However, it is infeasible to randomly generate translation parameters, which will lead to a negative transfer.
Therefore, a {\it knowledge complement} strategy is designed, which leverages the rotation parameters searched in Task $\beta$ to roughly estimate the translation parameters by means of the classical numerical optimization method.
First, we model this estimation as a weighted least squares (WLS) problem as follows:
\begin{equation} \label{equ:WLS}
  \mathop{\arg\min}_{\boldsymbol{t} \in \mathbb{R}^3}\sum_{i=1}^{n} {w_i\Vert \boldsymbol{R}_{srch}{\boldsymbol{p}}_i + \boldsymbol{t} - \boldsymbol{q}_{\phi(i)}\Vert ^2}
\end{equation}
where $\boldsymbol{R}_{srch}$ denotes the known rotation matrix searched by Task $\beta$.
$w_i$ represents weight function whose independent variable is residual.
Then, this problem is solved through \emph{iterative reweighted least squares} (IRLS).
Algorithm \ref{alg:trans_est} presents the procedure of the knowledge complement. 
In Algorithm \ref{alg:trans_est}, the source point cloud is first rotated to obtain the transformed source point cloud using searched rotation matrix (line 1). 
For each point in the transformed source point cloud, its closest point in the target point cloud is found to establish correspondence (line 3). 
Next, the weight of each correspondence is calculated through the Tucky weight function as follows (line 4):
\begin{equation} \label{equ:trans_weight}
  w_i = 
  \begin{cases} 
  (1-(\frac{r_i}{d})^2)^2, & r_i \leq d \\ 
  0, & r_i > d 
  \end{cases}, 
  \end{equation}
where $r_i= \Vert T(\boldsymbol{p}_i) -{\boldsymbol{q}}_{\phi(i)} \Vert$ and $d$ is the threshold for detecting outliers.
During optimization, this function assigns small weights (close to 0) to outliers while giving large weights (close to 1) to inliers. 
Therefore, the effect of outliers on the cost is largely discounted.
After that, the translation increment is estimated by weighted centroids of the transformed source and target point cloud (from lines 5 to 6).
In line 7, the transformed source point cloud is updated by translating itself using translation increment  (line 7).
Finally, the estimated translation is computed using all stored translation increments (line 9). 

\begin{table}[!t]
  \caption{Experimental Datasets \label{tab:dataset}}
  \centering
  \renewcommand\arraystretch{1.2}
  \begin{tabular}{cccc}
  \hline
  \textbf{Dataset}                                                            & \textbf{Name}    & \textbf{Point Number} & \textbf{Overlap Rate}  \\ \hline
  \textit{(i)object-scale}                                                    &                  &                       &                        \\
  \multirow{4}{*}{\begin{tabular}[c]{@{}c@{}}Stanford\end{tabular}}           & Armadillo        & 28.2k,27.3k           & 88.1\%,79.9\%          \\
                                                                              & Bunny            & 40.3k,40.1k           & 90.9\%,88.4\%          \\
                                                                              & Dragon           & 41.9k,34.8k           & 94.6\%,90.2\%          \\
                                                                              & Happy            & 78.1k,75.6k           & 86.5\%,85.0\%          \\
  \hdashline
  \multirow{4}{*}{\begin{tabular}[c]{@{}c@{}}U3M\end{tabular}}                & Chef             & 69.0k,68.7k           & 82.1\%,83.6\%          \\
                                                                              & Chicken          & 29.5k,28.8k           & 80.8\%,71.2\%          \\
                                                                              & Parasaurolophus  & 42.9k,37.1k           & 73.5\%,71.7\%          \\
                                                                              & T-rex            & 38.8k,27.3k           & 67.0\%,76.4\%          \\
  \hdashline
  \multirow{4}{*}{\begin{tabular}[c]{@{}c@{}}FGR\end{tabular}}                & Angel            & 13.2k,18.9k           & 61.9\%,51.0\%          \\
                                                                              & Bimba            & 14.0k,12.9k           & 65.2\%,65.0\%          \\
                                                                              & Chinese Dragon   & 16.0k,16.5k           & 62.7\%,59.7\%          \\
                                                                              & Dancing Children & 15.1k,14.5k           & 64.3\%,67.9\%          \\
  \hdashline
  \multirow{2}{*}{\begin{tabular}[c]{@{}c@{}}ModelNet40\end{tabular}}         & Clean            & 1024,1024             & 100\%,100\%            \\
                                                                              & Noisy            & 1024,1024             & 96.4\%,96.4\%          \\ \hline
  
  \textit{(ii)scene-scale}                                                    &                  &                       &                        \\
  \multirow{3}{*}{\begin{tabular}[c]{@{}c@{}}RESSO\end{tabular}}              & 7a               & 461.6k,413.2k         & 74.8\%,75.3\%          \\
                                                                              & 7d               & 246.8k,220.6k         & 80.3\%,83.7\%          \\
                                                                              & 7e               & 222.4k,260.2k         & 74.4\%,81.2\%          \\ \hline
  \end{tabular}
  \end{table}
  
\subsection{Sparse-to-Dense}
\label{subsec:S2D}

Obviously, using more points for evaluation will yield a more accurate transformation estimation.
On the other hand, the computational cost grows as the point number increases, since the nearest neighbor search (NNS) algorithm is called for each point in evaluation. 
Through observation, we found that the difference in accuracy between sparse and dense point clouds is only reflected later in the search process for evolving registration.
Compared with sparse point clouds, the evaluation with dense ones in the early phase consumes much more computational resources, which is unnecessary.
Even worse, the waste will be amplified by the large population and numerous iterations.
For Task $\alpha$, a \emph{sparse-to-dense} strategy based on the aforementioned observation is proposed, which aims to reduce the computational cost by using point clouds with different point densities in different evolution stages.
This strategy can ensure that in the rough (accurate) evolution stage the sparse (dense) point clouds are used for evaluation.

The evolution process of Task $\alpha$ is divided into two stages, and the switch parameter $\delta \in (0, 1)$ controls the change of the stage. 
In the first stage, sparse point clouds are used for evaluation to reduce unnecessary computational cost. 
In the remaining stage, dense point clouds are used to evaluate for obtaining transformation with high accuracy.
At the same time, we modify the fitness function of Task $\alpha$ as follows:
\begin{equation}
    \textit{f}^\alpha=
    \begin{cases}
    \sum\limits^{N_1}\limits_{i=1}\rho(\Vert \boldsymbol{R}\boldsymbol{p}_{1,i} + \boldsymbol{t} - \boldsymbol{q}_{1,\phi(i)} \Vert), 
    &\text{(S stage)}\\
    \sum\limits^{N_2}\limits_{i=1}\rho(\Vert \boldsymbol{R}\boldsymbol{p}_{2,i} + \boldsymbol{t} - \boldsymbol{q}_{2,\phi(i)} \Vert) \\
    \quad + \sum\limits^{M_2}\limits_{i=1}\rho(\Vert \boldsymbol{R}\boldsymbol{p}_{2,\phi(i)} + \boldsymbol{t} - \boldsymbol{q}_{2,i} \Vert), 
    &\text{(D stage)}
    \end{cases}
  \label{equ:S2D}
\end{equation}
where $N_1$, $N_2$ and $M_2$ are the number of points in the sparse source point cloud, dense source one and dense target one, respectively.
In the S stage, not only points are reduced, but also the chamfer distance is not used, which further reduces the computational cost.

When the number of iterations is equal to $\lceil \delta * \textit{MaxIt} \rceil$, the evolutionary process of Task $\alpha$ switches from the S stage to the D stage.
$\lceil \cdot \rceil$ represents the rounding up operation.
All current positions, {\it pbest}s and {\it gbest} are evaluated using fitness function of D stage and dense point clouds (Algorithm \ref{alg:framework} line 24). 
The position with the best fitness is selected to update {\it gbest} and {\it pbest} of each particle is replaced by the better one of its current position and {\it pbest} (line 25).

\begin{table}[!t]
  \centering
  \caption{Different types of rotation and translation and their corresponding ranges.\label{tab:range}}
  \renewcommand\arraystretch{1.4}
  \begin{tabular}{cc}
  \hline
  \textbf{Type}     & \textbf{Range}                                                                                                          \\ \hline
  small rotation($SR$)    & $\vert \alpha \vert, \vert \gamma \vert \leq \frac{\pi}{2},\quad \vert \beta \vert \leq \frac{\pi}{4}$                       \\
  large rotation($LR$)    & $\frac{\pi}{2} < \vert \alpha \vert, \vert \gamma \vert \leq \pi,\quad \frac{\pi}{4} < \vert \beta \vert \leq \frac{\pi}{2}$ \\
  small translation($ST$) & $|t_x|, |t_y|, |t_z| \leq \frac{\textit{diag}}{2}$                                                                    \\
  large translation($LT$) & $\frac{\textit{diag}}{2} < |t_x|, |t_y|, |t_z| \leq \textit{diag}$                                       \\ \hline
  \end{tabular}
  \end{table}

\section{Experimental Design}
\label{sec:ED}

For the purpose of examining the effectiveness of the proposed EMTR-SSC, a series of experiments have been conducted on the chosen datasets. 
This section includes the benchmark methods, the benchmark datasets, the metric for quantifying the performance, the preparations before evolution and the parameter settings.

\subsection{Benchmark Methods and Datasets}
The benchmark methods can be divided into three categories. 
Firstly, 1) PSO \cite{kennedy1995particle}; 2) JADE \cite{zhang2009jade}; 3) APSO \cite{zhan2009adaptive}, are \emph{evolutionary methods} in the single task way, which are applied to evaluate the effectiveness of multitasking. 
Next, 4) MFEA \cite{gupta2015multifactorial}; 5) MFEA-II \cite{bali2019multifactorial}; 6) MTEA-AD \cite{wang2021solving}; 7) DEORA \cite{li2022evolutionary}, belong to \emph{EMTO methods} employed to demonstrate the efficiency of our designed knowledge transfer. 
Notably, the latter three algorithms are state-of-the-art ones of EMTO.
The state-of-the-art evolving registration method, 8) EMFPCR \cite{wu2022evolutionary}, is also involved in the comparison.
Finally, the \emph{traditional methods}, i.e. 9) ICP \cite{besl1992method}; 10) NDT \cite{biber2003normal}; 11) TrICP \cite{chetverikov2005robust}; 12) FGR \cite{zhou2016fast}, and the \emph{deep learning methods}, i.e. 13) PointNetLK \cite{aoki2019pointnetlk}; 14) DCP \cite{wang2019deep}; 15) RPM-Net \cite{yew2020rpm}, will compare with the proposed approach in terms of accuracy.

More specifically, the evolutionary single-task methods employ the same fitness function as the D stage of Eq. \eqref{equ:S2D}. 
With regard to comparative EMTO methods, except for the sparse-to-dense mechanism, they have the same setting about fitness function as the proposed approach. 
For the sake of fair accuracy comparison, the initial pose is provided for traditional local registration methods, i.e., method 9) - 11), via RANSAC.

In these experiments, four broadly used PCR benchmark datasets with different difficulties are selected as the test set to examine the performance of EMTR-SSC.
Four of these datasets are object-scale ones, including Stanford\footnote{http://graphics.stanford.edu/data/3Dscanrep/}, FGR\footnote{https://github.com/isl-org/FastGlobalRegistration}, U3M\footnote{http://staffhome.ecm.uwa.edu.au/$\sim$00053650/databases.html} and ModelNet40\footnote{https://modelnet.cs.princeton.edu/}. 
ModelNet40 is a widely used dataset for deep learning registration, so it is only used in comparison with deep learning methods.
The remaining one is a scene-scale dataset, RESSO\footnote{https://3d.bk.tudelft.nl/liangliang/publications/2019/plade/resso.html}, which has much more points and greater measurement noise than the object-scale dataset.
Furthermore, the same point cloud of RESSO has varying point densities, due to different distances from the scanner.
With this dataset, the performance of the proposed method in handling large-scale outdoor scenes can be investigated. 
The detailed information of these datasets is shown in Table \ref{tab:dataset}.
The point clouds included in this test set are diverse, especially in shape, the number of points, and the overlap rate. 
It is worth noting that the overlap rate is evenly distributed from 60\% to 100\%.
Therefore, registration on this test set is a challenging task.
After this, for simplicity, P, CD and DC are the shorthand for Parasaurolophus, Chinese Dragon and Dancing Children, respectively.

\begin{figure}[!t]
  \centering
  \subfloat[]{\includegraphics[width=1.7in]{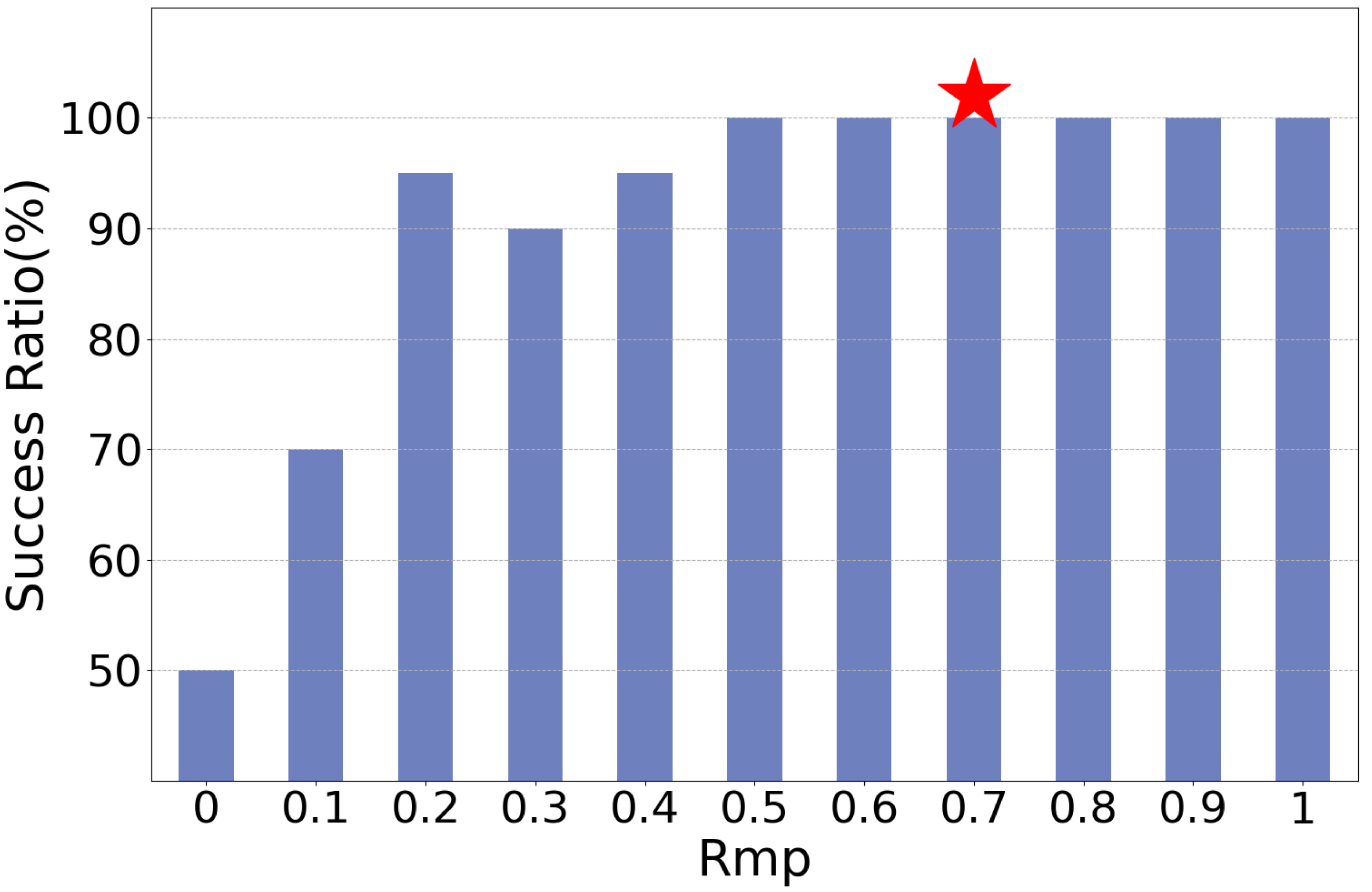}\label{fig:rmp_success}}
  \hspace{0.01\linewidth}
  \subfloat[]{\includegraphics[width=1.7in]{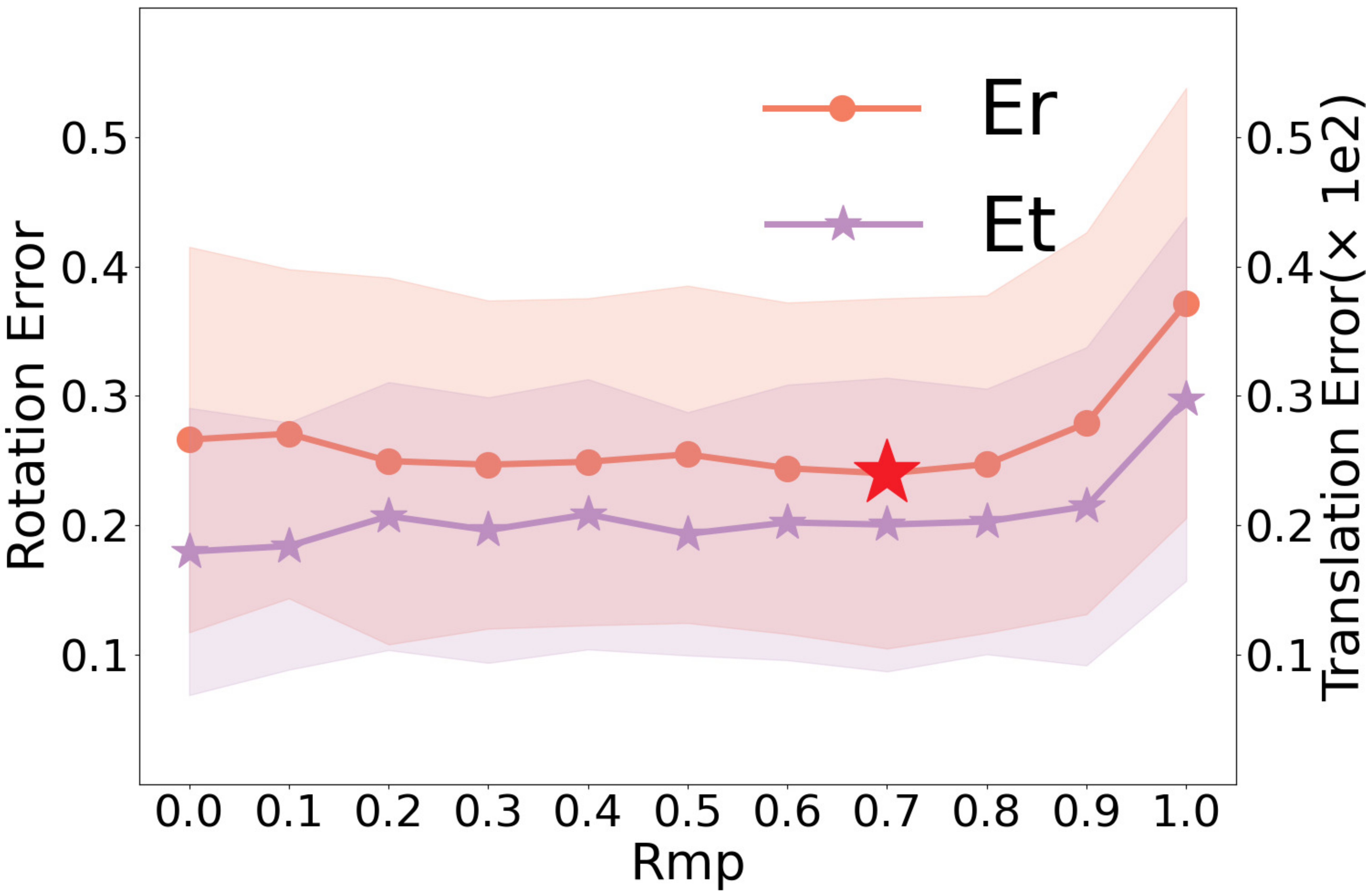}\label{rmp_error}}
  \caption{Comparison of success ratio, rotation error and translation error between different $rmp$s. \label{fig:subfig1}}
  \end{figure}

\subsection{Metrics}
\subsubsection{Error Metrics}
We use widespread used metrics \cite{li2021point, bai2021pointdsc} for quantitative assessment of error:
\begin{equation}
  \begin{cases}
  E_{\boldsymbol{R}} = \arccos ((\textit{tr}(\boldsymbol{R}_{gt}\boldsymbol{R}_{est}^T)-1)/2)  \\
  E_{\boldsymbol{t}} = \Vert \boldsymbol{t}_{gt} - \boldsymbol{t}_{est} \Vert
  \end{cases}
\end{equation}
where $E_{\boldsymbol{R}}$ and $E_{\boldsymbol{t}}$ represent rotation and translation error, subscript {\it gt} and {\it est} are used to denote the ground-truth value offered by dataset and the estimated value calculated by the algorithm, respectively.
$\textit{tr}(\cdot)$ represents the trace of a matrix. 
If $E_{\boldsymbol{R}}$ is below $5^{\circ}$, the corresponding registration is regard as a successful one \cite{li2021point}.
For evolving registration, this is also an indicator of whether the swarm is stuck in the local optima.
In addition, when calculating the average error, we only consider the successful registration, because the failed ones mostly with large errors can result in unreliable mean metrics \cite{chen2022sc2}.

\subsubsection{Novel Computational Cost Metric}
Due to the introduction of the sparse-to-dense strategy, the computational cost of one evaluation differs considerably between the S and D stages. 
Therefore, the number of calls to the nearest neighbor search (NNS) algorithm, which accounts for a large proportion of evaluation, is used to quantify computational cost rather than the number of evaluations. 
The formula of the new metric is as follows:
\begin{equation}
  \begin{aligned}
  & \textit{Computational Cost} = \\
    & \begin{cases}
    P * I(N_d + M_d)\textit{nn}_d,  & -S2D\\
    P * I(\delta N_s \textit{nn}_s + (1 - \delta)(N_d + M_d)\textit{nn}_d), & +S2D
    \end{cases}
  \end{aligned}
\end{equation}
where $P$ and $I$ represent the number of particles and iterations, respectively.
$\textit{nn}$ denotes the computational cost consumed in one call to the NNS algorithm, the subscript indicates whether the used point cloud is sparse or dense.
$\delta$ is the switch parameter in sparse-to-dense. 
$+S2D$ and $-S2D$ represent the cases with and without sparse-to-dense, respectively.
For simplicity, we assume $\textit{nn}_s$ is equivalent to $\textit{nn}_d$.
In actuality, $\textit{nn}_d$ with more candidate points costs more computation resources than $\textit{nn}_s$.
In the subsequent comparison of evolving approaches, whether single-task or multi-task, the same computational cost is guaranteed by setting the same number of $nn$.

\subsection{Preparations}
\subsubsection{Datasets except ModelNet40}
In the experiment, we simulate unaligned point clouds in practice by applying a random transformation on the already aligned source point cloud.
In order to make solutions representative, the solution space is divided into subspaces and the transformations are generated randomly in the subspaces.
By segmenting the rotation and translation space, as shown in Table \ref{tab:range}, four subspaces can be obtained, namely $SR \times ST, SR \times LT, LR \times ST$ and $LR \times LT$, where $\times$ is the Cartesian product.
In Table \ref{tab:range}, {\it diag} denotes the diagonal length of the bounding box of the source point cloud.
Four transformations will be generated randomly in each subspace, and four ones will be generated completely randomly.
Therefore, each experiment is repeated 20 times using generated representative solutions.

\begin{figure}[!t]
  \centering
  \subfloat[S stage]{\includegraphics[width=1.7in]{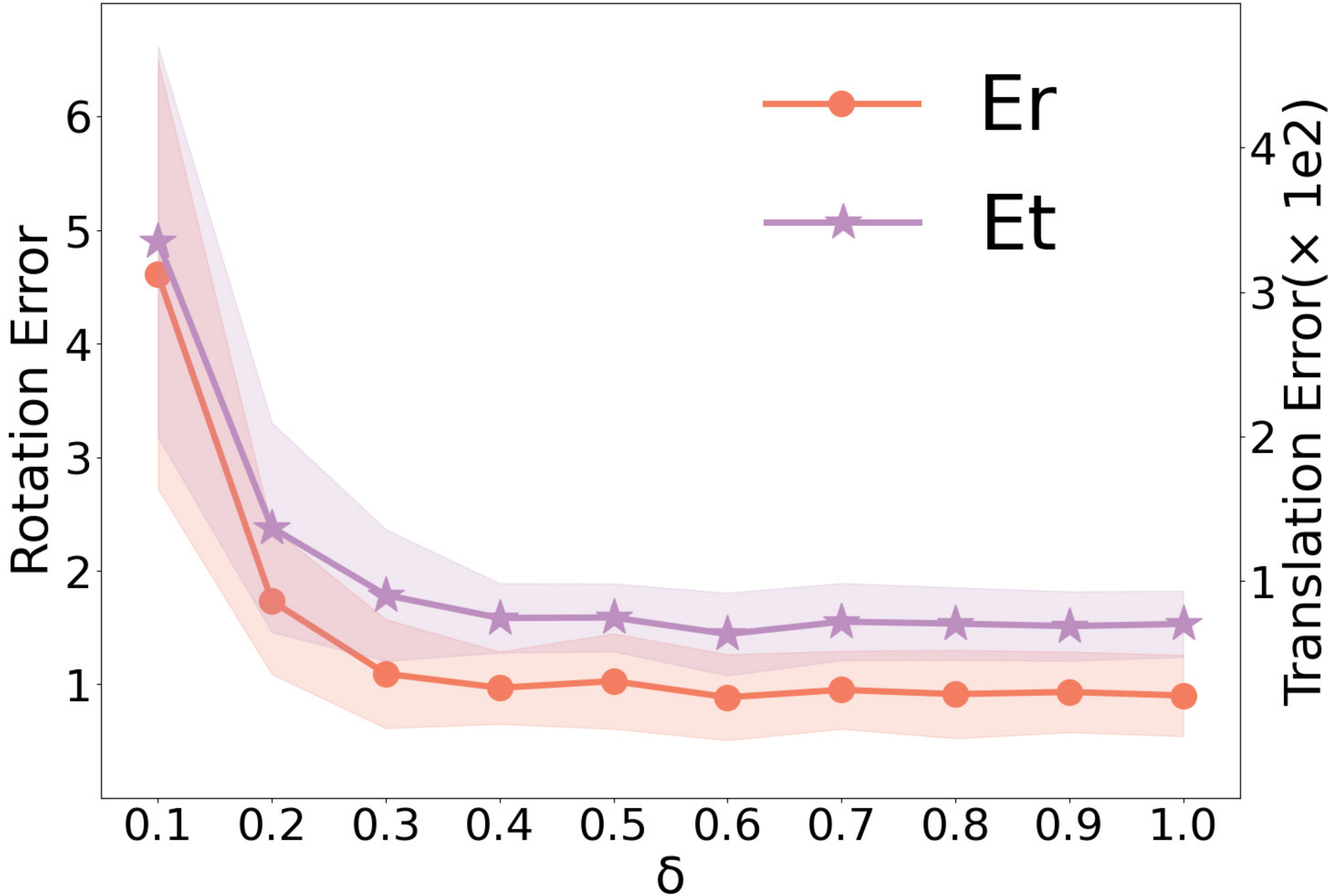}\label{TP_stage1}}
  \hspace{0.01\linewidth}
  \subfloat[D stage]{\includegraphics[width=1.7in]{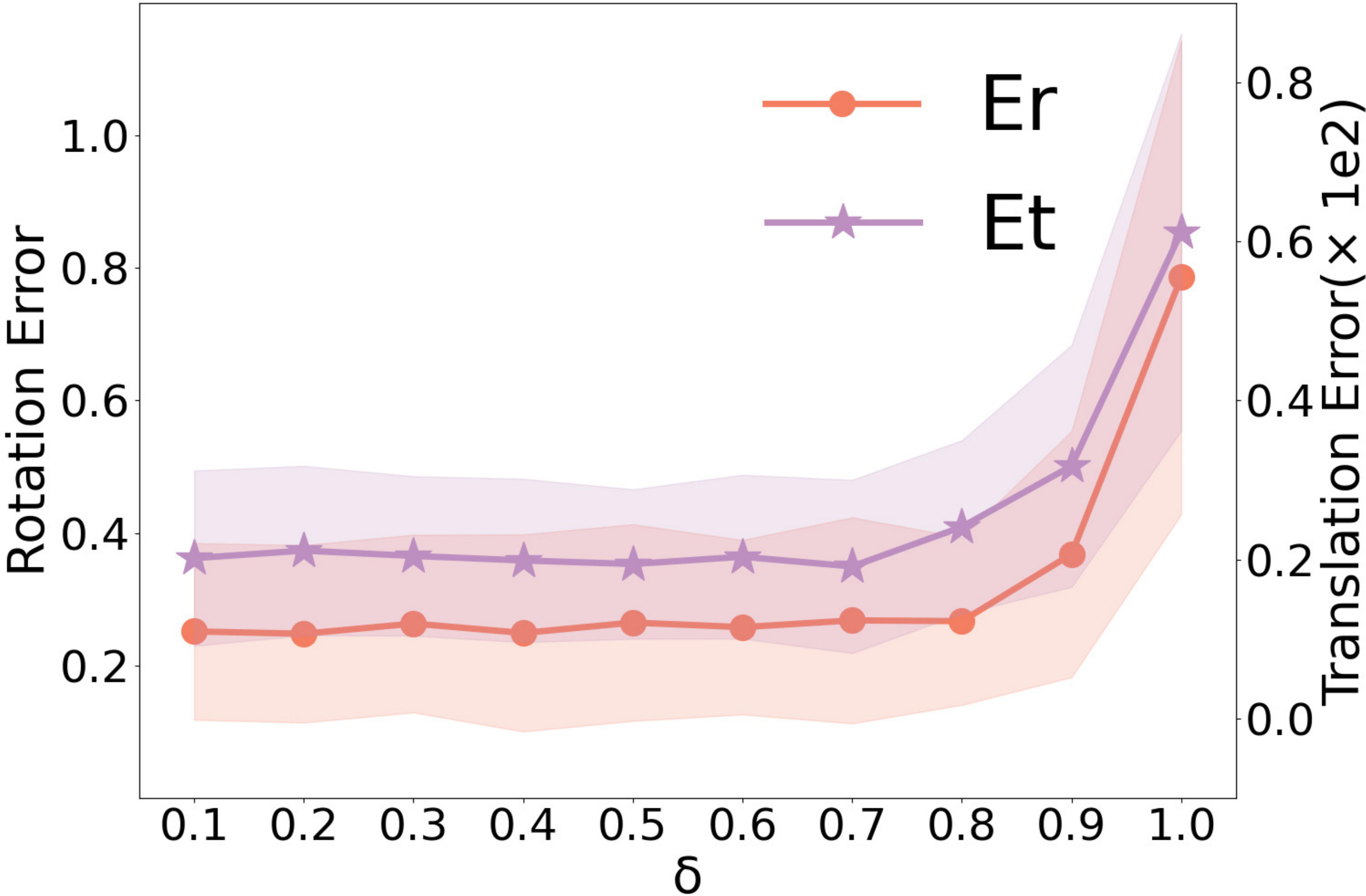}\label{TP_stage2}}
  \caption{Comparison of rotation error and translation error between different switch parameters ($\delta$) in two stages.}
  \label{fig:subfig2}
  \end{figure}
  
In addition, some pre-processing operations need to be introduced.
For example, when the source and target point clouds are far apart, especially they do not intersect at all, the search space about translation is unnecessarily enlarged a lot. 
Therefore, these two point clouds are independently centralized to the origin at first. 
Next, points in two point clouds are normalized to fit in $[-1,1]^3$ and $[-0.5,0.5]^3$ is set to translation space.
With respect to feature extraction, we choose {\it FPFH} as its enough discrimination and easy calculation.

\subsubsection{ModelNet40}
The preparations are the same as those in RPM-Net \cite{yew2020rpm}.
To summarize, in the clean dataset, each point in the source point cloud has an exact correspondence in the target one, while point clouds in the noisy dataset have Gaussian noise and no one-to-one correspondences.
Similar to other datasets, a random rigid transformation is applied on source point cloud. 
The ranges of rotation and translation in this transformation are $[0^\circ, {45}^\circ]$ and $[-0.5, 0.5]$, respectively.
The number of point clouds to be aligned in both datasets is 1266.
The proposed method is executed only once for each pair of point clouds.

\begin{figure*}[!t]
  \centering
  \includegraphics[width=6.5in]{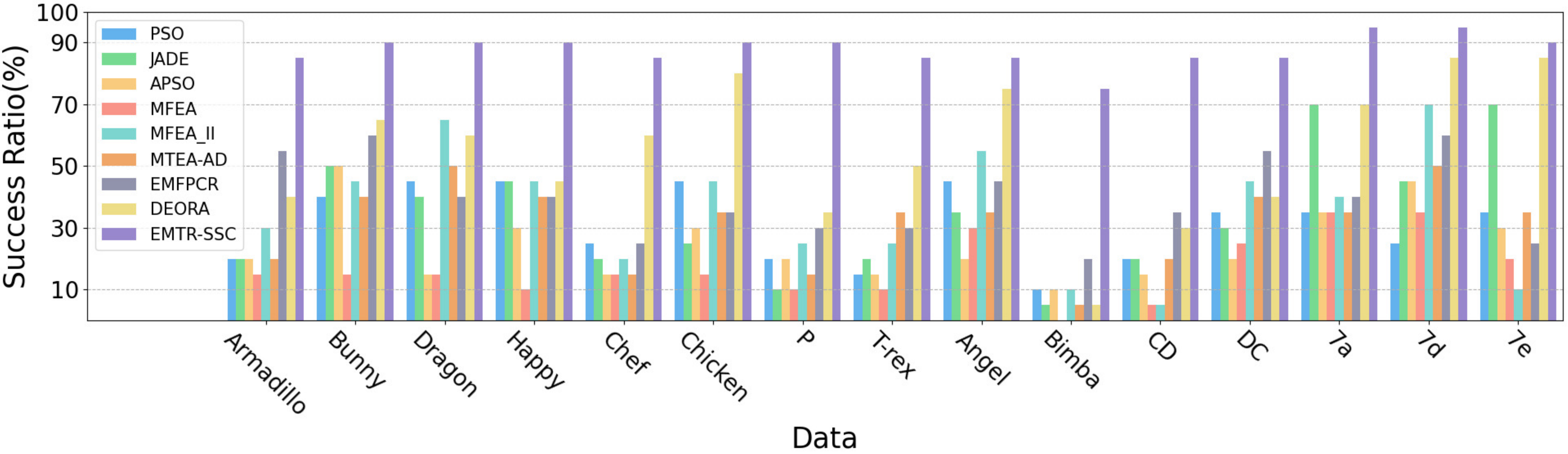}
  \caption{The success ratio obtained by compared evolutionary algorithms and proposed algorithm over the 20 independent runs on the 15 data.}
  \label{SN_bar}
  \end{figure*}

\begin{table}[!t]
  \caption{
  Rotation error (Avg$\pm$Std) and translation error (Avg$\pm$Std, $\times 10^2$) for two tasks. 
  The superior performance is highlighted in {\bf BOLD}.
  \label{tab:table_AO_cmp}}
  \centering
  \renewcommand\arraystretch{1.1}
  \resizebox{0.8\columnwidth}{!}{
  \begin{tabular}{cccccc}
    \hline
                      & \multicolumn{2}{c}{$E_{\boldsymbol{R}}$}&                & \multicolumn{2}{c}{$E_{\boldsymbol{t}}$}   \\ \cline{2-3} \cline{5-6}
                      & \textbf{$\beta$}  & \textbf{$\alpha$}           & \textbf{}        & \textbf{$\beta$}  & \textbf{$\alpha$}          \\
  \hline
  \textit{(i)object-scale} \\
  \textbf{Armadillo} & 1.328$\pm$0.751     & \textbf{0.161$\pm$0.073} & \textbf{}        & 0.276$\pm$0.181 & \textbf{0.084$\pm$0.034} \\
  \textbf{Bunny}     & 1.284$\pm$0.847     & \textbf{0.233$\pm$0.120} & \textbf{}        & 0.745$\pm$0.482 & \textbf{0.175$\pm$0.085} \\
  \textbf{Dragon}    & 1.093$\pm$0.596     & \textbf{0.190$\pm$0.074} & \textbf{}        & 0.477$\pm$0.249 & \textbf{0.168$\pm$0.074} \\
  \textbf{Happy}     & 2.430$\pm$1.182     & \textbf{0.369$\pm$0.195} & \textbf{}        & 0.725$\pm$0.530 & \textbf{0.220$\pm$0.087} \\
  \textbf{Chef}      & 2.264$\pm$1.112     & \textbf{0.379$\pm$0.208} & \textbf{}        & 0.872$\pm$0.425 & \textbf{0.197$\pm$0.137} \\
  \textbf{Chicken}   & 2.508$\pm$1.140     & \textbf{0.271$\pm$0.178} & \textbf{}        & 1.202$\pm$0.796 & \textbf{0.331$\pm$0.201} \\
  \textbf{P}         & 2.334$\pm$1.050     & \textbf{0.335$\pm$0.186} & \textbf{}        & 3.265$\pm$1.888 & \textbf{0.299$\pm$0.148} \\
  \textbf{T-rex}     & 2.108$\pm$1.142     & \textbf{0.296$\pm$0.165} & \textbf{}        & 1.920$\pm$0.987 & \textbf{0.366$\pm$0.176} \\
  \textbf{Angel}     & 2.531$\pm$0.705     & \textbf{0.178$\pm$0.076} & \textbf{}        & 1.461$\pm$0.626 & \textbf{0.277$\pm$0.182} \\
  \textbf{Bimba}     & 2.493$\pm$1.115     & \textbf{0.417$\pm$0.237} & \textbf{}        & 1.238$\pm$0.587 & \textbf{0.419$\pm$0.146} \\
  \textbf{CD}        & 2.116$\pm$0.914     & \textbf{0.258$\pm$0.123} & \textbf{}        & 1.352$\pm$0.447 & \textbf{0.251$\pm$0.149} \\
  \textbf{DC}        & 2.002$\pm$0.945     & \textbf{0.199$\pm$0.108} & \textbf{}        & 0.646$\pm$0.332 & \textbf{0.265$\pm$0.267} \\
  \hline
  \textit{(ii)scene-scale} \\                                                                             
  \textbf{7a}        & 2.053$\pm$0.789     & \textbf{0.147$\pm$0.063} & \textbf{}        & 0.768$\pm$0.379 & \textbf{0.051$\pm$0.029} \\
  \textbf{7d}        & 1.231$\pm$0.795     & \textbf{0.192$\pm$0.075} & \textbf{}        & 0.201$\pm$0.095 & \textbf{0.035$\pm$0.020} \\
  \textbf{7e}        & 1.321$\pm$0.657     & \textbf{0.549$\pm$0.220} & \textbf{}        & 0.193$\pm$0.132 & \textbf{0.135$\pm$0.064} \\
  \hline
  \end{tabular}}
  \end{table}

\subsection{Parameter Settings}
For the proposed method, the subpopulation size for each task is set to 50 and the total iteration number is specified as 100.
In velocity update, the acceleration constant $c_1 = c_2 = 1.49445$, inertia weight $\omega = 0.9 - 0.5 * it / MaxIt$, where $it$ and $MaxIt$ denote the number of the current and maximum iterations, respectively. 
$\lambda^\alpha = 0.1 + 0.6 * it / MaxIt, \, \lambda^\beta = 0.999 - 0.8 * it / MaxIt$ in arithmetic crossover of knowledge transfer.
The factor $\eta$, which is used to adjust the order of magnitude in fitness $\beta$, is set to 50$\tau$.

Furthermore, two important parameters need to be determined through experiments.
One is the random mating probability ($\textit{rmp}$) in the knowledge transfer mechanism, and the other one is the switch parameter ($\delta$) in the sparse-to-dense strategy.
Sensitivity analysis for two parameters is conducted based on \emph{Bunny} data.
Fig. \ref{fig:subfig1} shows the variation of success ratio, rotation error and translation error at different $rmp$s.
In Fig. \ref{fig:rmp_success}, when $rmp$ is below 0.5, due to the insufficient knowledge transfer, the success ratio is fluctuant. 
As shown in Fig. \ref{rmp_error}, the errors are small when $rmp$ is below 0.5, because the mean calculation does not consider failed registration.
Due to the poor reachable precision of Task $\beta$, excessively frequent transfer (i.e. $rmp \geq 0.9$) leads to increased error.
In conclusion, the $rmp$ is set to 0.7, at which time the success ratio is stable, and the rotation error is minimized, which are marked by red five-pointed stars in subfigures.
Fig. \ref{fig:subfig2} shows the variation of rotation error and translation error at different switch parameters in two stages.
When $\delta$ is above 0.5 in the S stage or below 0.7 in the D stage, rotation and translation errors remain stable.
Hence, $\delta$ takes the average of two numbers, i.e. 0.6, in order to reduce computational cost as much as possible while ensuring accurate results. 

\begin{table}[!t]
  \caption{The average success ratio for each method.\label{tab:aver_SR}}
  \renewcommand\arraystretch{1.5}
  \resizebox{1\columnwidth}{!}{
  \begin{tabular}{c|ccccccccc}
  \hline
        & \textbf{PSO} & \textbf{JADE}   & \textbf{APSO} & \textbf{MFEA} & \textbf{MFEA-II} & \textbf{MTEA-AD} & \textbf{EMFPCR} & \textbf{DEORA} & \textbf{EMTR-SSC} \\
  \hline
  Avg SR  & 30.7\%       & 33.3\%          & 24.7\%        & 17.0\%        & 35.7\%         & 31.3\%           & 39.7\%           & 55.0\%         & \textbf{87.7\%}
  \\
  \hline
  \end{tabular}}
  \end{table}

\section{Results and Discussion}
\label{sec:RD}
In this section, several experiments are performed to illustrate the effectiveness and efficiency of EMTR-SSC. 
The purpose of the first comparison is to verify whether the result of Task $\alpha$ is better than that of Task $\beta$. 
The second comparison aims to validate whether the proposed method outperforms other evolving approaches in terms of success ratio. 
The goal of the third and fourth comparisons is to test whether this method surpasses compared traditional and deep learning approaches in terms of registration precision. 
Finally, we conduct the fifth comparison to prove the effectiveness of two designed strategies.

\subsection{Comparisons of Task $\alpha$ and Task $\beta$}
The rotation error and translation error are important criteria to measure the registration precision. 
During these experiments, the registration with low errors suggests that the corresponding approach is effective at solving PCR problems.
 
Table \ref{tab:table_AO_cmp} compares Task $\alpha$ with Task $\beta$ in terms of the average (Avg) and standard deviation (Std) of rotation error and those of translation error over the 20 independent runs on the 4 datasets. 
The column $E_{\boldsymbol{R}}$ and $E_{\boldsymbol{t}}$ provide rotation error and translation error, respectively.
For the sake of easy comparison, the Avg and Std of all translation errors are magnified 100 times.
Notably, the translation error of Task $\beta$ is calculated using the translation obtained by the knowledge complement strategy.
As can be seen from Table \ref{tab:table_AO_cmp}, the mean errors of Task $\alpha$ are decreased by an order of magnitude compared with Task $\beta$ in most cases.
Furthermore, the Std of errors is smaller in Task $\alpha$, indicating its result is more stable.
Therefore, these experiment results justify that the result of Task $\alpha$ is taken as the output of the whole approach.

\begin{table}[!t]
  \caption{
  Rotation error (Avg$\pm$Std) and translation error (Avg$\pm$Std, $\times 10^2$) for each method on the 15 data. 
  The superior performance is highlighted in {\bf BOLD}. \label{tab:table_accu_trad}
  }
  \centering
  \renewcommand\arraystretch{1} 
  \resizebox{1\columnwidth}{!}{
  \begin{tabular}{ccccccc}
    \hline
                                      &                      & \textbf{ICP$^+$} & \textbf{NDT$^+$} & \textbf{TrICP$^+$}       & \textbf{FGR}          & \textbf{EMTR-SSC}        \\
  \hline
  \textit{(i)object-scale} \\
  \hdashline
  \textit{a.high overlap} \\
  \multirow{2}{*}{\textbf{Armadillo}} & $E_{\boldsymbol{R}}$ & -                & 4.392$\pm$0.484  & 1.425$\pm$0.190          & 0.391$\pm$0.093 & \textbf{0.161$\pm$0.073} \\
                                      & $E_{\boldsymbol{t}}$ & -                & 1.434$\pm$0.429  & 0.622$\pm$0.070          & 0.256$\pm$0.061 & \textbf{0.084$\pm$0.034} \\ \\
  \multirow{2}{*}{\textbf{Bunny}}     & $E_{\boldsymbol{R}}$ & 4.381$\pm$0.129  & 4.014$\pm$0.595  & 0.597$\pm$0.168          & 0.374$\pm$0.156 & \textbf{0.233$\pm$0.120} \\
                                      & $E_{\boldsymbol{t}}$ & 2.913$\pm$0.315  & 4.161$\pm$0.519  & 0.449$\pm$0.174          & 0.294$\pm$0.080 & \textbf{0.175$\pm$0.085} \\ \\
  \multirow{2}{*}{\textbf{Dragon}}    & $E_{\boldsymbol{R}}$ & 0.208$\pm$0.053  & 1.516$\pm$0.837  & \textbf{0.162$\pm$0.064} & 0.385$\pm$0.135       & 0.190$\pm$0.074    \\
                                      & $E_{\boldsymbol{t}}$ & 0.638$\pm$0.036  & 3.150$\pm$1.460  & 0.284$\pm$0.089    & 0.657$\pm$0.115       & \textbf{0.168$\pm$0.074} \\ \\
  \multirow{2}{*}{\textbf{Happy}}     & $E_{\boldsymbol{R}}$ & 3.638$\pm$0.254  & -                & 1.352$\pm$0.071    & 3.171$\pm$1.093       & \textbf{0.369$\pm$0.195} \\
                                      & $E_{\boldsymbol{t}}$ & 1.027$\pm$0.093  & -                & 0.236$\pm$0.021    & 0.596$\pm$0.214       & \textbf{0.220$\pm$0.087} \\ \\
  \multirow{2}{*}{\textbf{Chef}}      & $E_{\boldsymbol{R}}$ & -                & -                & 2.415$\pm$0.263          & 1.161$\pm$0.349 & \textbf{0.379$\pm$0.208} \\
                                      & $E_{\boldsymbol{t}}$ & -                & -                & 1.342$\pm$0.147          & 0.410$\pm$0.120 & \textbf{0.197$\pm$0.137} \\ \\
  \multirow{2}{*}{\textbf{Chicken}}   & $E_{\boldsymbol{R}}$ & 2.956$\pm$0.095  & 3.478$\pm$1.317  & 1.214$\pm$0.199          & 0.470$\pm$0.119 & \textbf{0.262$\pm$0.181} \\
                                      & $E_{\boldsymbol{t}}$ & 3.252$\pm$0.103  & 4.642$\pm$1.006  & 1.428$\pm$0.269          & 0.277$\pm$0.082 & \textbf{0.266$\pm$0.120} \\ \\
  \multirow{2}{*}{\textbf{P}}         & $E_{\boldsymbol{R}}$ & -                & 3.000$\pm$1.127  & 1.262$\pm$0.202    & 1.732$\pm$0.769       & \textbf{0.335$\pm$0.186} \\
                                      & $E_{\boldsymbol{t}}$ & -                & 3.349$\pm$1.195  & 1.896$\pm$0.247          & 0.591$\pm$0.427 & \textbf{0.299$\pm$0.148} \\
  \hdashline
  \textit{b.middle overlap} \\
  \multirow{2}{*}{\textbf{T-rex}}     & $E_{\boldsymbol{R}}$ & -                & 3.155$\pm$1.100  & 1.269$\pm$0.101          & 1.108$\pm$0.445 & \textbf{0.296$\pm$0.165} \\
                                      & $E_{\boldsymbol{t}}$ & -                & 4.198$\pm$2.125  & 2.811$\pm$0.169          & 1.167$\pm$0.255 & \textbf{0.366$\pm$0.176} \\ \\
  \multirow{2}{*}{\textbf{Angel}}     & $E_{\boldsymbol{R}}$ & 4.354$\pm$0.128  & 4.164$\pm$0.574  & 3.027$\pm$0.208          & 1.628$\pm$0.789 & \textbf{0.178$\pm$0.076} \\
                                      & $E_{\boldsymbol{t}}$ & 4.469$\pm$0.101  & 2.648$\pm$0.815  & 2.738$\pm$0.315          & 1.344$\pm$0.289 & \textbf{0.277$\pm$0.182} \\ \\
  \multirow{2}{*}{\textbf{Bimba}}     & $E_{\boldsymbol{R}}$ & -                & -                & -                        & 0.853$\pm$0.394 & \textbf{0.417$\pm$0.237} \\
                                      & $E_{\boldsymbol{t}}$ & -                & -                & -                        & 0.533$\pm$0.240 & \textbf{0.419$\pm$0.146} \\ \\
  \multirow{2}{*}{\textbf{CD}}        & $E_{\boldsymbol{R}}$ & -                & 3.820$\pm$0.735  & 3.038$\pm$0.257          & 0.734$\pm$0.253 & \textbf{0.258$\pm$0.123} \\
                                      & $E_{\boldsymbol{t}}$ & -                & 4.978$\pm$1.238  & 2.609$\pm$0.216          & 0.896$\pm$0.256 & \textbf{0.251$\pm$0.149} \\ \\
  \multirow{2}{*}{\textbf{DC}}        & $E_{\boldsymbol{R}}$ & 1.347$\pm$0.054  & 3.365$\pm$0.418  & 0.596$\pm$0.068          & 0.589$\pm$0.321 & \textbf{0.199$\pm$0.108} \\
                                      & $E_{\boldsymbol{t}}$ & 1.905$\pm$0.048  & 4.670$\pm$1.630  & 1.140$\pm$0.097          & 0.637$\pm$0.280 & \textbf{0.265$\pm$0.267} \\
  \hline
  \textit{(ii)scene-scale} \\
  \multirow{2}{*}{\textbf{7a}}        & $E_{\boldsymbol{R}}$ & 3.139$\pm$0.121  & 2.101$\pm$0.698  & 0.167$\pm$0.047    & 1.310$\pm$0.396       & \textbf{0.147$\pm$0.063} \\ 
                                      & $E_{\boldsymbol{t}}$ & 1.929$\pm$0.093  & 3.920$\pm$0.786  & 0.098$\pm$0.036    & 0.305$\pm$0.135       & \textbf{0.051$\pm$0.029} \\ \\
  \multirow{2}{*}{\textbf{7d}}        & $E_{\boldsymbol{R}}$ & 1.150$\pm$0.068  & 3.765$\pm$0.838  & 0.235$\pm$0.056    & 0.948$\pm$0.435       & \textbf{0.192$\pm$0.075} \\
                                      & $E_{\boldsymbol{t}}$ & 1.401$\pm$0.083  & 2.003$\pm$0.824  & 0.066$\pm$0.030    & 0.180$\pm$0.055       & \textbf{0.035$\pm$0.020} \\ \\
  \multirow{2}{*}{\textbf{7e}}        & $E_{\boldsymbol{R}}$ & 2.097$\pm$0.232  & 4.213$\pm$0.199  & 0.857$\pm$0.064    & 0.927$\pm$0.386       & \textbf{0.549$\pm$0.220} \\
                                      & $E_{\boldsymbol{t}}$ & 8.347$\pm$1.191  & 1.475$\pm$0.286  & 0.543$\pm$0.115          & 0.250$\pm$0.088 & \textbf{0.135$\pm$0.064} \\
  \hline
  \end{tabular}}
  \end{table}

\subsection{Comparisons with Evolving Approaches}
The success ratio is an essential indicator to quantify the ability to tackle local optima.
It is calculated by $\frac{\#Successful}{\#All}$, where $\#Successful$ and $\#All$ are the number of successful registrations and all independent runs, respectively.
Fig.\ref{SN_bar} shows the success ratio of EMTR-SSC and other seven compared evolving approaches based on the 20 independent runs in the 15 data. 
As can be seen from Fig.\ref{SN_bar},  the proposed EMTR-SSC achieves 90\% and above success ratio in 8 out of the 15 data and 75\% and above in all data.
However, for the compared approaches, except for JADE, MFEA-II, EMFPCR and DEORA, the 50\% success ratio is hard to reach.
This reflects evolving approaches generally cannot handle local optima well.
In this figure, EMTR-SSC achieves the highest improvement in \emph{Bimba} data compared with the best of the other approaches, i.e., EMFPCR, where the success ratio is enhanced by 60\%. 
These comparisons confirm the effectiveness of the proposed method in reducing falling into local optima. 

\begin{figure}[!t]
  \centering
  \scalebox{0.5}{
  \begin{tabular}{cccccccc}
  Input & ICP$^+$ & NDT$^+$ & TrICP$^+$ & FGR & EMTR-SSC & Ground Truth &
  \\
  \begin{minipage}[b]{0.168\columnwidth}
    \centering
    \raisebox{-.5\height}{\includegraphics[width=\linewidth]{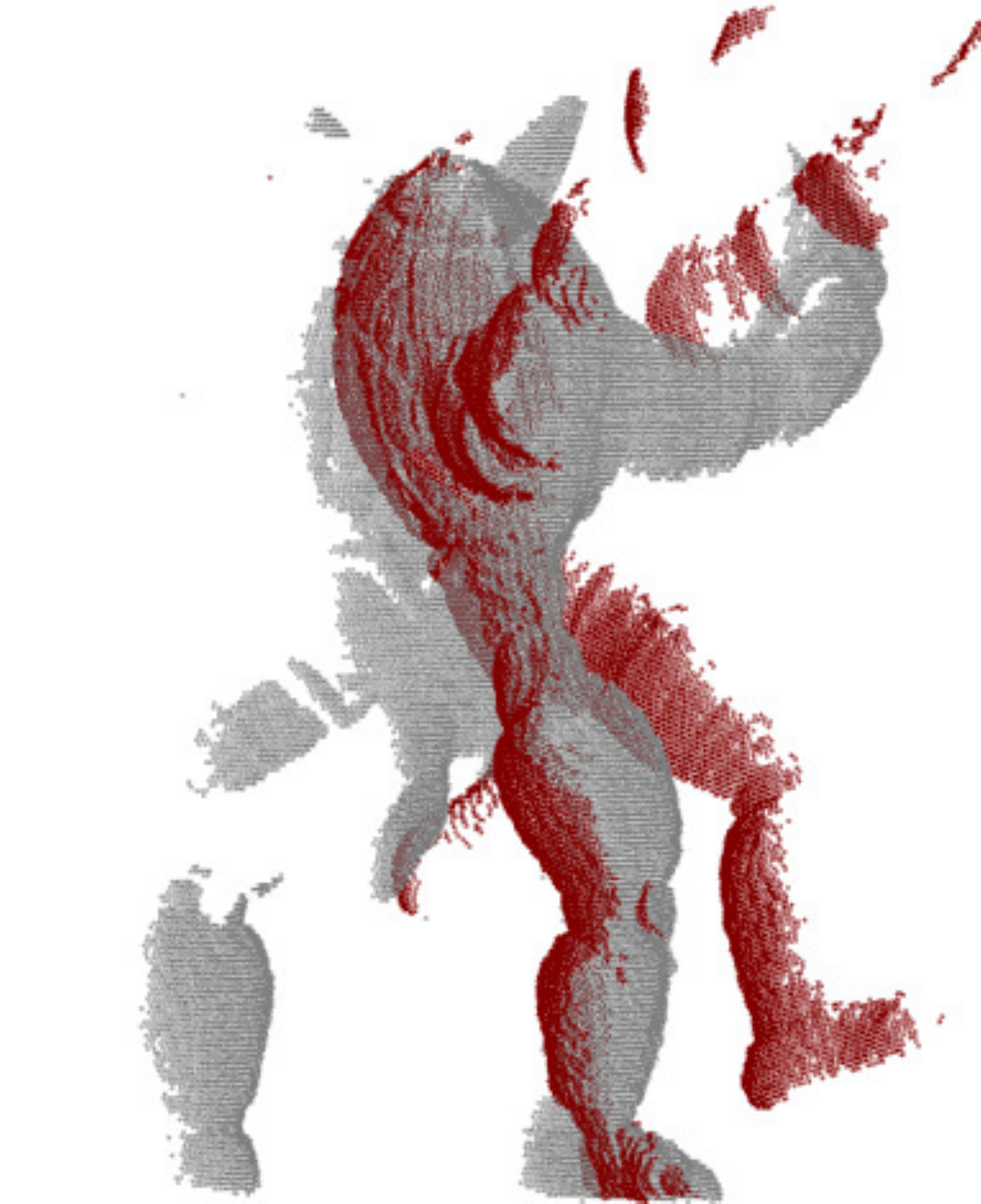}}
  \end{minipage} 
  & \begin{minipage}[b]{0.168\columnwidth}
    \centering
    \raisebox{-.5\height}{\includegraphics[width=\linewidth]{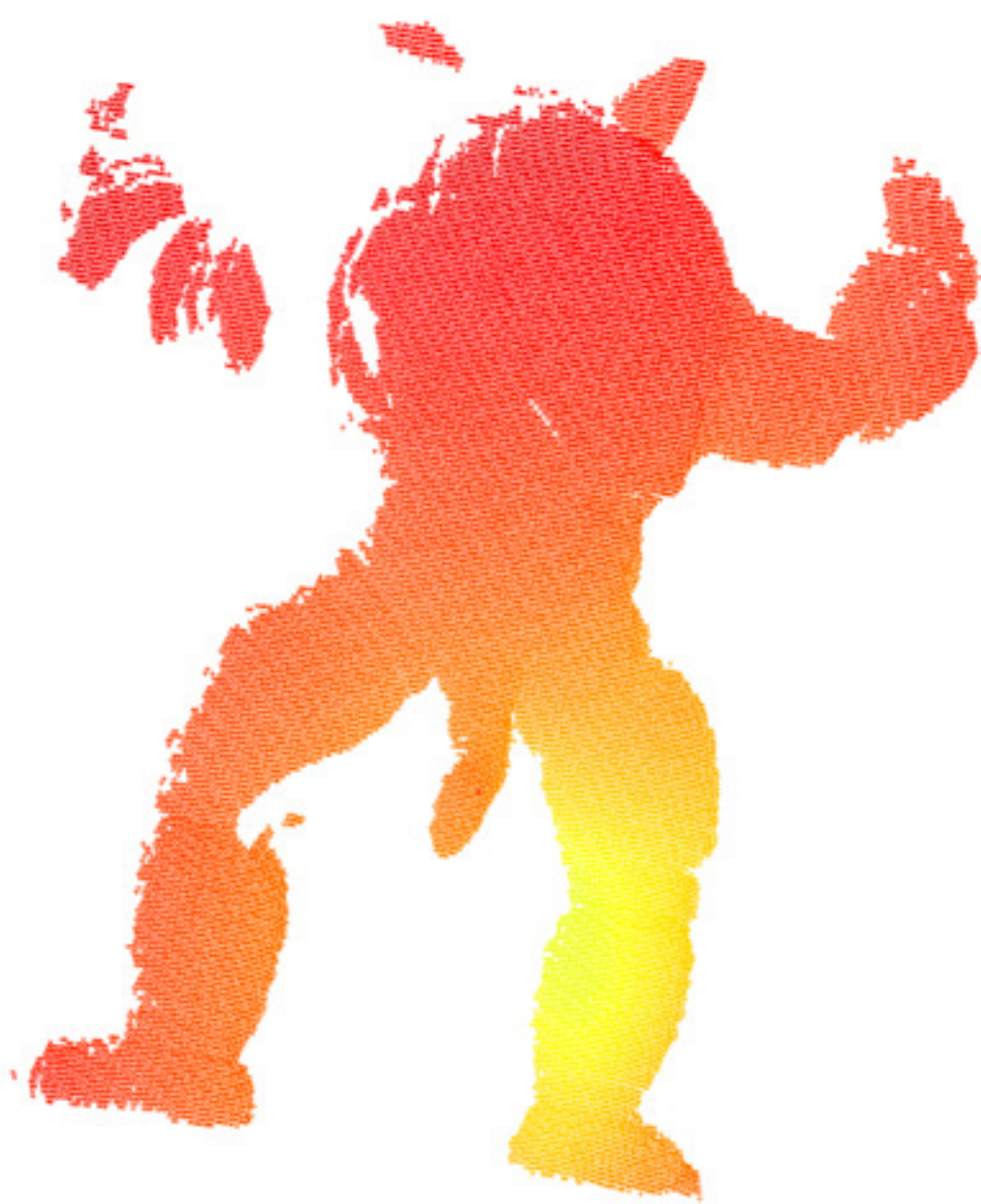}}
  \end{minipage}
  & \begin{minipage}[b]{0.168\columnwidth}
    \centering
    \raisebox{-.5\height}{\includegraphics[width=\linewidth]{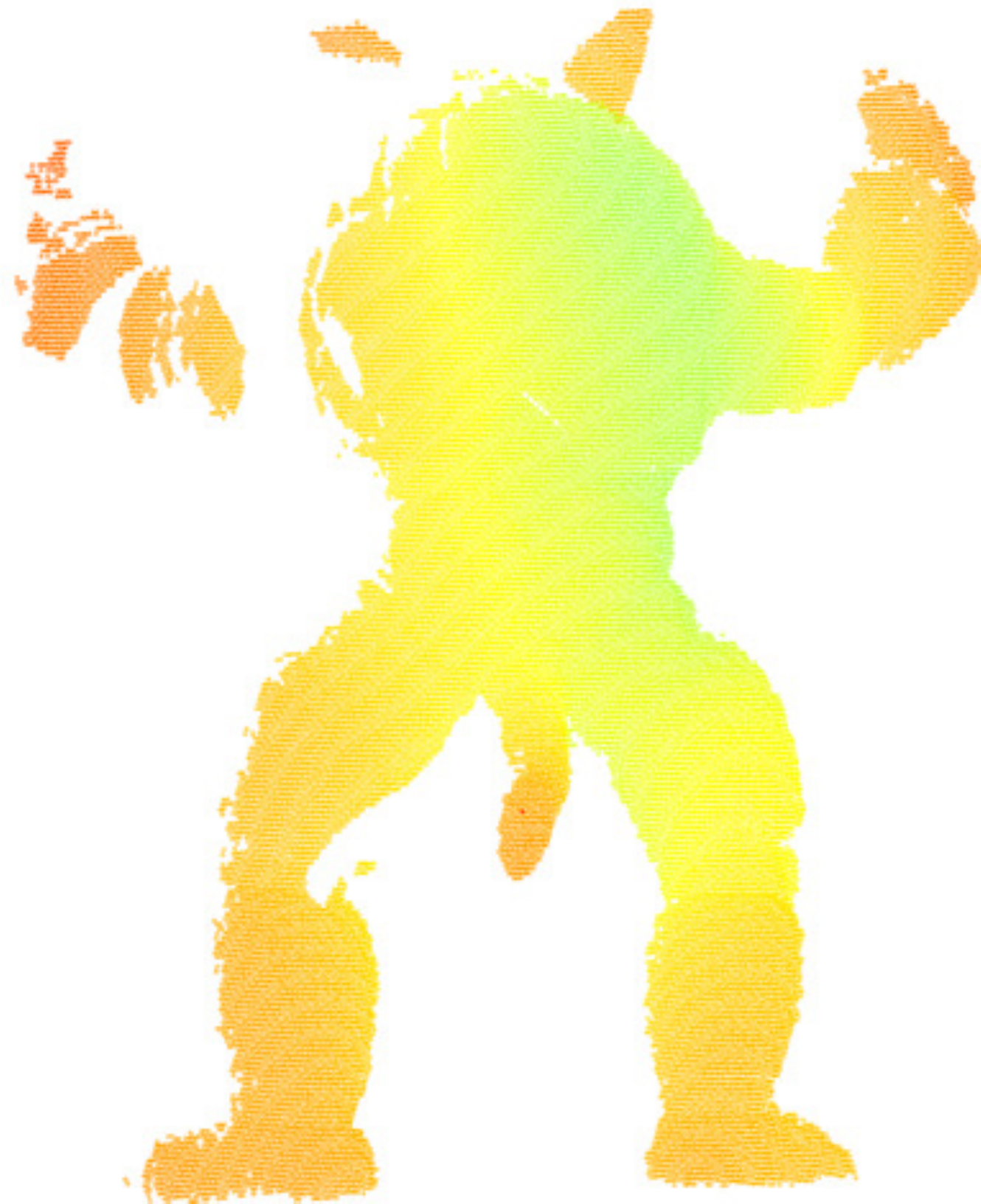}}
  \end{minipage} 
  & \begin{minipage}[b]{0.168\columnwidth}
    \centering
    \raisebox{-.5\height}{\includegraphics[width=\linewidth]{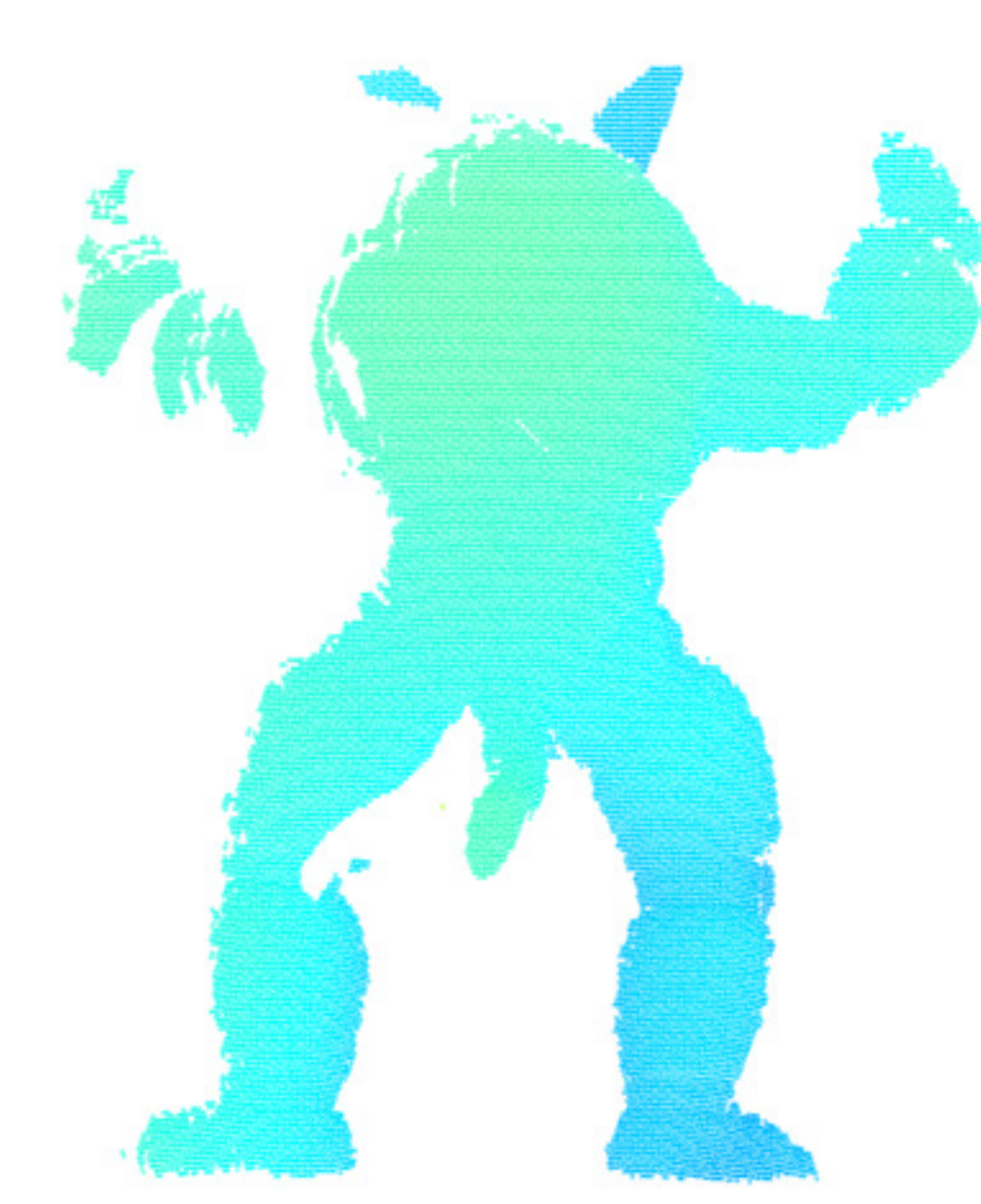}}
  \end{minipage} 
  & \begin{minipage}[b]{0.168\columnwidth}
    \centering
    \raisebox{-.5\height}{\includegraphics[width=\linewidth]{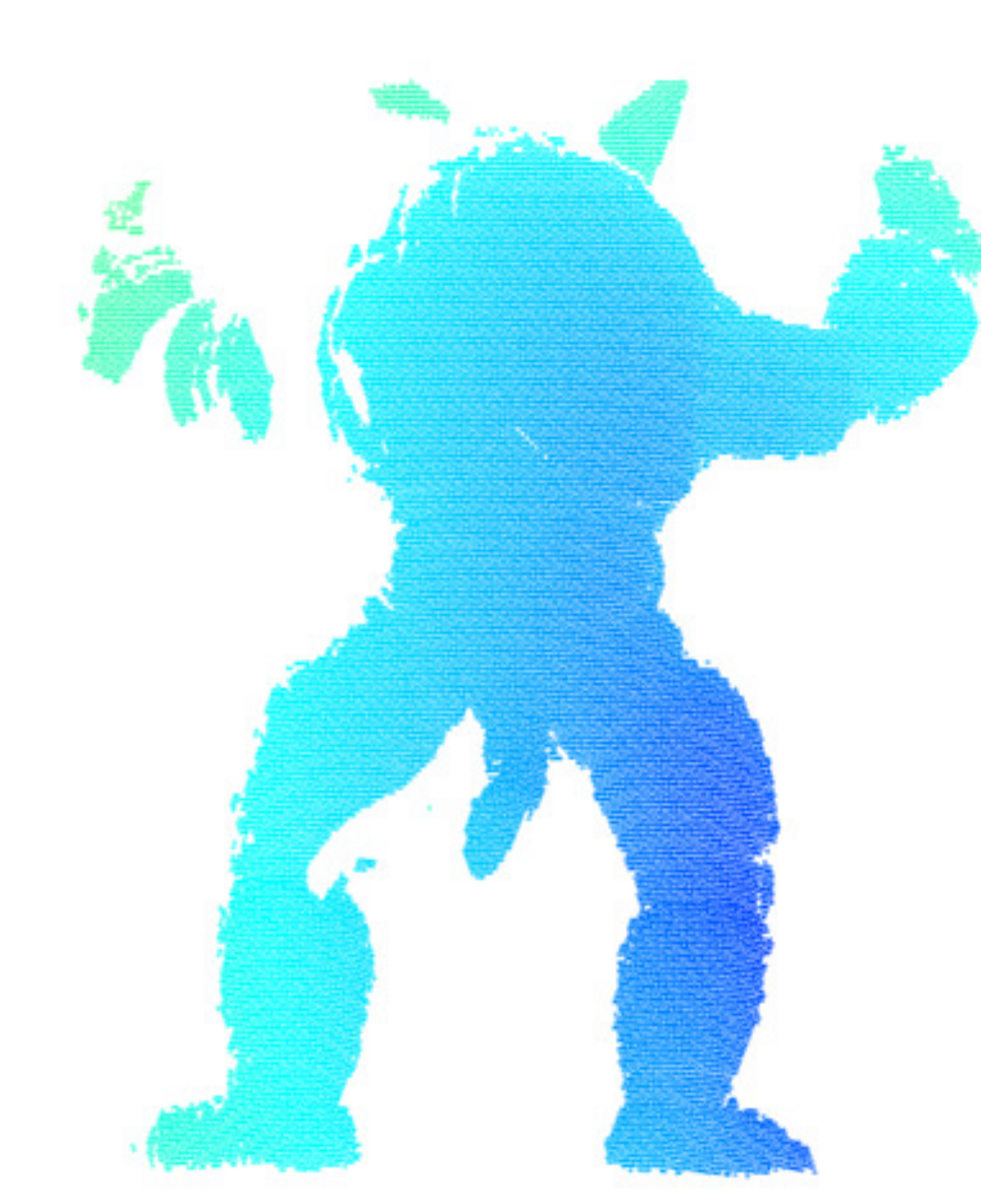}}
  \end{minipage} 
  & \begin{minipage}[b]{0.168\columnwidth}
    \centering
    \raisebox{-.5\height}{\includegraphics[width=\linewidth]{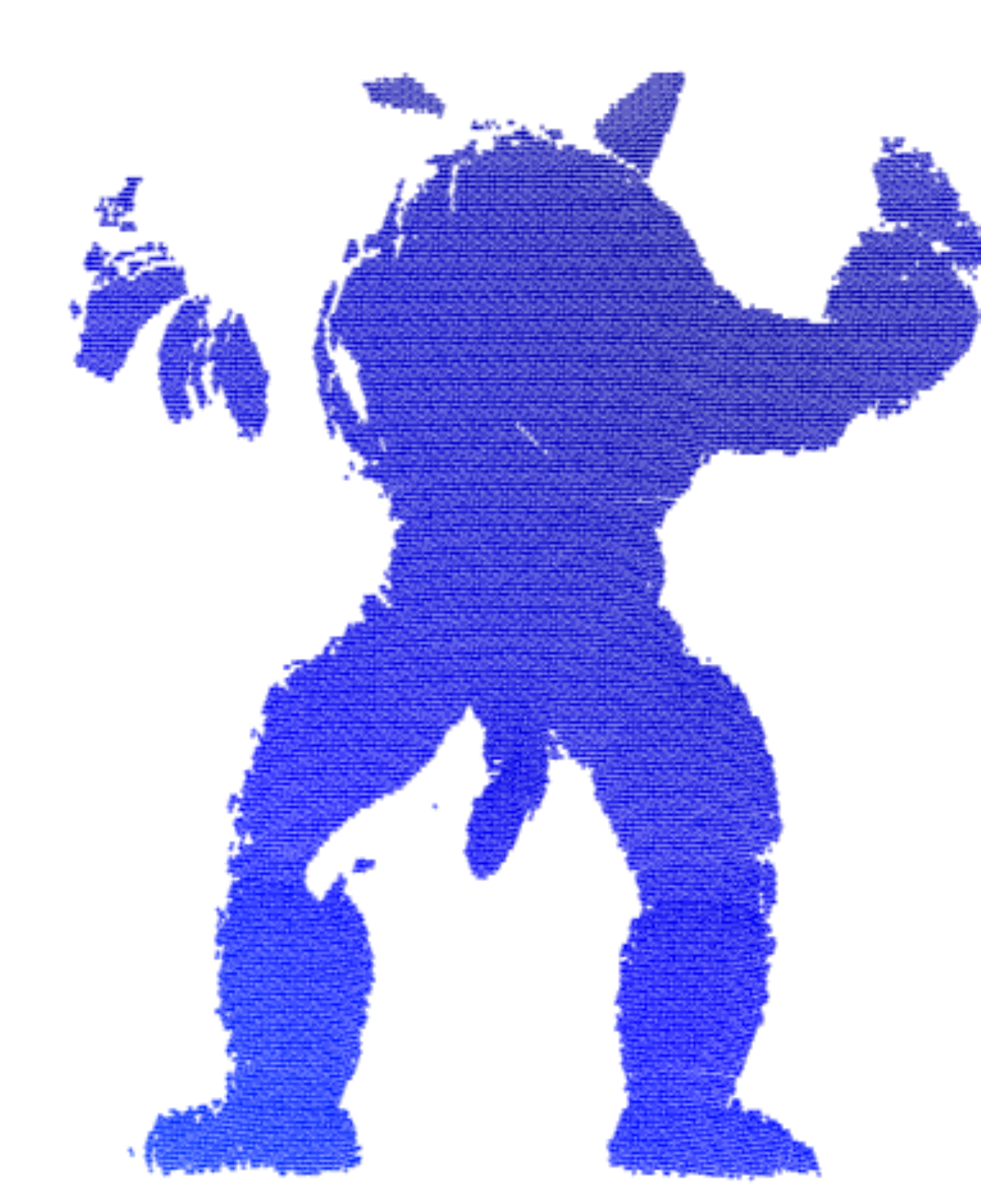}}
  \end{minipage} 
  & \begin{minipage}[b]{0.168\columnwidth}
    \centering
    \raisebox{-.5\height}{\includegraphics[width=\linewidth]{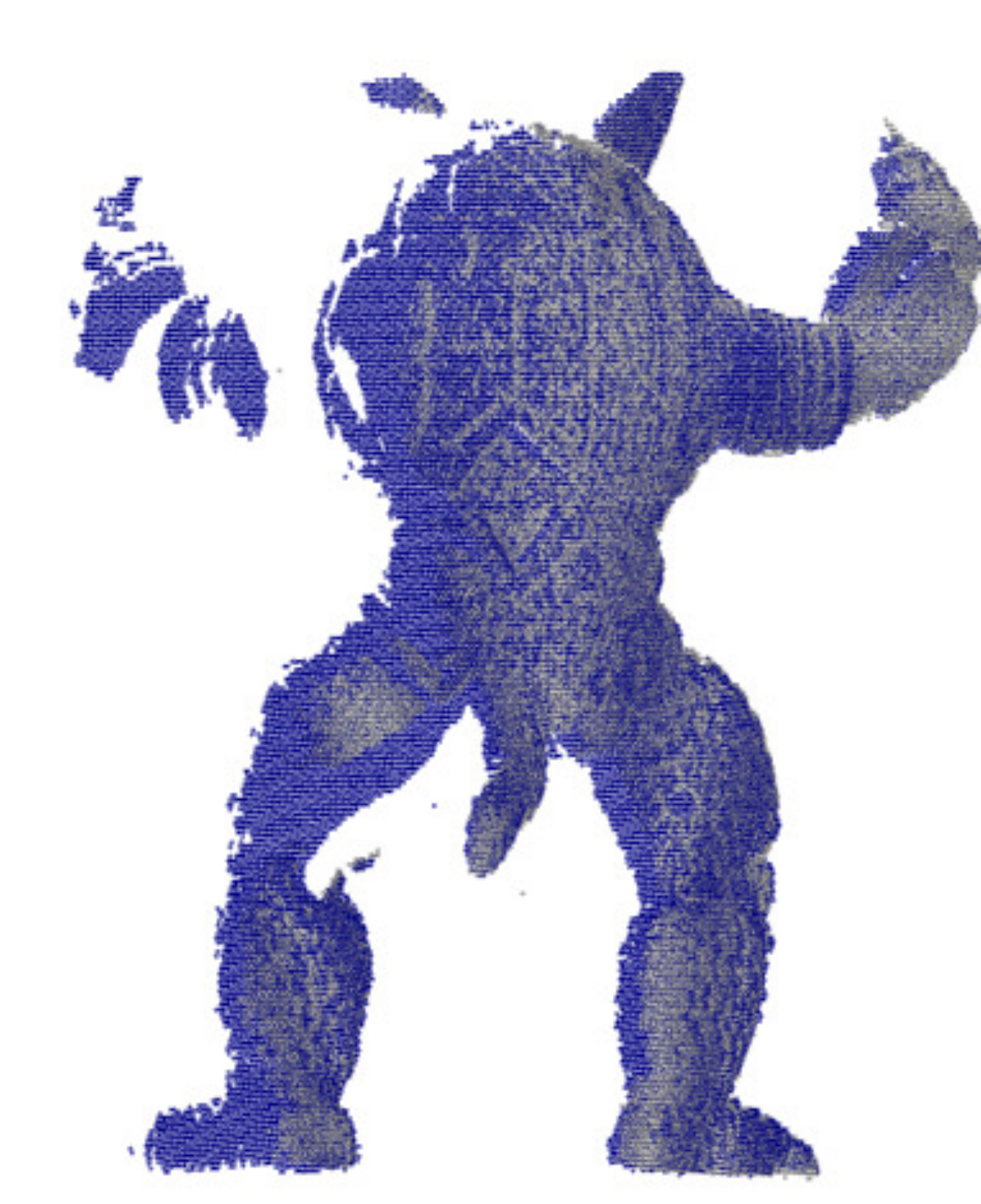}}
  \end{minipage} 
  & \begin{minipage}[b]{0.08\columnwidth}
    \centering
    \raisebox{-.5\height}{\includegraphics[width=\linewidth]{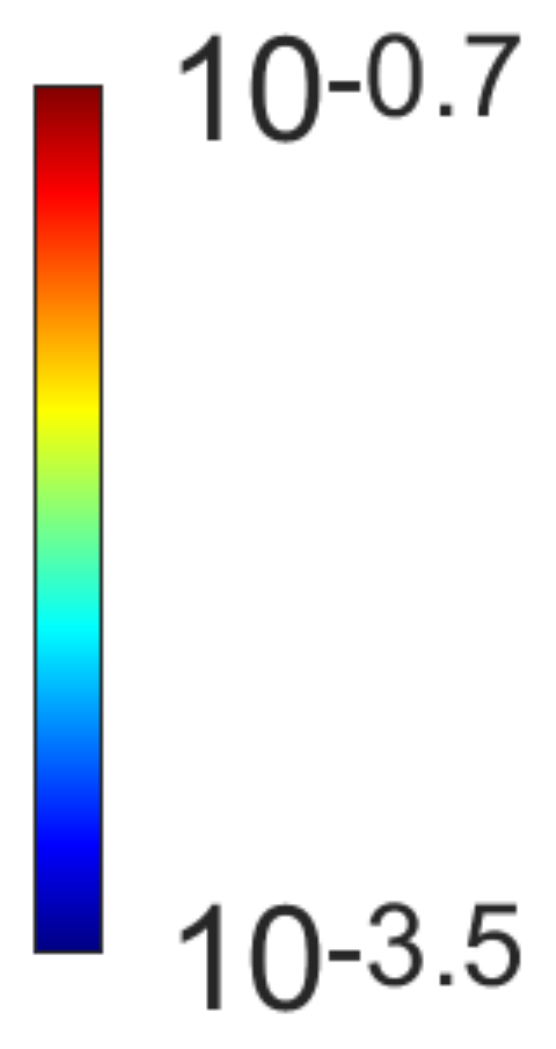}}
  \end{minipage} 
  \\

  \begin{minipage}[b]{0.16\columnwidth}
    \centering
    \raisebox{-.5\height}{\includegraphics[width=\linewidth]{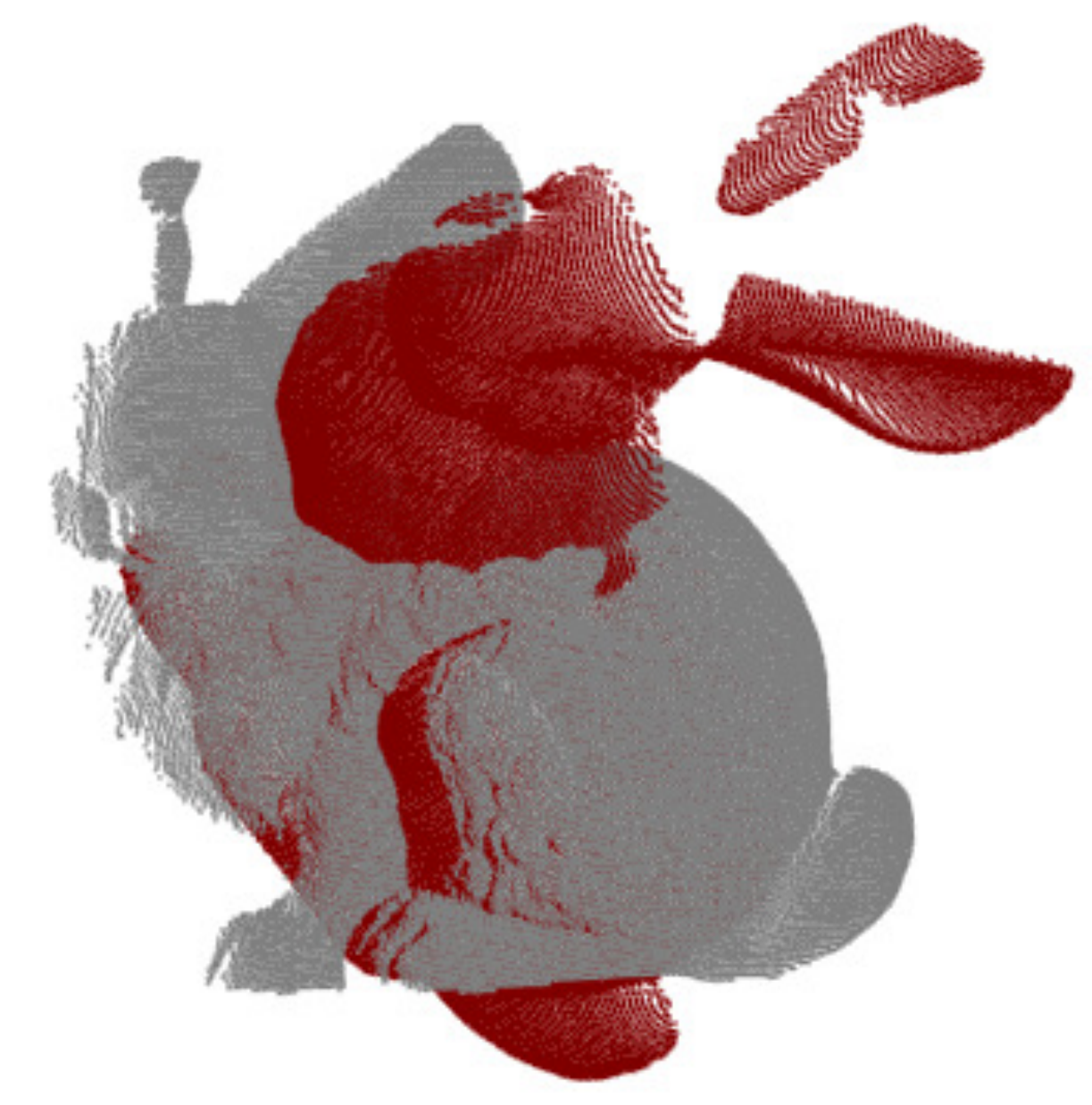}}
  \end{minipage} 
  & \begin{minipage}[b]{0.16\columnwidth}
    \centering
    \raisebox{-.5\height}{\includegraphics[width=\linewidth]{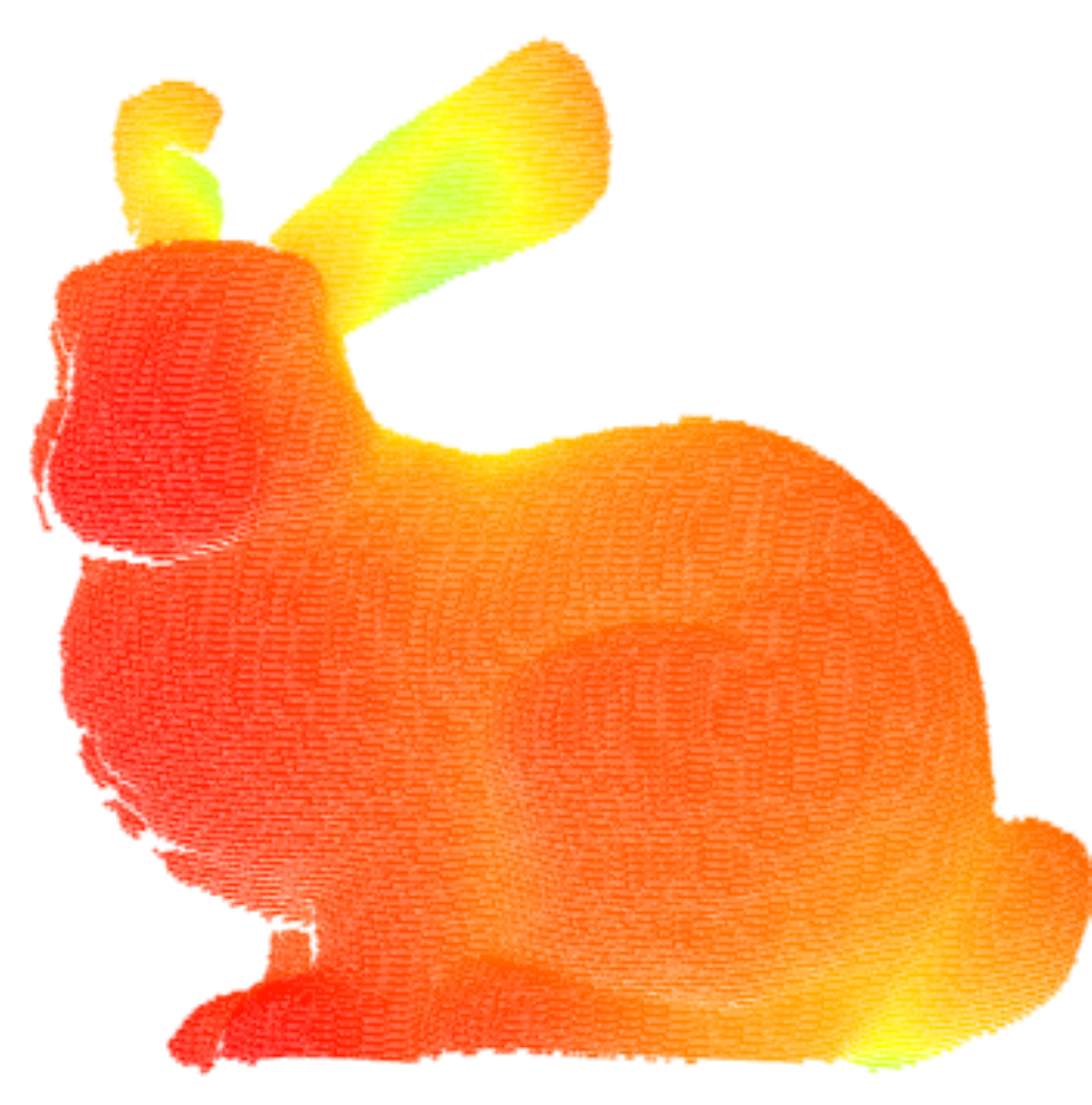}}
  \end{minipage}
  & \begin{minipage}[b]{0.16\columnwidth}
    \centering
    \raisebox{-.5\height}{\includegraphics[width=\linewidth]{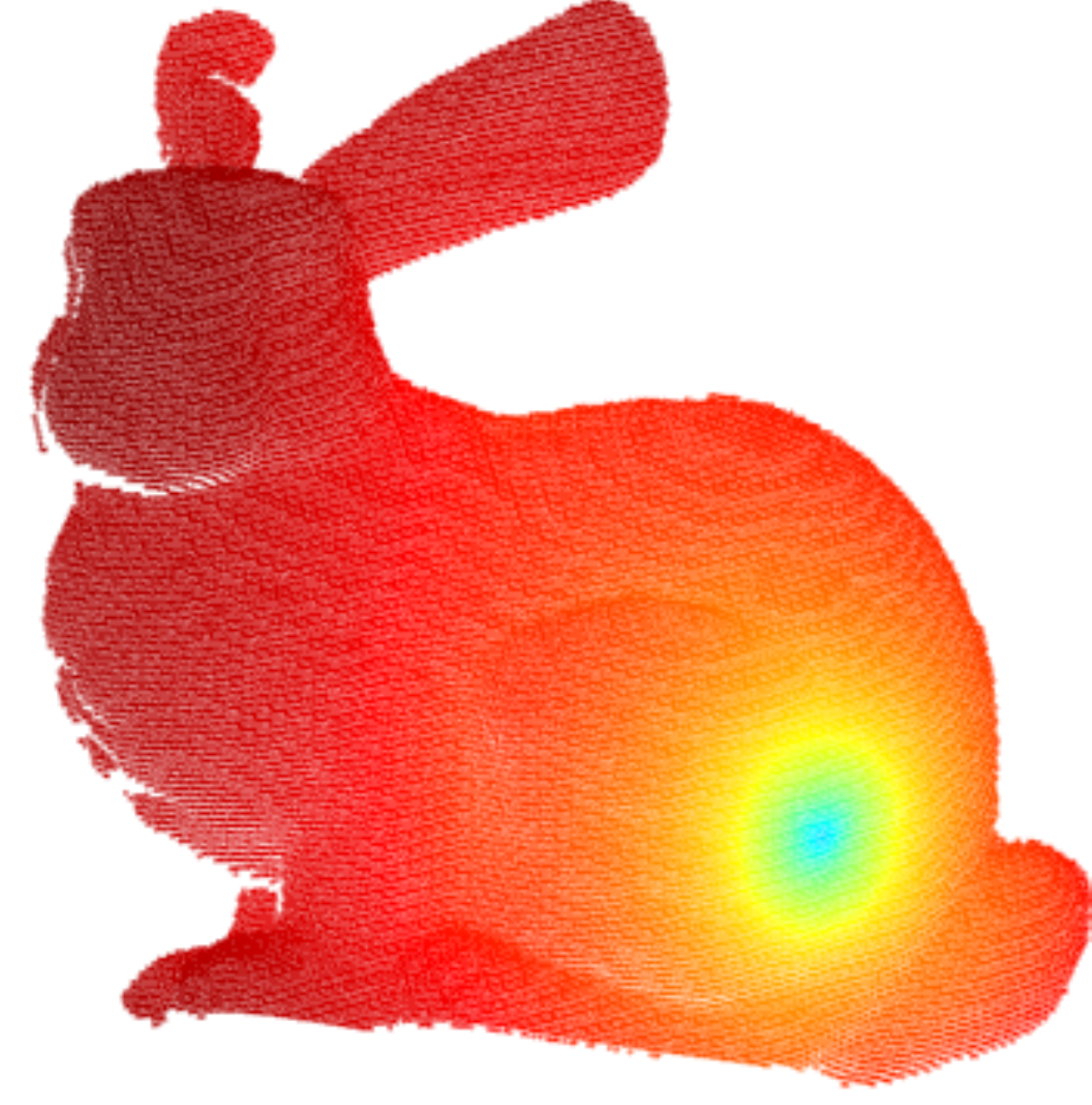}}
  \end{minipage} 
  & \begin{minipage}[b]{0.16\columnwidth}
    \centering
    \raisebox{-.5\height}{\includegraphics[width=\linewidth]{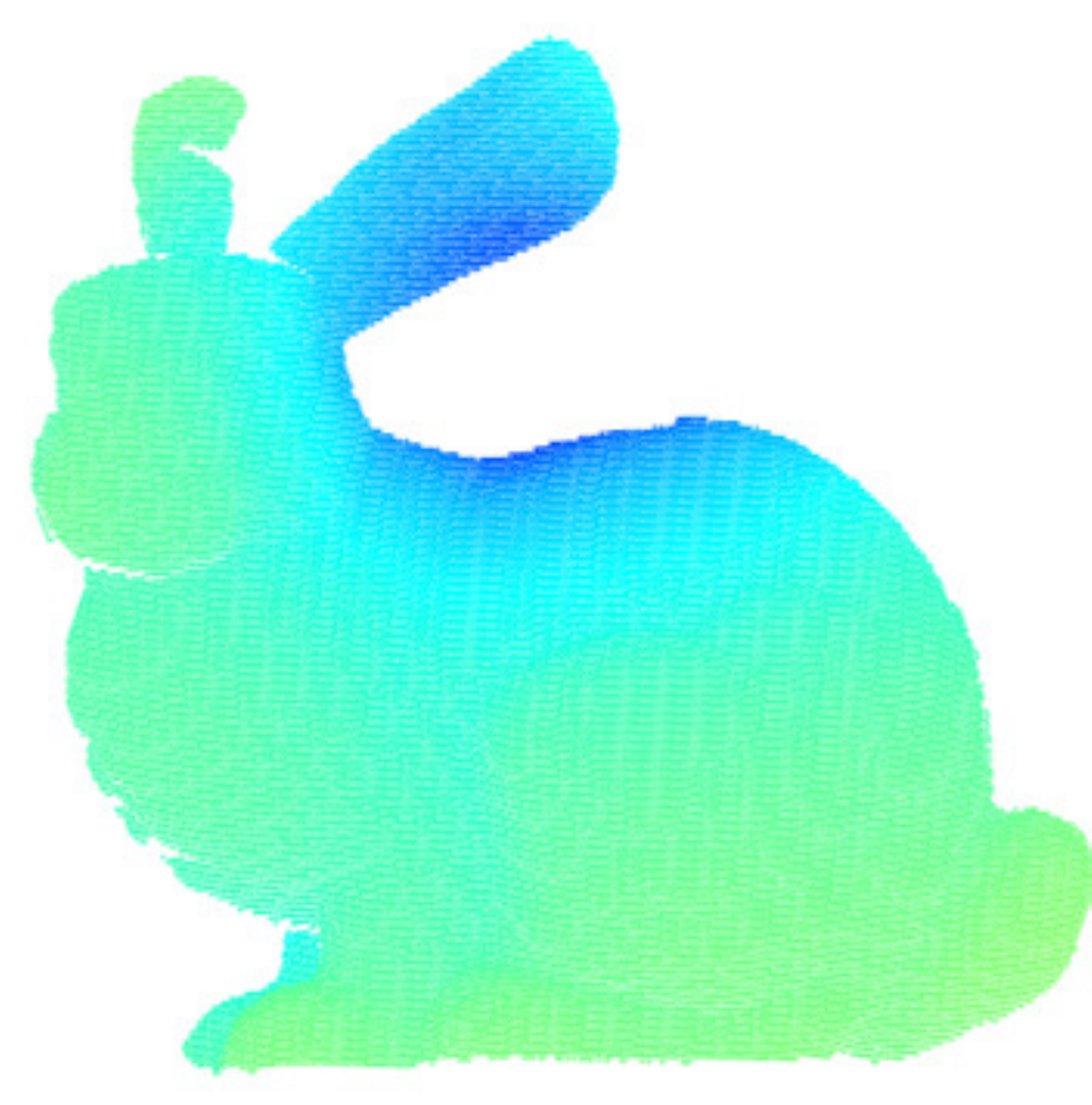}}
  \end{minipage} 
  & \begin{minipage}[b]{0.16\columnwidth}
    \centering
    \raisebox{-.5\height}{\includegraphics[width=\linewidth]{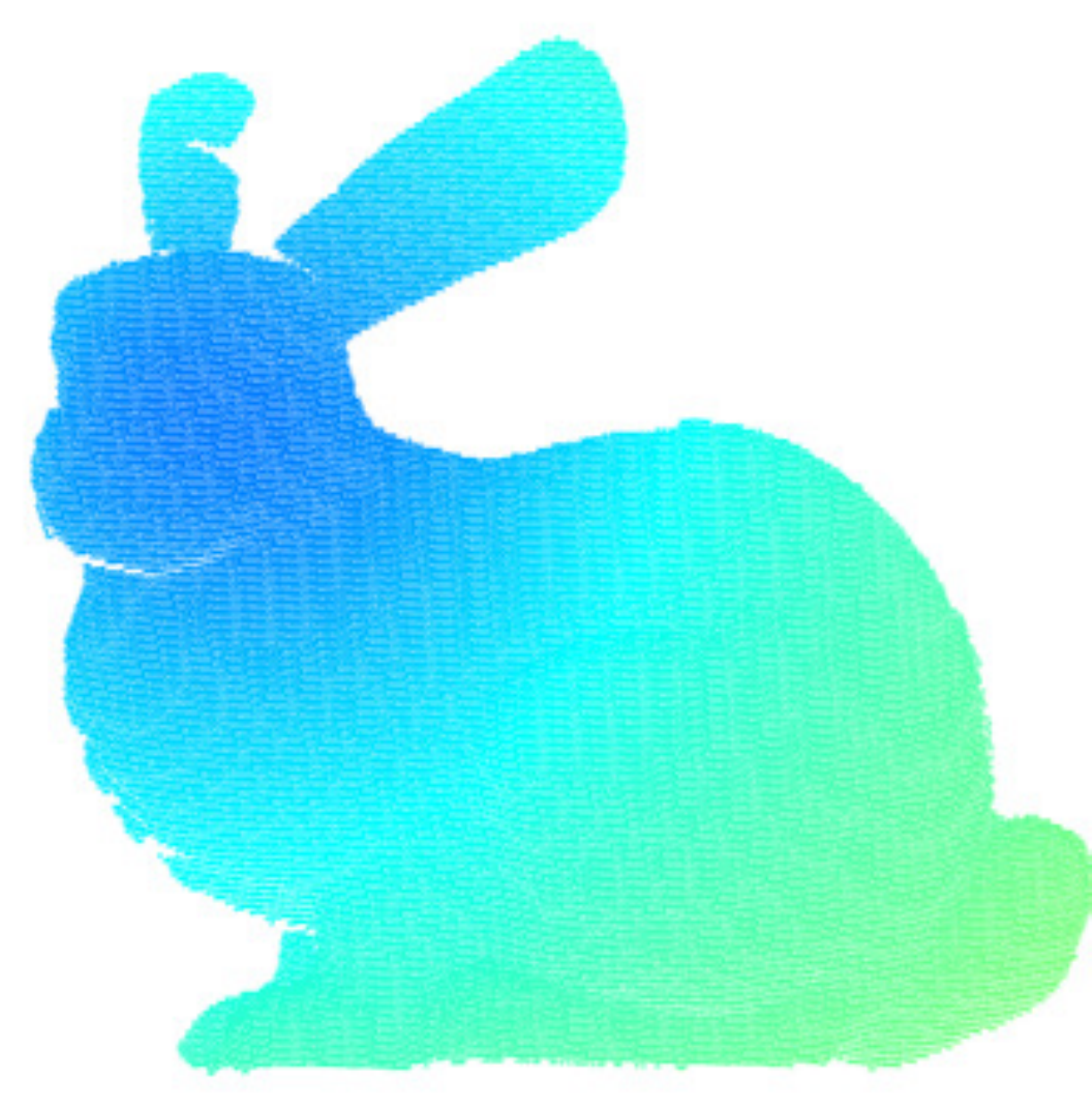}}
  \end{minipage} 
  & \begin{minipage}[b]{0.16\columnwidth}
    \centering
    \raisebox{-.5\height}{\includegraphics[width=\linewidth]{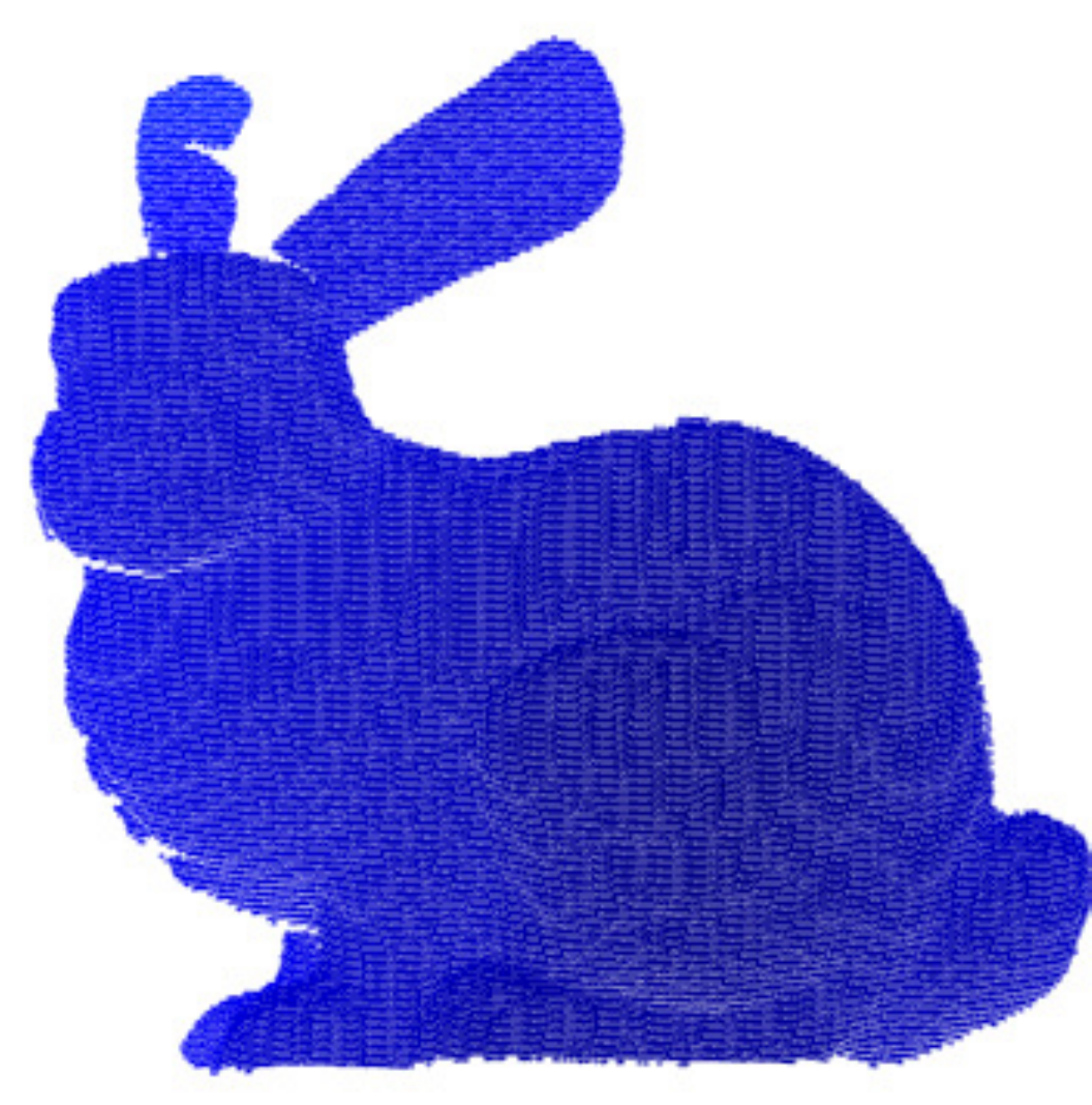}}
  \end{minipage} 
  & \begin{minipage}[b]{0.16\columnwidth}
    \centering
    \raisebox{-.5\height}{\includegraphics[width=\linewidth]{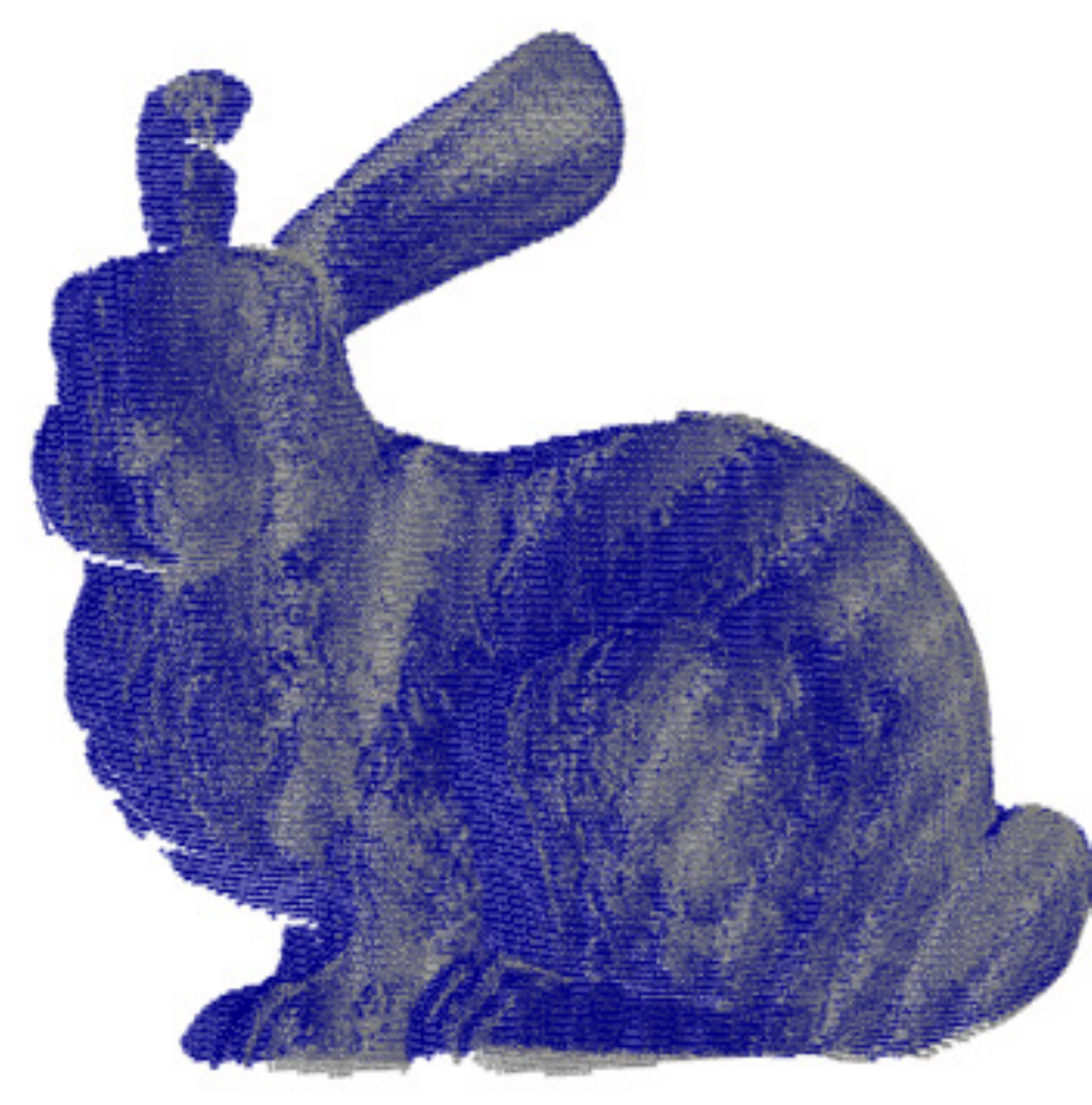}}
  \end{minipage} 
  & \begin{minipage}[b]{0.08\columnwidth}
    \centering
    \raisebox{-.5\height}{\includegraphics[width=\linewidth]{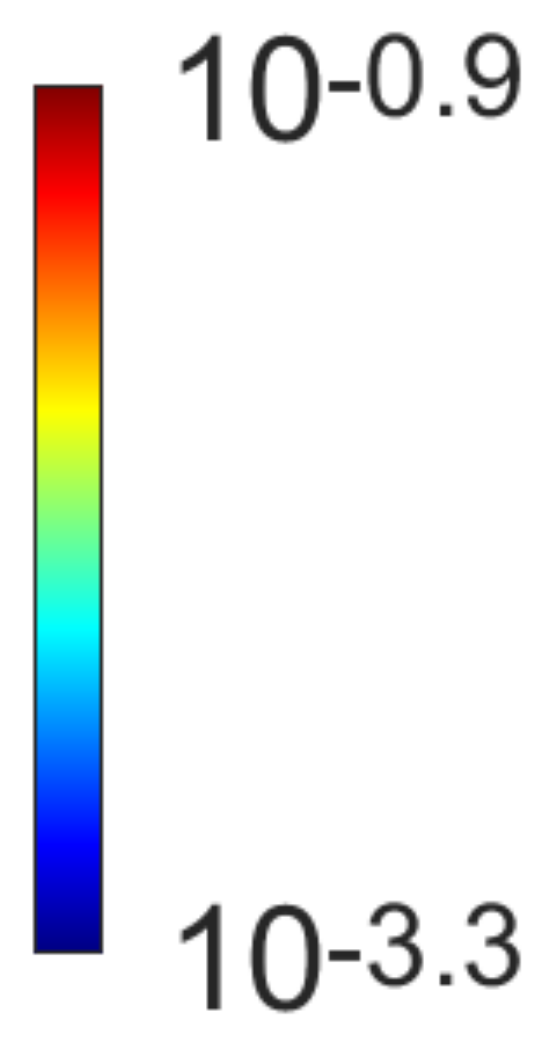}}
  \end{minipage} 
  \\

  \begin{minipage}[b]{0.2\columnwidth}
    \centering
    \raisebox{-.5\height}{\includegraphics[width=\linewidth]{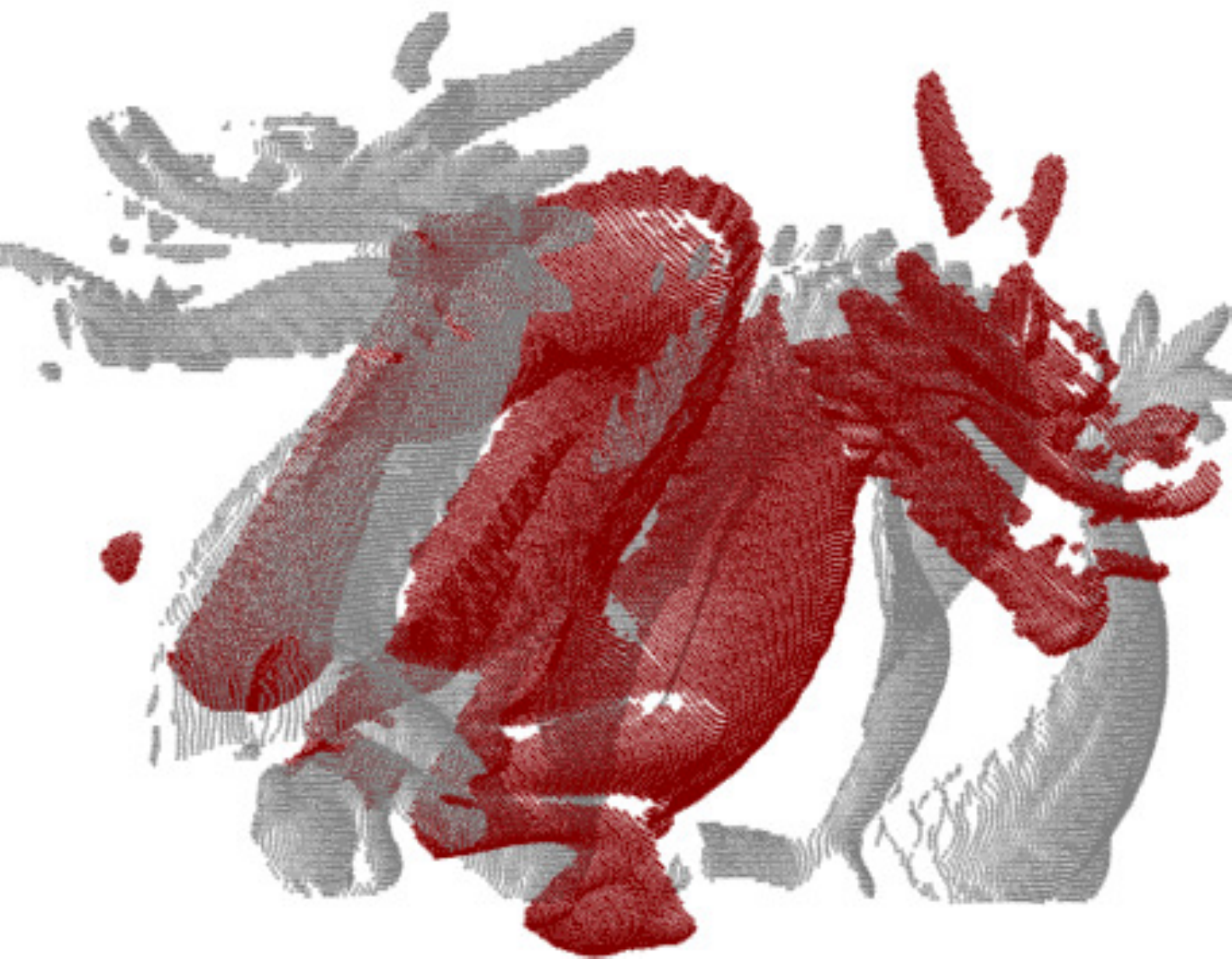}}
  \end{minipage} 
  & \begin{minipage}[b]{0.2\columnwidth}
    \centering
    \raisebox{-.5\height}{\includegraphics[width=\linewidth]{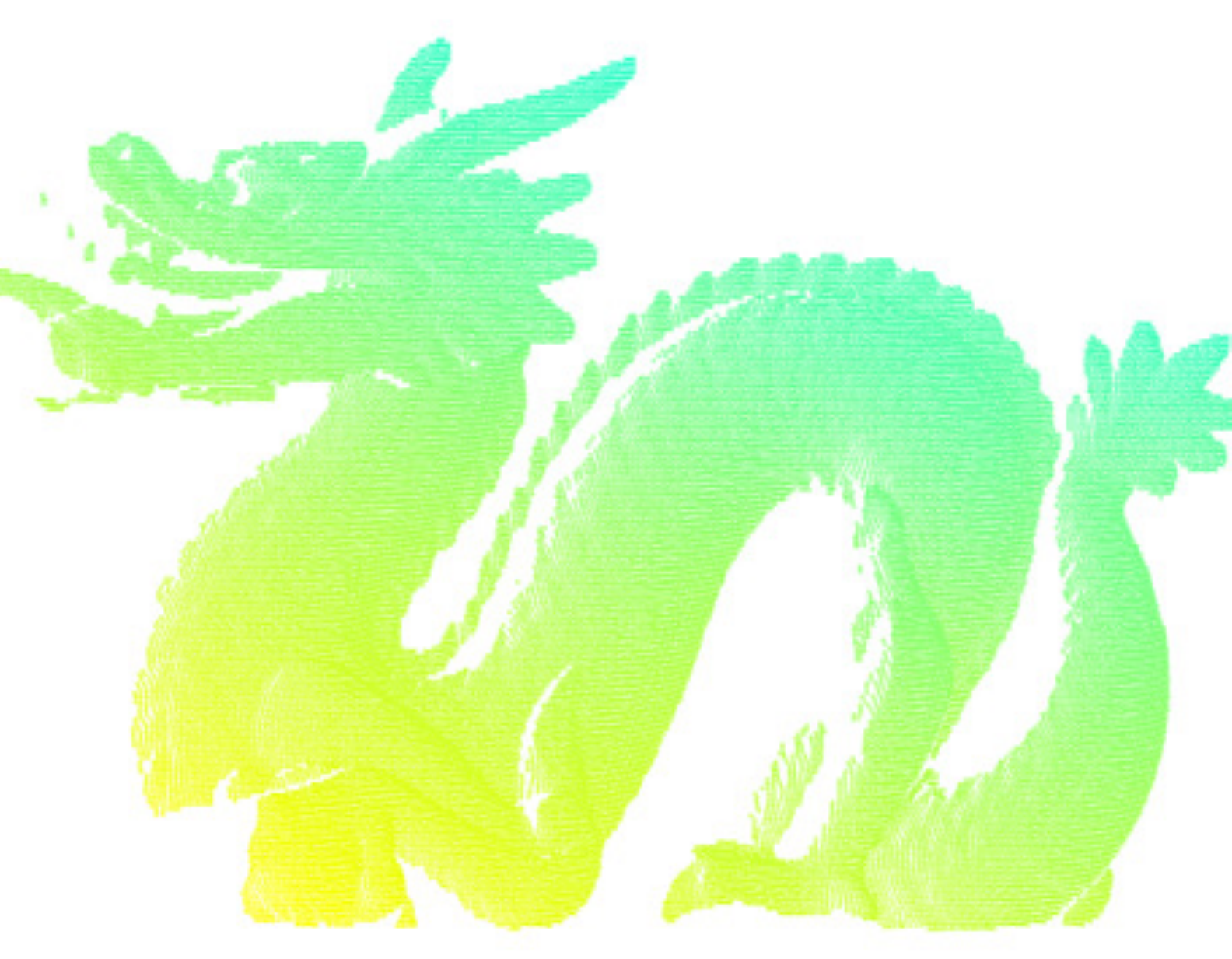}}
  \end{minipage}
  & \begin{minipage}[b]{0.2\columnwidth}
    \centering
    \raisebox{-.5\height}{\includegraphics[width=\linewidth]{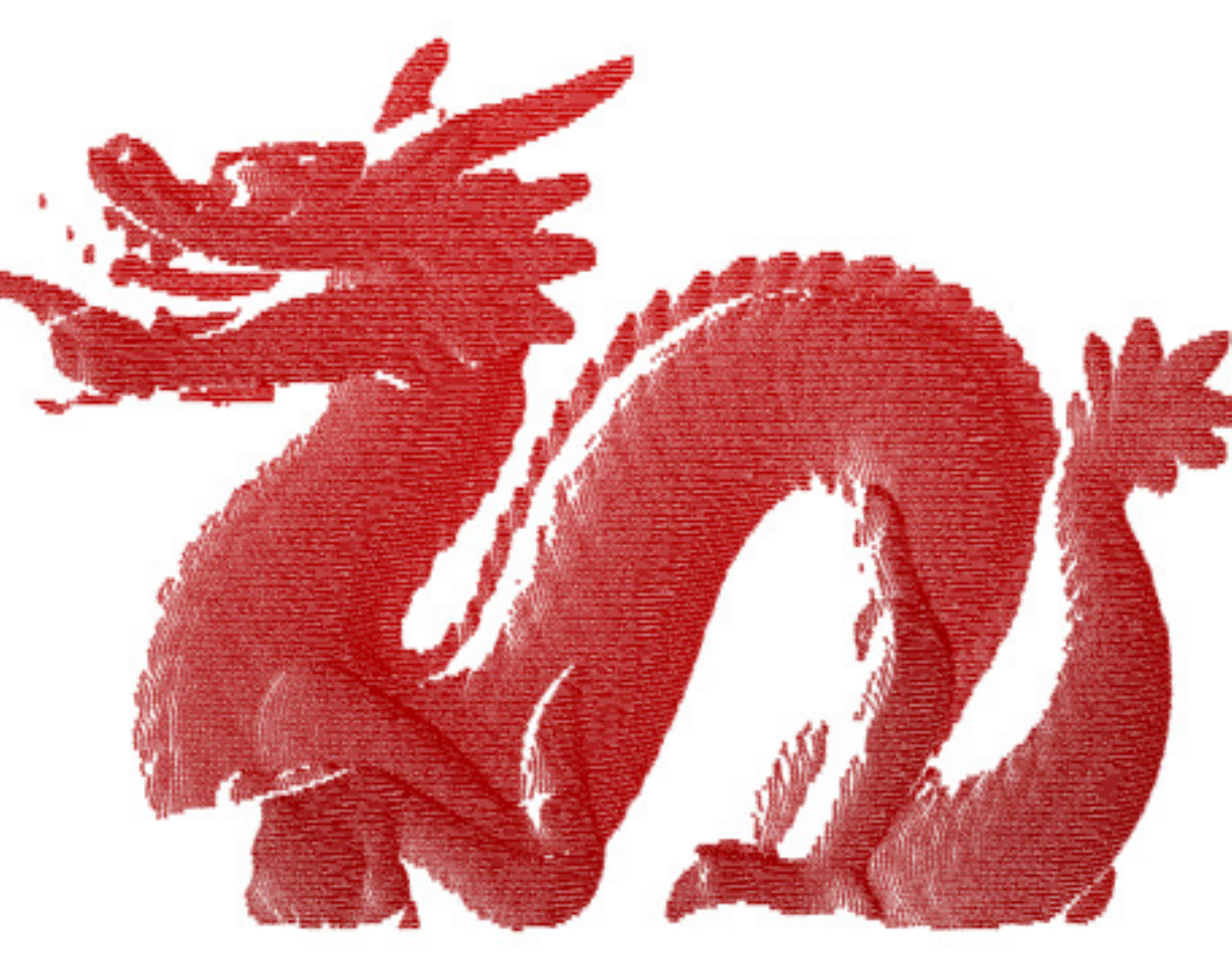}}
  \end{minipage} 
  & \begin{minipage}[b]{0.2\columnwidth}
    \centering
    \raisebox{-.5\height}{\includegraphics[width=\linewidth]{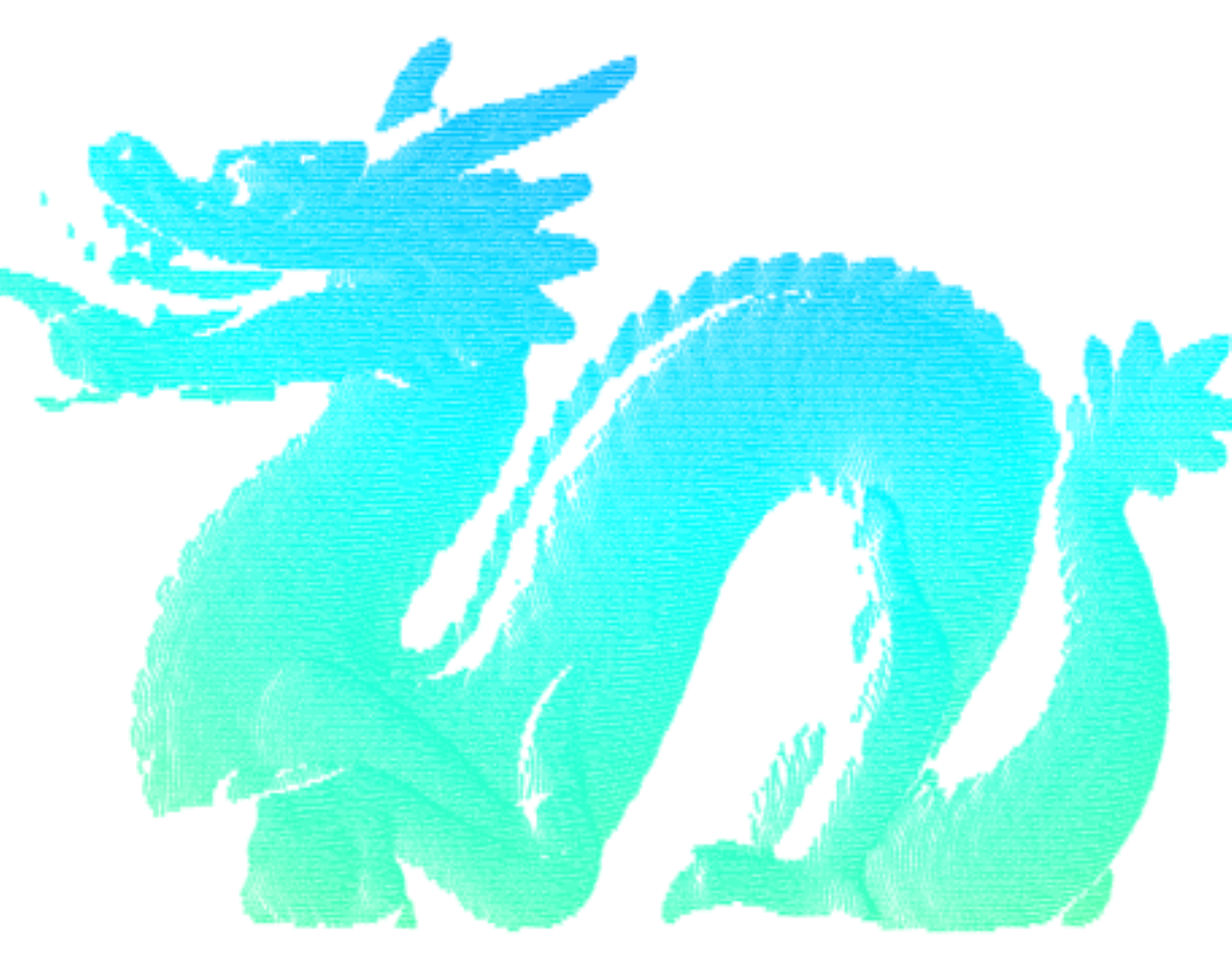}}
  \end{minipage} 
  & \begin{minipage}[b]{0.2\columnwidth}
    \centering
    \raisebox{-.5\height}{\includegraphics[width=\linewidth]{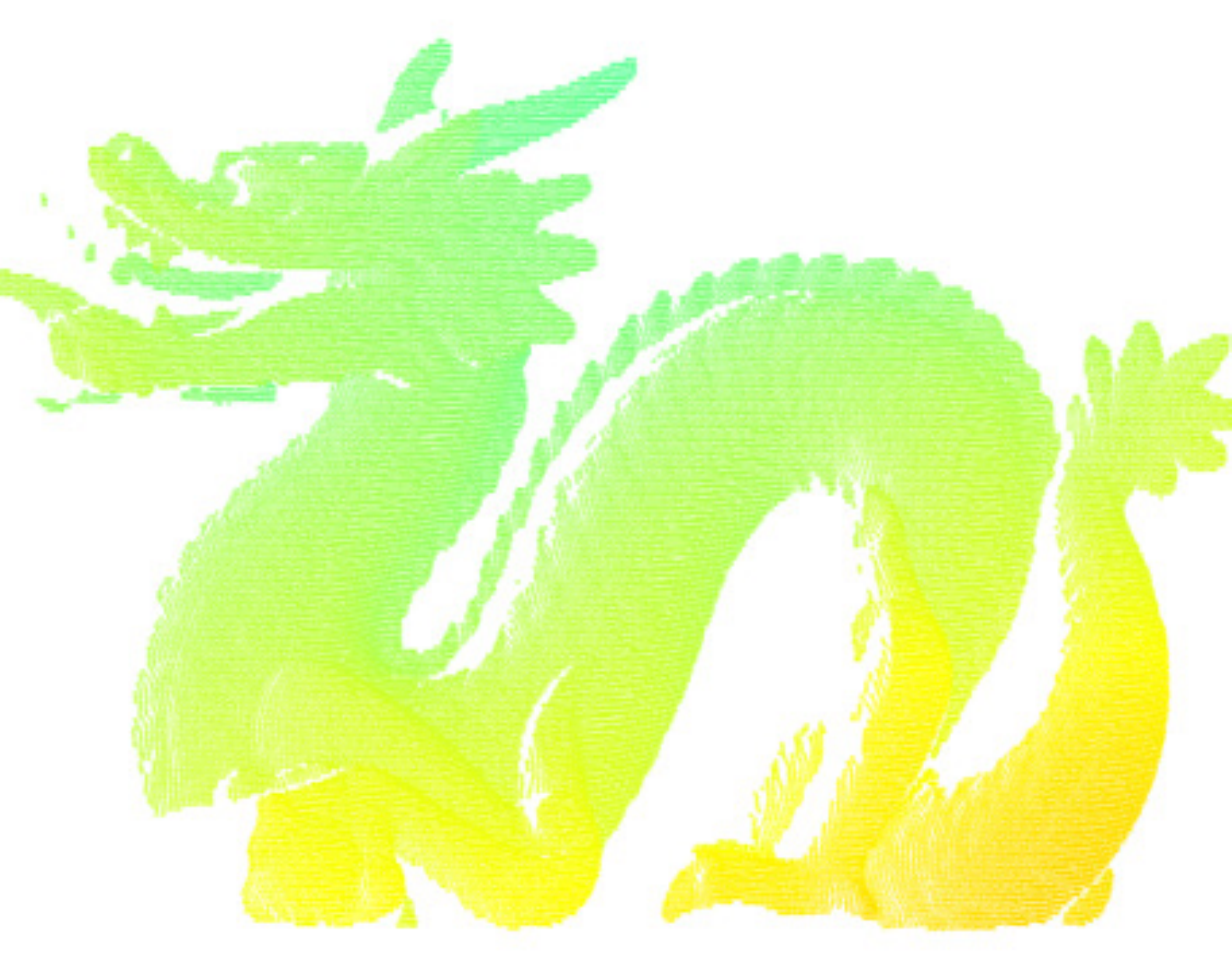}}
  \end{minipage} 
  & \begin{minipage}[b]{0.2\columnwidth}
    \centering
    \raisebox{-.5\height}{\includegraphics[width=\linewidth]{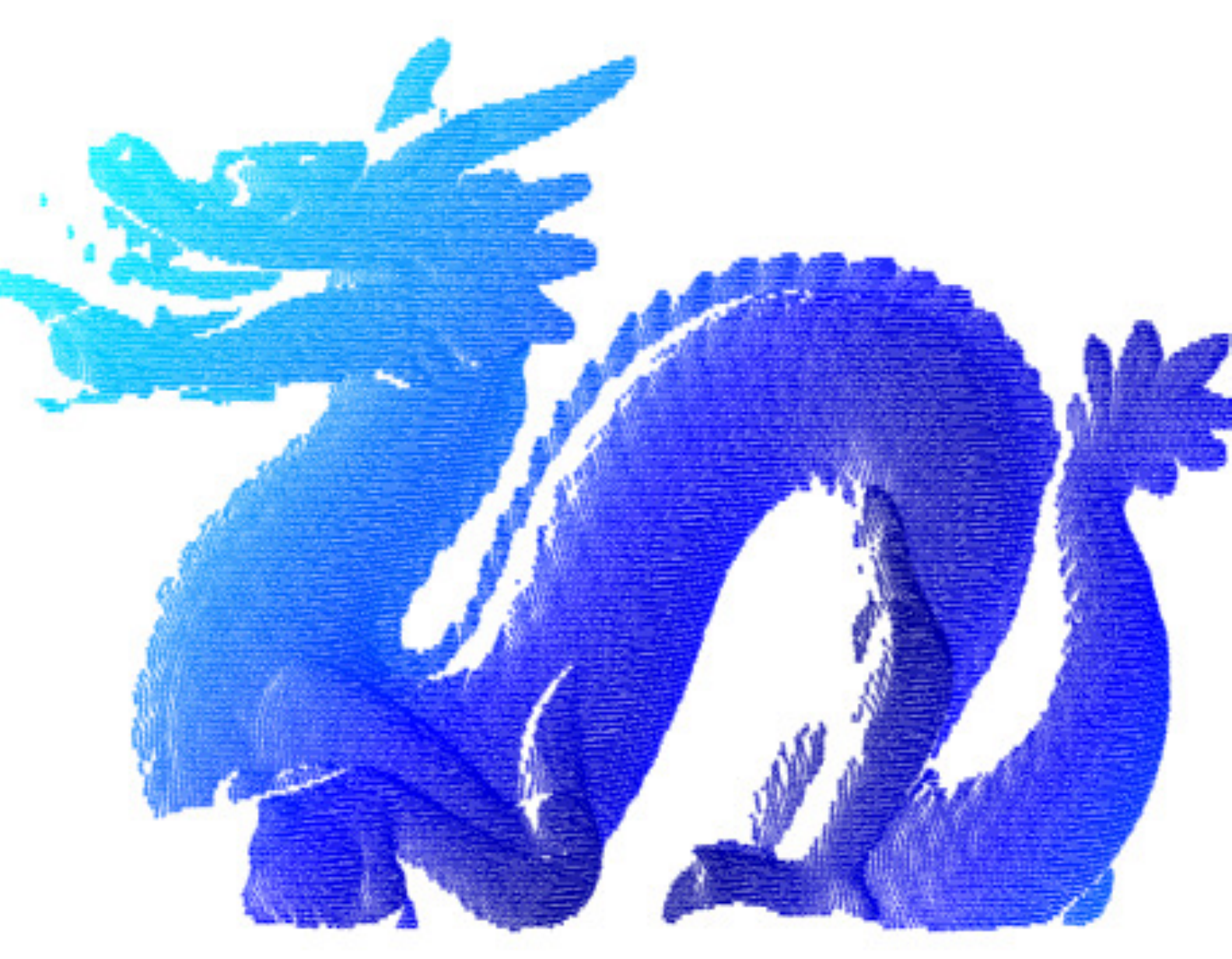}}
  \end{minipage} 
  & \begin{minipage}[b]{0.2\columnwidth}
    \centering
    \raisebox{-.5\height}{\includegraphics[width=\linewidth]{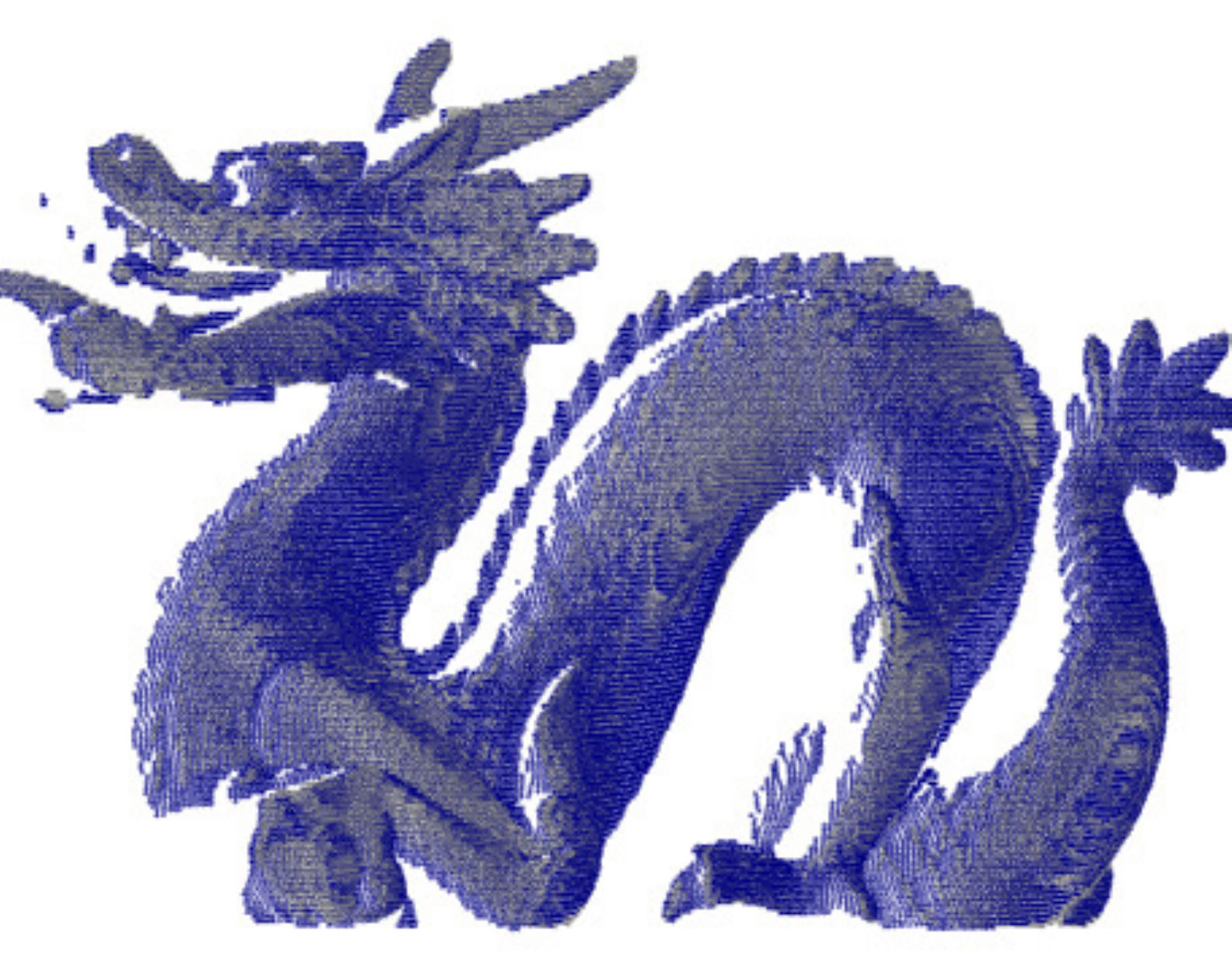}}
  \end{minipage}
  & \begin{minipage}[b]{0.08\columnwidth}
    \centering
    \raisebox{-.5\height}{\includegraphics[width=\linewidth]{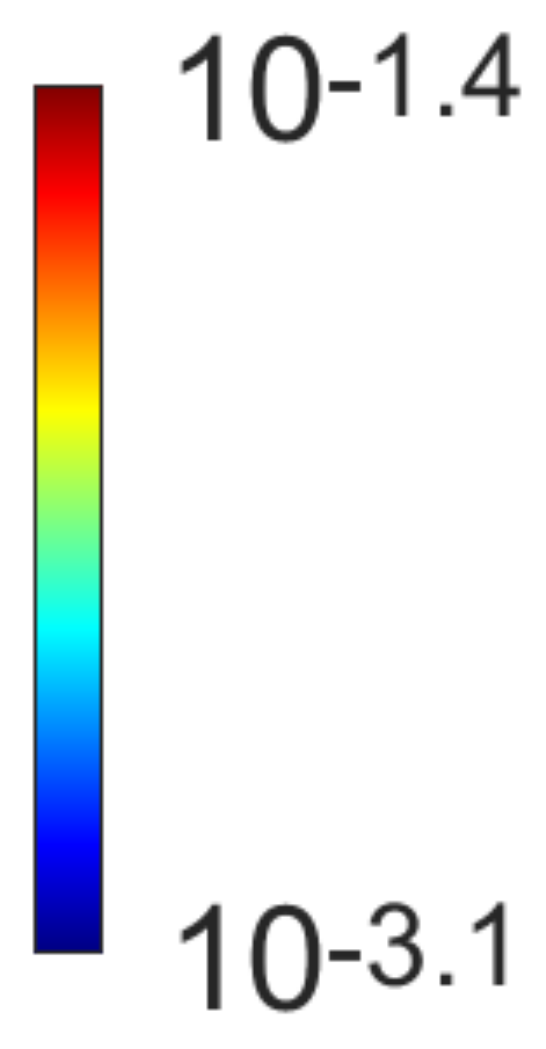}}
  \end{minipage} 
  \\

  \begin{minipage}[b]{0.112\columnwidth}
    \centering
    \raisebox{-.5\height}{\includegraphics[width=\linewidth]{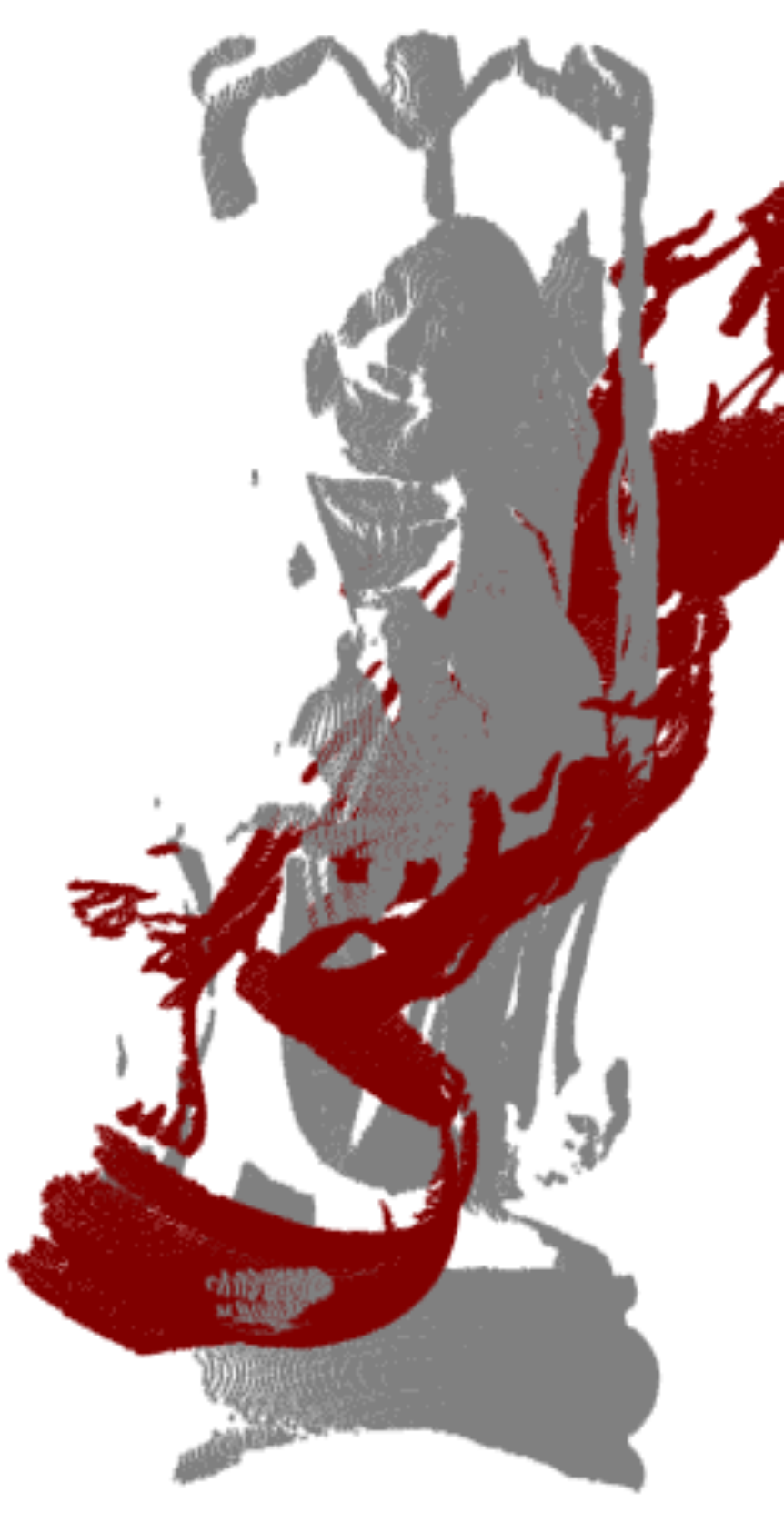}}
  \end{minipage} 
  & \begin{minipage}[b]{0.112\columnwidth}
    \centering
    \raisebox{-.5\height}{\includegraphics[width=\linewidth]{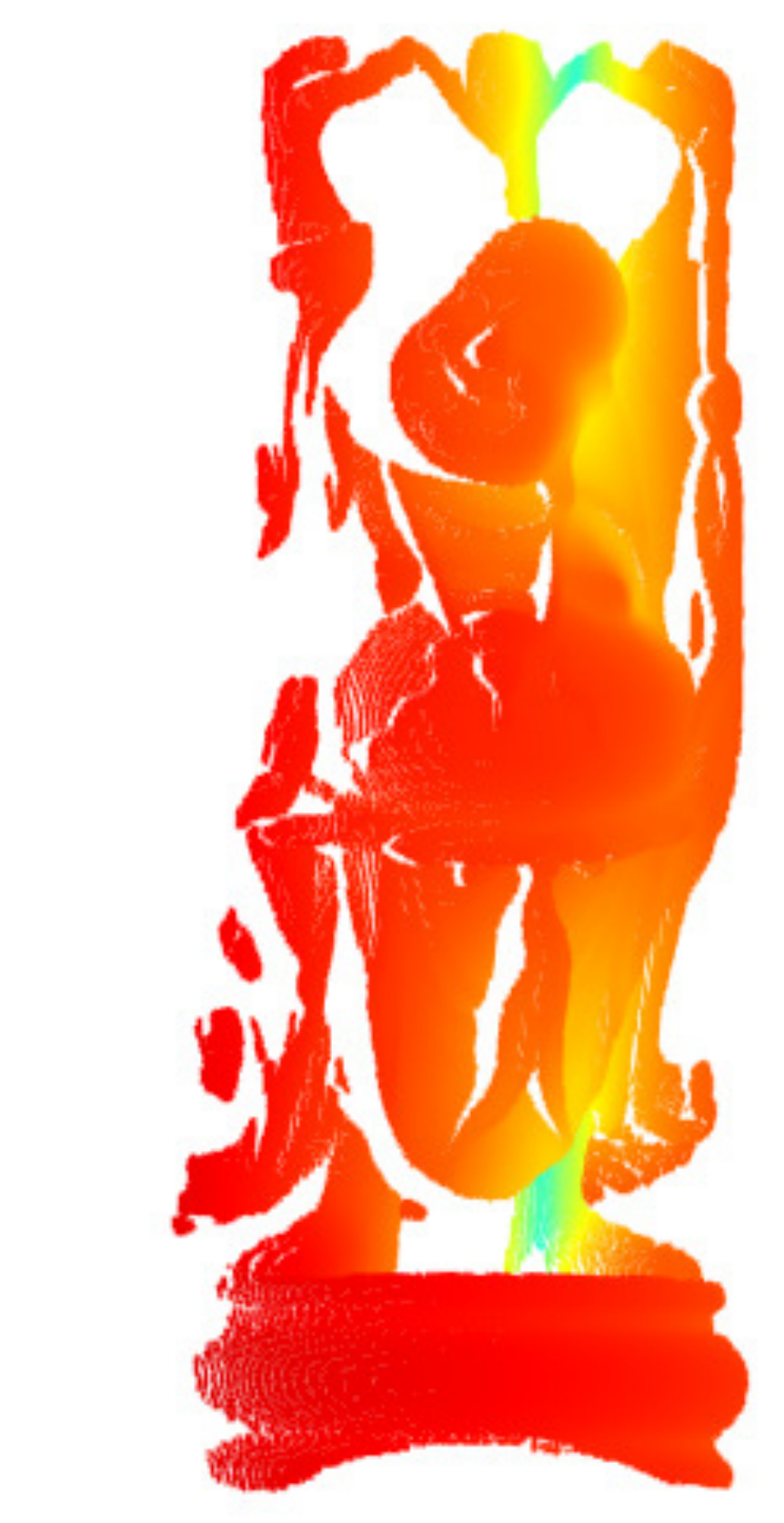}}
  \end{minipage}
  & \begin{minipage}[b]{0.112\columnwidth}
    \centering
    \raisebox{-.5\height}{\includegraphics[width=\linewidth]{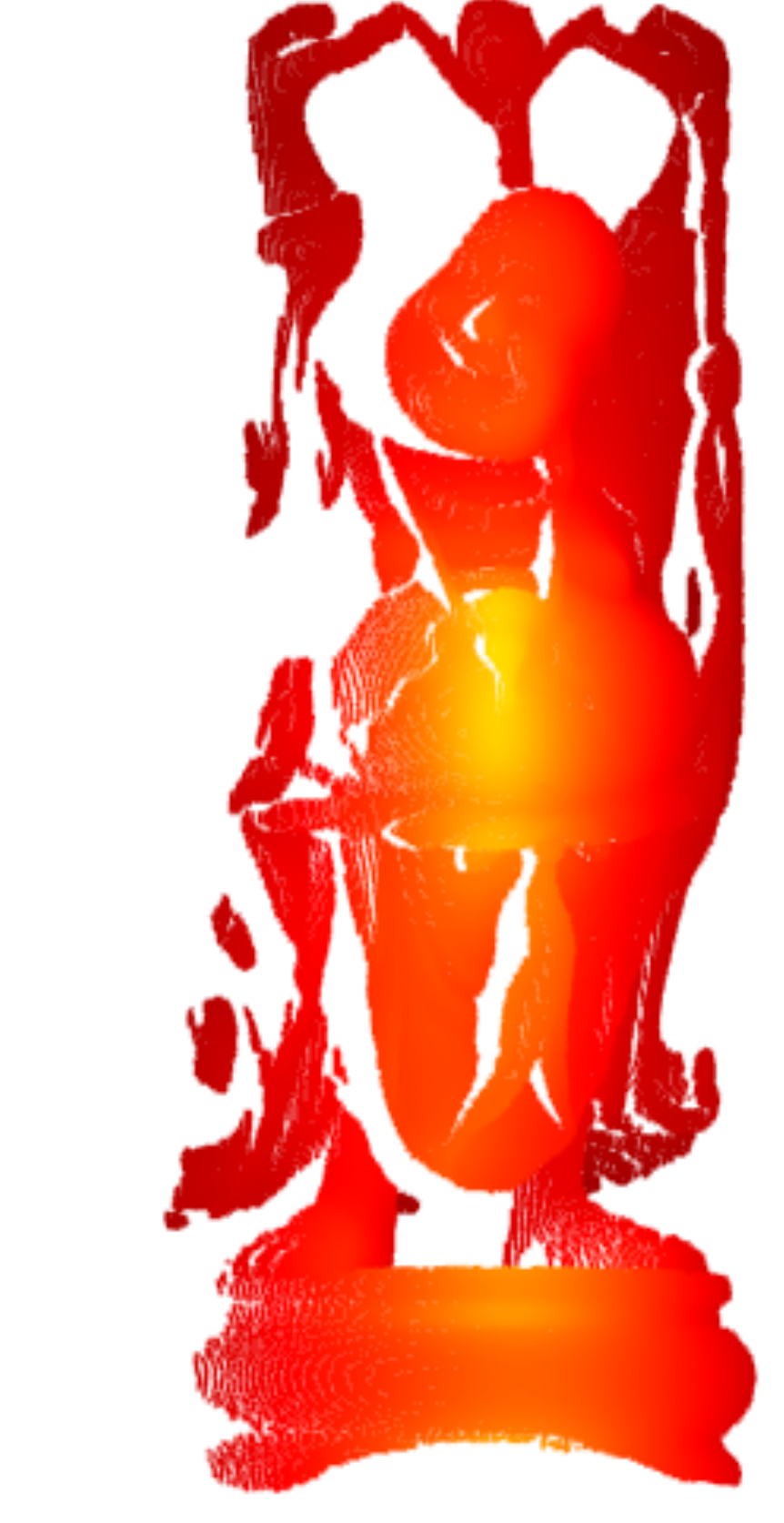}}
  \end{minipage} 
  & \begin{minipage}[b]{0.112\columnwidth}
    \centering
    \raisebox{-.5\height}{\includegraphics[width=\linewidth]{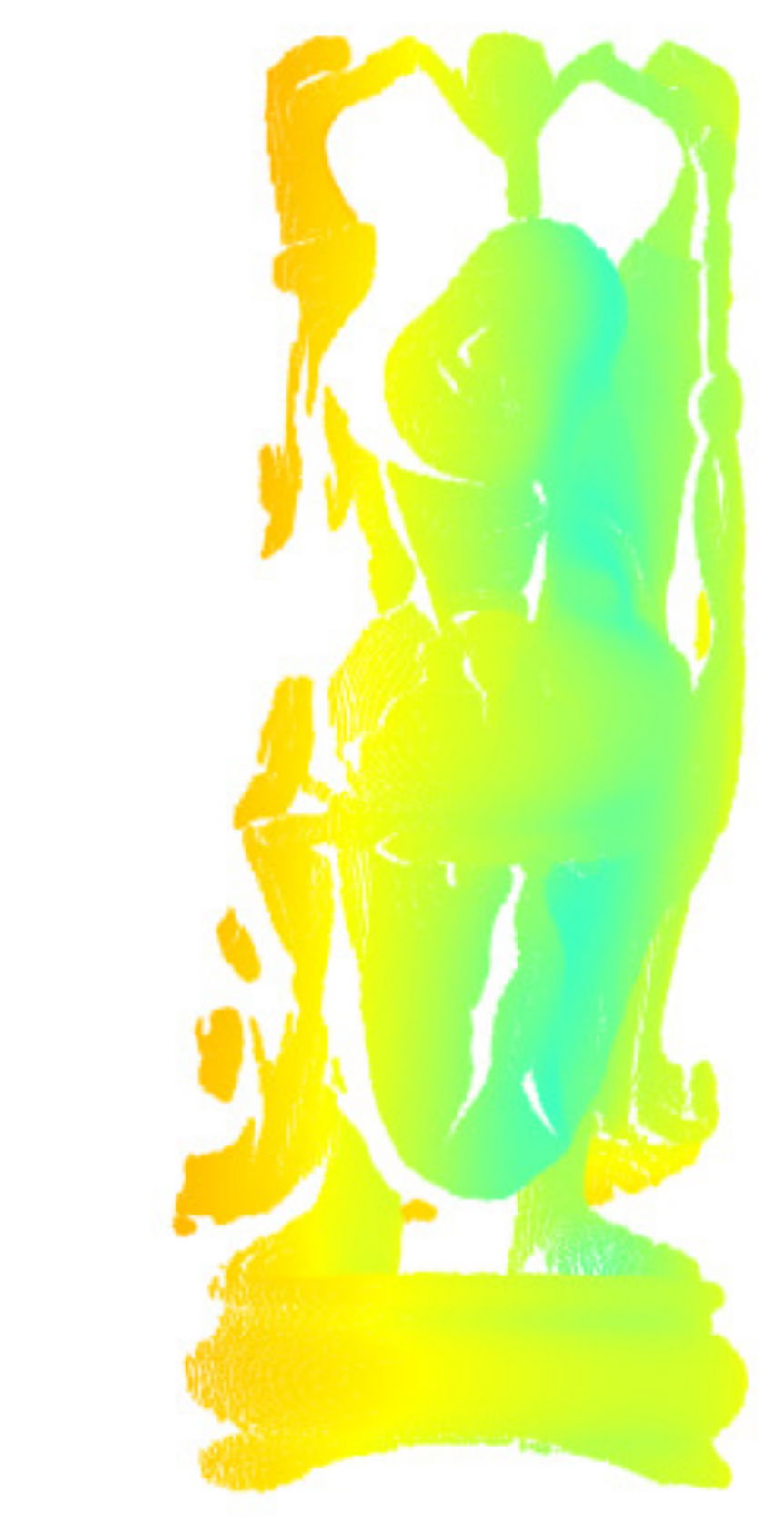}}
  \end{minipage} 
  & \begin{minipage}[b]{0.112\columnwidth}
    \centering
    \raisebox{-.5\height}{\includegraphics[width=\linewidth]{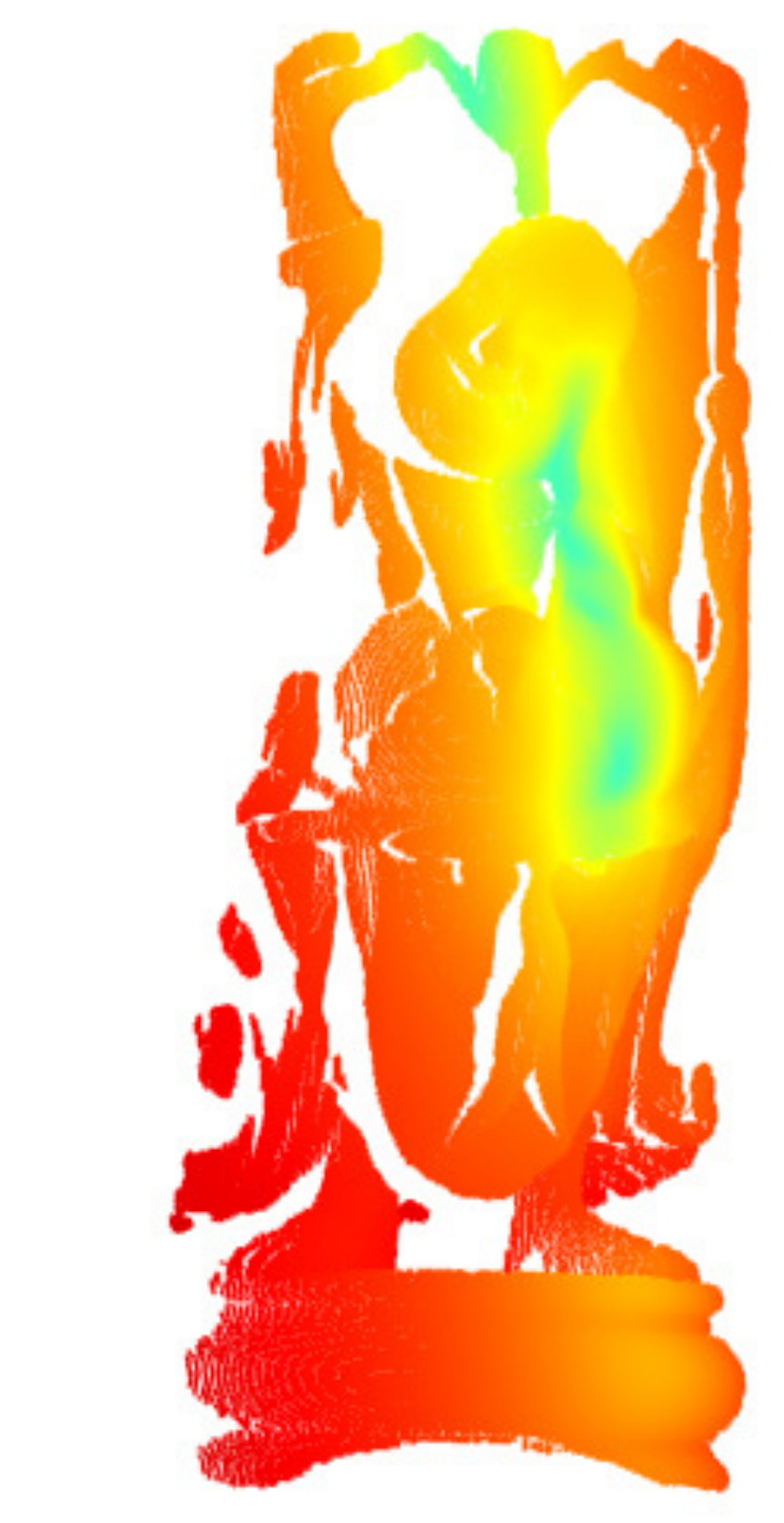}}
  \end{minipage} 
  & \begin{minipage}[b]{0.112\columnwidth}
    \centering
    \raisebox{-.5\height}{\includegraphics[width=\linewidth]{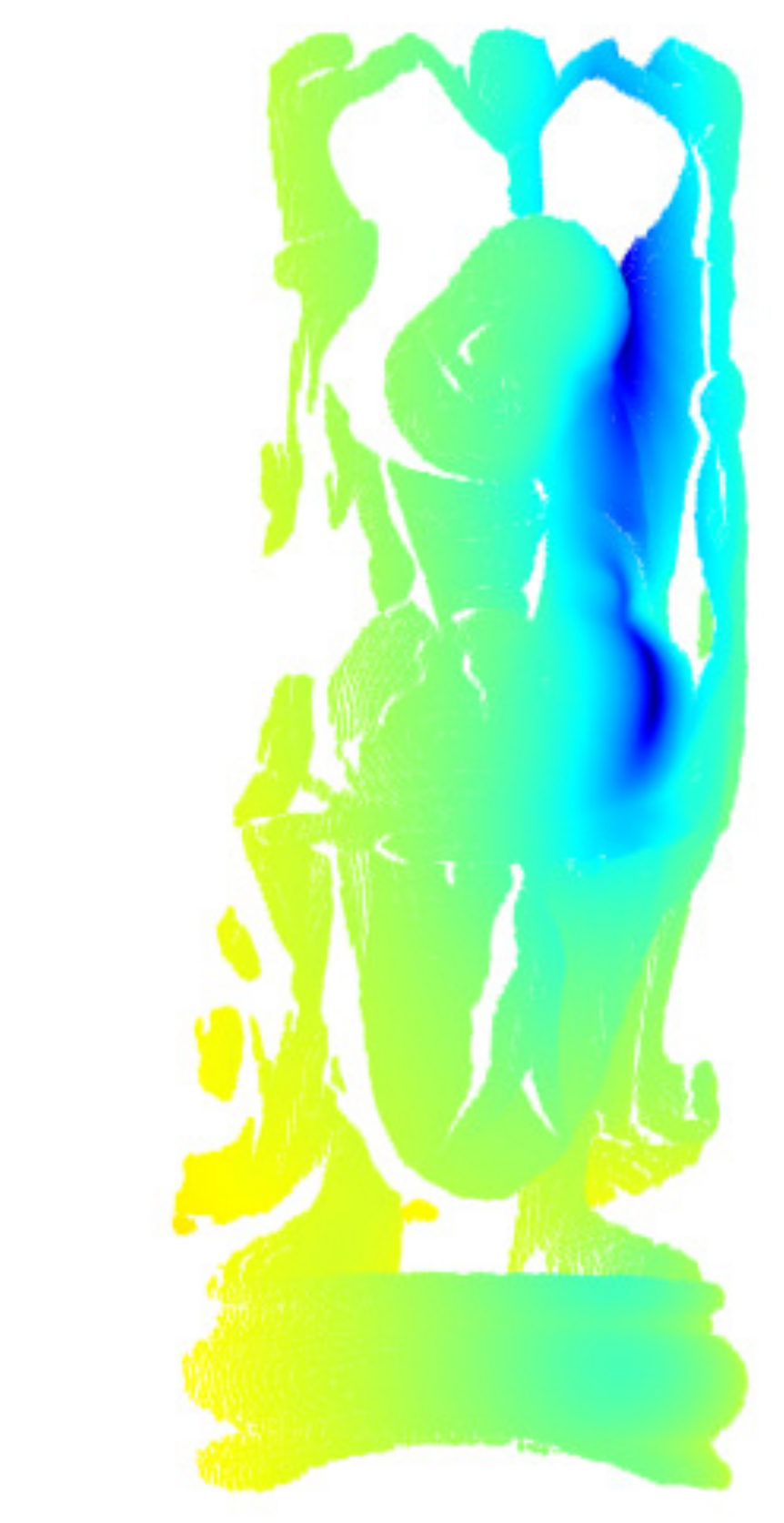}}
  \end{minipage} 
  & \begin{minipage}[b]{0.112\columnwidth}
    \centering
    \raisebox{-.5\height}{\includegraphics[width=\linewidth]{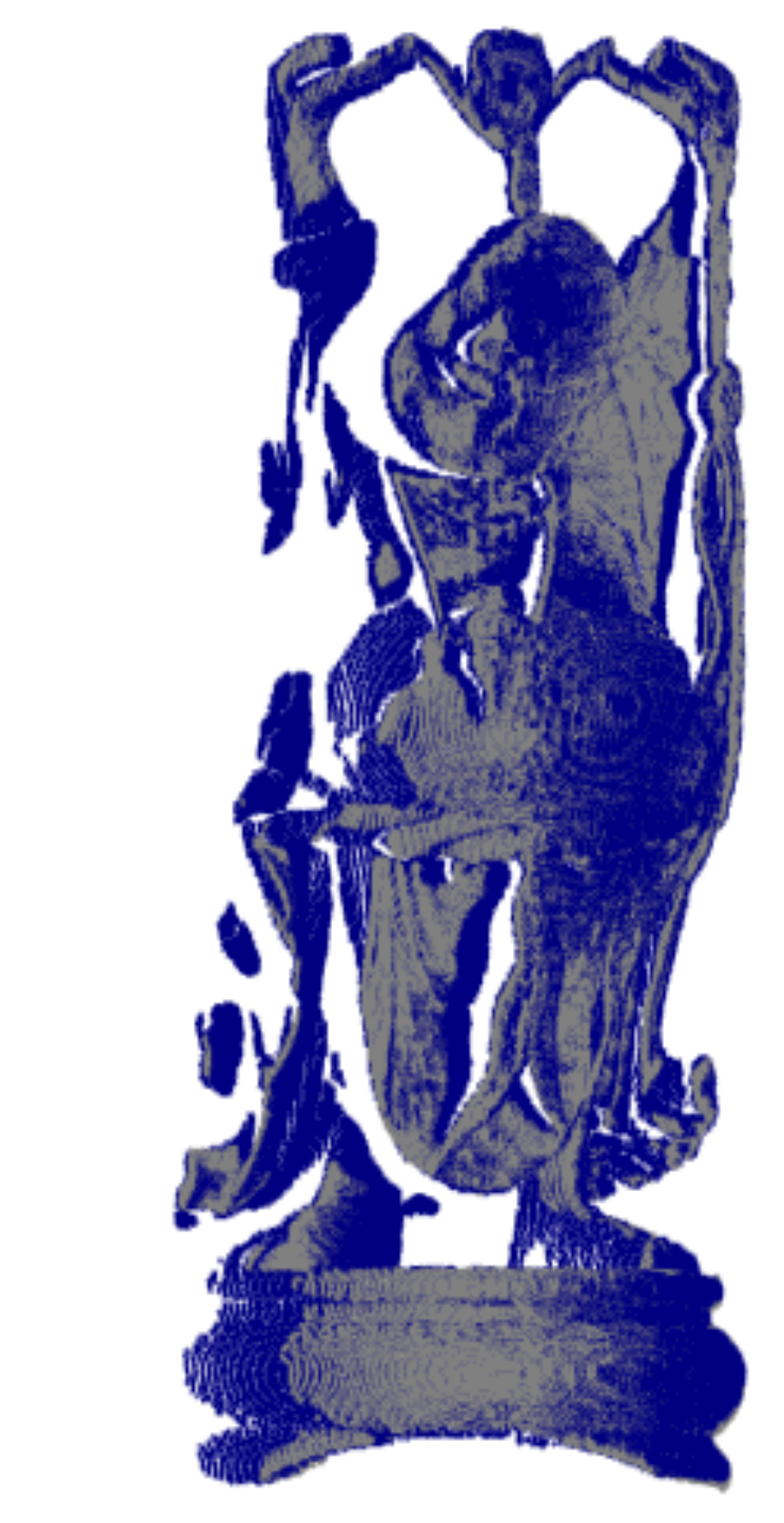}}
  \end{minipage} 
  & \begin{minipage}[b]{0.08\columnwidth}
    \centering
    \raisebox{-.5\height}{\includegraphics[width=\linewidth]{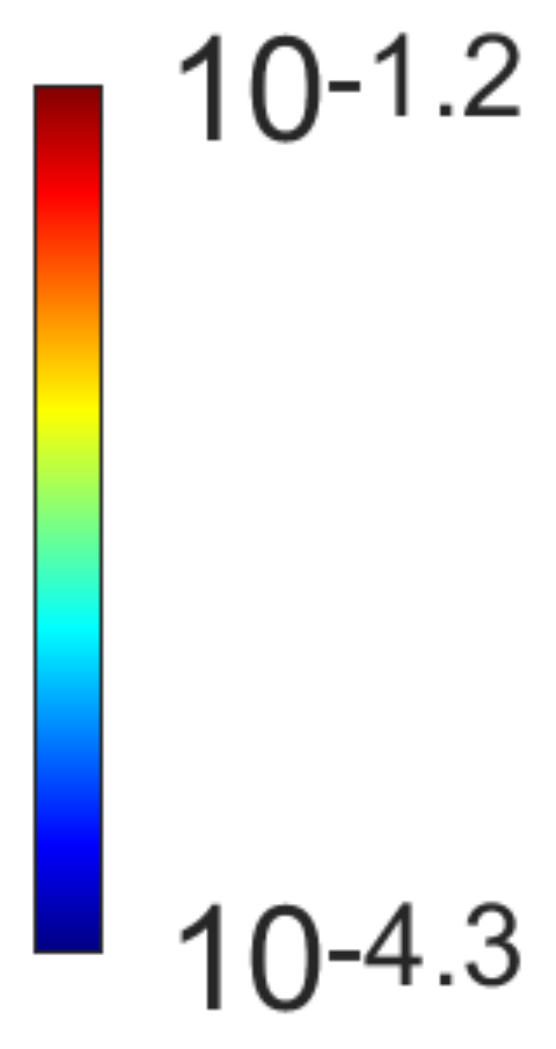}}
  \end{minipage} 
  \\

  \begin{minipage}[b]{0.096\columnwidth}
    \centering
    \raisebox{-.5\height}{\includegraphics[width=\linewidth]{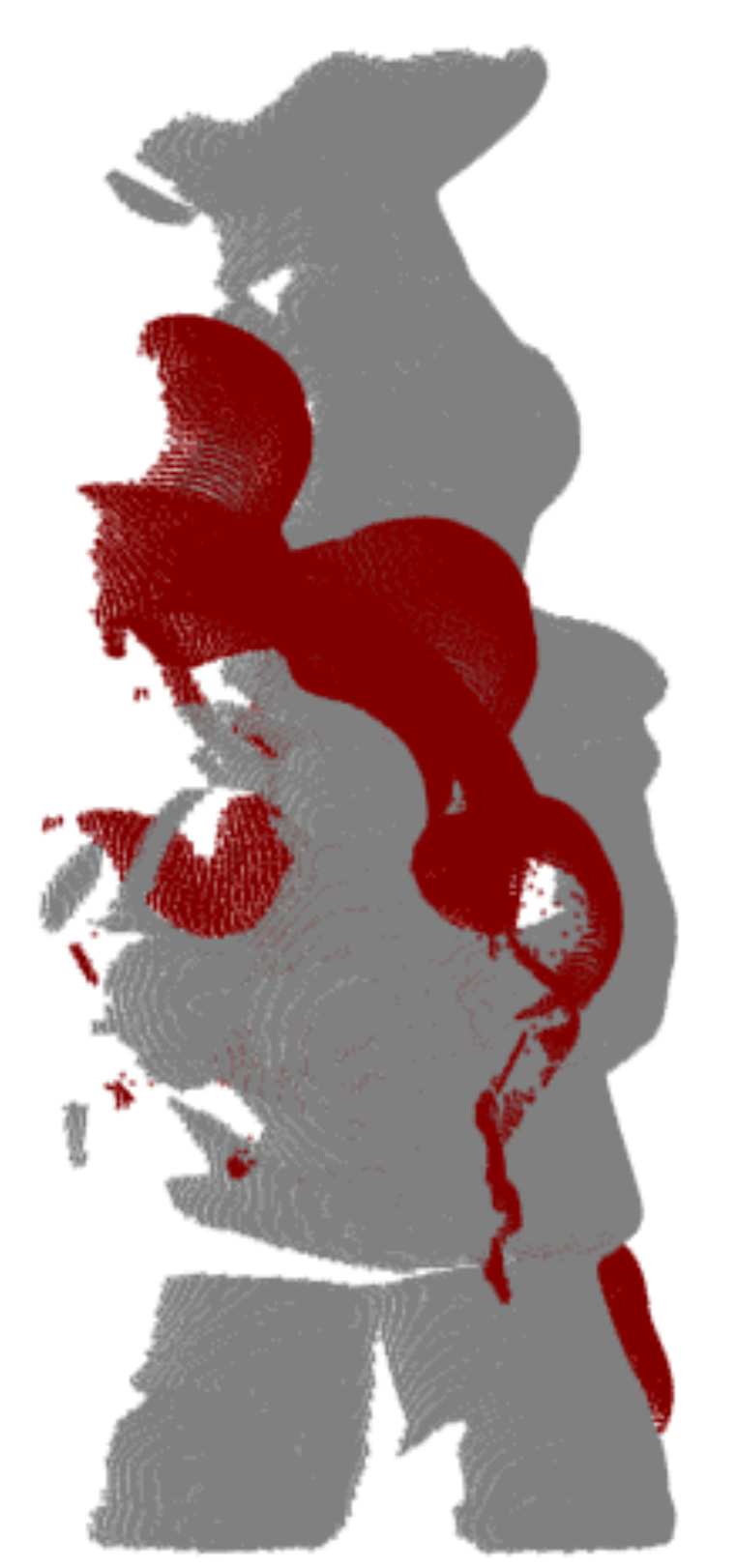}}
  \end{minipage} 
  & \begin{minipage}[b]{0.096\columnwidth}
    \centering
    \raisebox{-.5\height}{\includegraphics[width=\linewidth]{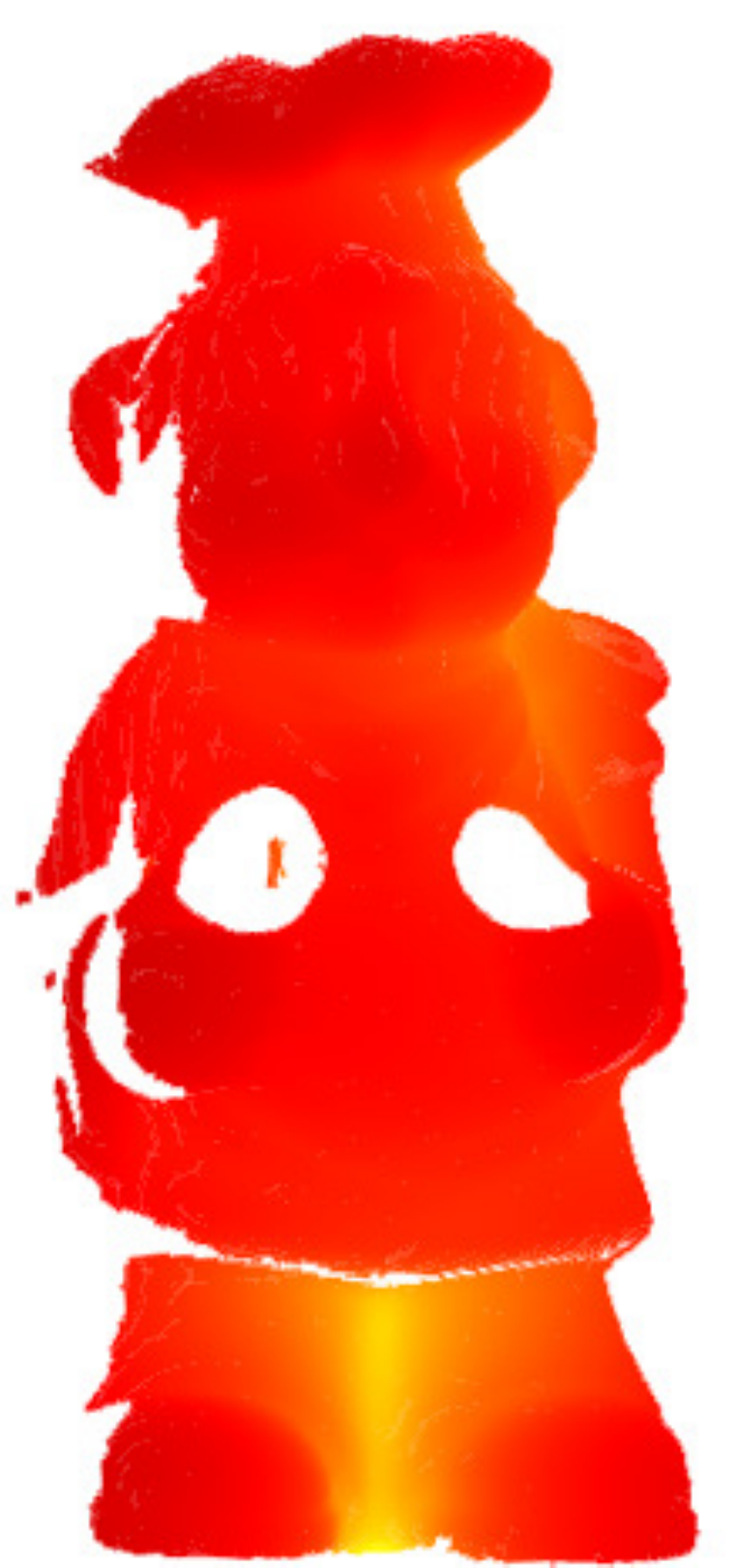}}
  \end{minipage}
  & \begin{minipage}[b]{0.096\columnwidth}
    \centering
    \raisebox{-.5\height}{\includegraphics[width=\linewidth]{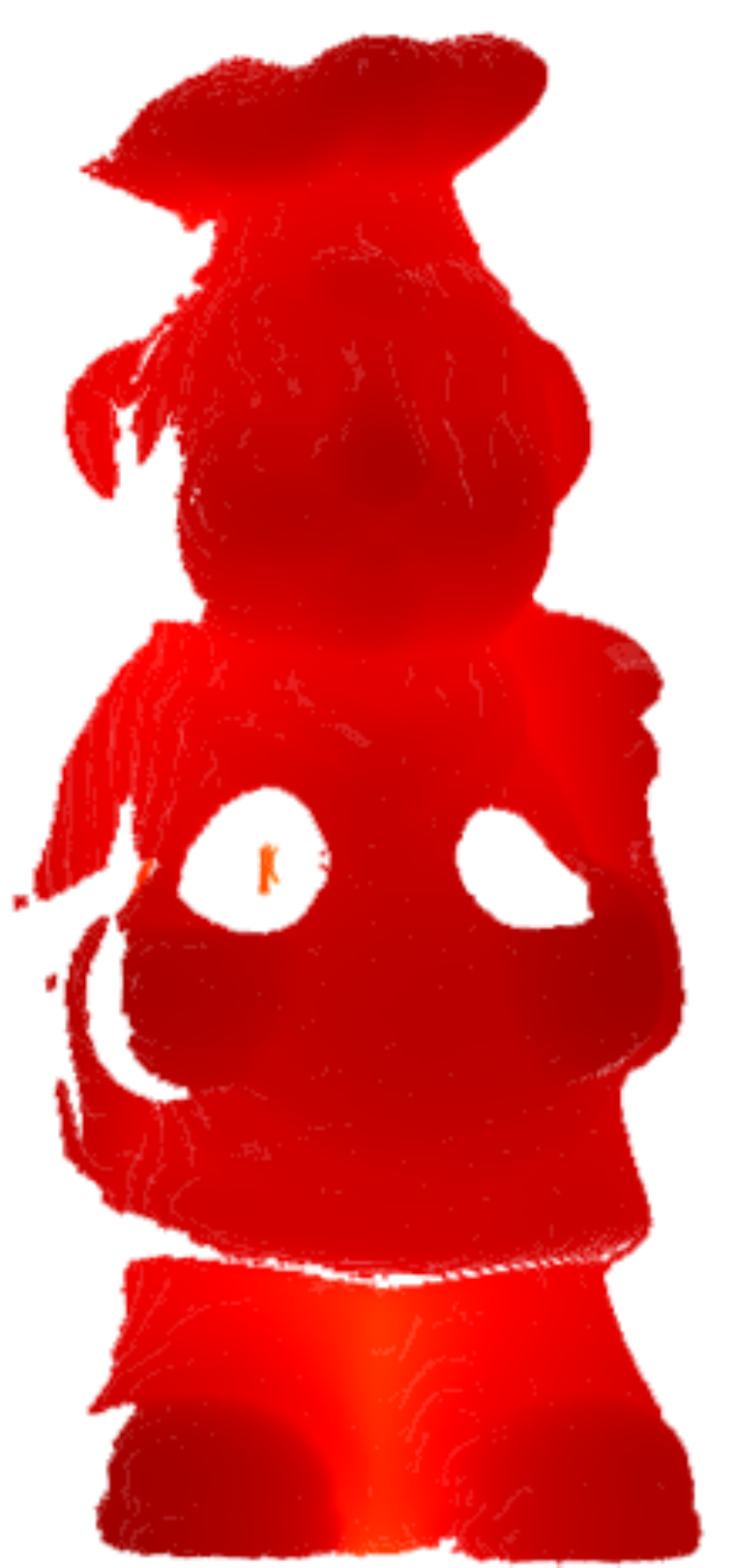}}
  \end{minipage} 
  & \begin{minipage}[b]{0.096\columnwidth}
    \centering
    \raisebox{-.5\height}{\includegraphics[width=\linewidth]{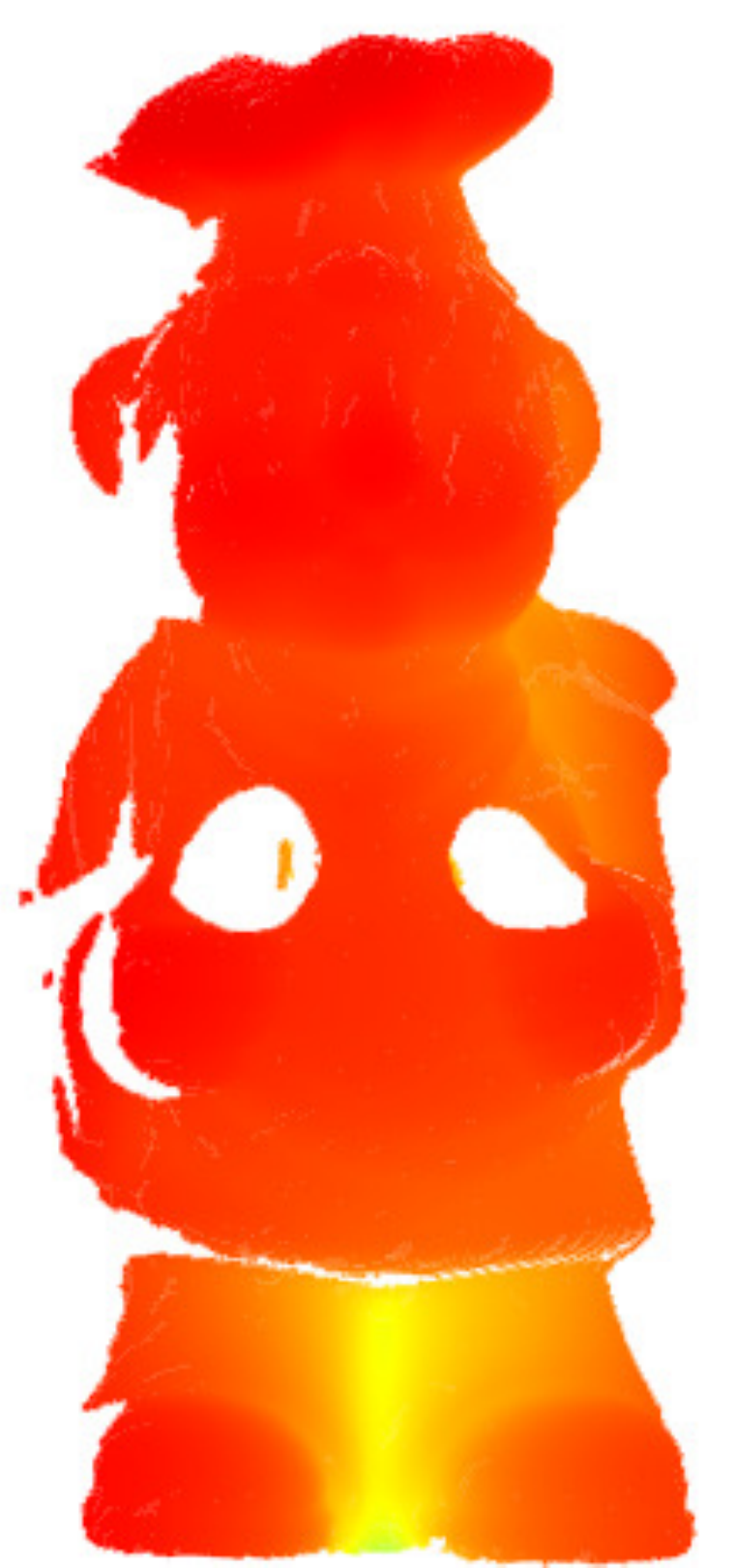}}
  \end{minipage} 
  & \begin{minipage}[b]{0.096\columnwidth}
    \centering
    \raisebox{-.5\height}{\includegraphics[width=\linewidth]{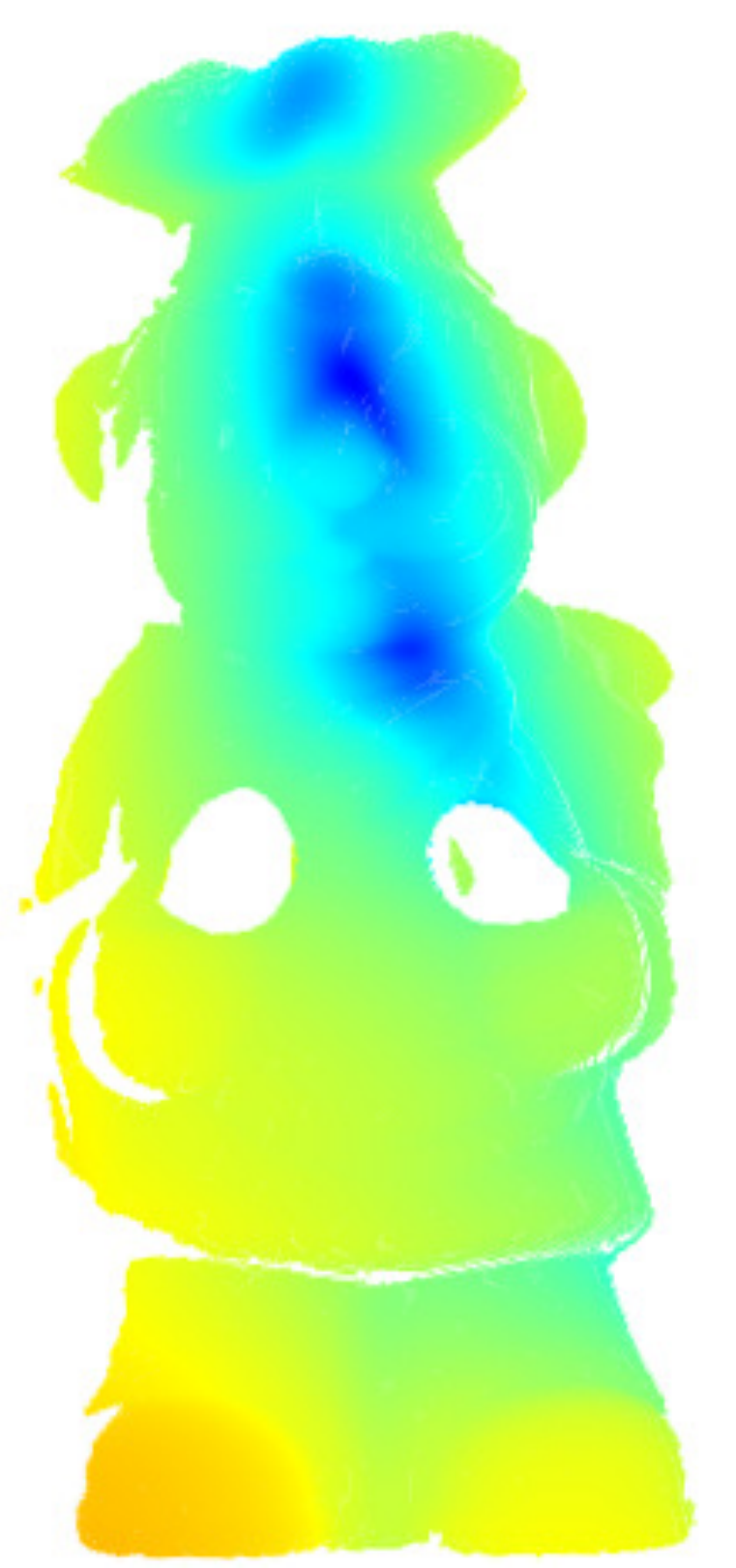}}
  \end{minipage}  
  & \begin{minipage}[b]{0.096\columnwidth}
    \centering
    \raisebox{-.5\height}{\includegraphics[width=\linewidth]{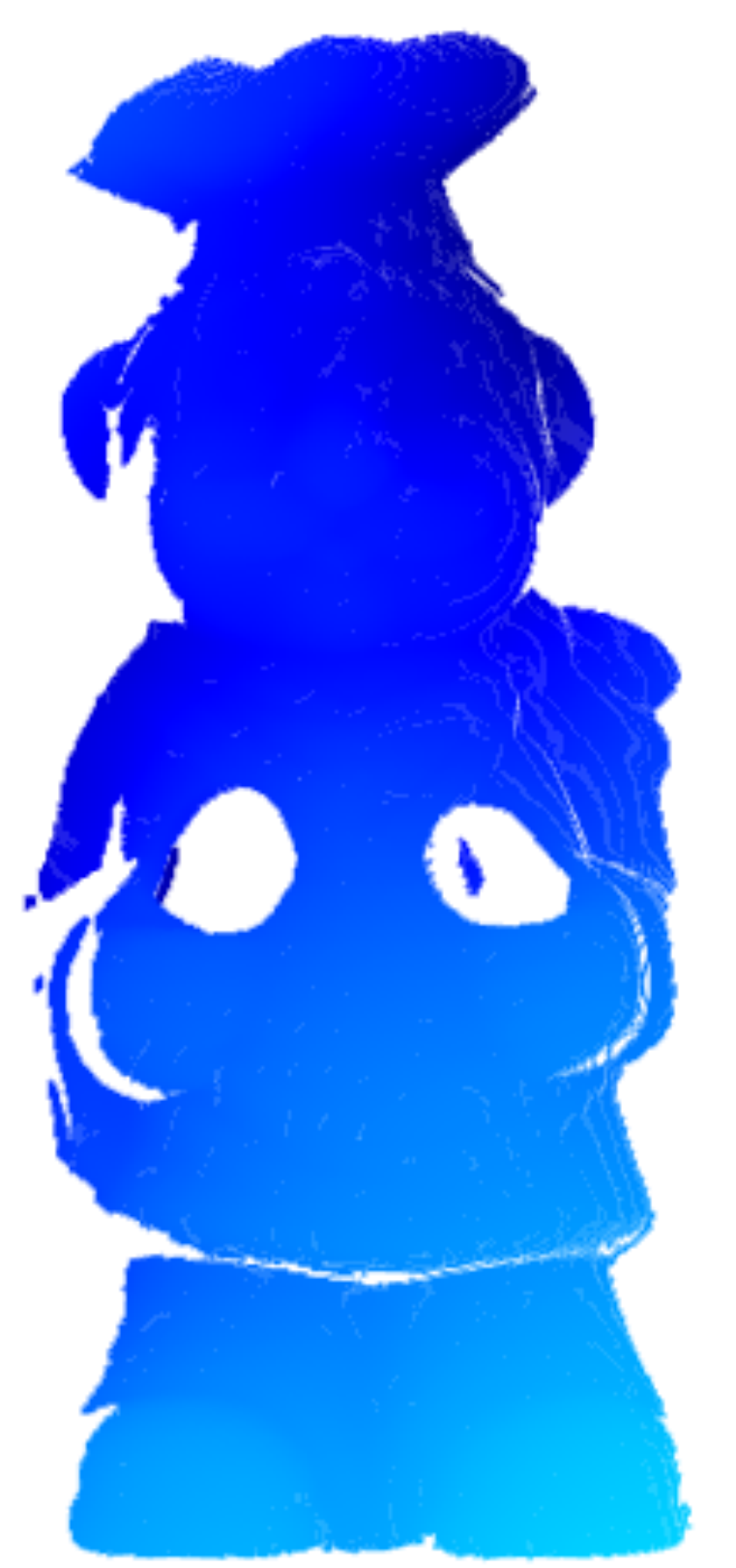}}
  \end{minipage} 
  & \begin{minipage}[b]{0.096\columnwidth}
    \centering
    \raisebox{-.5\height}{\includegraphics[width=\linewidth]{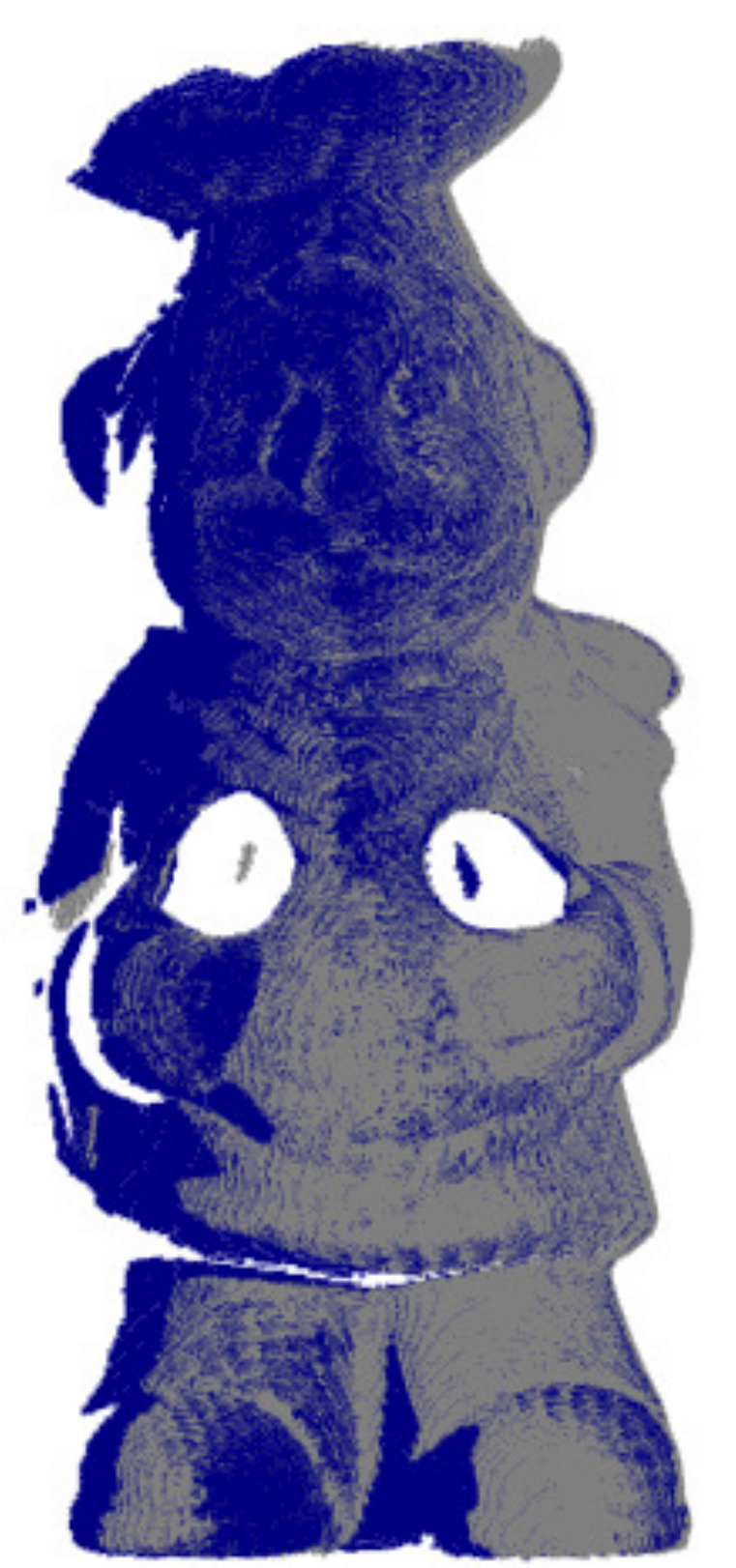}}
  \end{minipage}
  & \begin{minipage}[b]{0.08\columnwidth}
    \centering
    \raisebox{-.5\height}{\includegraphics[width=\linewidth]{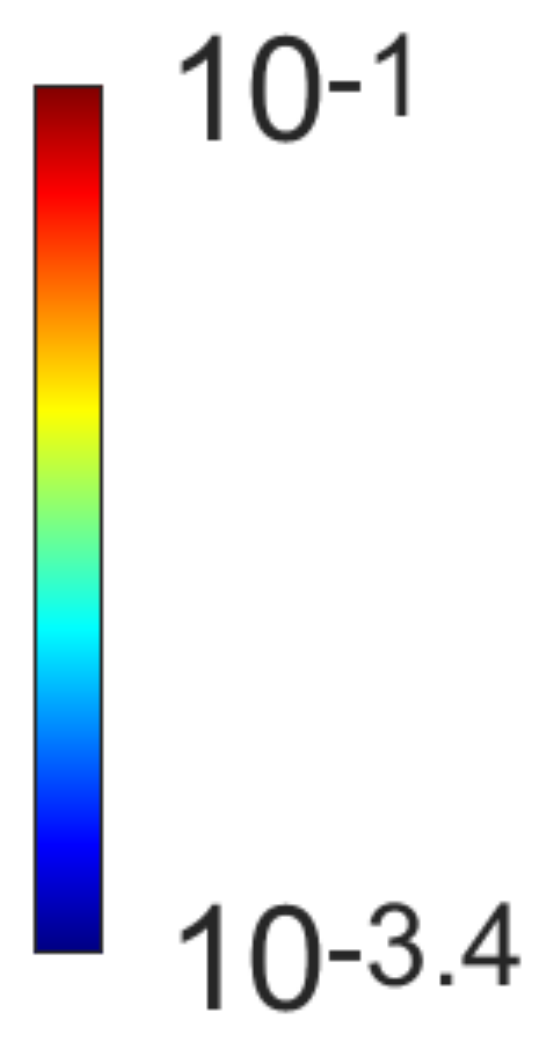}}
  \end{minipage} 
  \\

  \begin{minipage}[b]{0.152\columnwidth}
    \centering
    \raisebox{-.5\height}{\includegraphics[width=\linewidth]{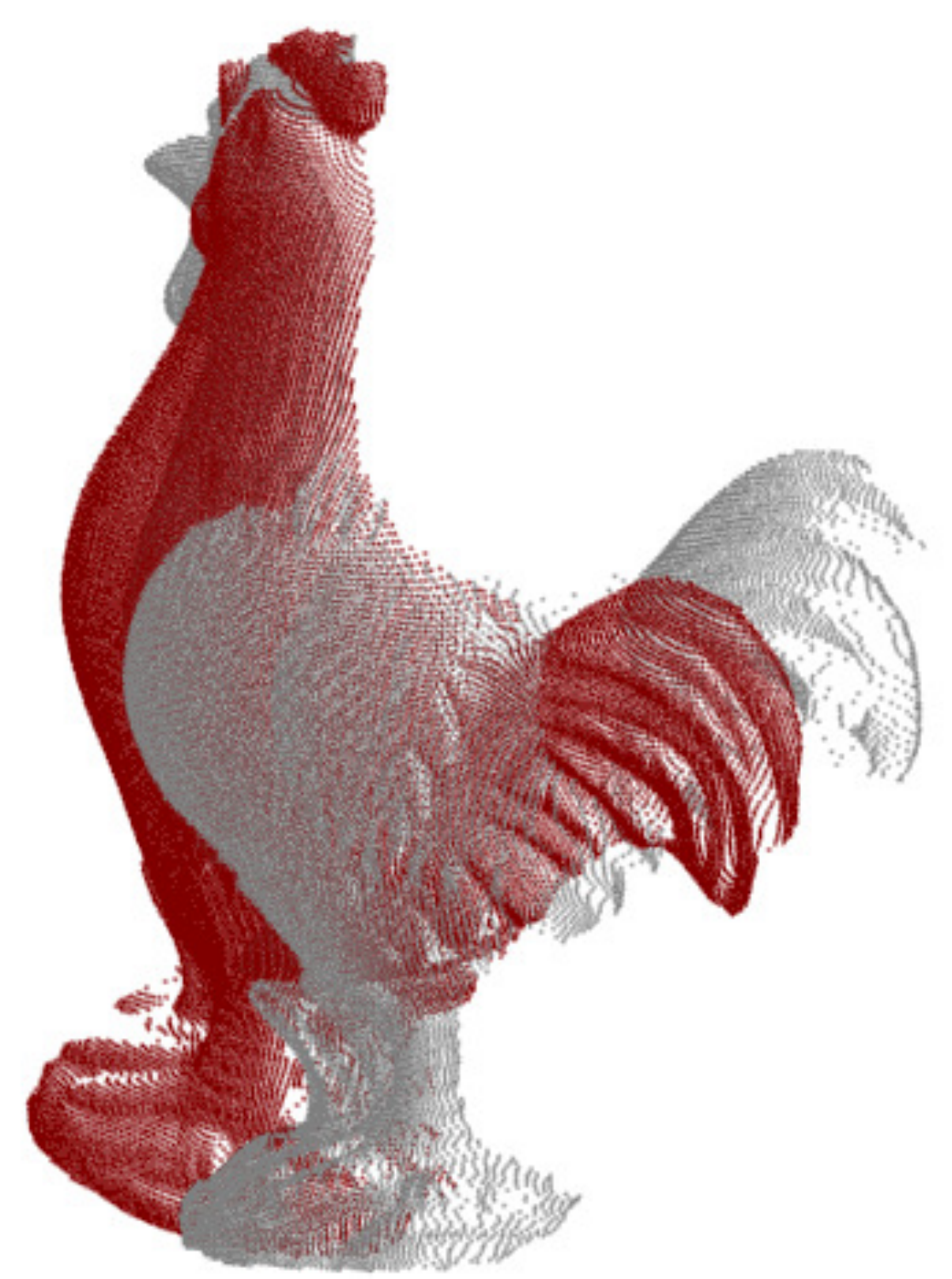}}
  \end{minipage} 
  & \begin{minipage}[b]{0.152\columnwidth}
    \centering
    \raisebox{-.5\height}{\includegraphics[width=\linewidth]{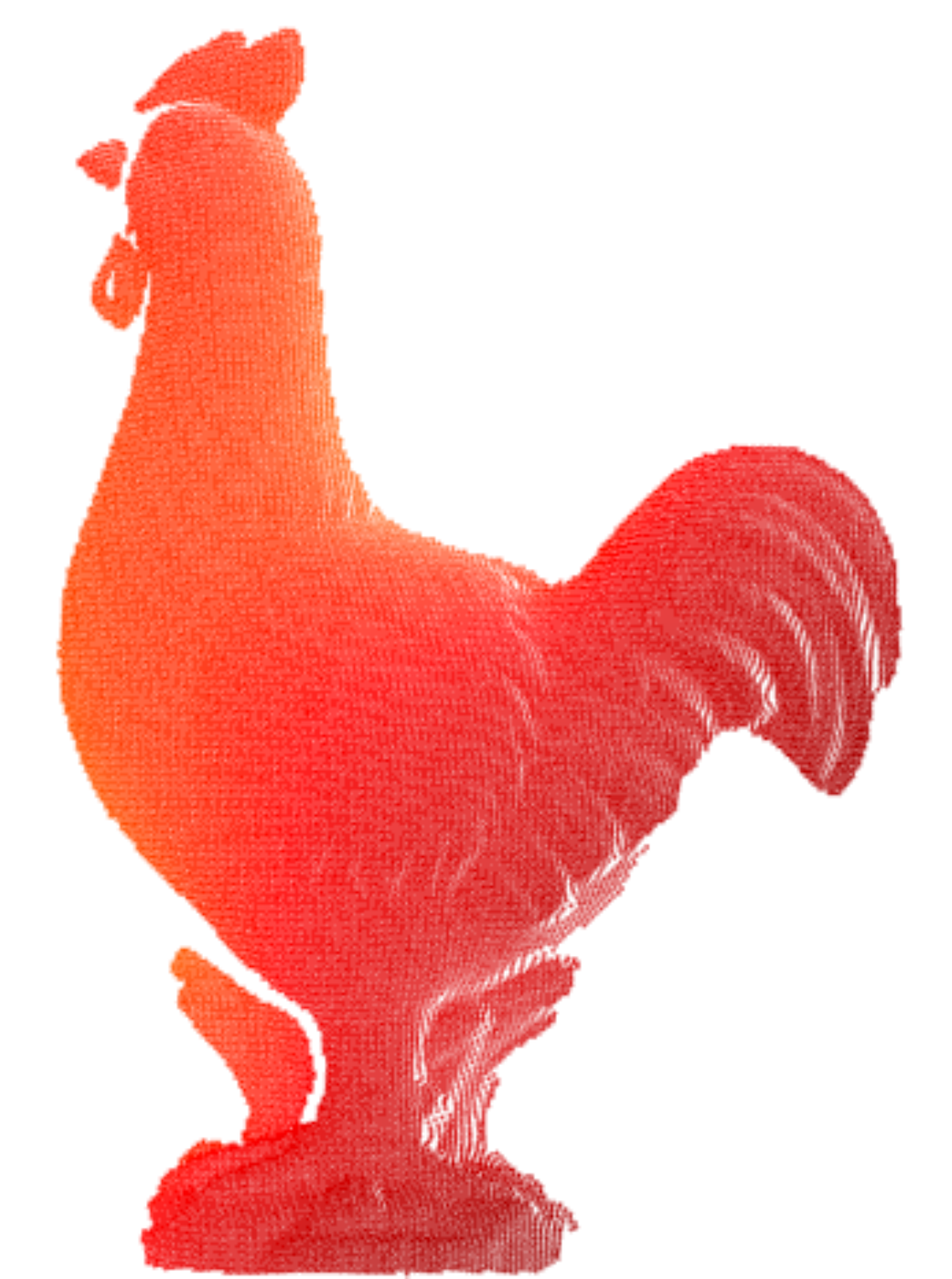}}
  \end{minipage}
  & \begin{minipage}[b]{0.152\columnwidth}
    \centering
    \raisebox{-.5\height}{\includegraphics[width=\linewidth]{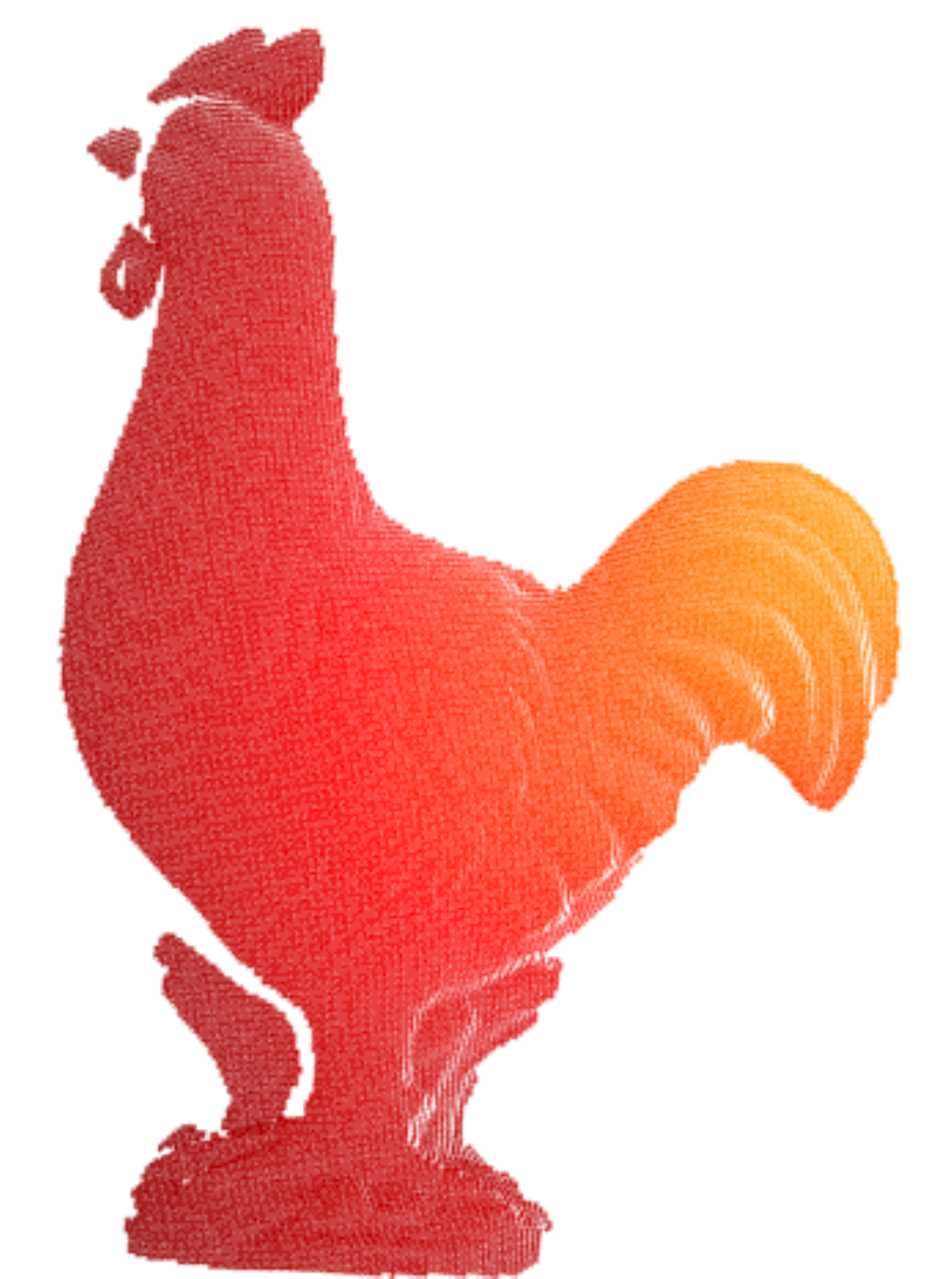}}
  \end{minipage} 
  & \begin{minipage}[b]{0.152\columnwidth}
    \centering
    \raisebox{-.5\height}{\includegraphics[width=\linewidth]{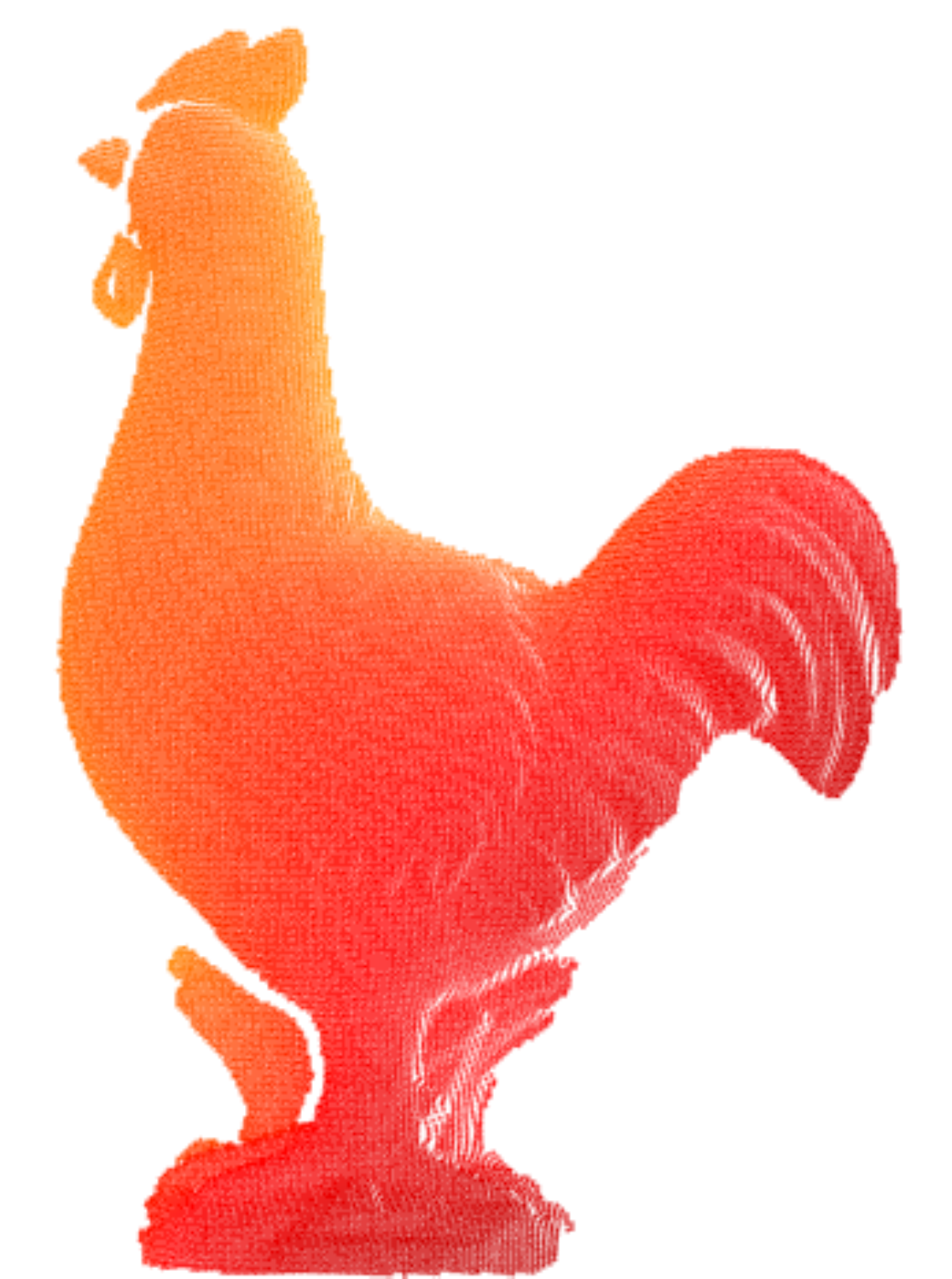}}
  \end{minipage} 
  & \begin{minipage}[b]{0.152\columnwidth}
    \centering
    \raisebox{-.5\height}{\includegraphics[width=\linewidth]{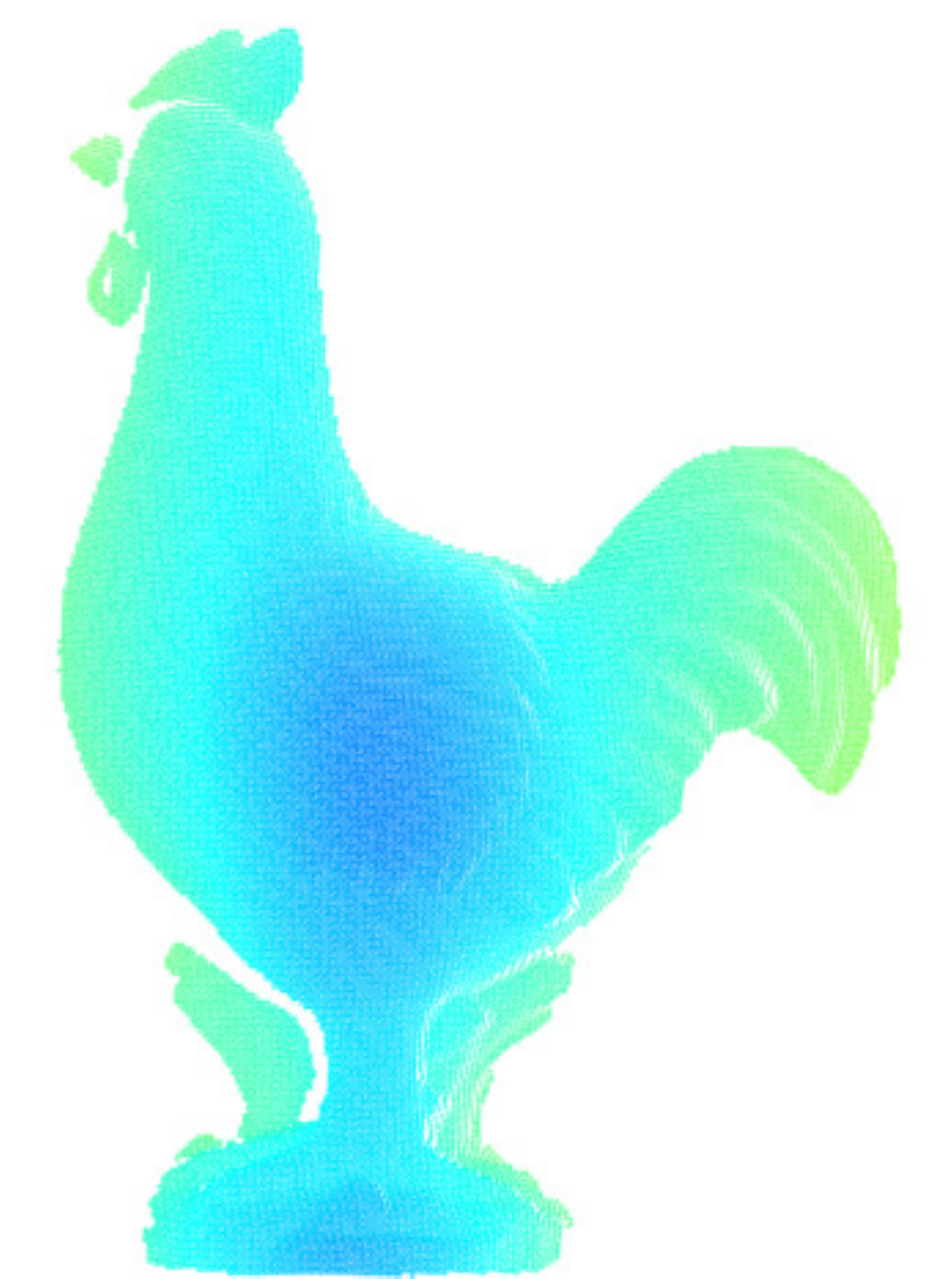}}
  \end{minipage}  
  & \begin{minipage}[b]{0.152\columnwidth}
    \centering
    \raisebox{-.5\height}{\includegraphics[width=\linewidth]{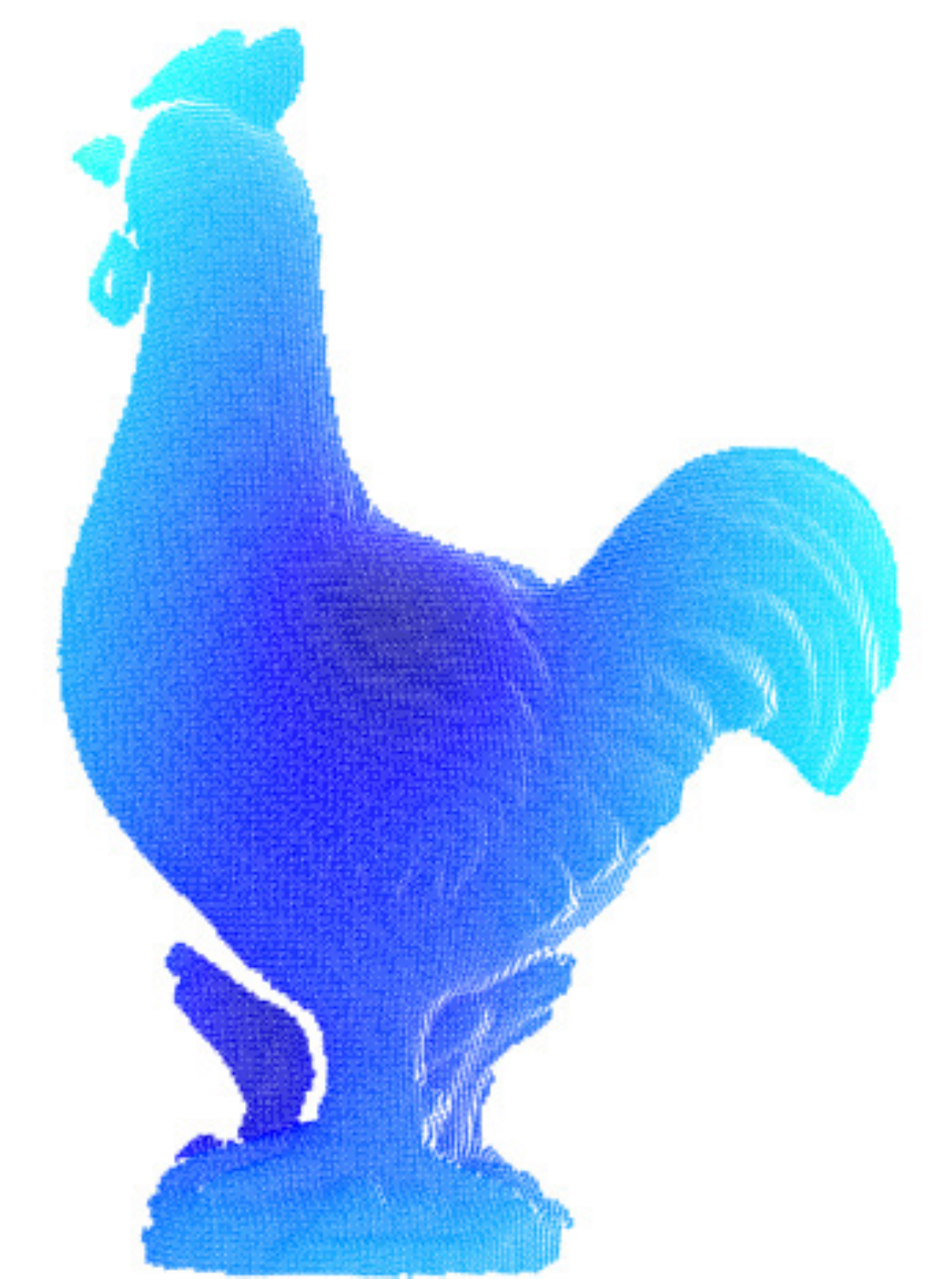}}
  \end{minipage} 
  & \begin{minipage}[b]{0.152\columnwidth}
    \centering
    \raisebox{-.5\height}{\includegraphics[width=\linewidth]{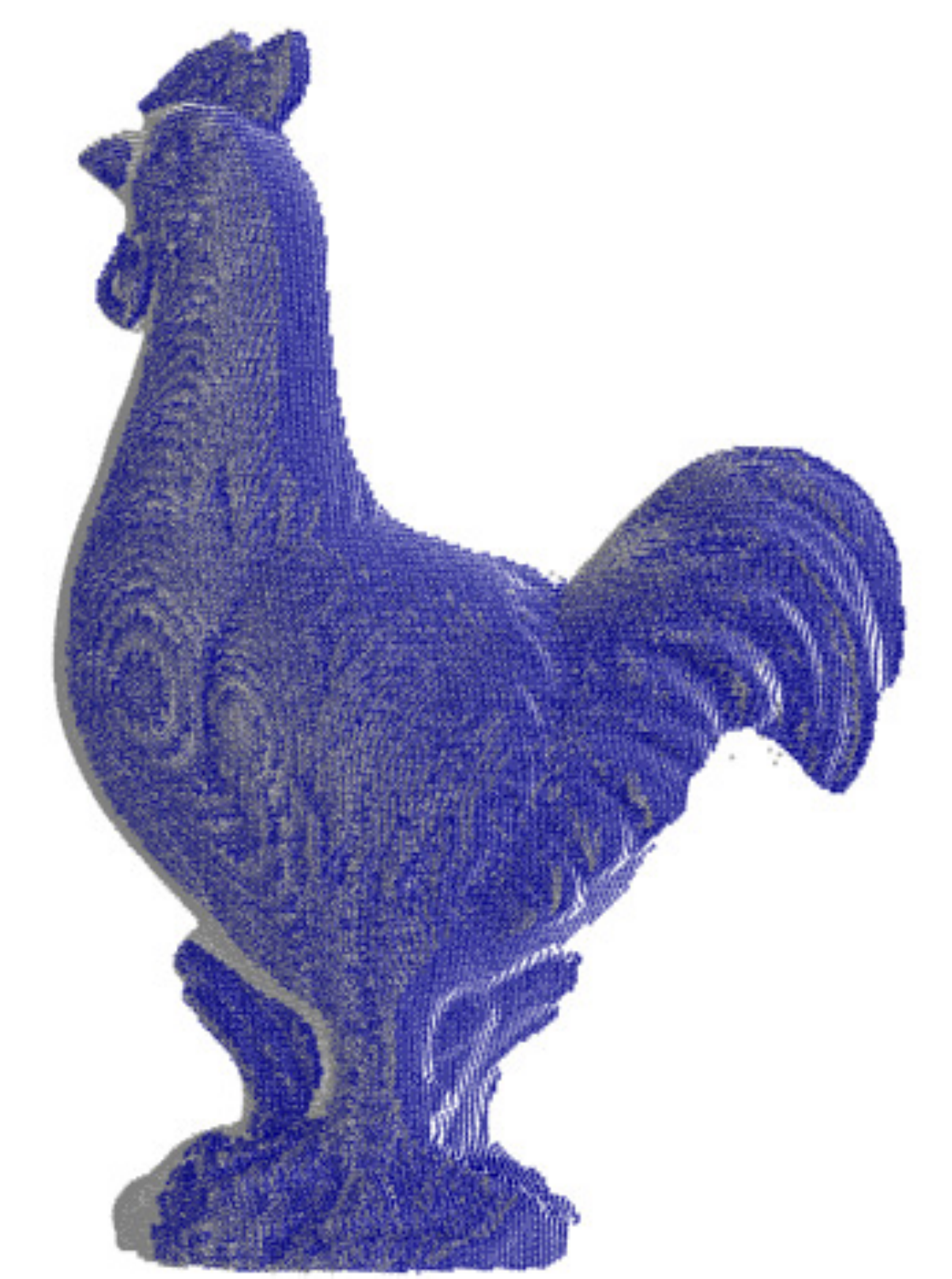}}
  \end{minipage}
  & \begin{minipage}[b]{0.08\columnwidth}
    \centering
    \raisebox{-.5\height}{\includegraphics[width=\linewidth]{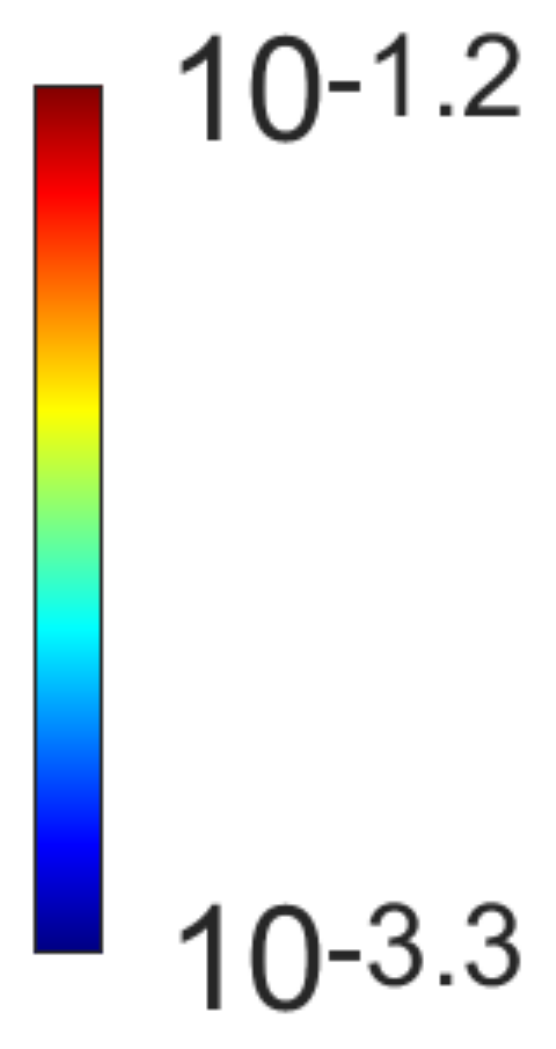}}
  \end{minipage} 
  \\

  \begin{minipage}[b]{0.2\columnwidth}
    \centering
    \raisebox{-.5\height}{\includegraphics[width=\linewidth]{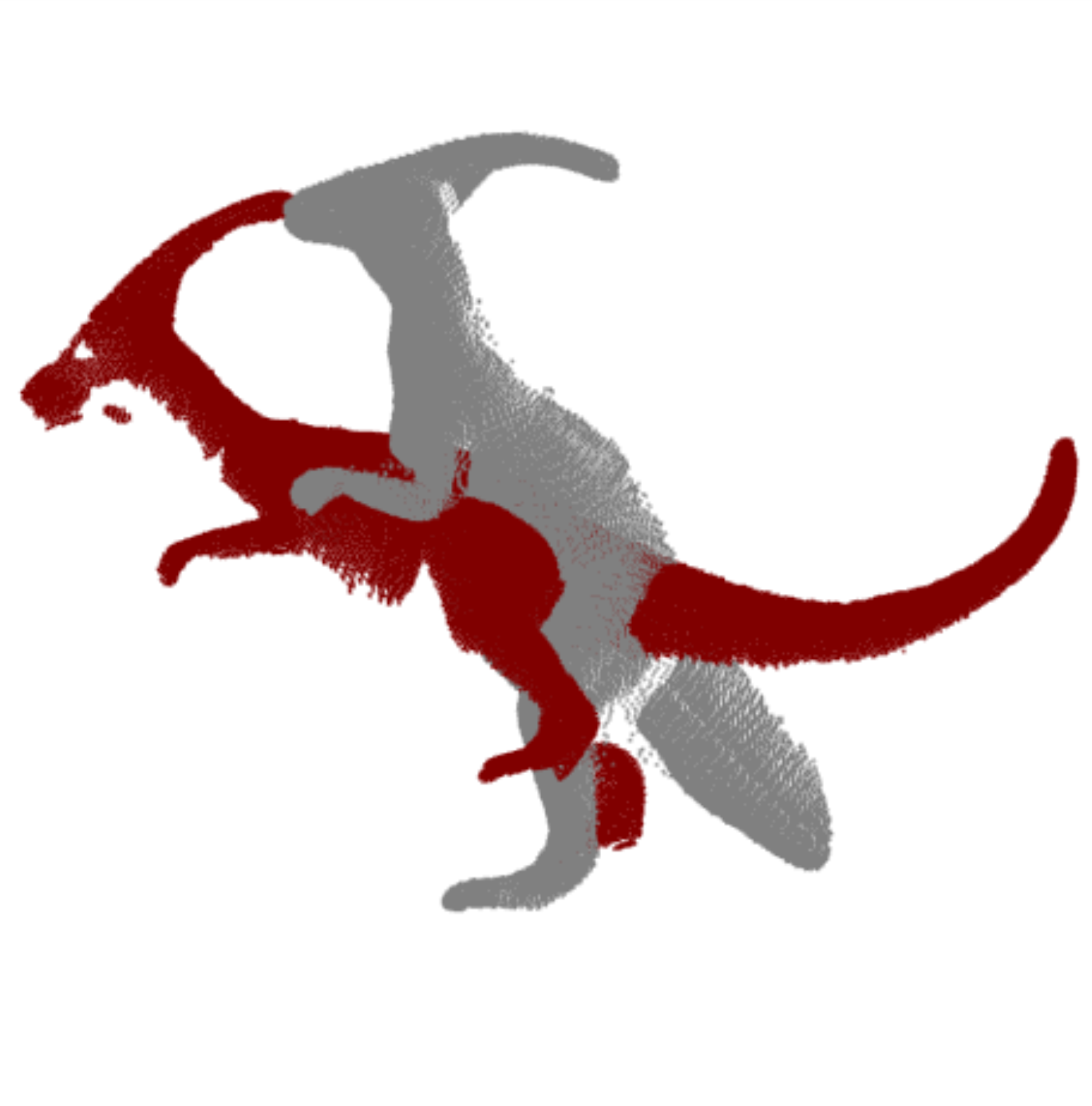}}
  \end{minipage} 
  & \begin{minipage}[b]{0.2\columnwidth}
    \centering
    \raisebox{-.5\height}{\includegraphics[width=\linewidth]{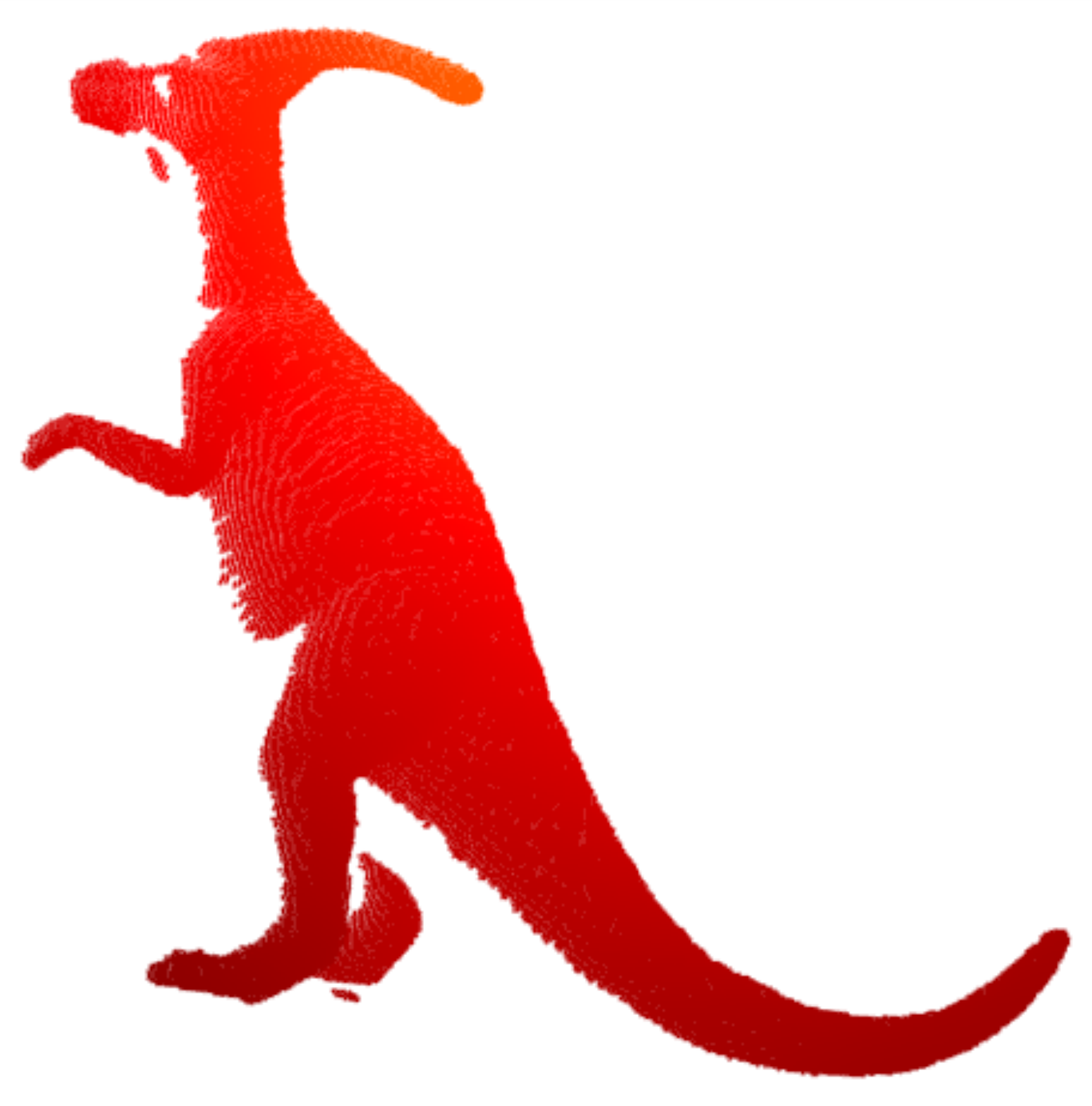}}
  \end{minipage}
  & \begin{minipage}[b]{0.2\columnwidth}
    \centering
    \raisebox{-.5\height}{\includegraphics[width=\linewidth]{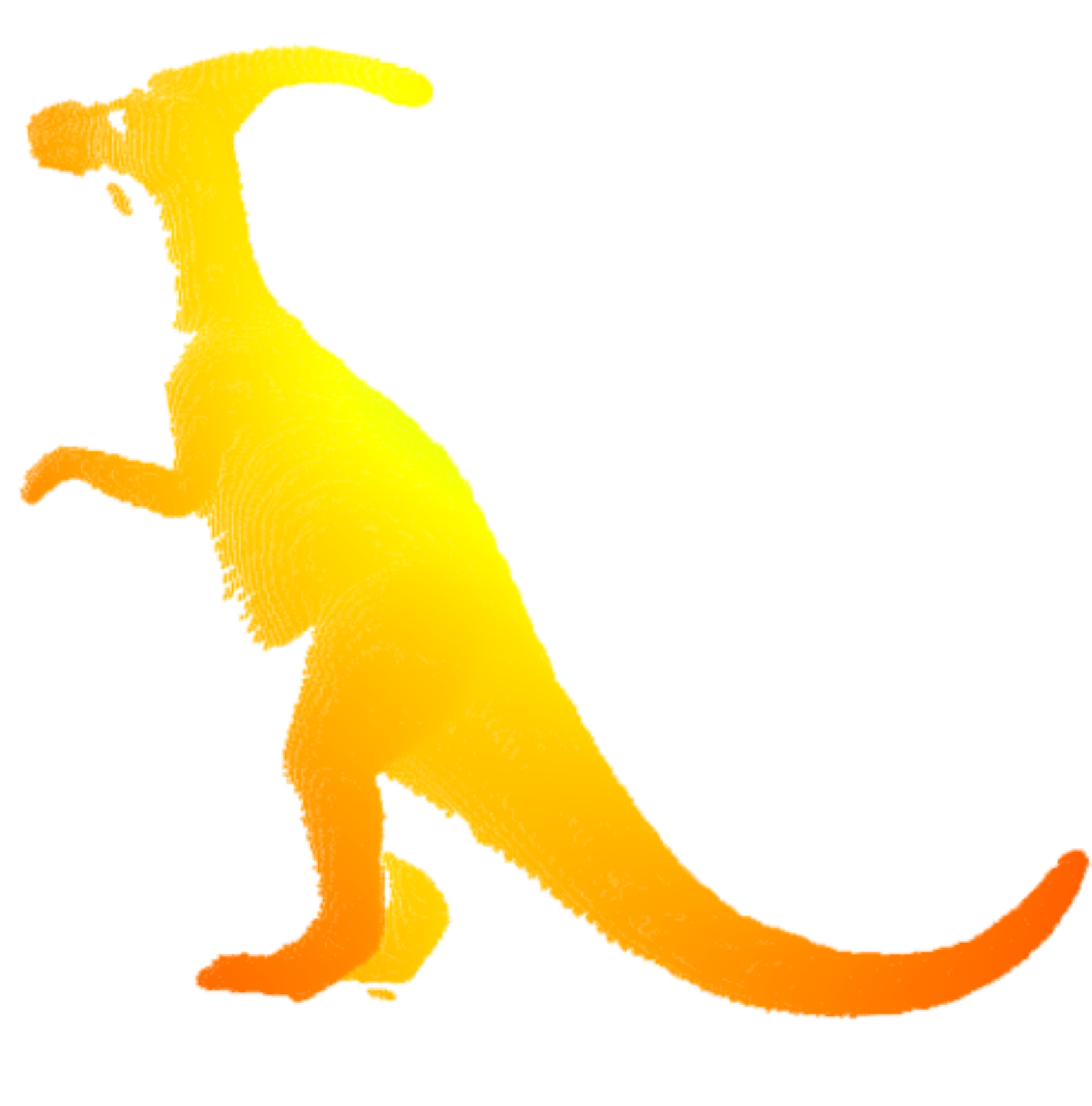}}
  \end{minipage} 
  & \begin{minipage}[b]{0.2\columnwidth}
    \centering
    \raisebox{-.5\height}{\includegraphics[width=\linewidth]{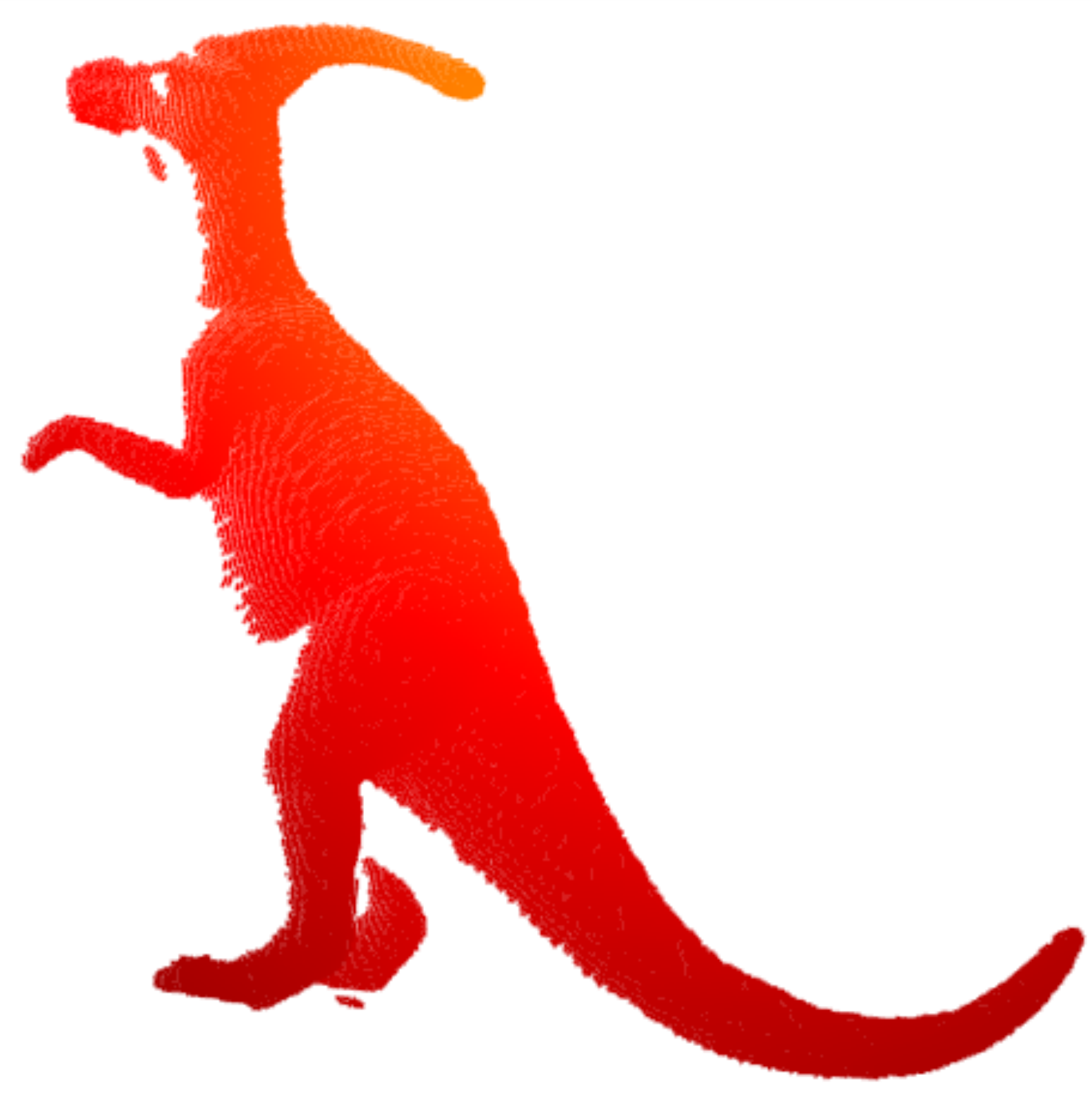}}
  \end{minipage} 
  & \begin{minipage}[b]{0.2\columnwidth}
    \centering
    \raisebox{-.5\height}{\includegraphics[width=\linewidth]{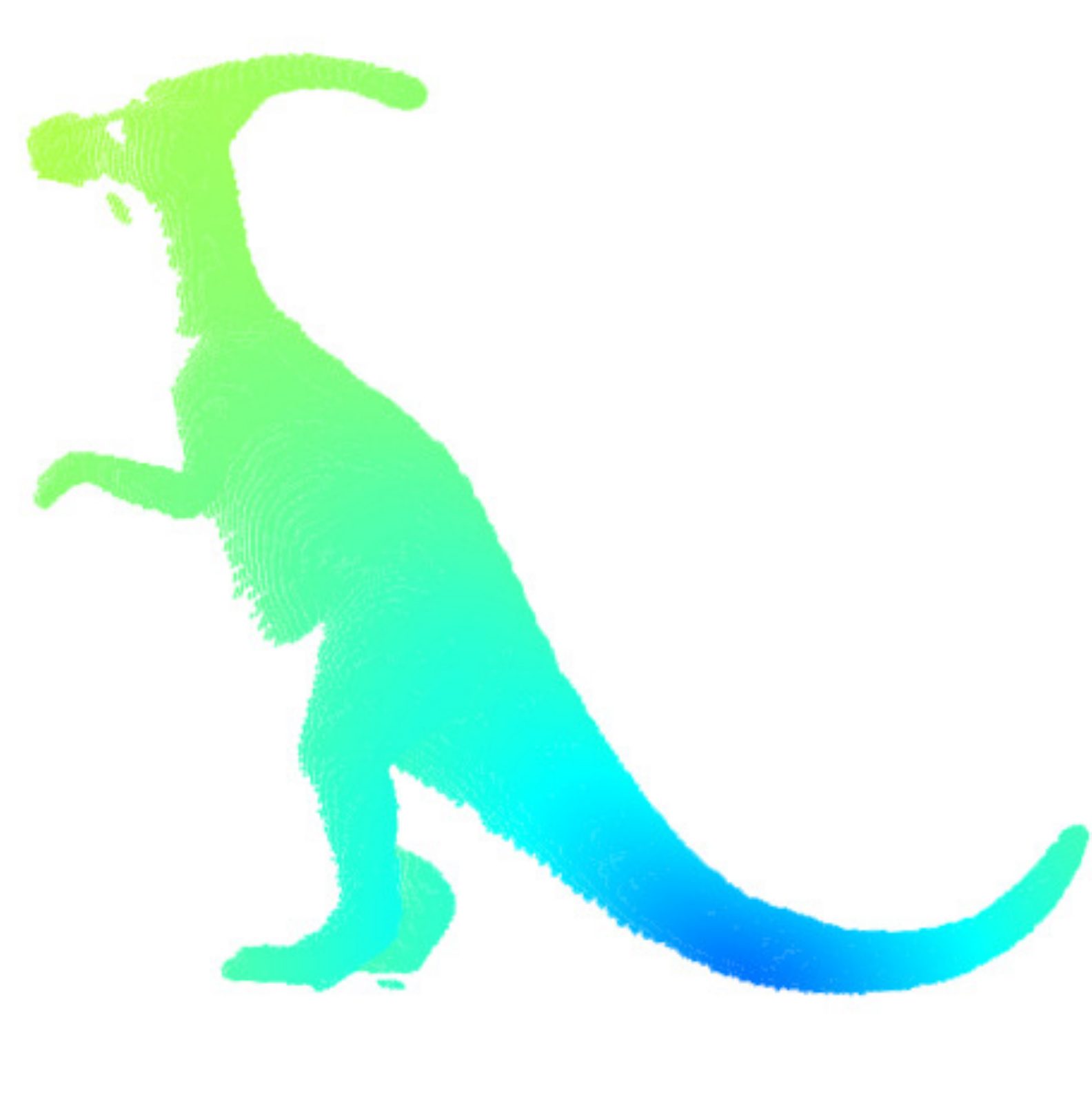}}
  \end{minipage}  
  & \begin{minipage}[b]{0.2\columnwidth}
    \centering
    \raisebox{-.5\height}{\includegraphics[width=\linewidth]{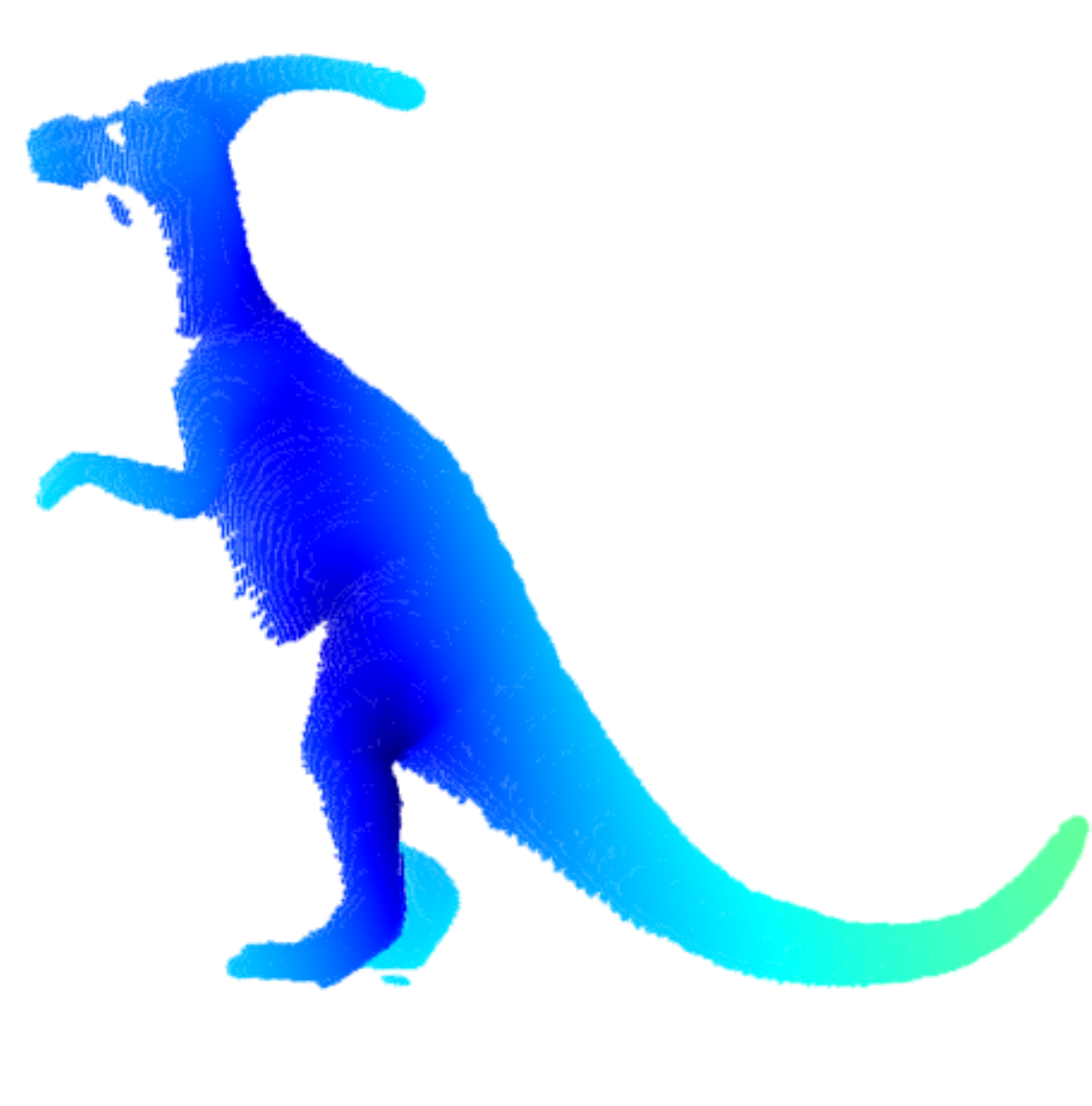}}
  \end{minipage} 
  & \begin{minipage}[b]{0.2\columnwidth}
    \centering
    \raisebox{-.5\height}{\includegraphics[width=\linewidth]{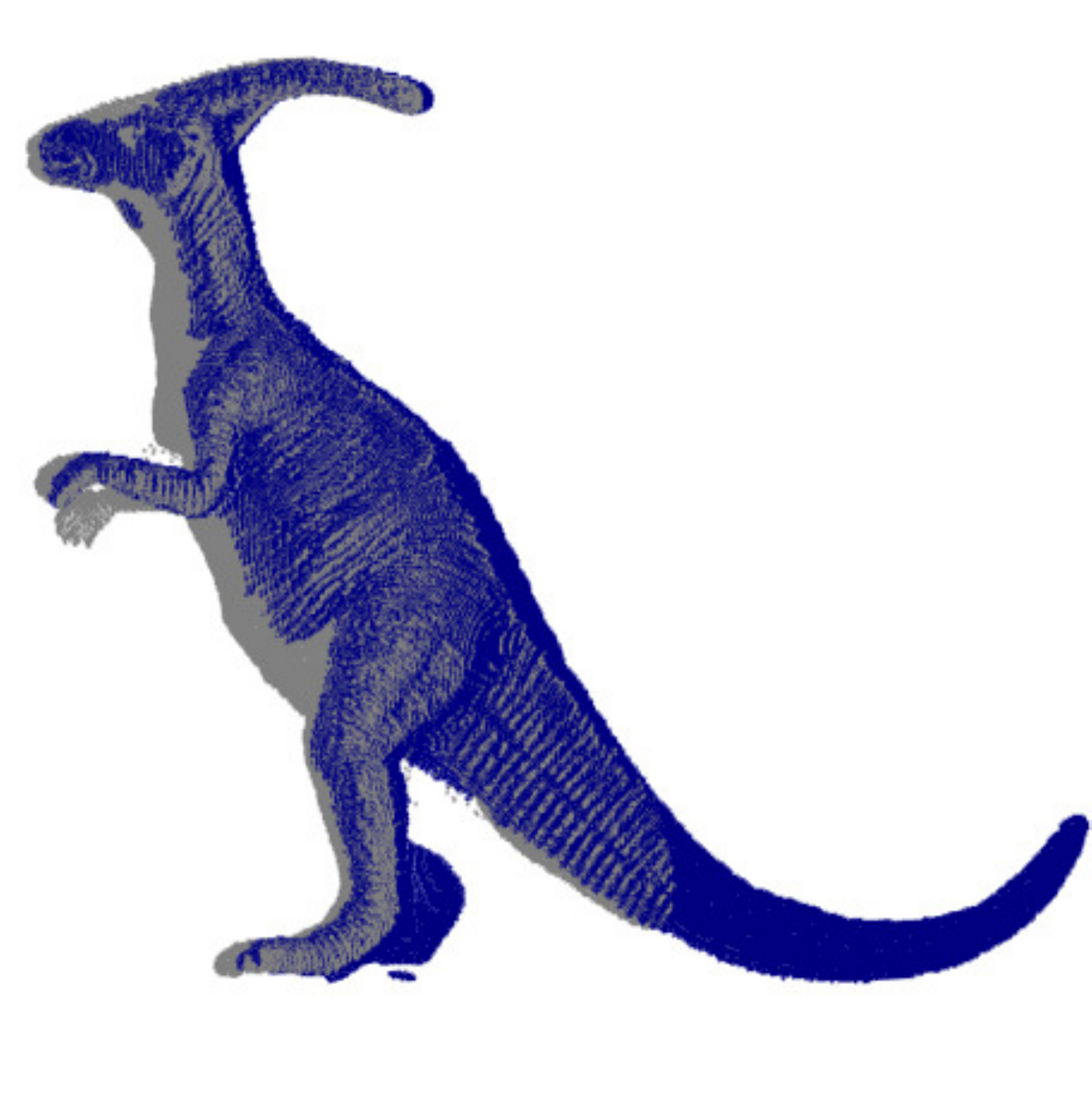}}
  \end{minipage}
  & \begin{minipage}[b]{0.08\columnwidth}
    \centering
    \raisebox{-.5\height}{\includegraphics[width=\linewidth]{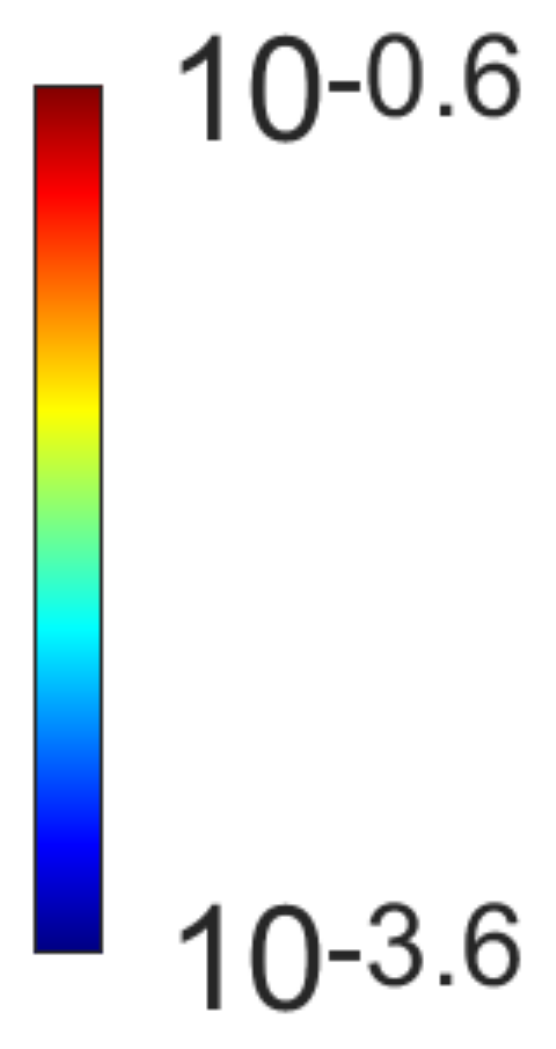}}
  \end{minipage} 
  \\

  \begin{minipage}[b]{0.2\columnwidth}
    \centering
    \raisebox{-.5\height}{\includegraphics[width=\linewidth]{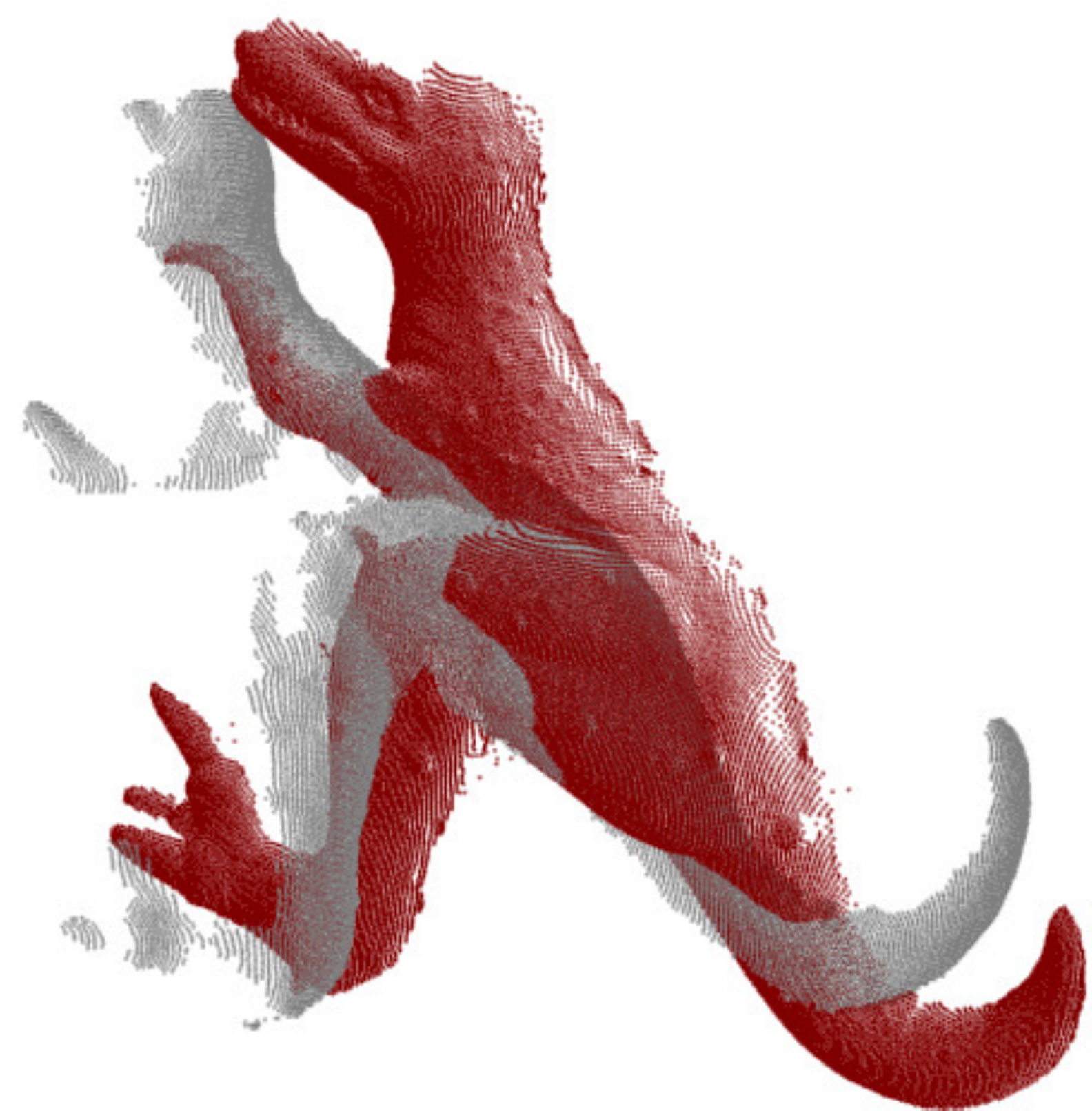}}
  \end{minipage} 
  & \begin{minipage}[b]{0.2\columnwidth}
    \centering
    \raisebox{-.5\height}{\includegraphics[width=\linewidth]{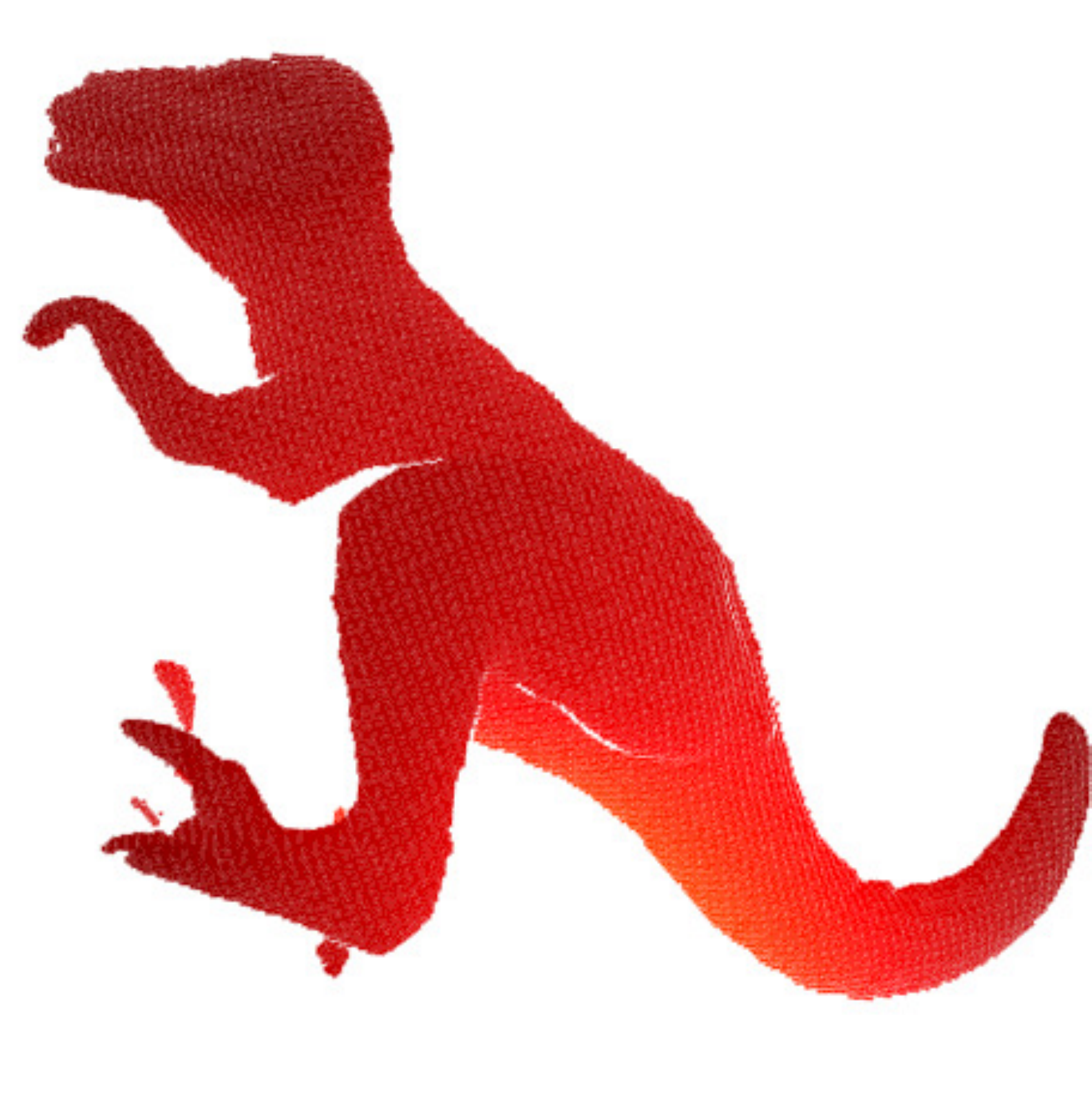}}
  \end{minipage}
  & \begin{minipage}[b]{0.2\columnwidth}
    \centering
    \raisebox{-.5\height}{\includegraphics[width=\linewidth]{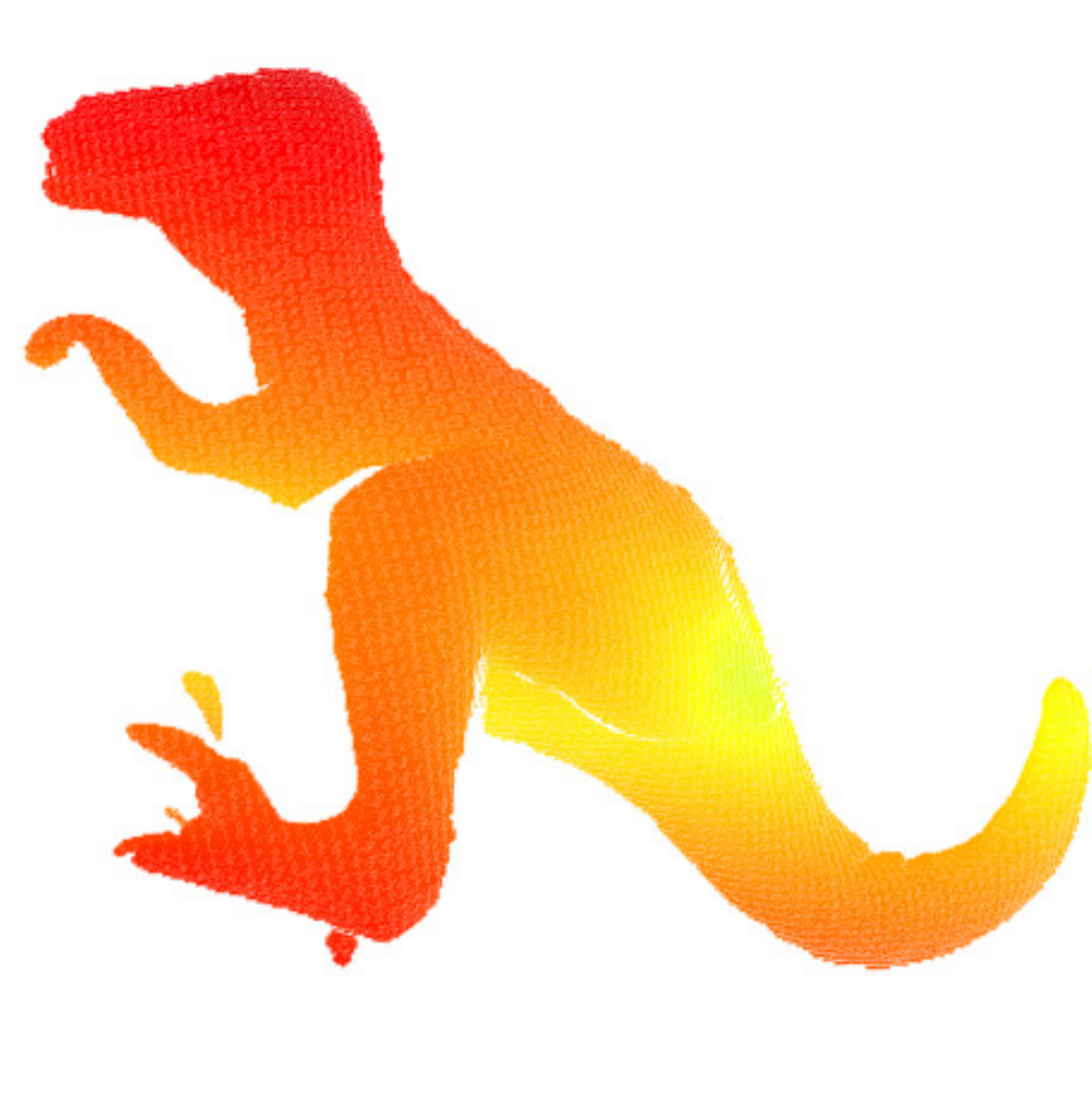}}
  \end{minipage} 
  & \begin{minipage}[b]{0.2\columnwidth}
    \centering
    \raisebox{-.5\height}{\includegraphics[width=\linewidth]{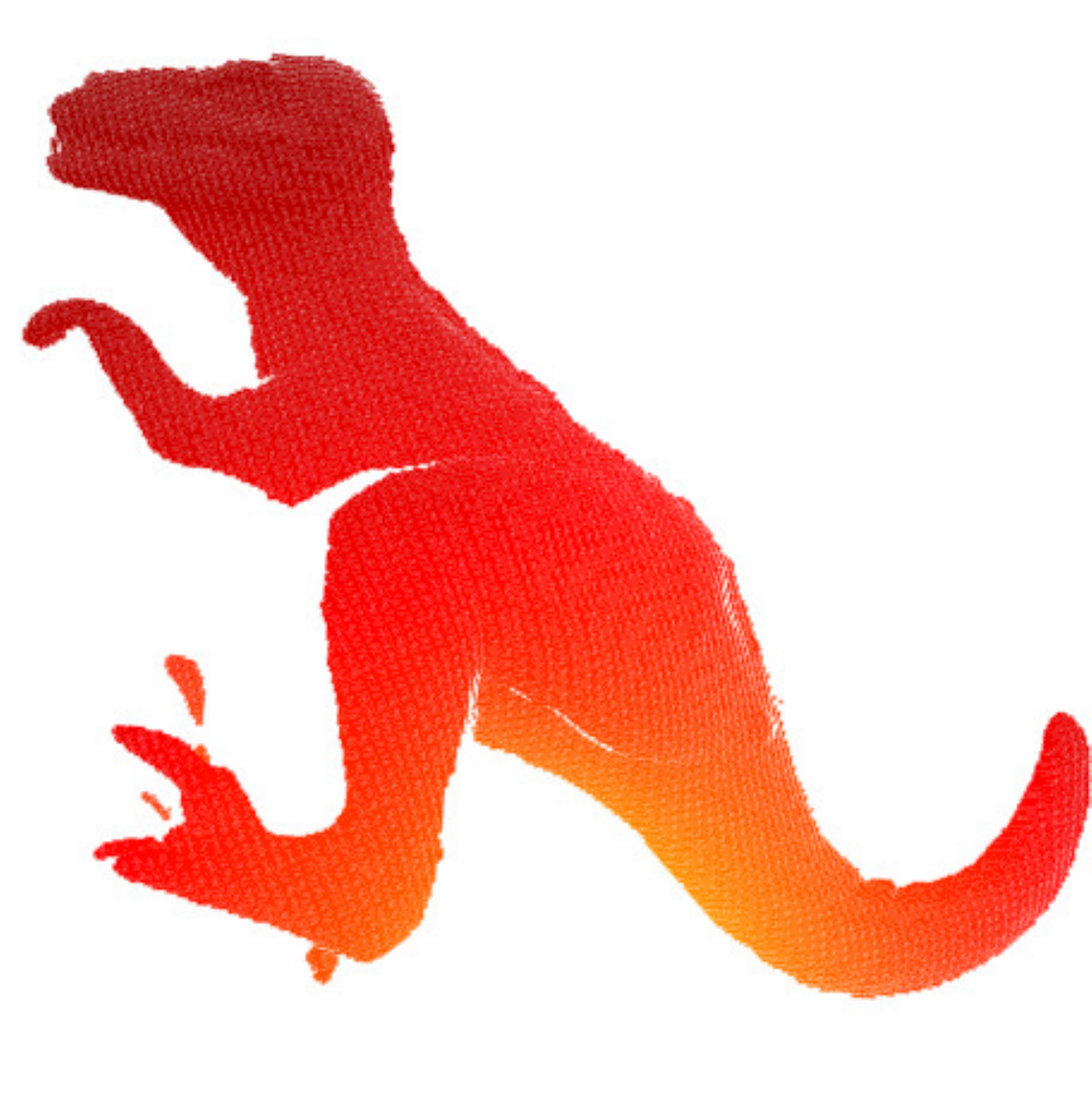}}
  \end{minipage} 
  & \begin{minipage}[b]{0.2\columnwidth}
    \centering
    \raisebox{-.5\height}{\includegraphics[width=\linewidth]{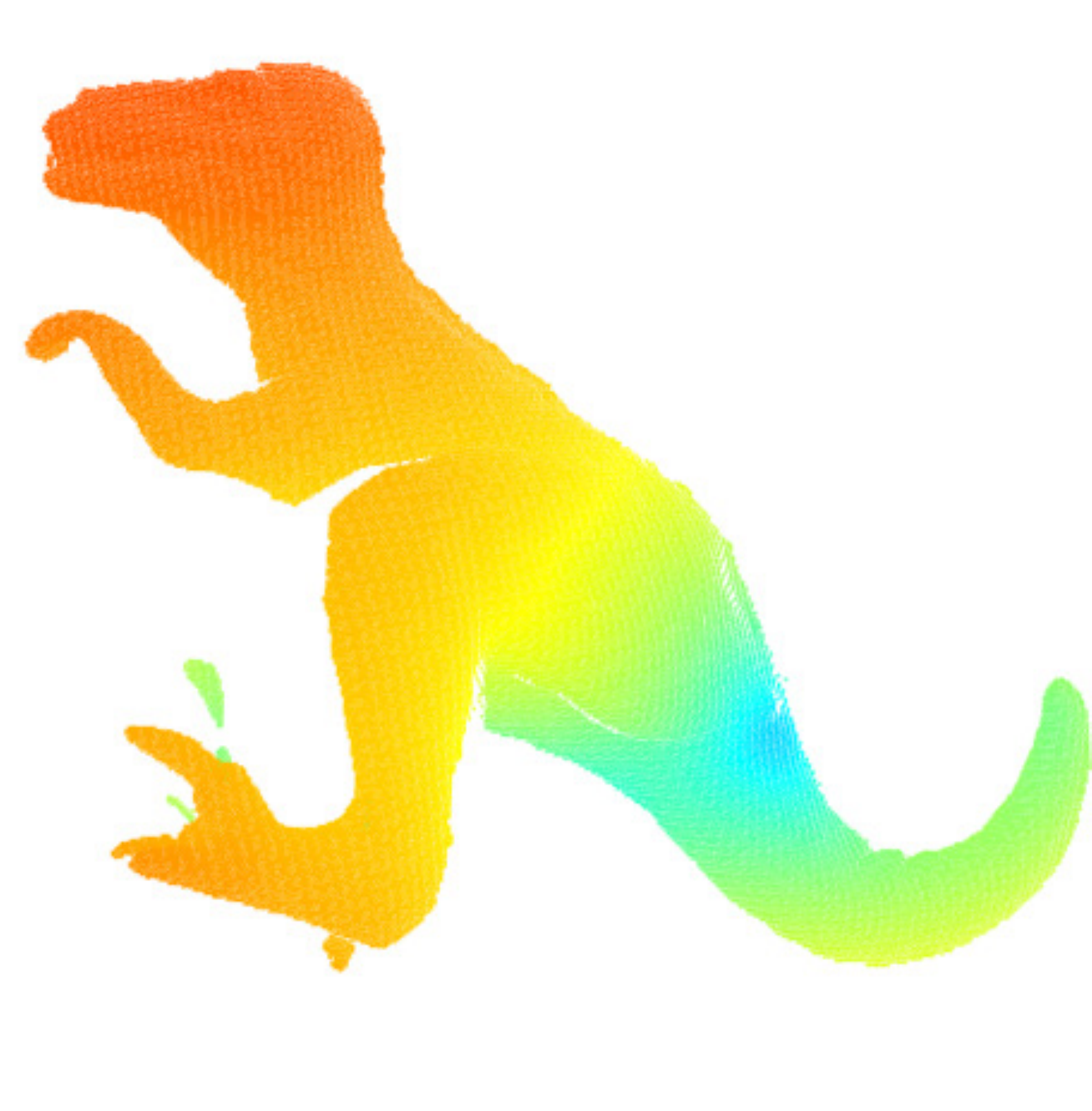}}
  \end{minipage}  
  & \begin{minipage}[b]{0.2\columnwidth}
    \centering
    \raisebox{-.5\height}{\includegraphics[width=\linewidth]{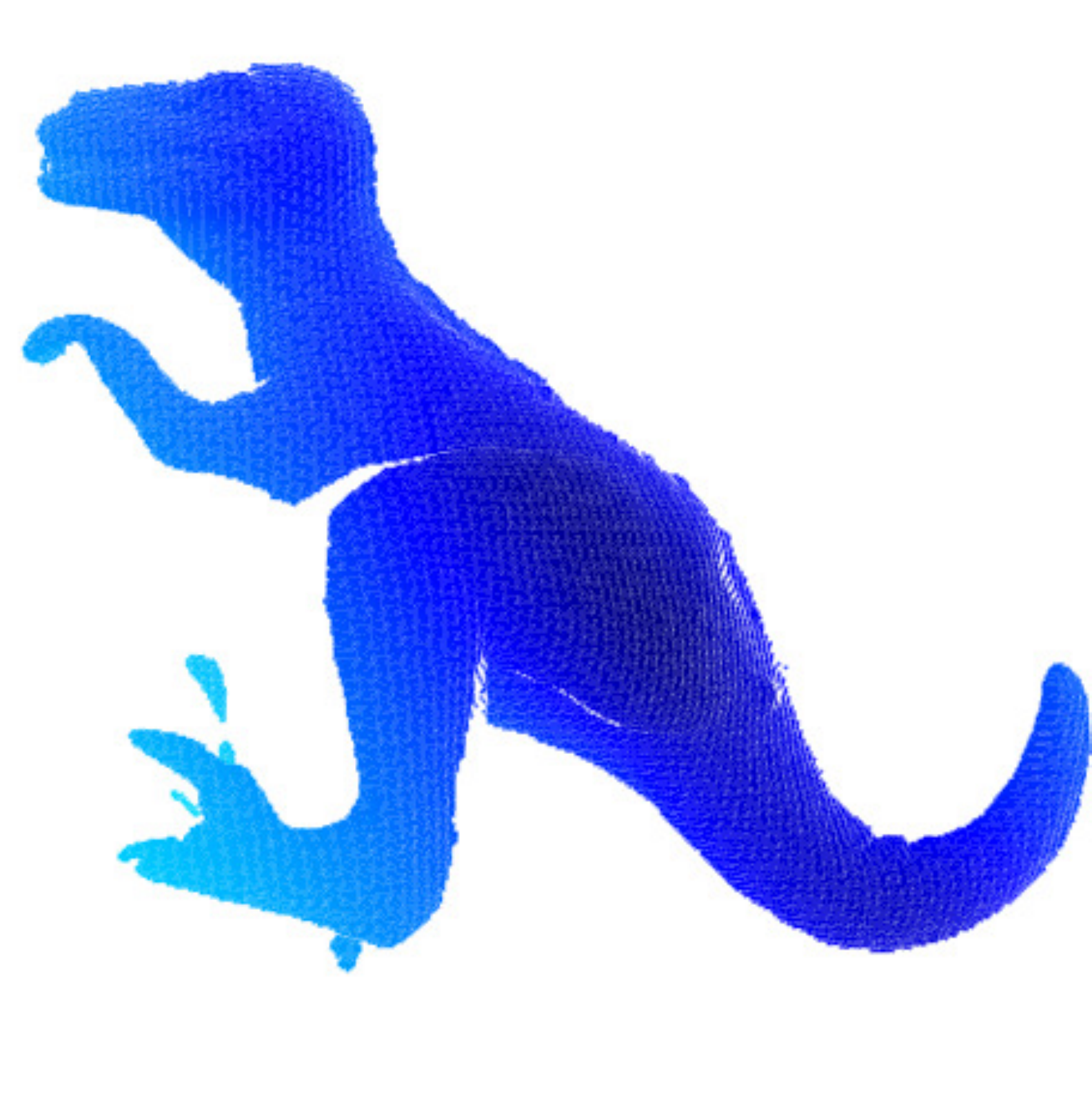}}
  \end{minipage} 
  & \begin{minipage}[b]{0.2\columnwidth}
    \centering
    \raisebox{-.5\height}{\includegraphics[width=\linewidth]{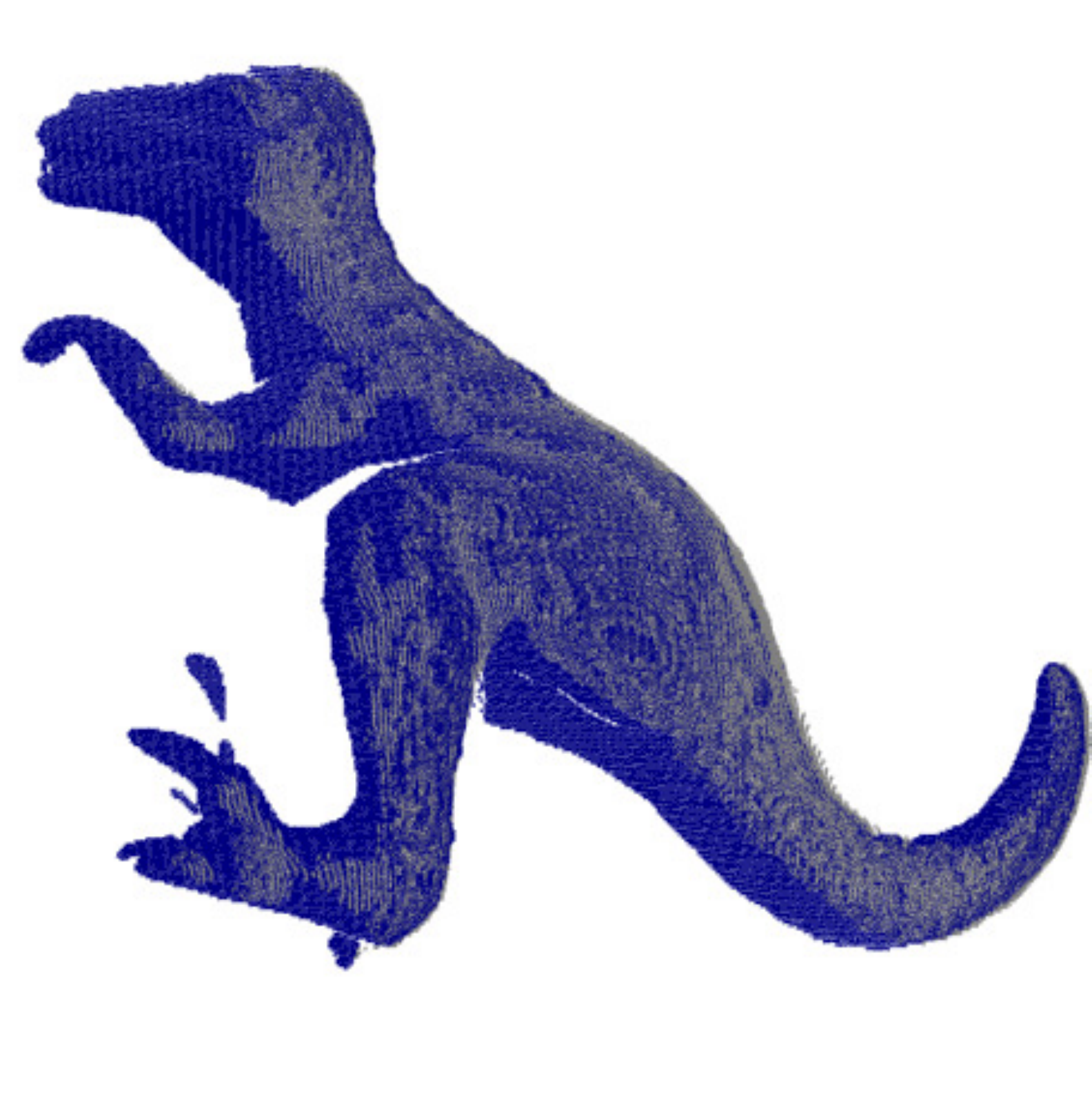}}
  \end{minipage}
  & \begin{minipage}[b]{0.08\columnwidth}
    \centering
    \raisebox{-.5\height}{\includegraphics[width=\linewidth]{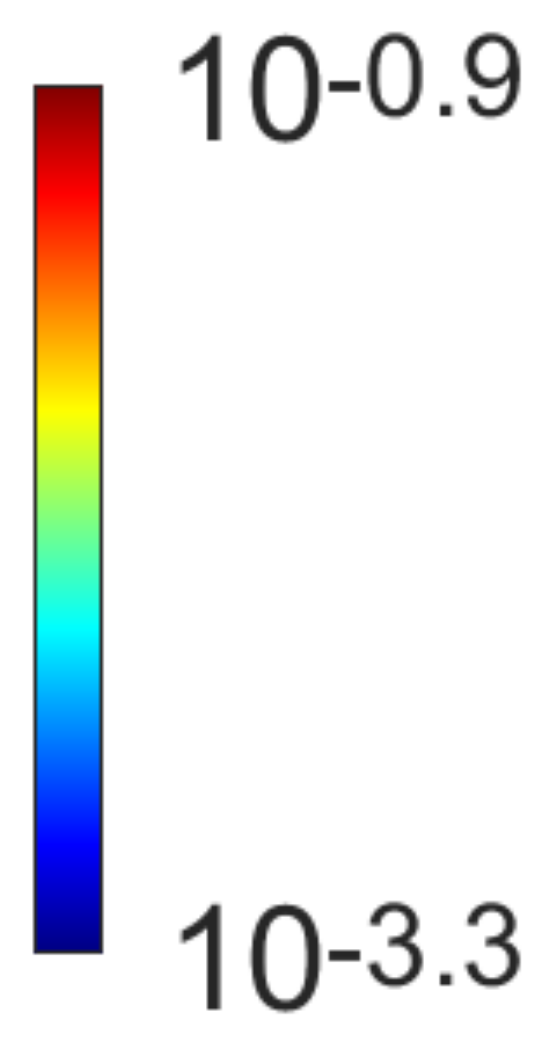}}
  \end{minipage} 
  \\

  \begin{minipage}[b]{0.144\columnwidth}
    \centering
    \raisebox{-.5\height}{\includegraphics[width=\linewidth]{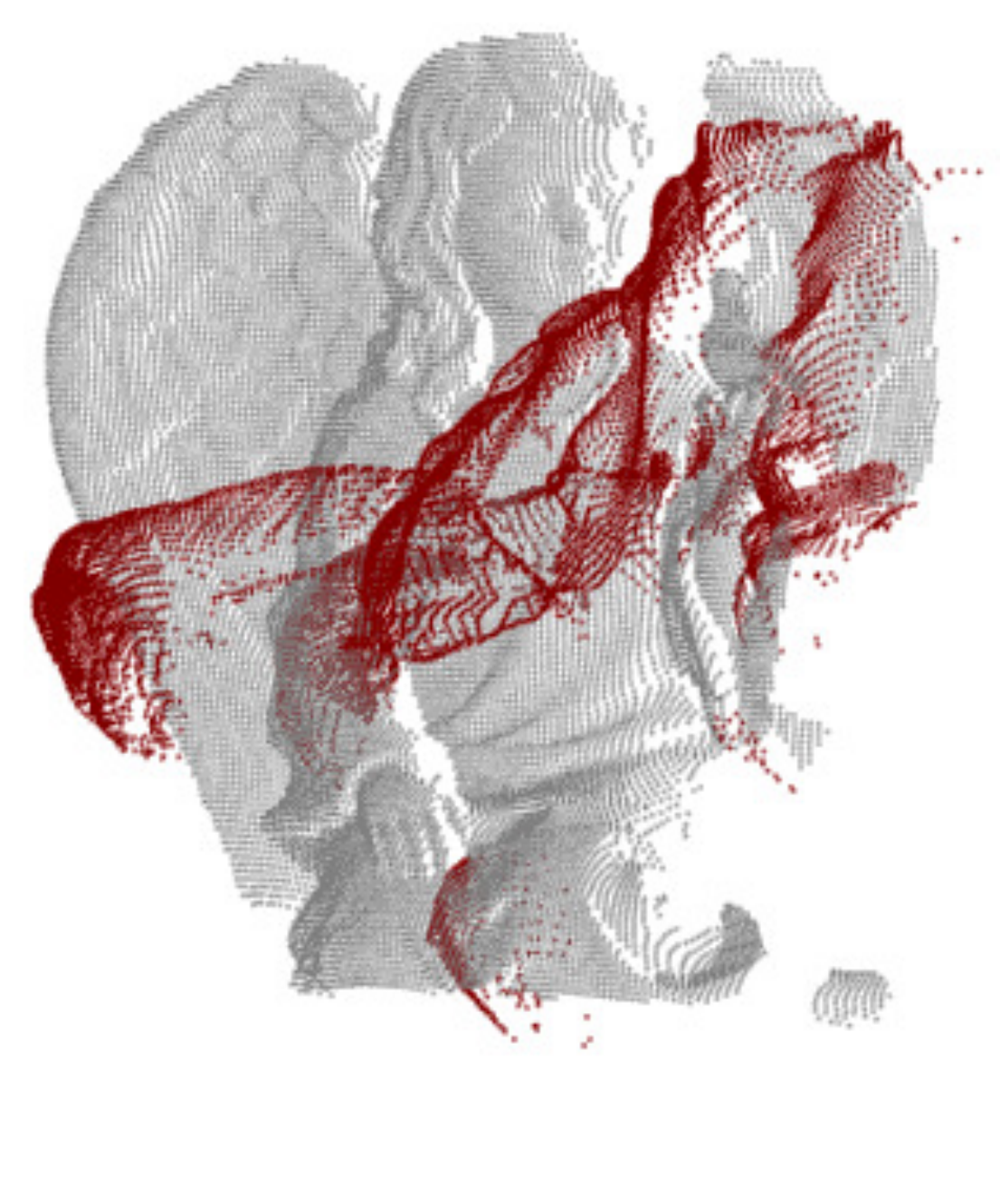}}
  \end{minipage} 
  & \begin{minipage}[b]{0.144\columnwidth}
    \centering
    \raisebox{-.5\height}{\includegraphics[width=\linewidth]{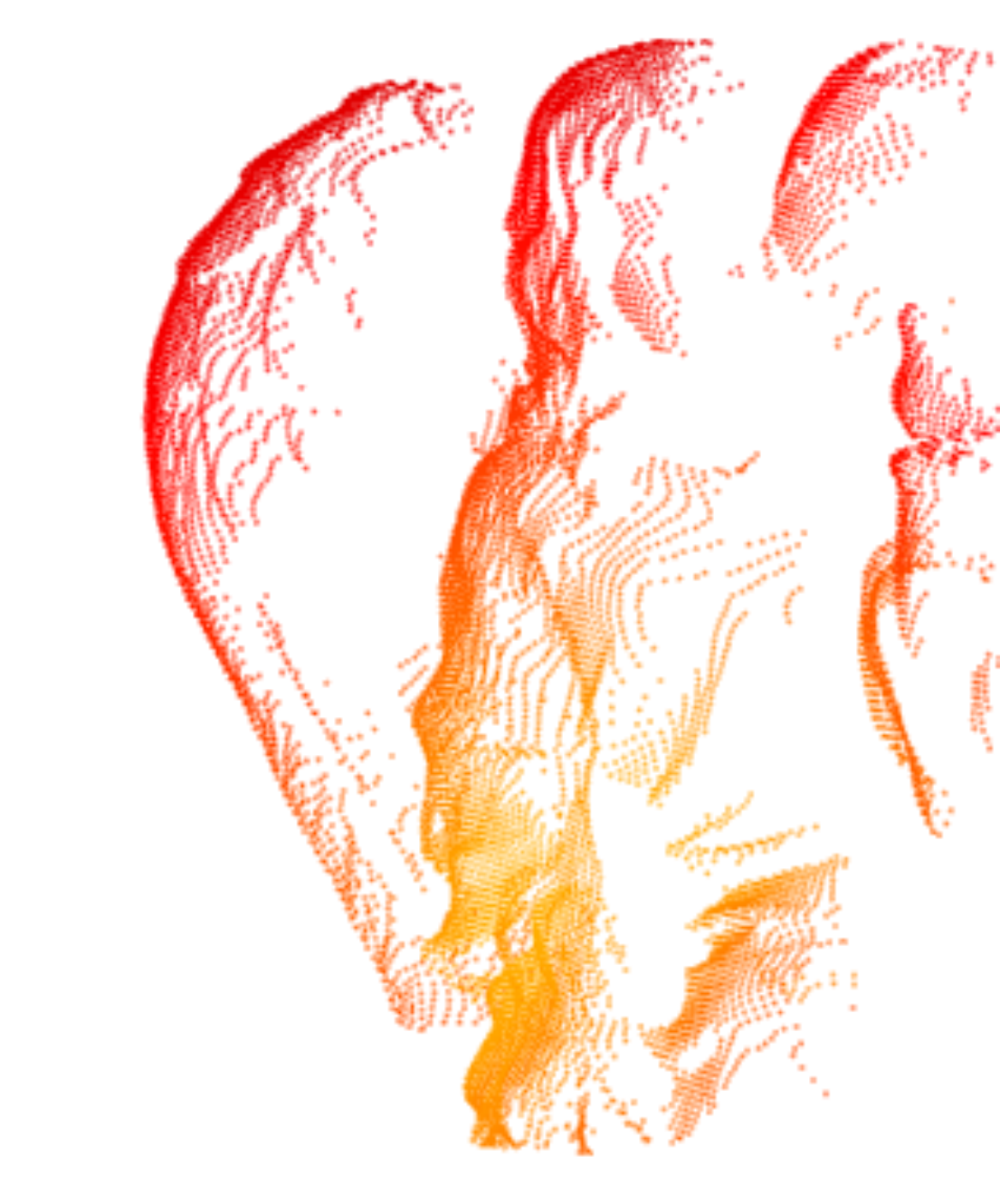}}
  \end{minipage}
  & \begin{minipage}[b]{0.144\columnwidth}
    \centering
    \raisebox{-.5\height}{\includegraphics[width=\linewidth]{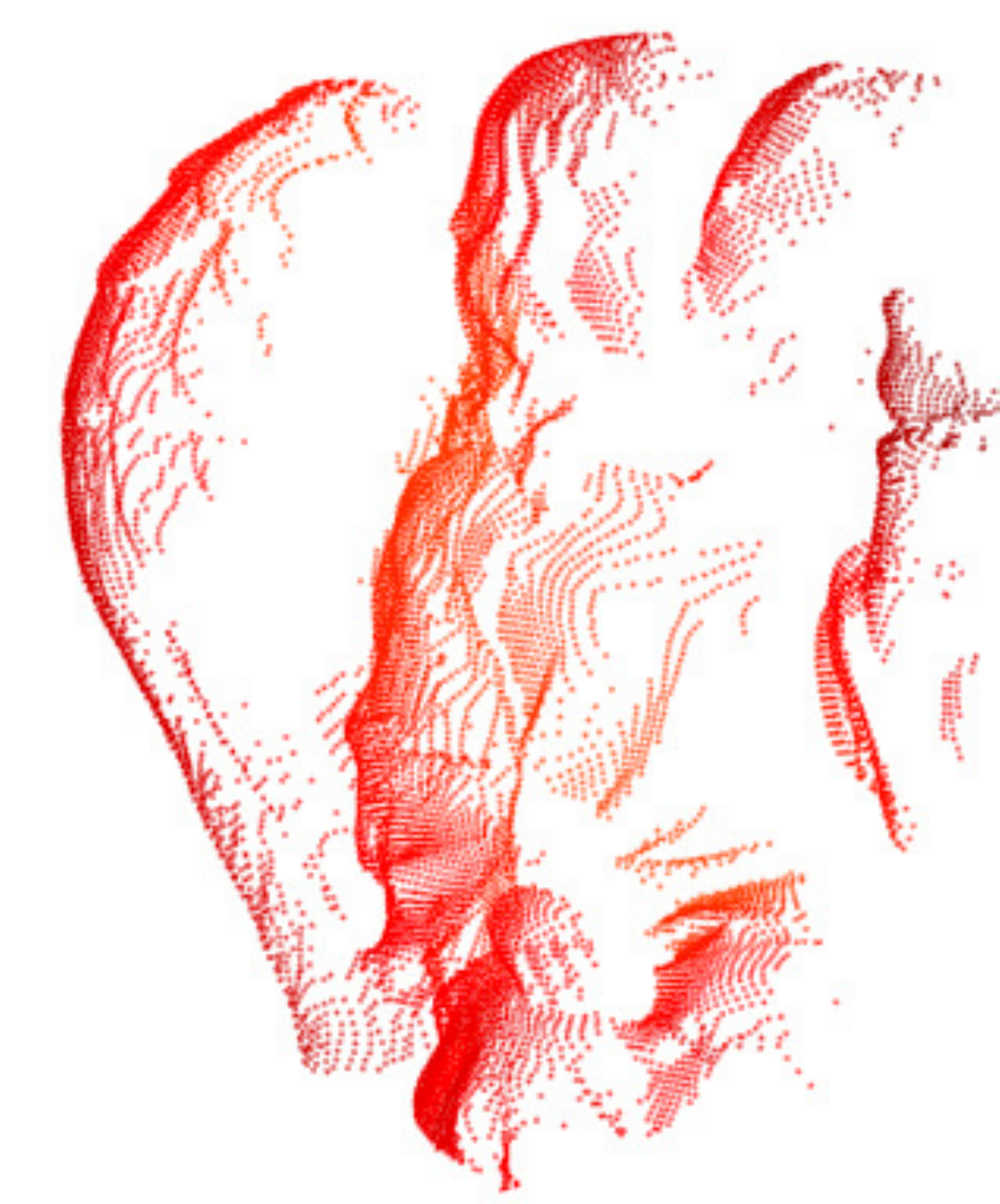}}
  \end{minipage} 
  & \begin{minipage}[b]{0.144\columnwidth}
    \centering
    \raisebox{-.5\height}{\includegraphics[width=\linewidth]{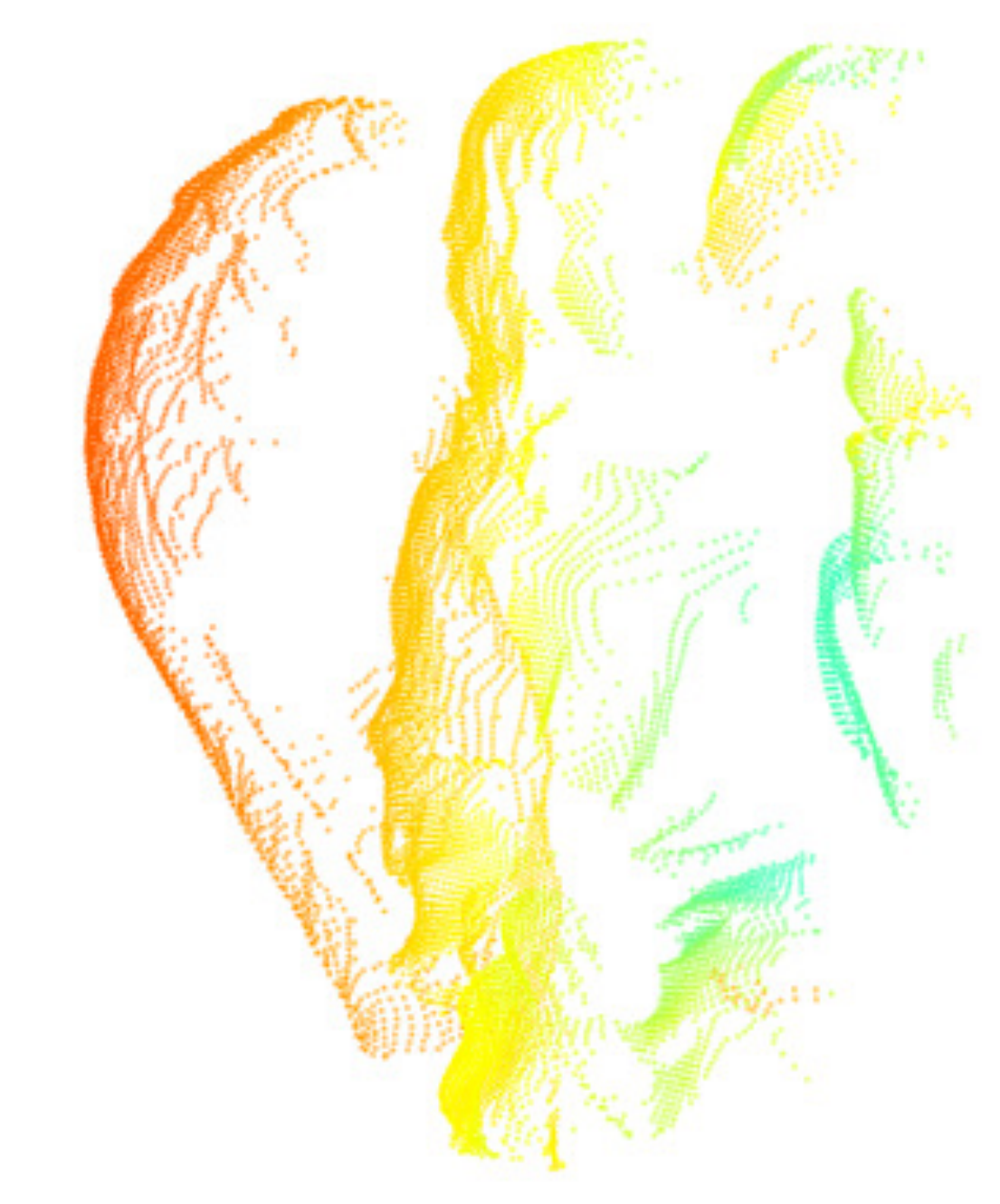}}
  \end{minipage} 
  & \begin{minipage}[b]{0.144\columnwidth}
    \centering
    \raisebox{-.5\height}{\includegraphics[width=\linewidth]{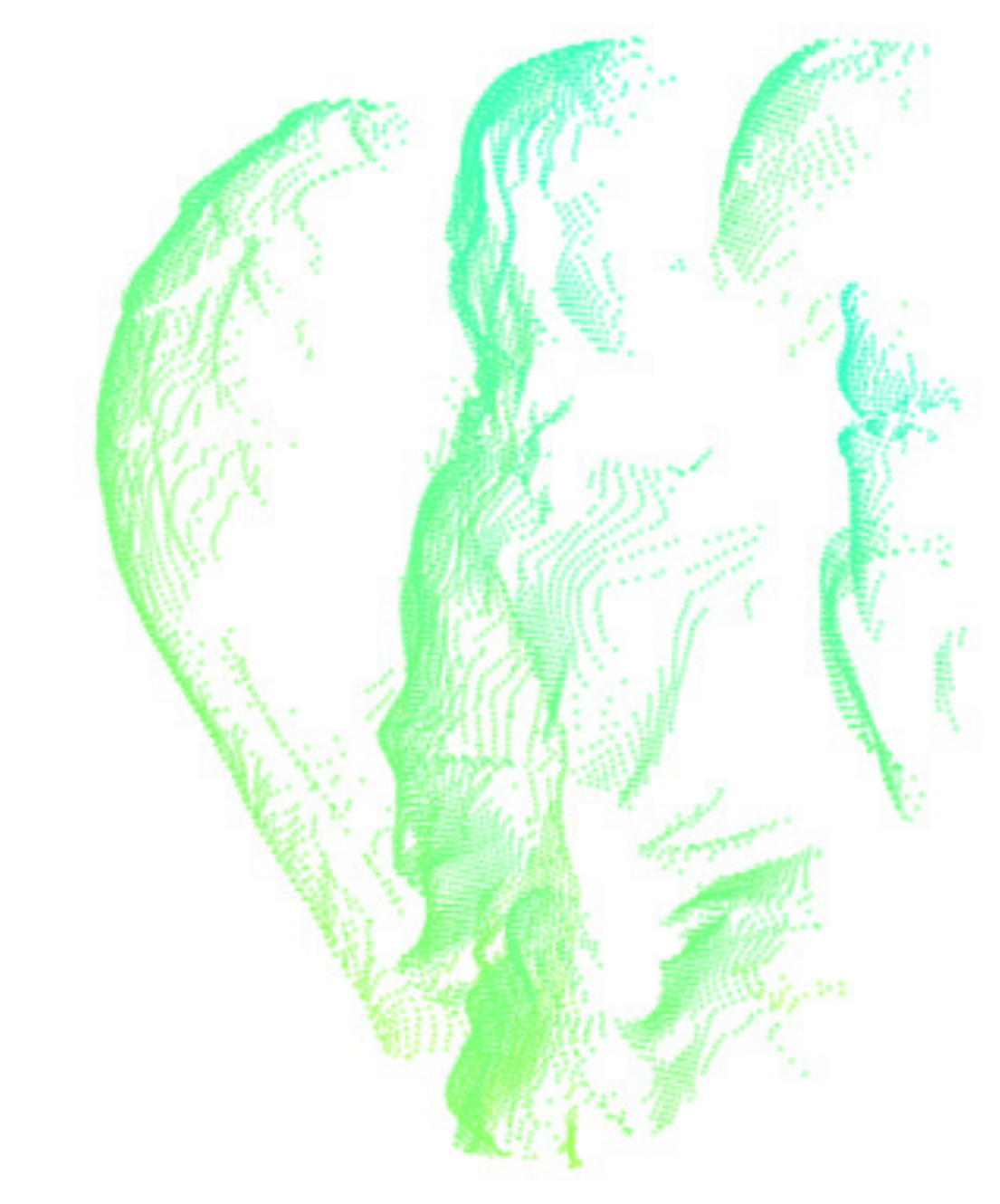}}
  \end{minipage} 
  & \begin{minipage}[b]{0.144\columnwidth}
    \centering
    \raisebox{-.5\height}{\includegraphics[width=\linewidth]{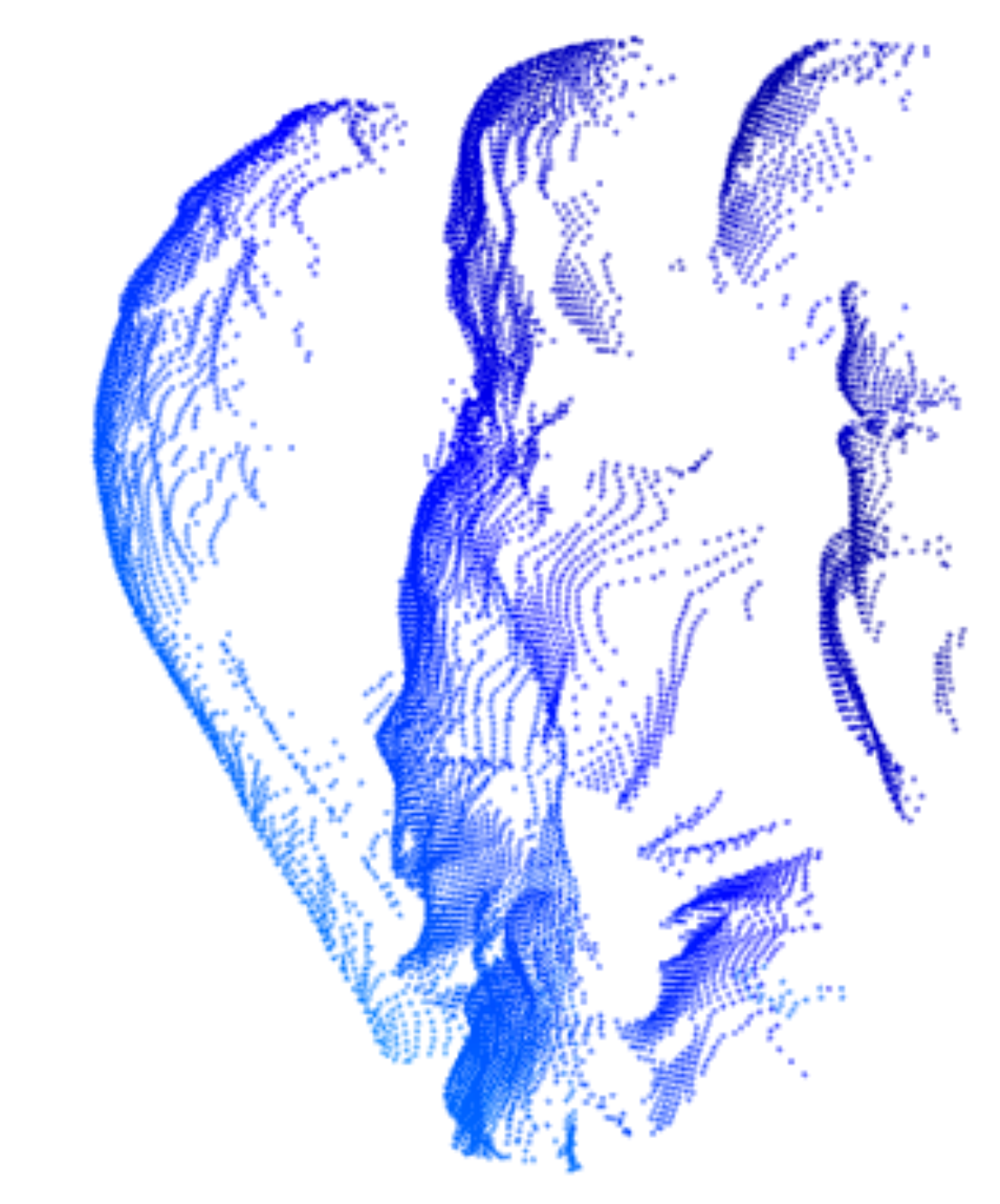}}
  \end{minipage} 
  & \begin{minipage}[b]{0.144\columnwidth}
    \centering
    \raisebox{-.5\height}{\includegraphics[width=\linewidth]{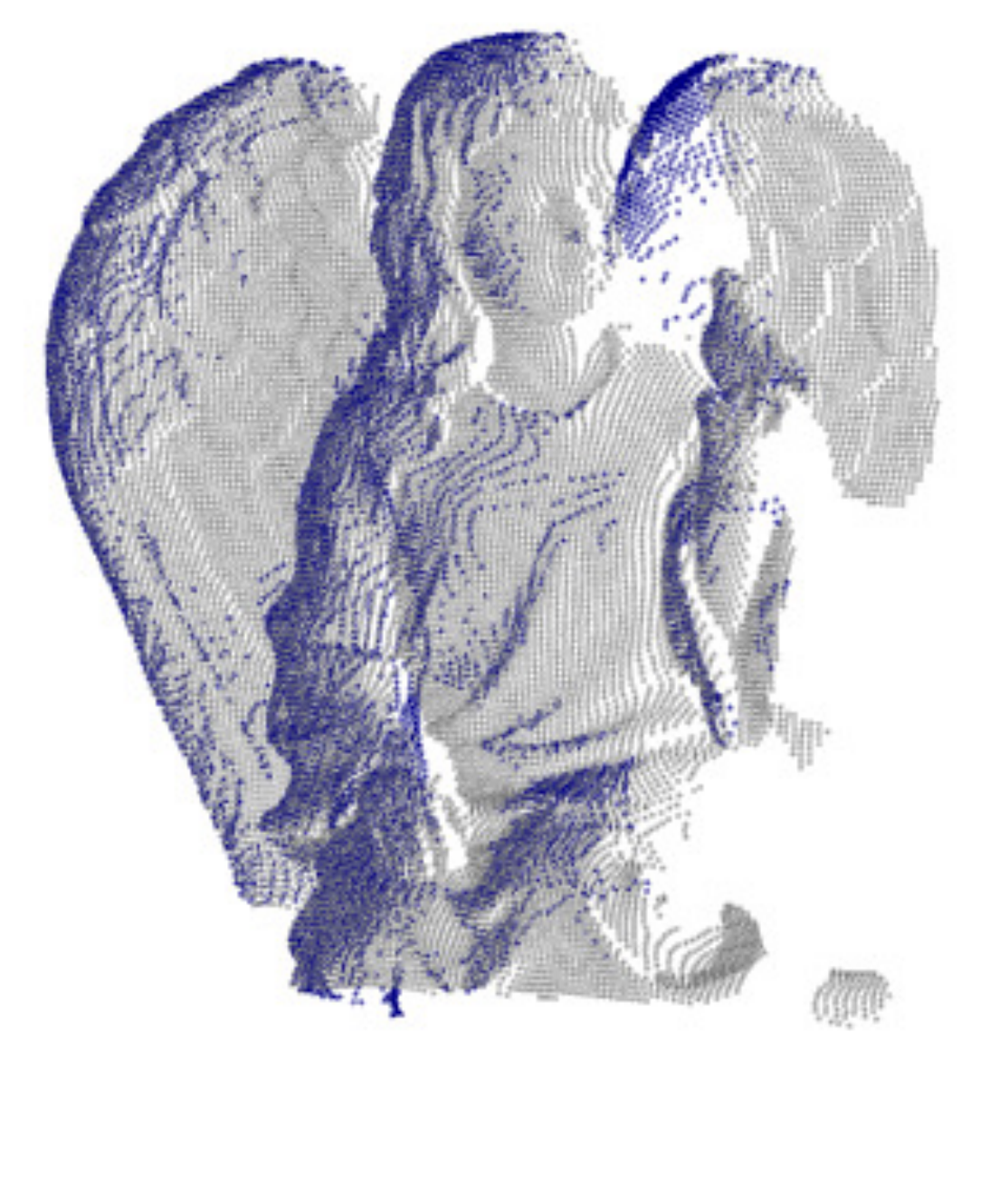}}
  \end{minipage}
  & \begin{minipage}[b]{0.08\columnwidth}
    \centering
    \raisebox{-.5\height}{\includegraphics[width=\linewidth]{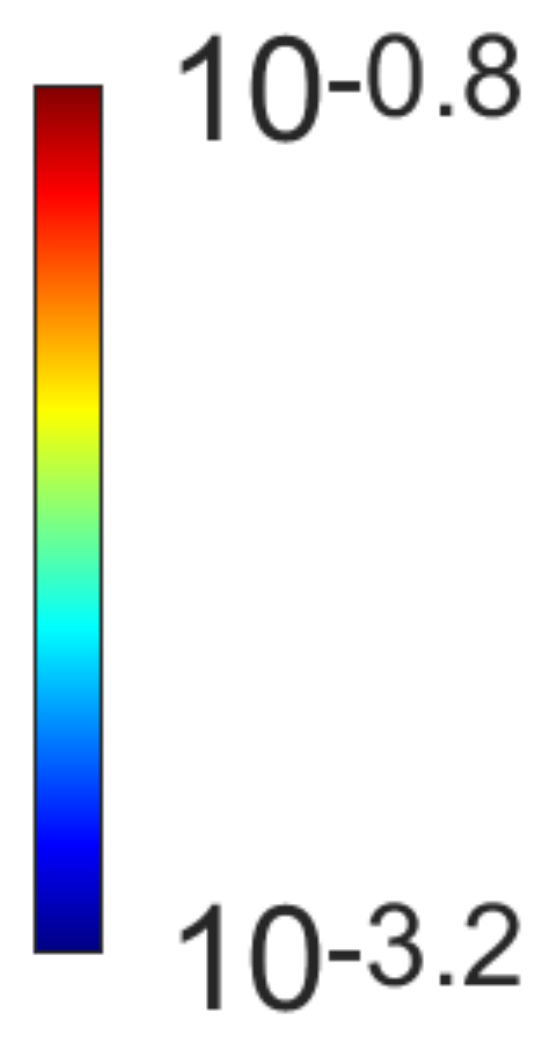}}
  \end{minipage} 
  \\

  \begin{minipage}[b]{0.128\columnwidth}
    \centering
    \raisebox{-.5\height}{\includegraphics[width=\linewidth]{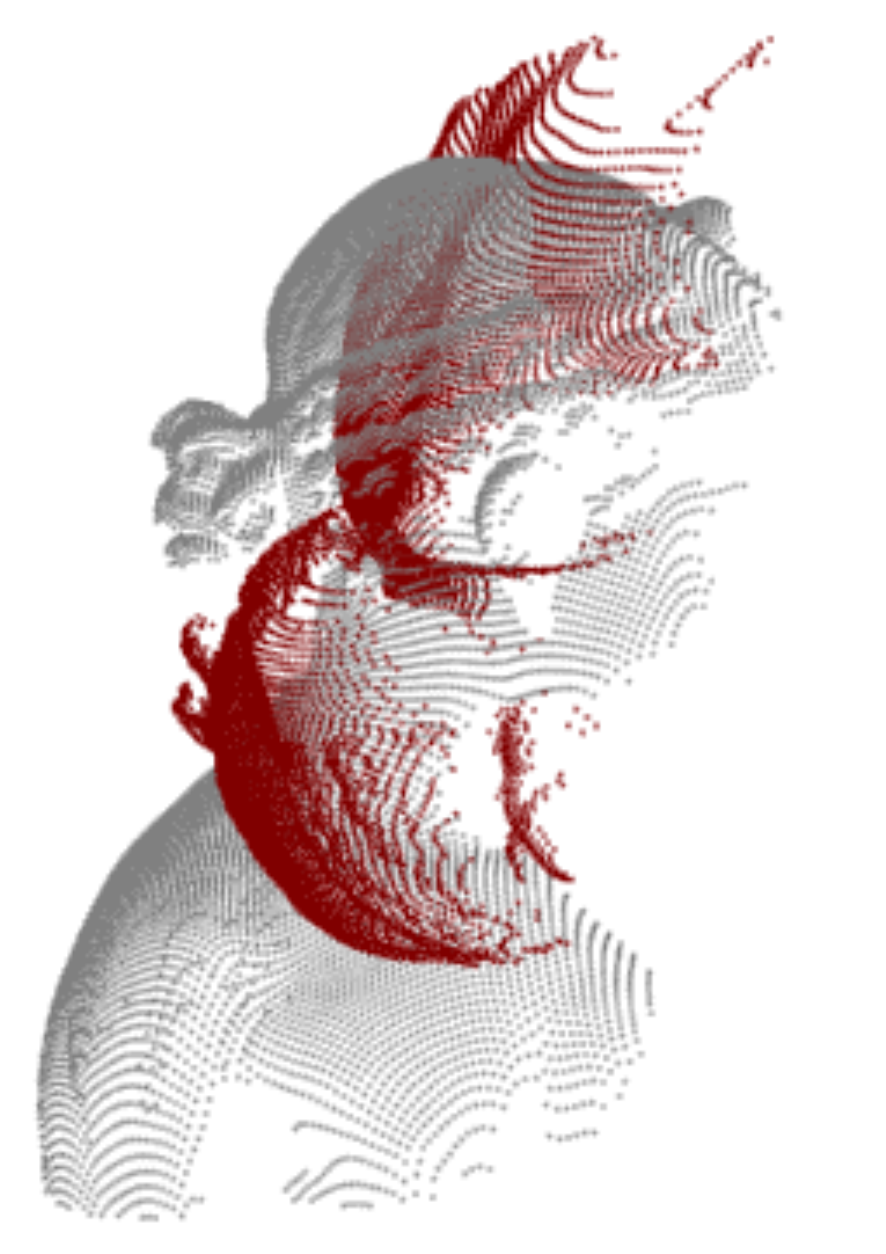}}
  \end{minipage} 
  & \begin{minipage}[b]{0.128\columnwidth}
    \centering
    \raisebox{-.5\height}{\includegraphics[width=\linewidth]{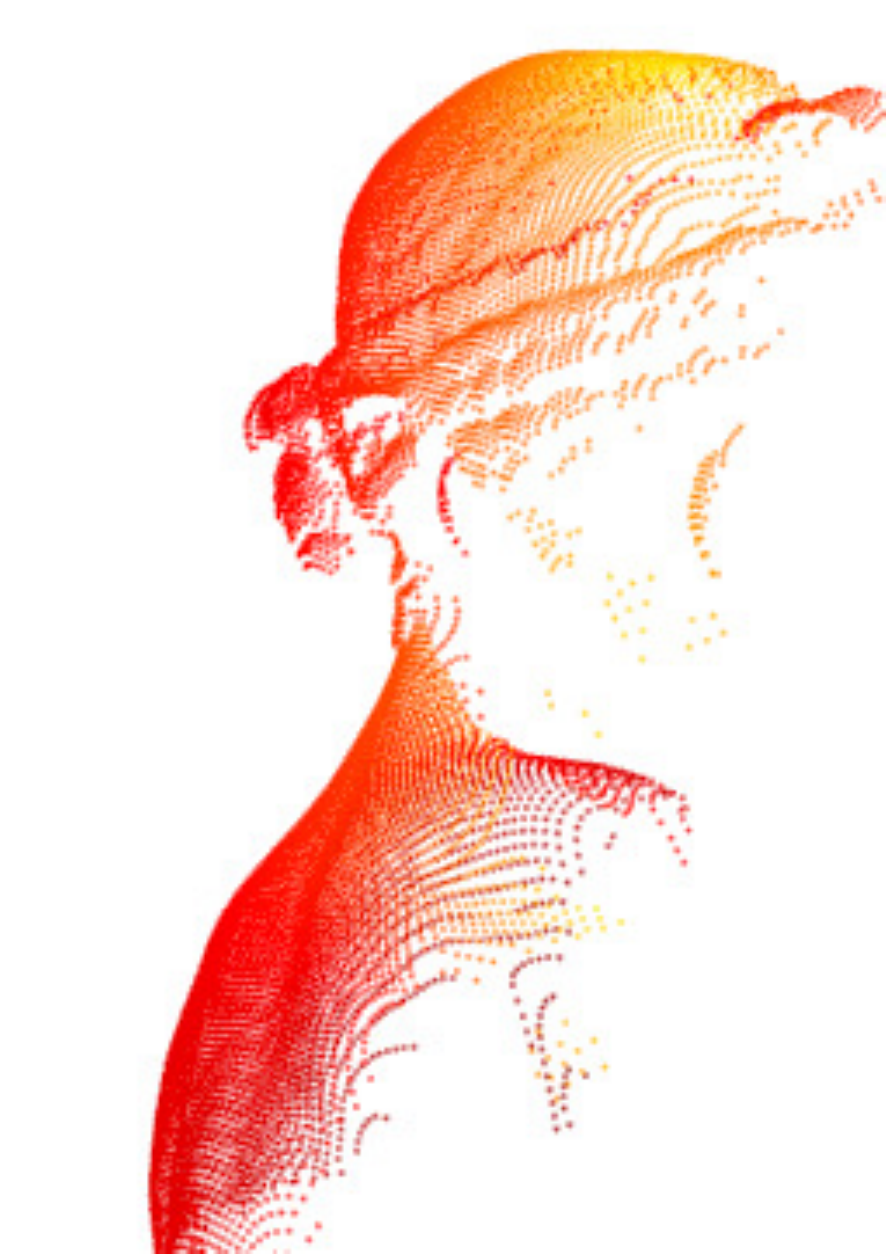}}
  \end{minipage}
  & \begin{minipage}[b]{0.128\columnwidth}
    \centering
    \raisebox{-.5\height}{\includegraphics[width=\linewidth]{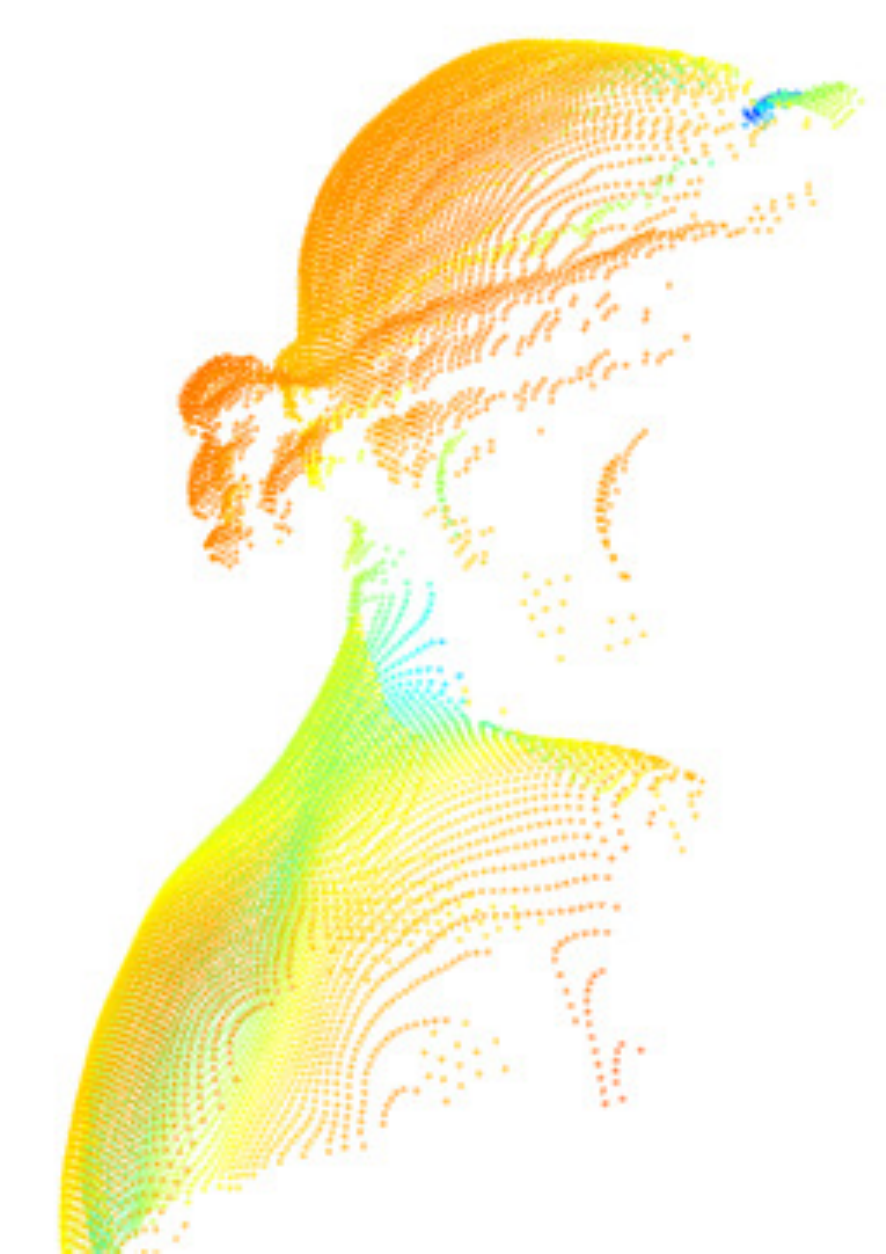}}
  \end{minipage} 
  & \begin{minipage}[b]{0.128\columnwidth}
    \centering
    \raisebox{-.5\height}{\includegraphics[width=\linewidth]{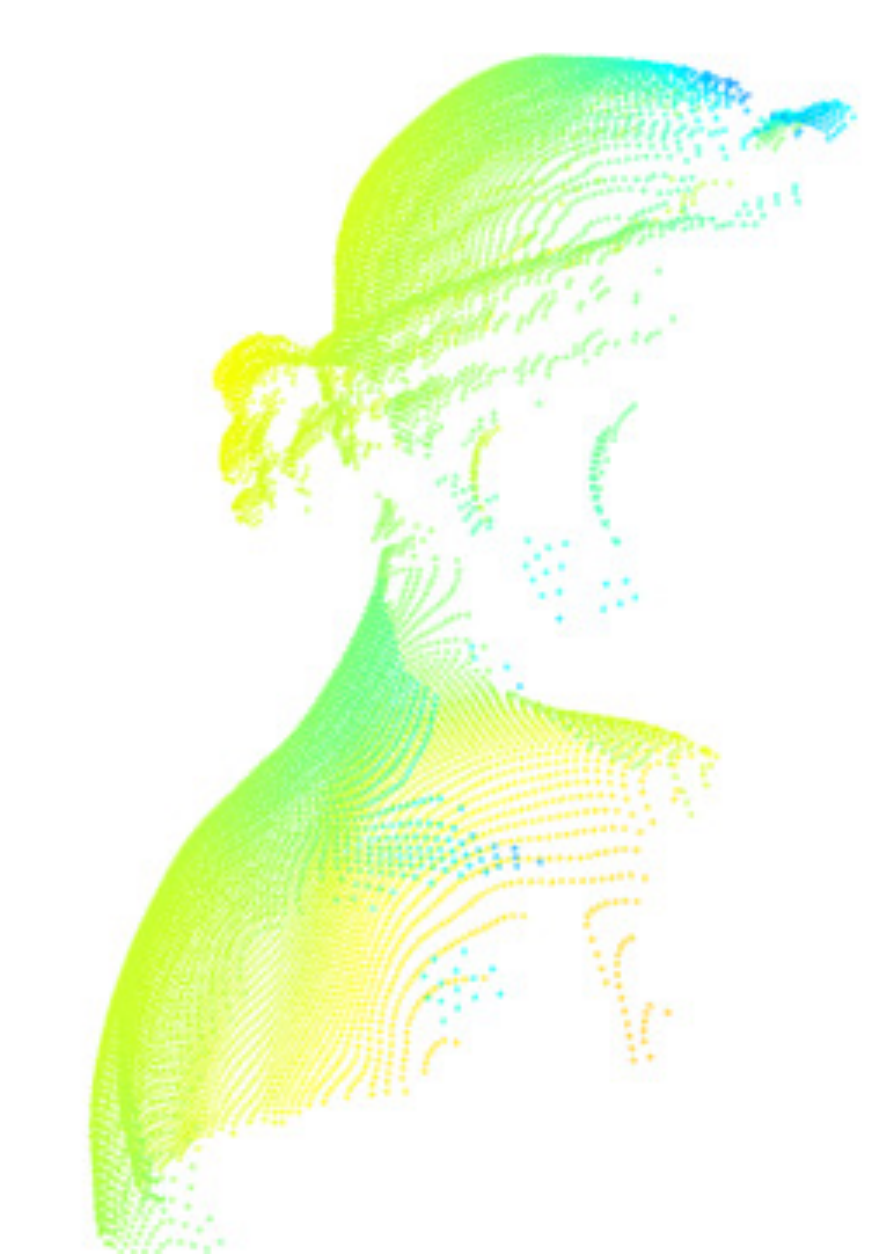}}
  \end{minipage} 
  & \begin{minipage}[b]{0.128\columnwidth}
    \centering
    \raisebox{-.5\height}{\includegraphics[width=\linewidth]{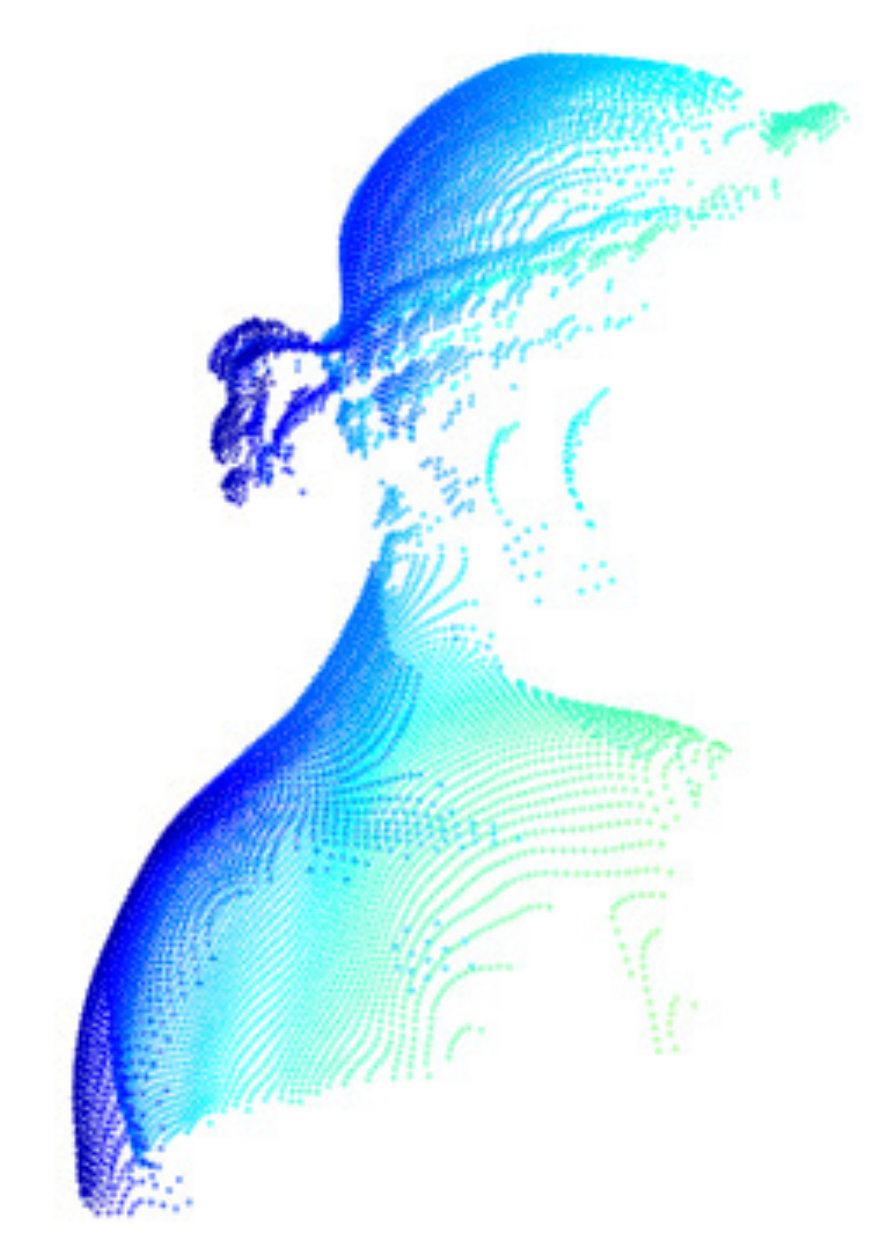}}
  \end{minipage} 
  & \begin{minipage}[b]{0.128\columnwidth}
    \centering
    \raisebox{-.5\height}{\includegraphics[width=\linewidth]{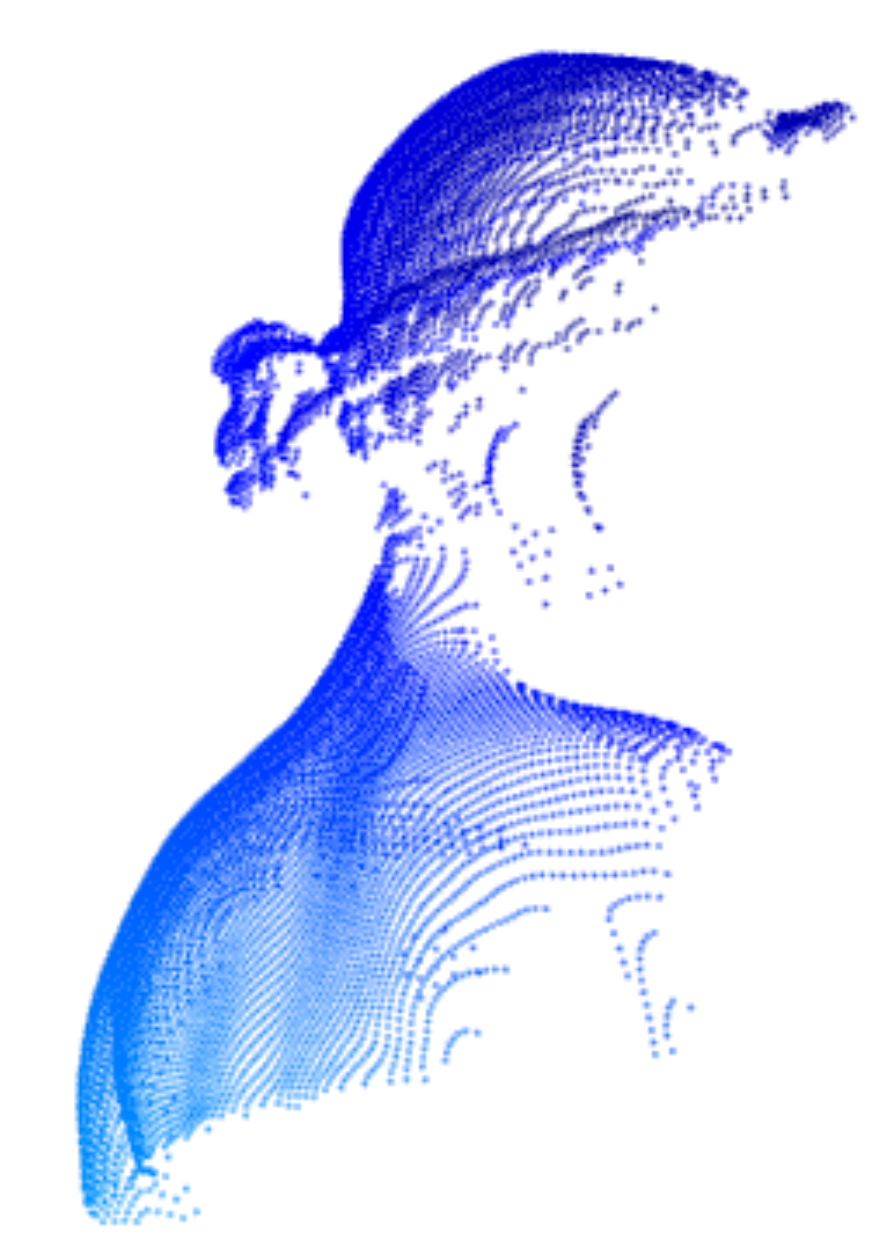}}
  \end{minipage} 
  & \begin{minipage}[b]{0.128\columnwidth}
    \centering
    \raisebox{-.5\height}{\includegraphics[width=\linewidth]{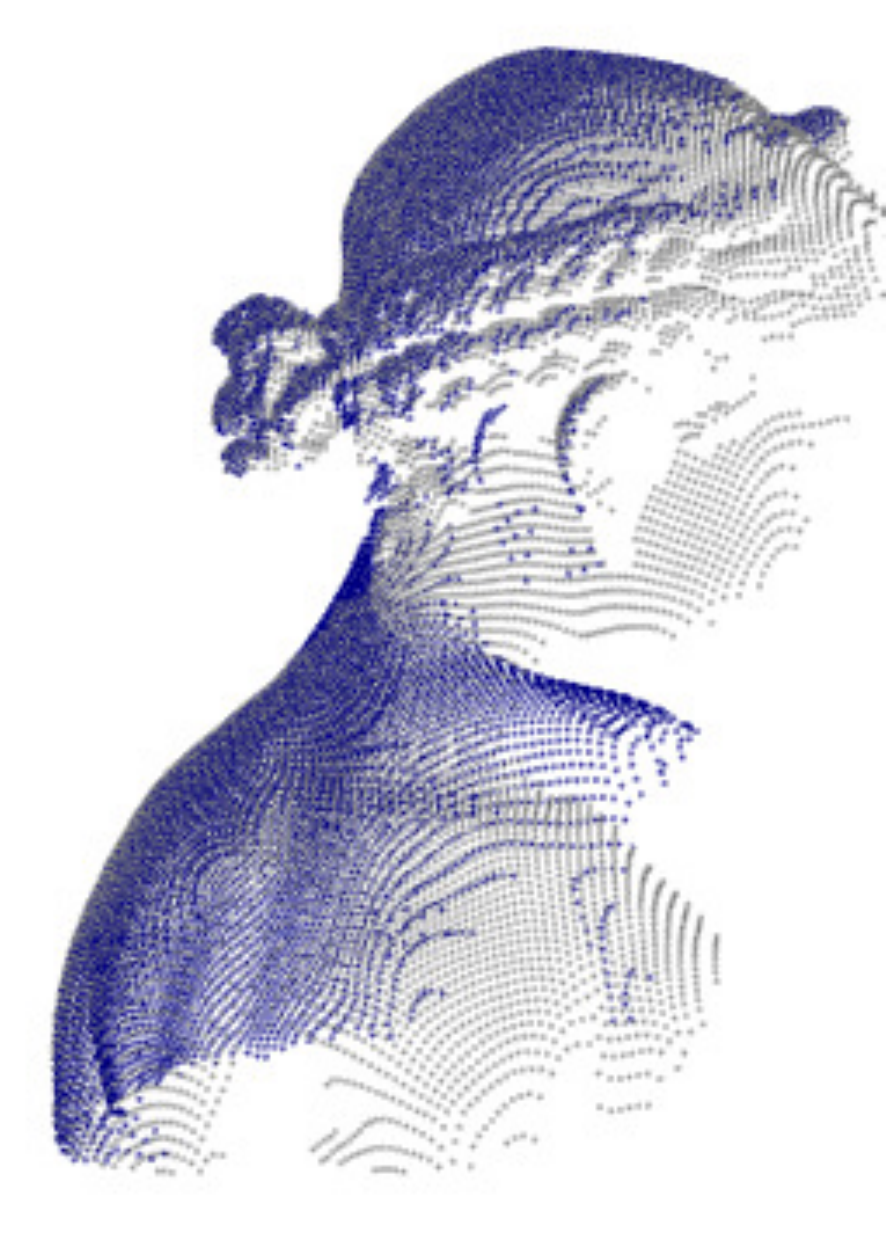}}
  \end{minipage} 
  & \begin{minipage}[b]{0.08\columnwidth}
    \centering
    \raisebox{-.5\height}{\includegraphics[width=\linewidth]{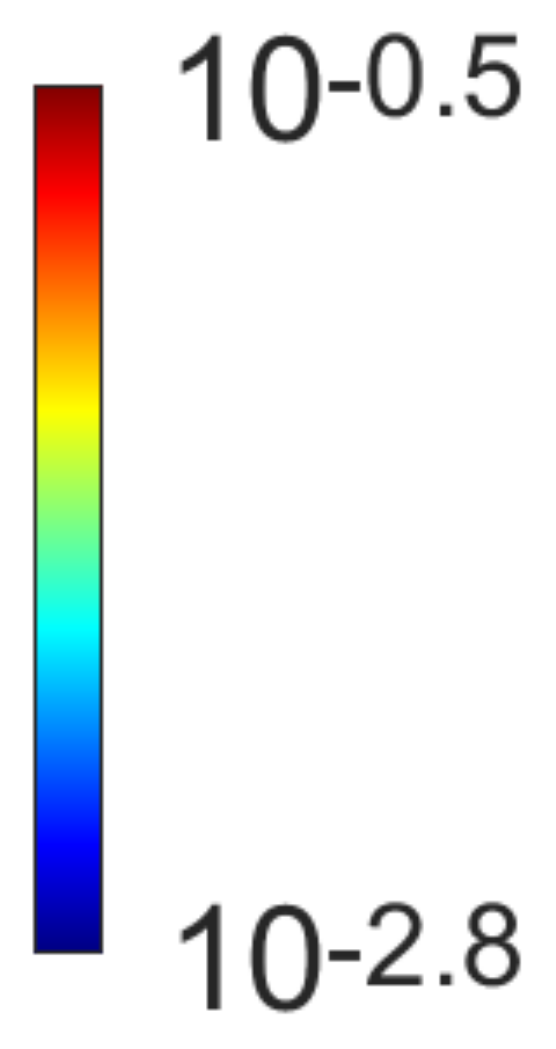}}
  \end{minipage} 
  \\

  \begin{minipage}[b]{0.128\columnwidth}
    \centering
    \raisebox{-.5\height}{\includegraphics[width=\linewidth]{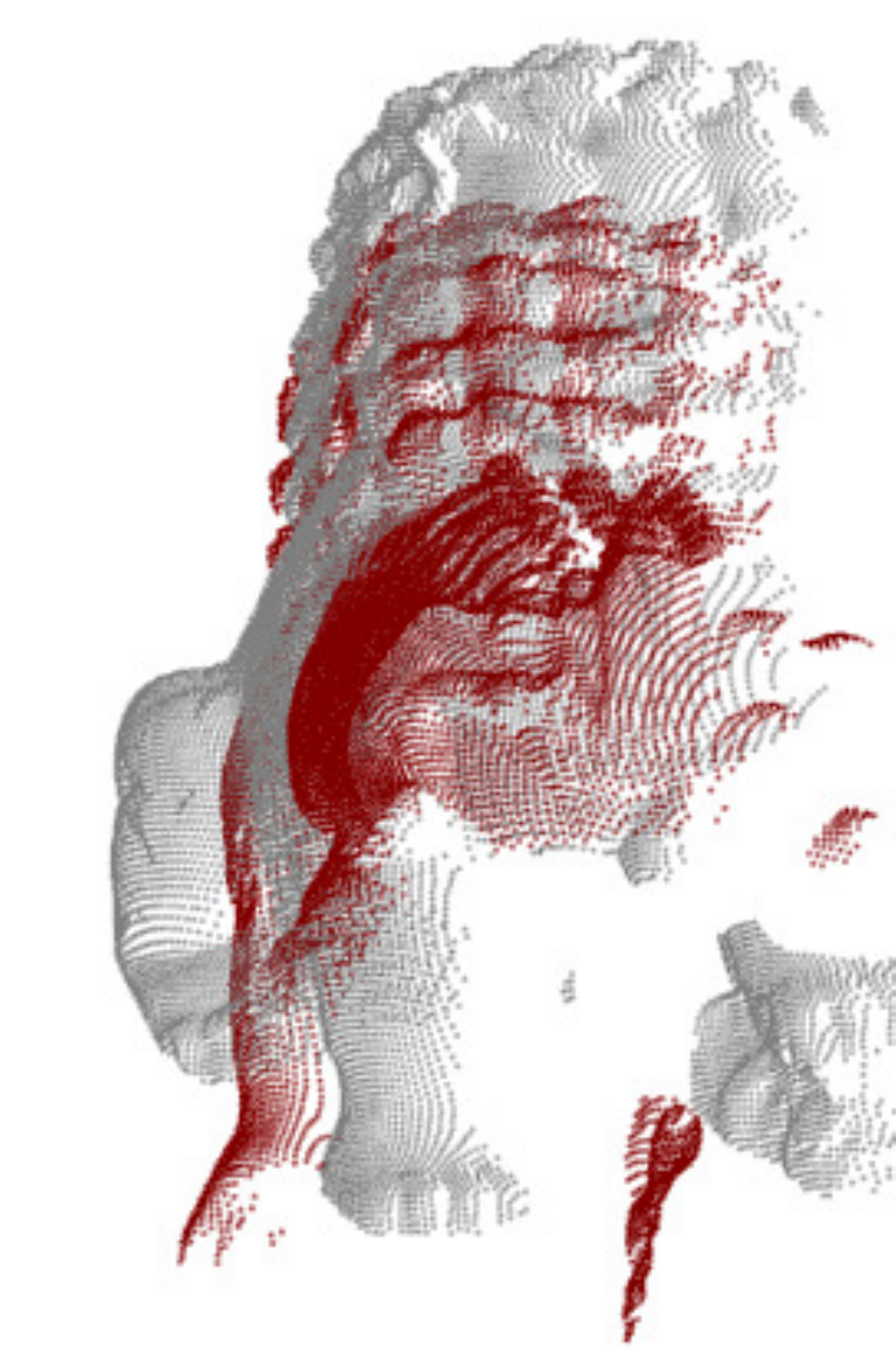}}
  \end{minipage} 
  & \begin{minipage}[b]{0.128\columnwidth}
    \centering
    \raisebox{-.5\height}{\includegraphics[width=\linewidth]{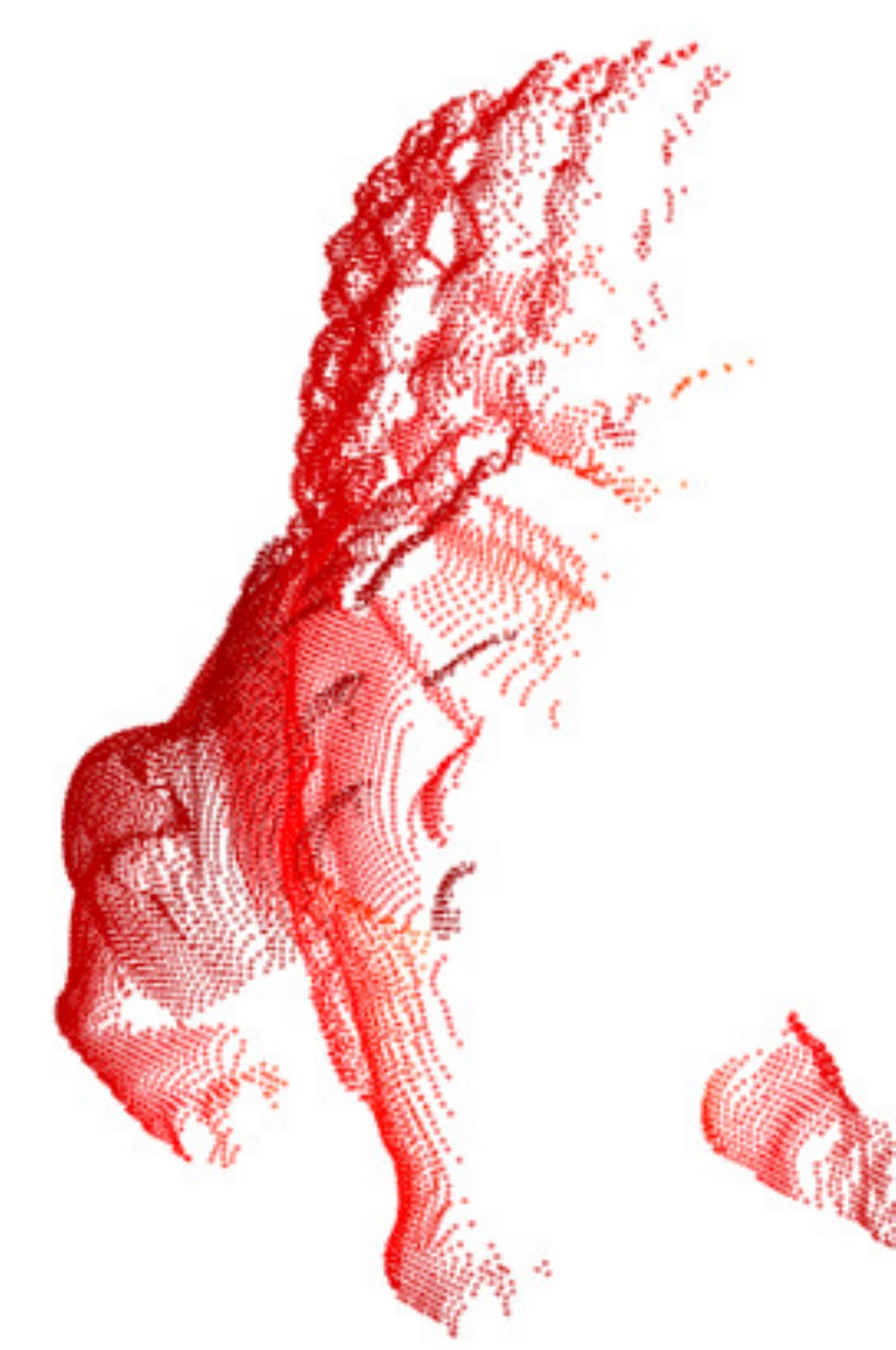}}
  \end{minipage}
  & \begin{minipage}[b]{0.128\columnwidth}
    \centering
    \raisebox{-.5\height}{\includegraphics[width=\linewidth]{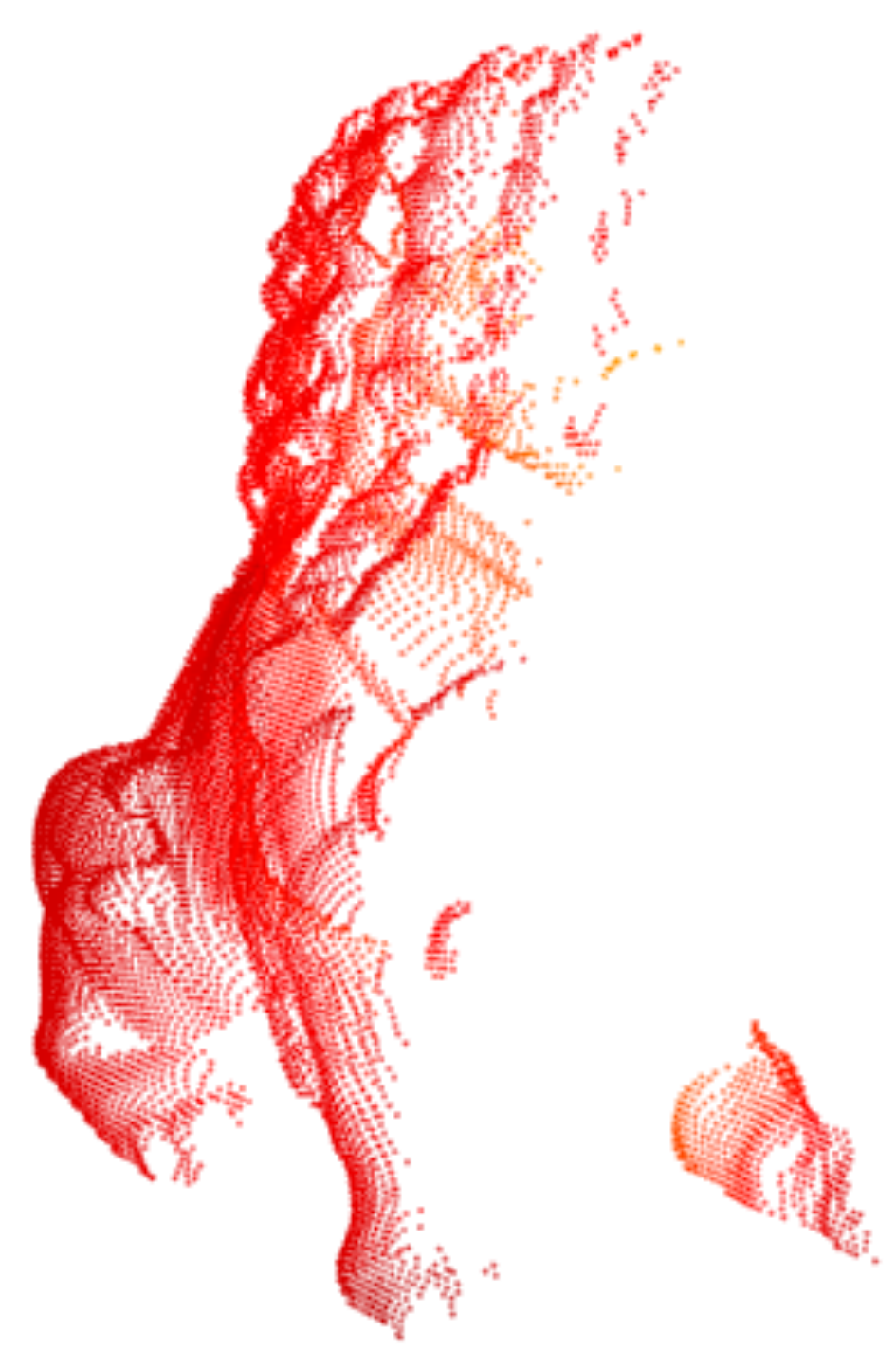}}
  \end{minipage} 
  & \begin{minipage}[b]{0.128\columnwidth}
    \centering
    \raisebox{-.5\height}{\includegraphics[width=\linewidth]{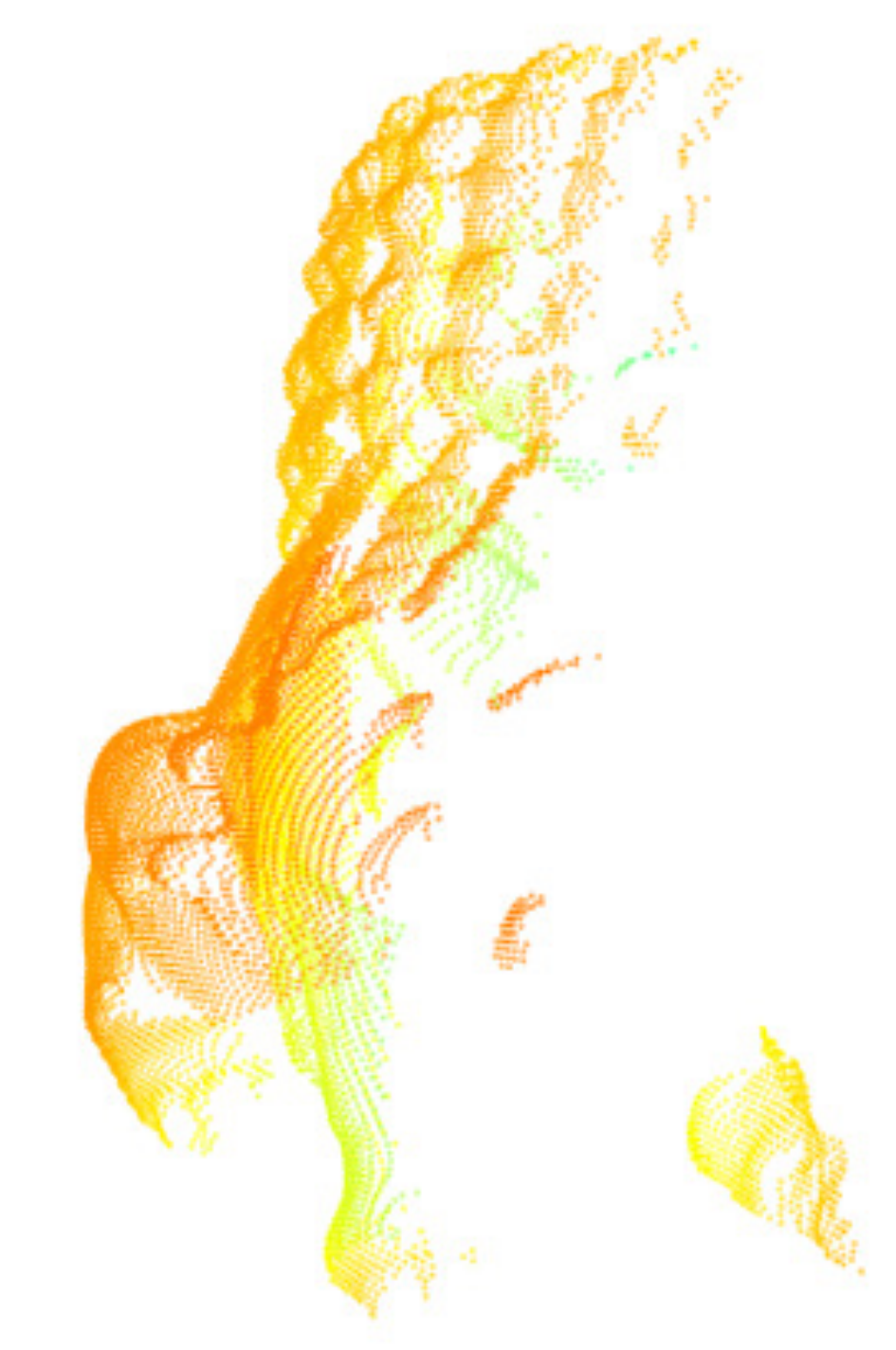}}
  \end{minipage} 
  & \begin{minipage}[b]{0.128\columnwidth}
    \centering
    \raisebox{-.5\height}{\includegraphics[width=\linewidth]{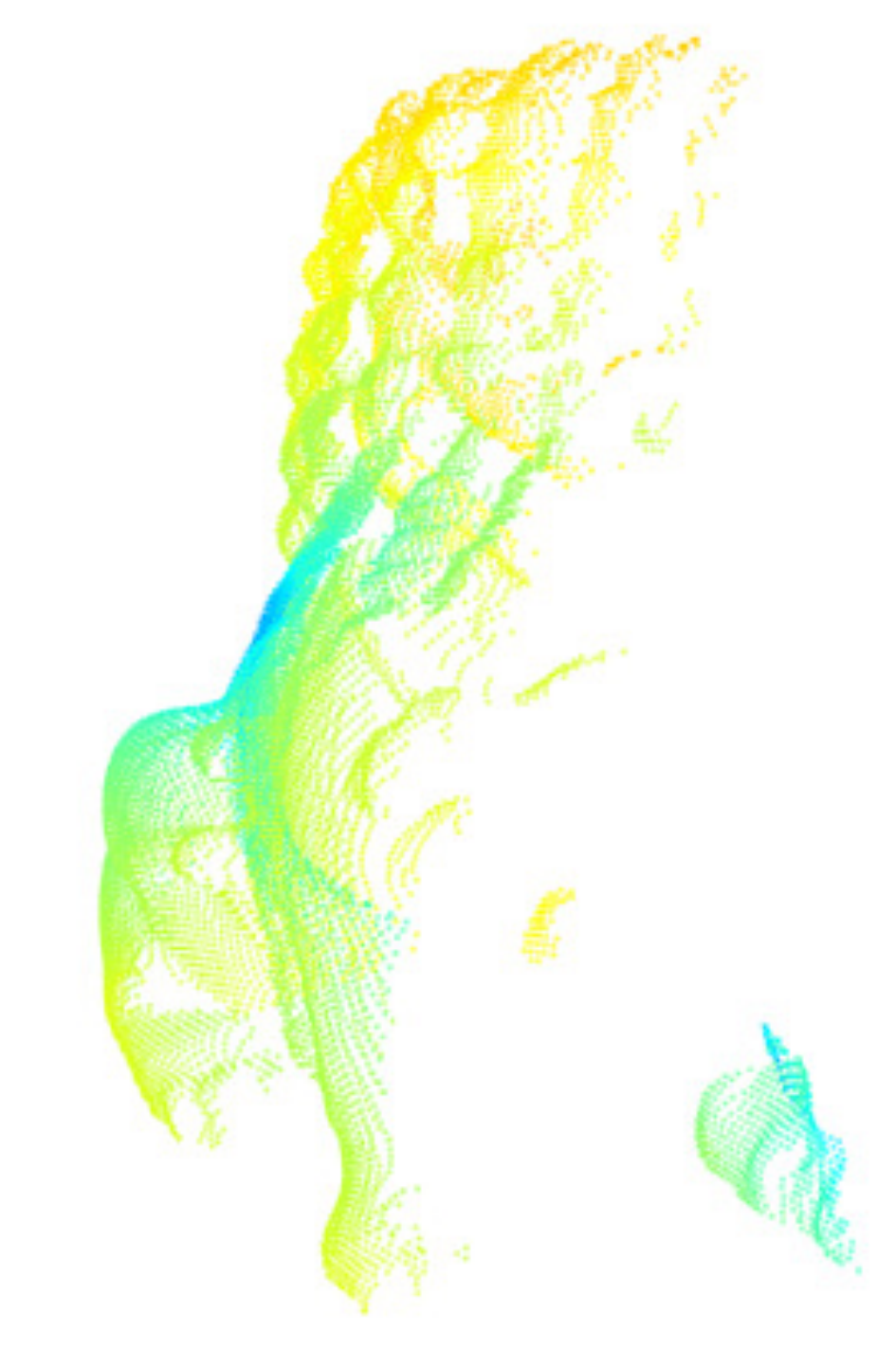}}
  \end{minipage} 
  & \begin{minipage}[b]{0.128\columnwidth}
    \centering
    \raisebox{-.5\height}{\includegraphics[width=\linewidth]{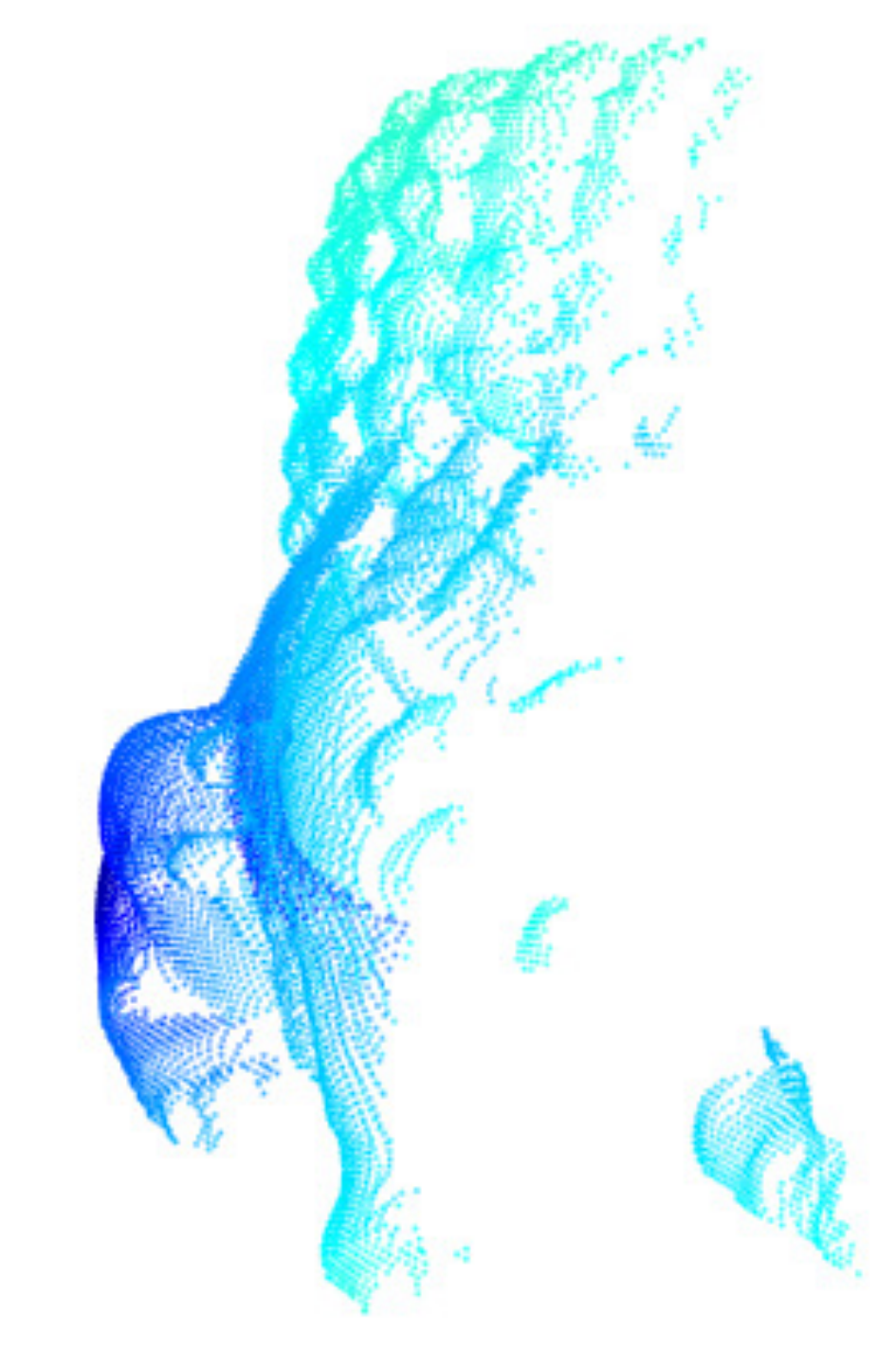}}
  \end{minipage} 
  & \begin{minipage}[b]{0.16\columnwidth}
    \centering
    \raisebox{-.5\height}{\includegraphics[width=\linewidth]{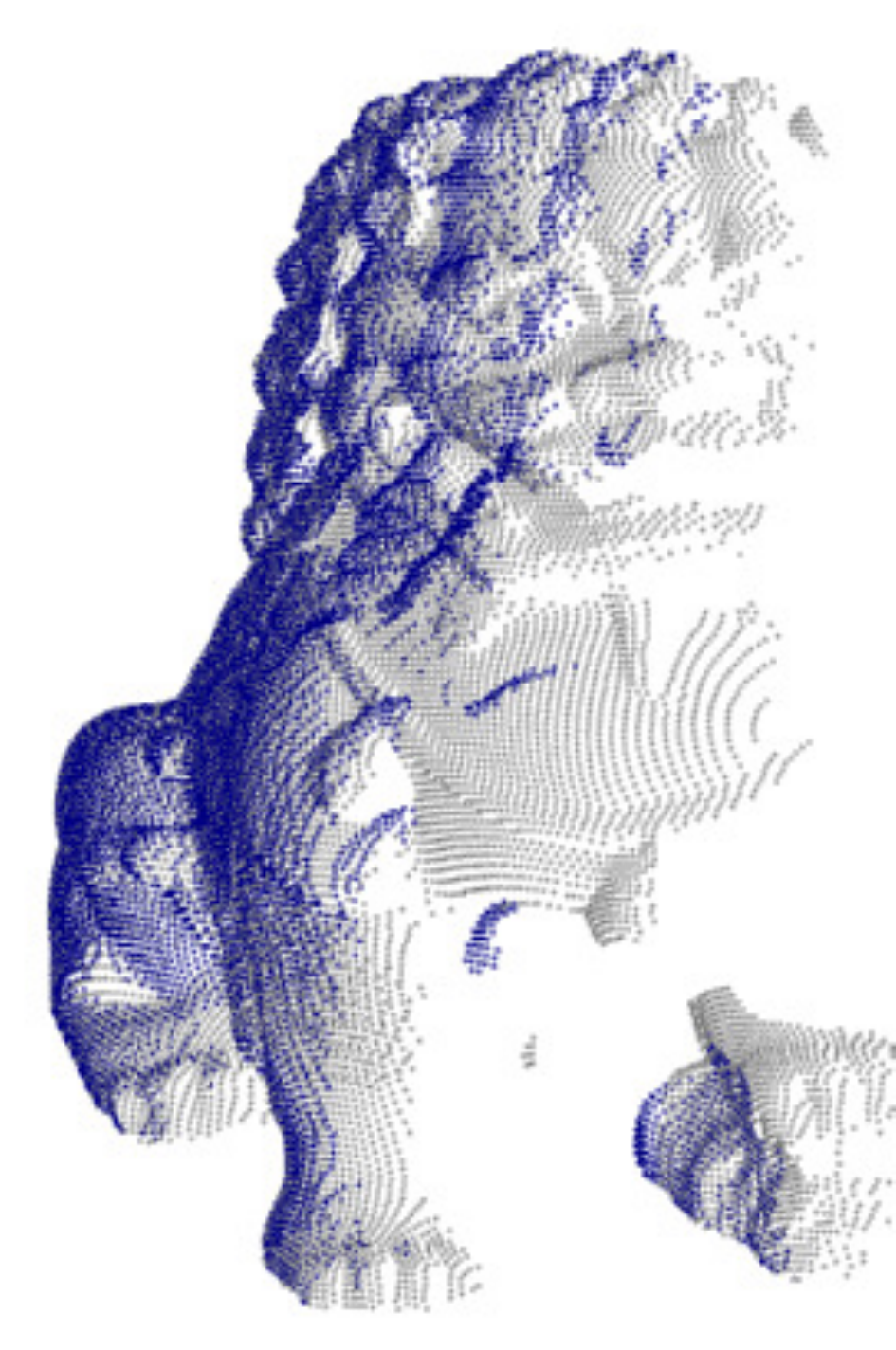}}
  \end{minipage} 
  & \begin{minipage}[b]{0.08\columnwidth}
    \centering
    \raisebox{-.5\height}{\includegraphics[width=\linewidth]{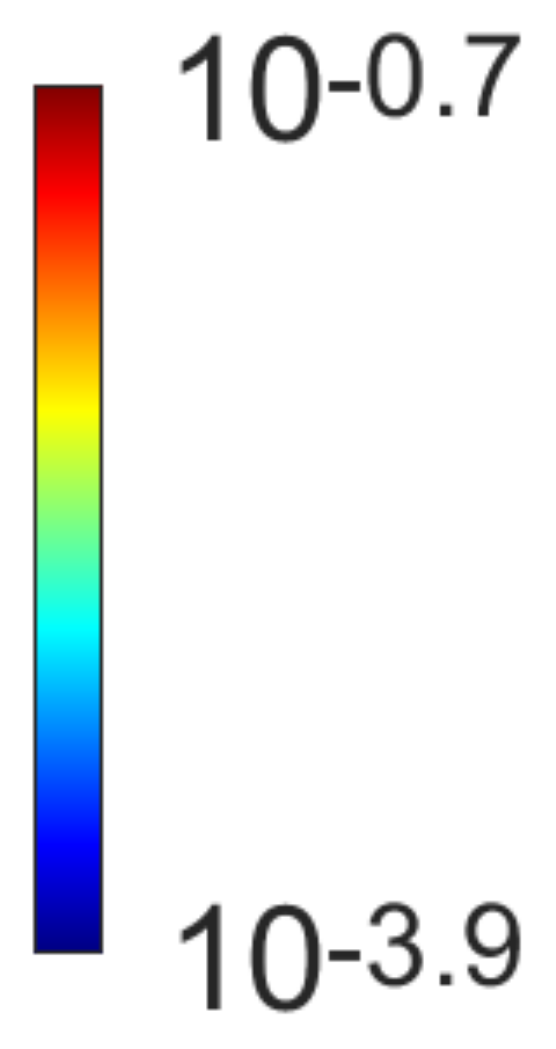}}
  \end{minipage} 
  \\

  \begin{minipage}[b]{0.184\columnwidth}
    \centering
    \raisebox{-.5\height}{\includegraphics[width=\linewidth]{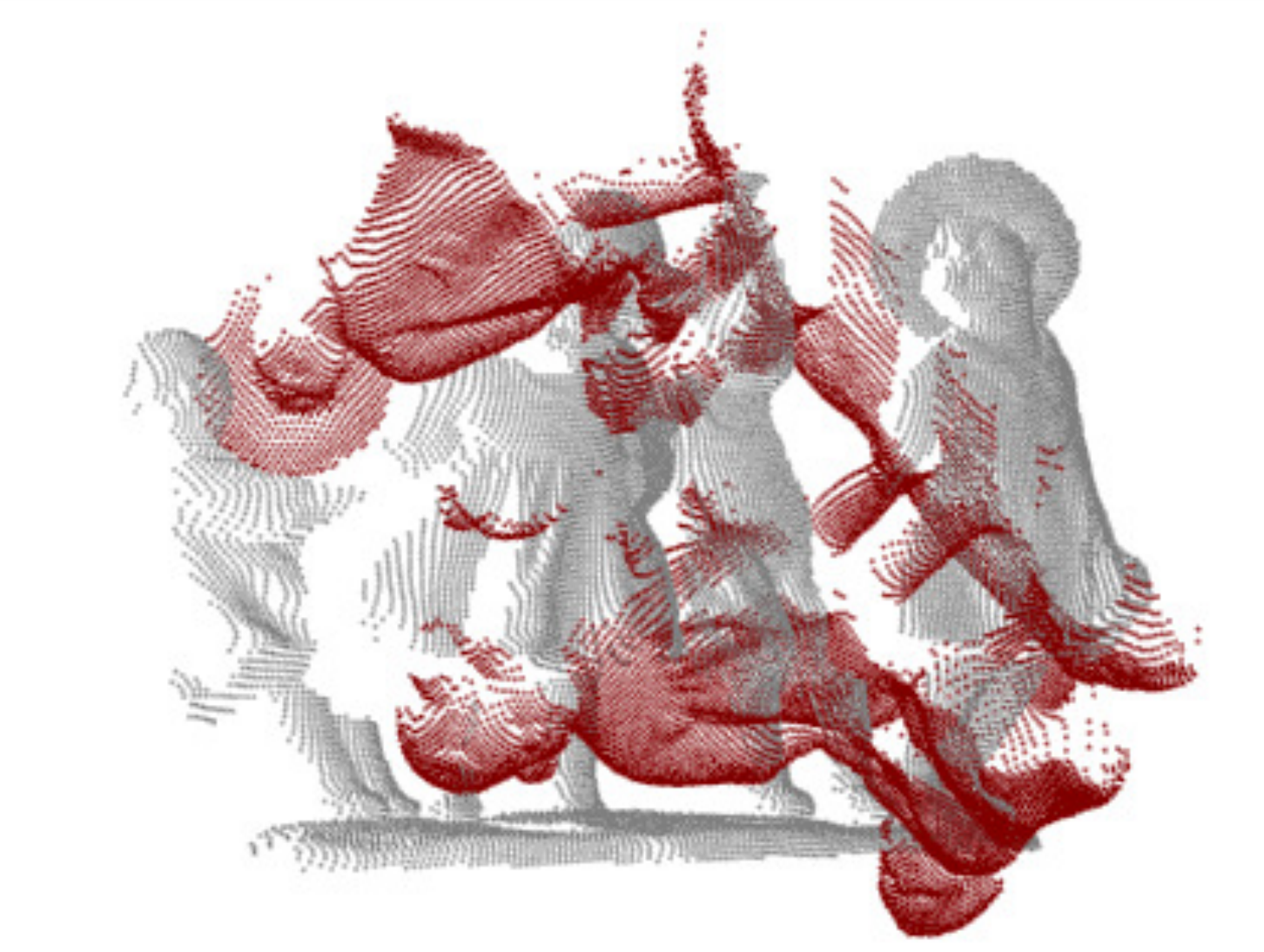}}
  \end{minipage} 
  & \begin{minipage}[b]{0.184\columnwidth}
    \centering
    \raisebox{-.5\height}{\includegraphics[width=\linewidth]{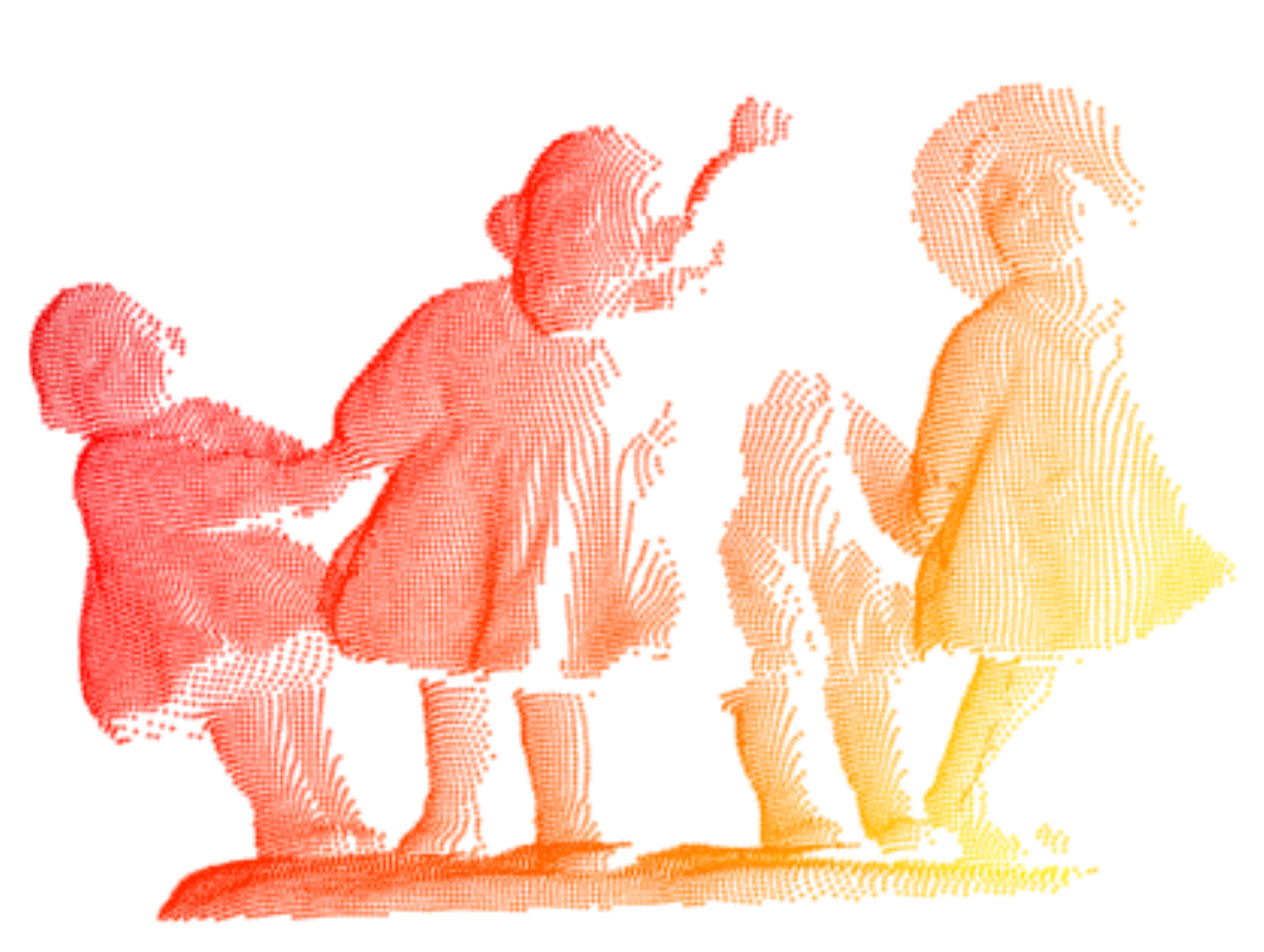}}
  \end{minipage}
  & \begin{minipage}[b]{0.184\columnwidth}
    \centering
    \raisebox{-.5\height}{\includegraphics[width=\linewidth]{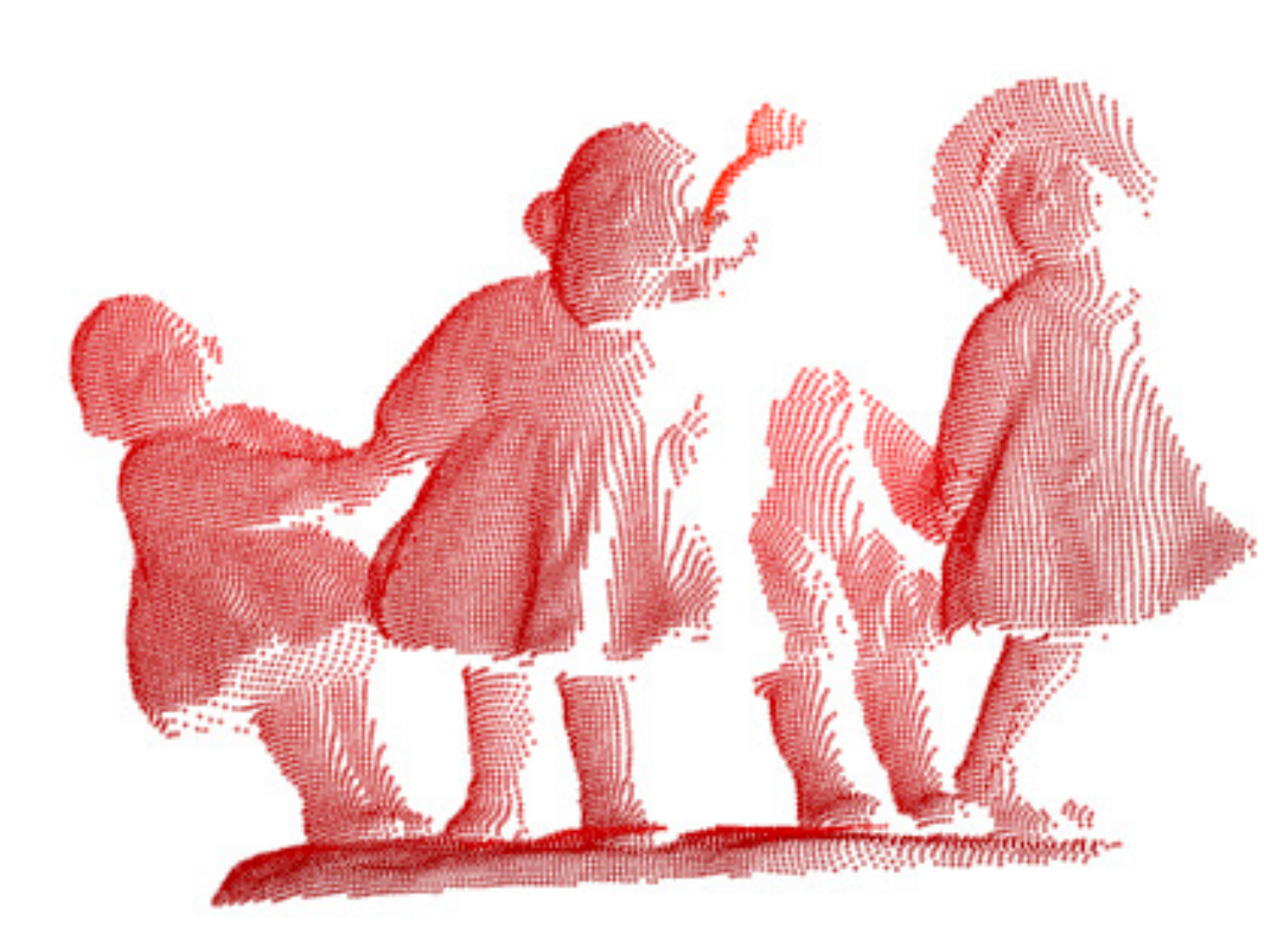}}
  \end{minipage} 
  & \begin{minipage}[b]{0.184\columnwidth}
    \centering
    \raisebox{-.5\height}{\includegraphics[width=\linewidth]{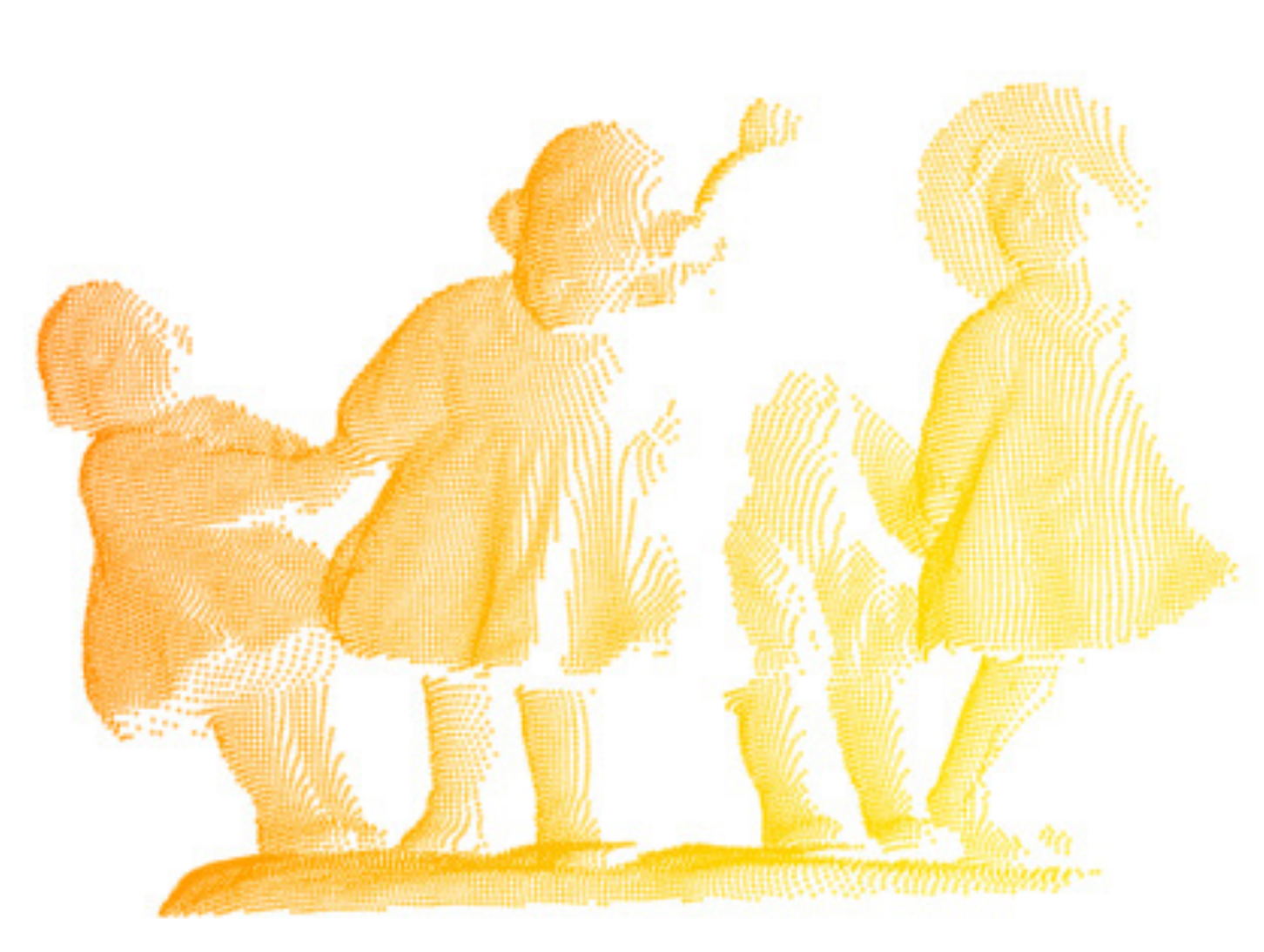}}
  \end{minipage} 
  & \begin{minipage}[b]{0.184\columnwidth}
    \centering
    \raisebox{-.5\height}{\includegraphics[width=\linewidth]{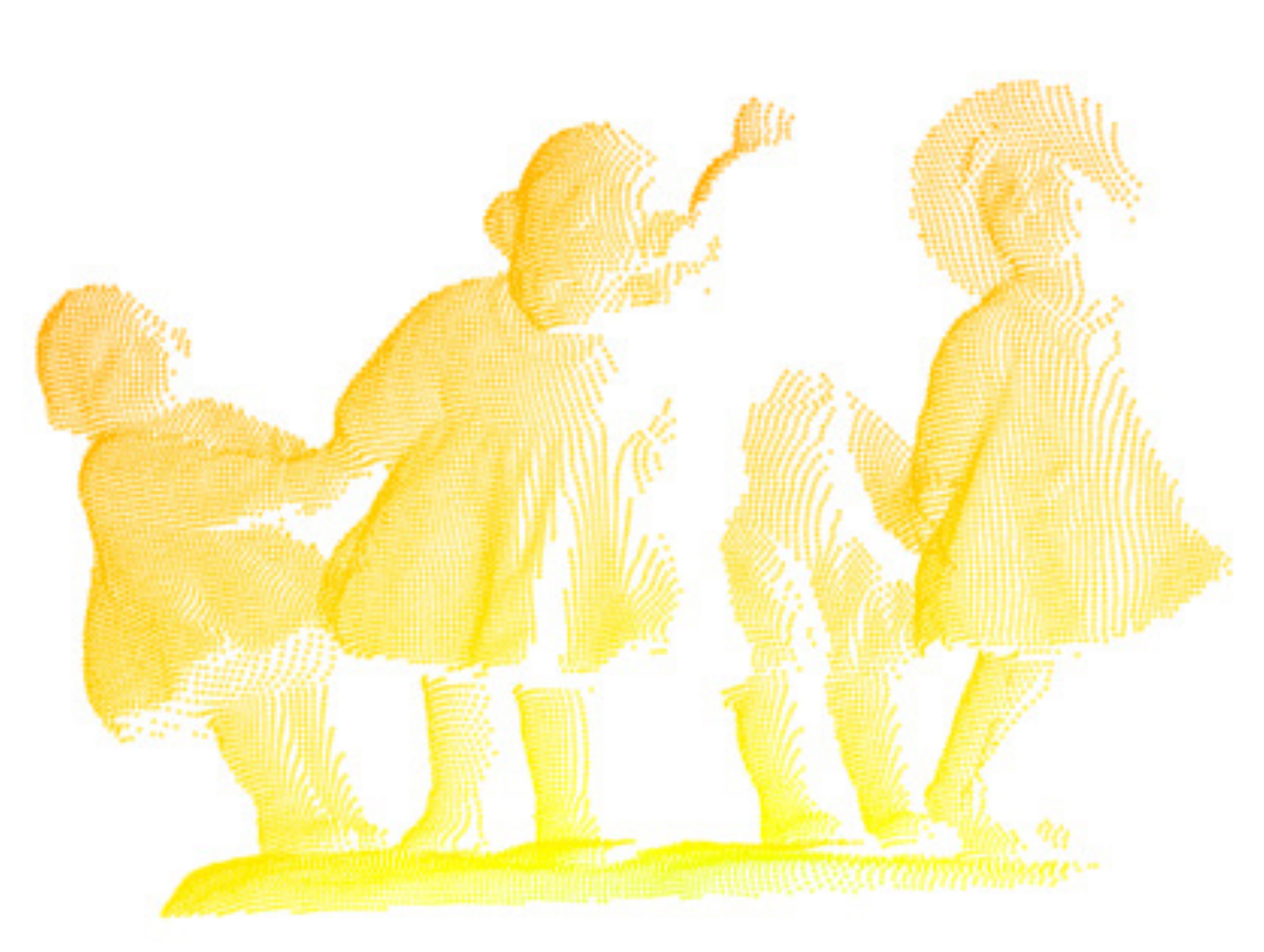}}
  \end{minipage} 
  & \begin{minipage}[b]{0.184\columnwidth}
    \centering
    \raisebox{-.5\height}{\includegraphics[width=\linewidth]{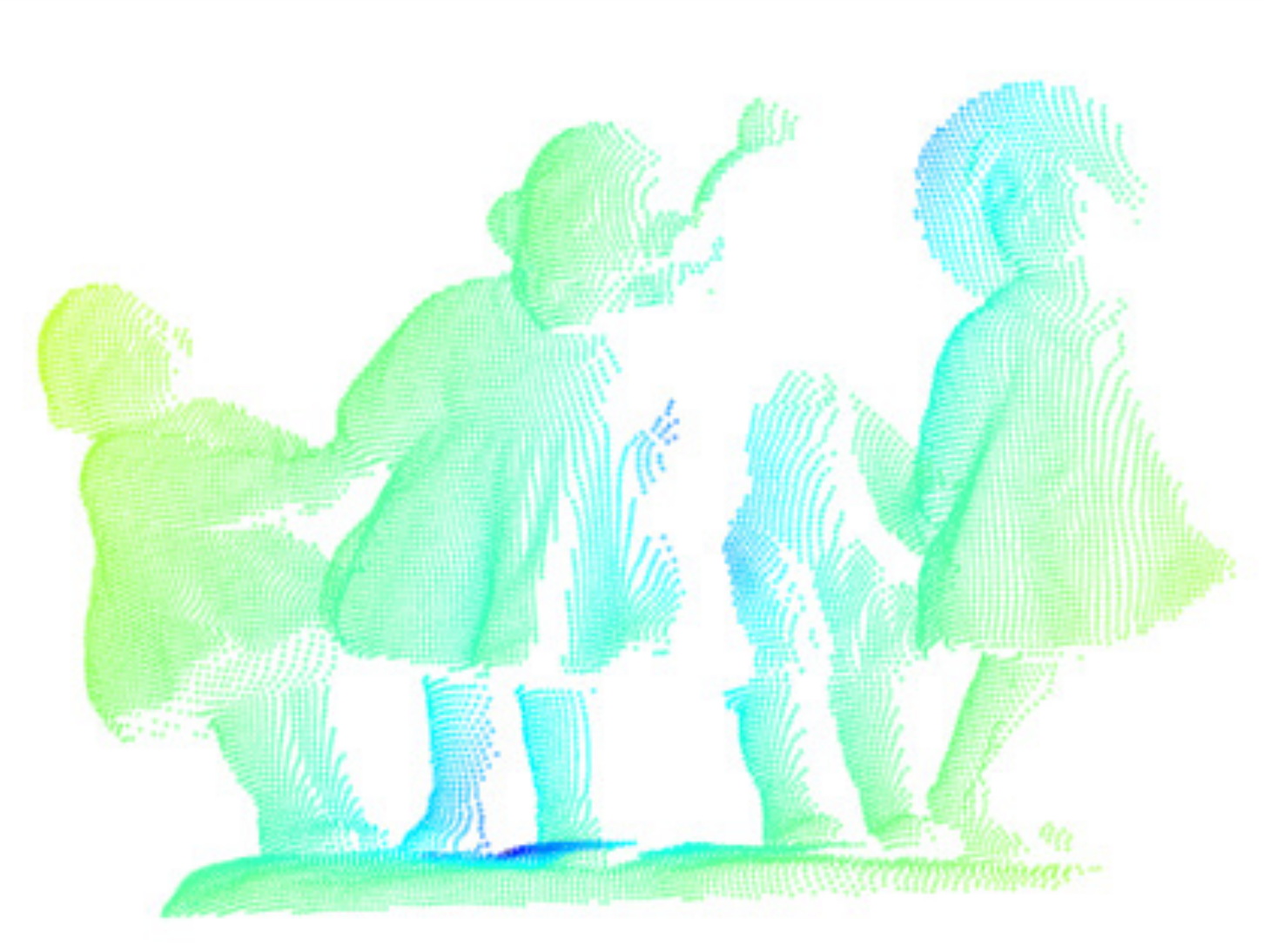}}
  \end{minipage} 
  & \begin{minipage}[b]{0.184\columnwidth}
    \centering
    \raisebox{-.5\height}{\includegraphics[width=\linewidth]{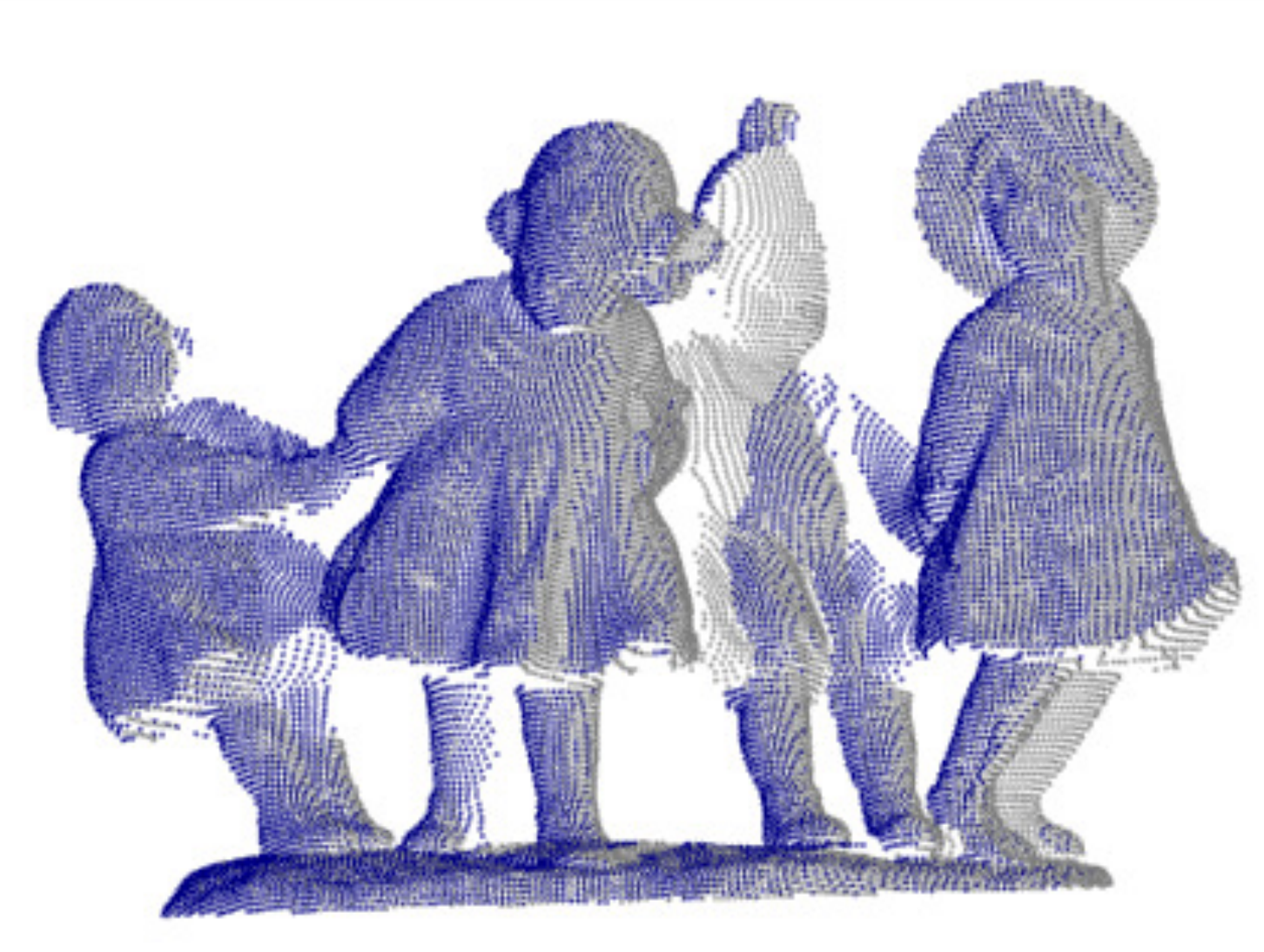}}
  \end{minipage} 
  & \begin{minipage}[b]{0.08\columnwidth}
    \centering
    \raisebox{-.5\height}{\includegraphics[width=\linewidth]{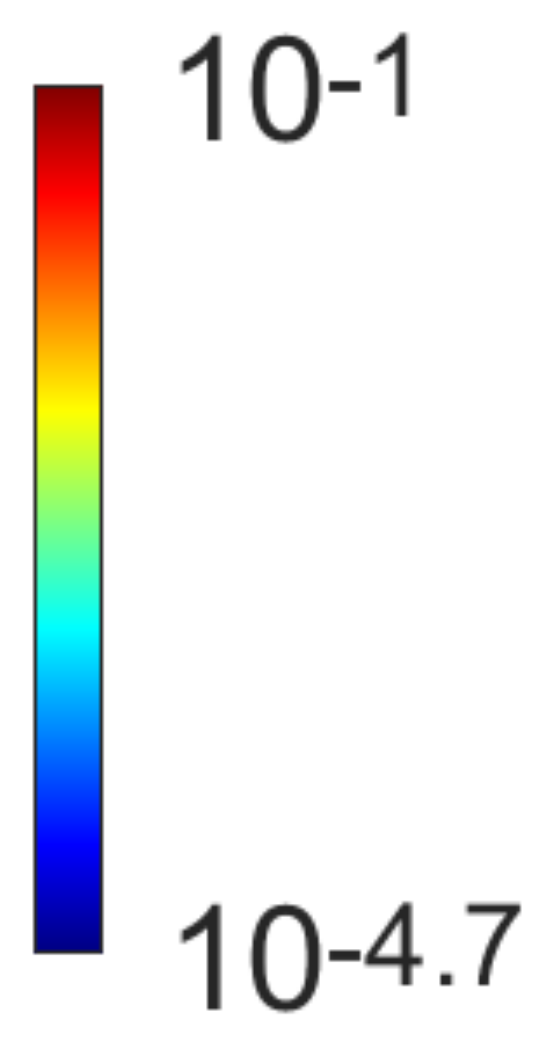}}
  \end{minipage} 
  \\
  \end{tabular}}
  \caption{Visualized registration results on object-scale example instances. 
  The RMSE is reflected by the error map using log-scale color-coding. 
  The more blue areas in visualized point cloud means the better alignment and the more red areas means the worse alignment.
  The visualized results from top to bottom are Armadillo, Bunny, Dragon, Happy, Chef, Chicken, P, T-rex, Angel, Bimba, CD and DC.
  \label{fig:vslz_object}}
\end{figure}

\begin{figure}[!t]
  \centering
  \Huge
  \resizebox{0.9\columnwidth}{!}{
  \begin{tabular}{cccc}
    \textbf{FGR} & \textbf{TrICP$^+$} & \textbf{EMTR-SSC} &
  \\
  \begin{minipage}[b]{1\linewidth}
    \centering
    \raisebox{-.5\height}{\includegraphics[width=\linewidth]{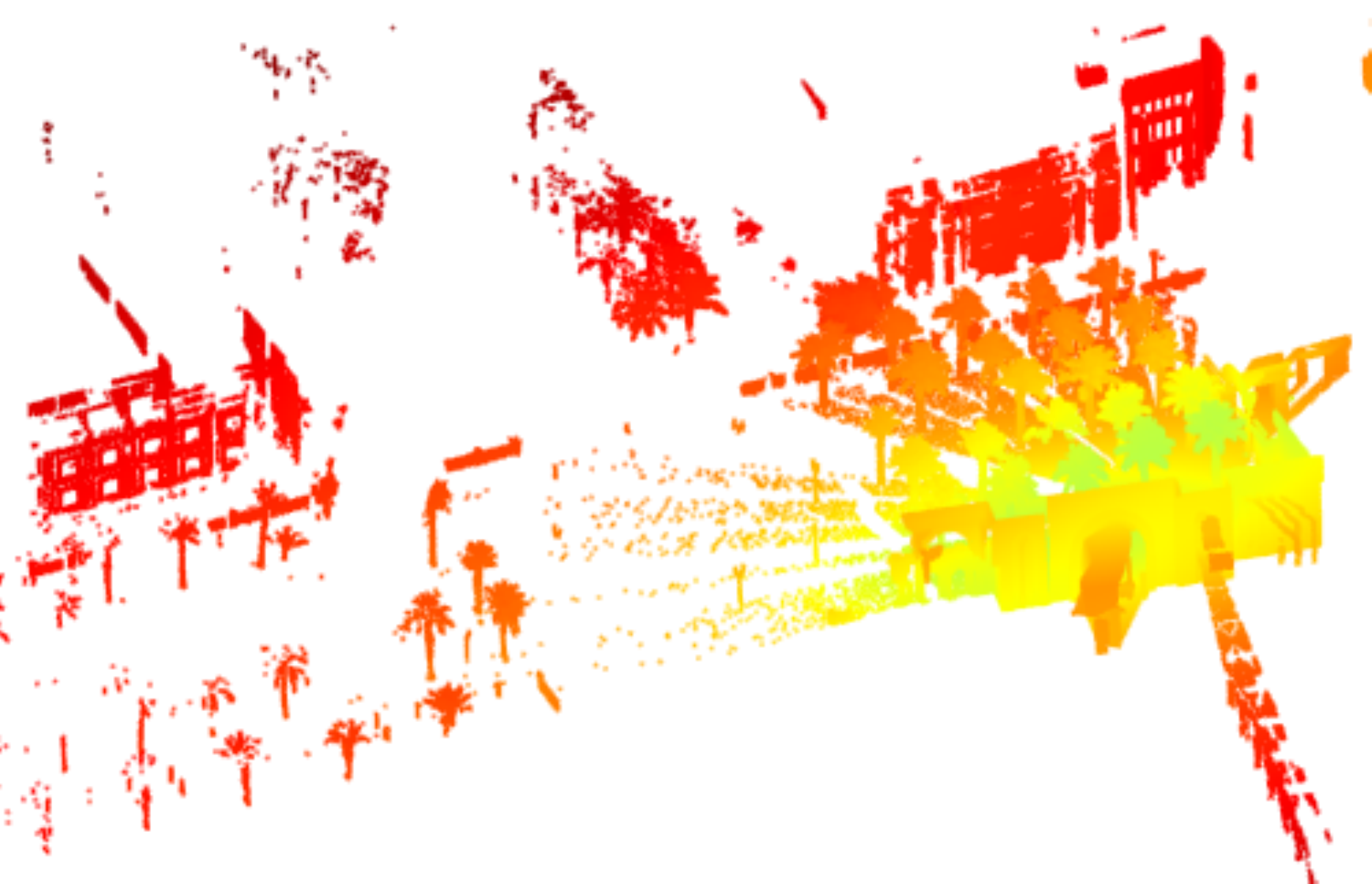}}
  \end{minipage}
  & \begin{minipage}[b]{1\linewidth}
    \centering
    \raisebox{-.5\height}{\includegraphics[width=\linewidth]{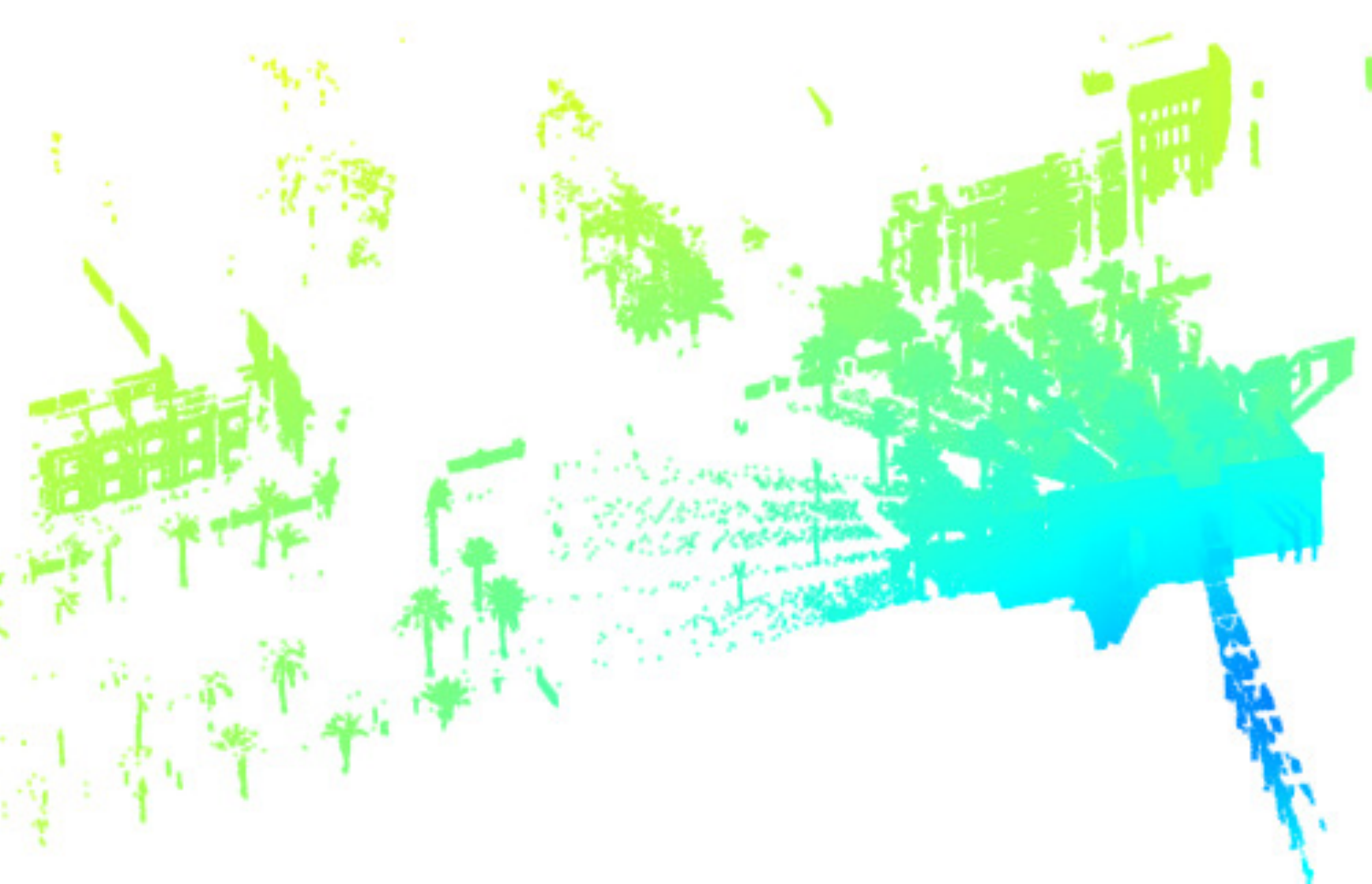}}
  \end{minipage} 
  & \begin{minipage}[b]{1\linewidth}
    \centering
    \raisebox{-.5\height}{\includegraphics[width=\linewidth]{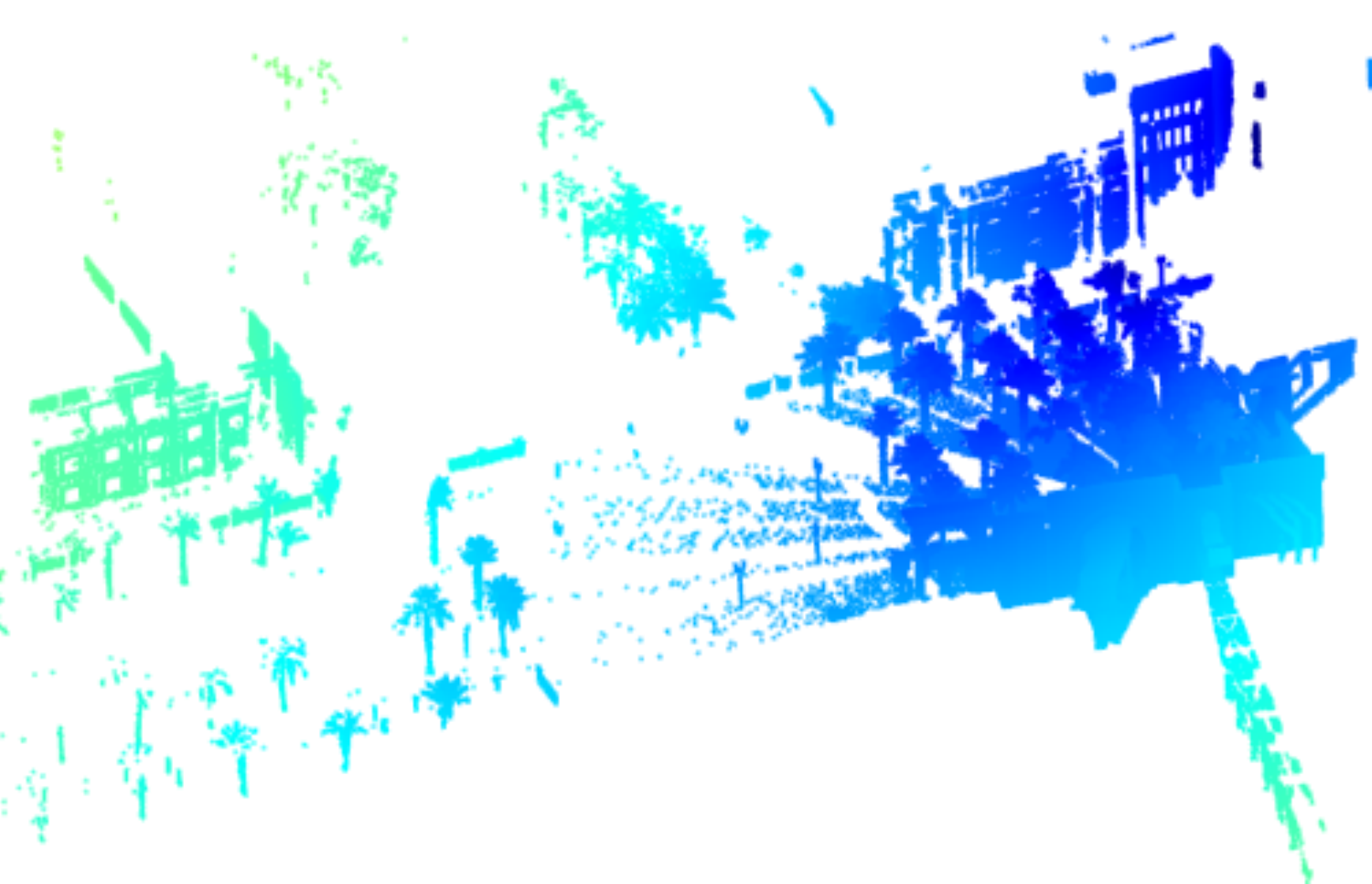}}
  \end{minipage} 
  & \begin{minipage}[b]{0.3\linewidth}
    \centering
    \raisebox{-.5\height}{\includegraphics[width=\linewidth]{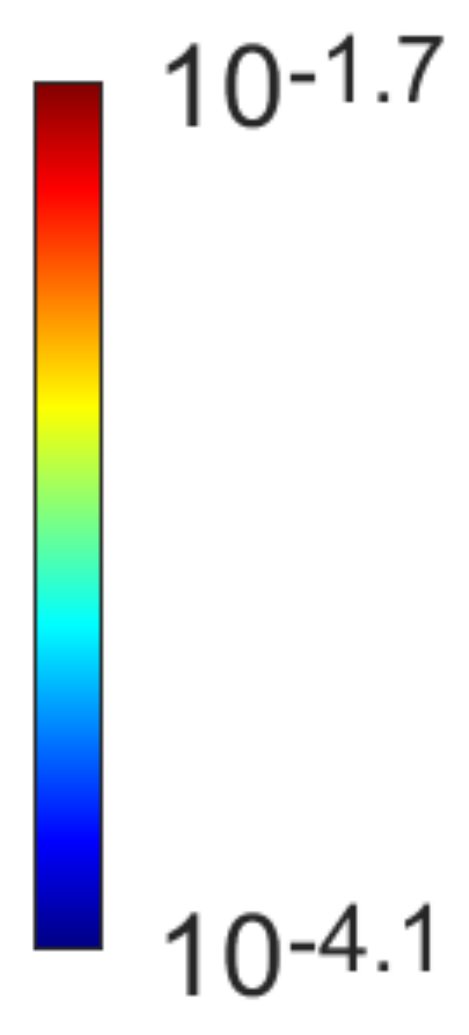}}
  \end{minipage} 
  \\
  \begin{minipage}[b]{1\linewidth}
    \centering
    \raisebox{-.5\height}{\includegraphics[width=\linewidth]{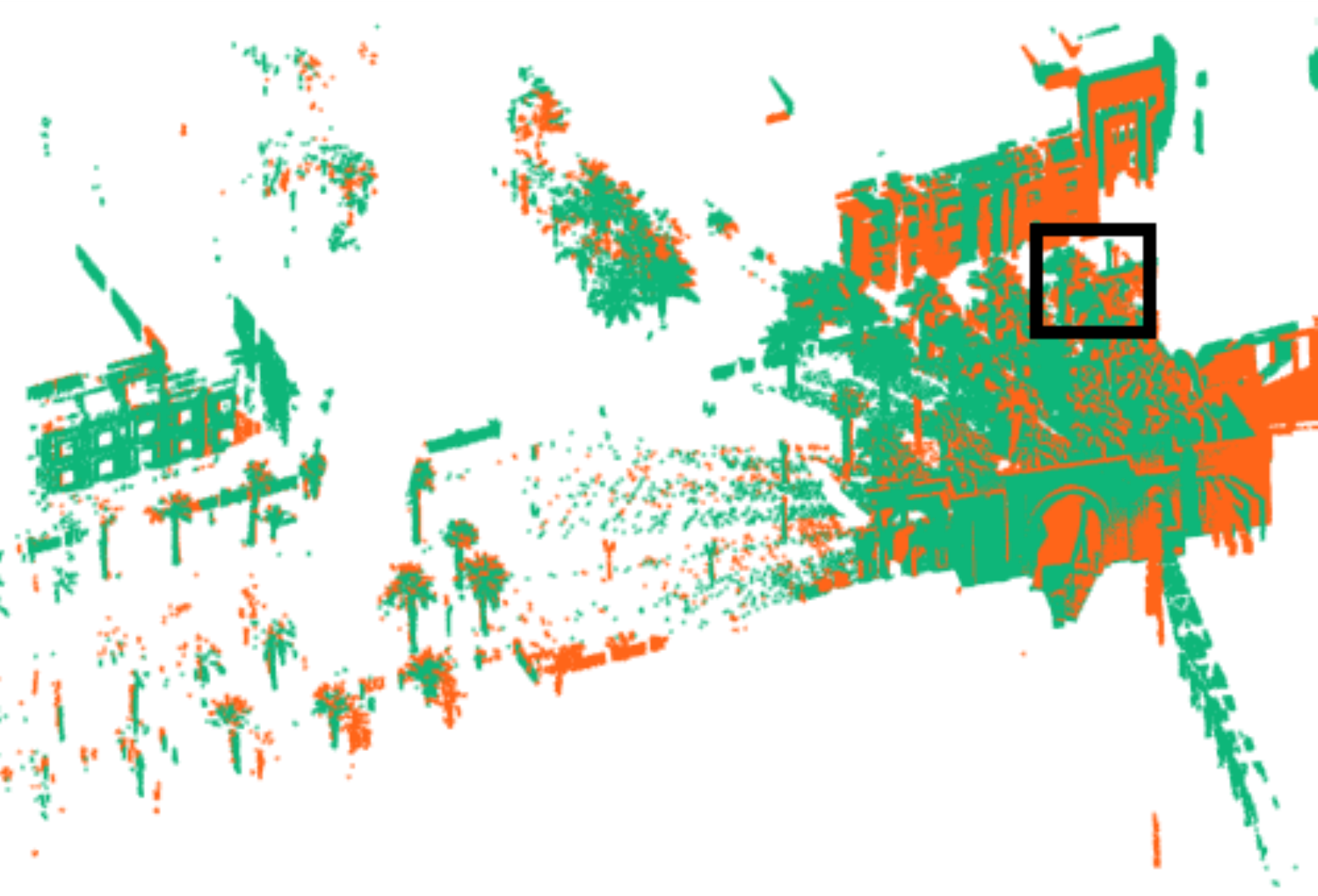}}
  \end{minipage} 
  & \begin{minipage}[b]{1\linewidth}
    \centering
    \raisebox{-.5\height}{\includegraphics[width=\linewidth]{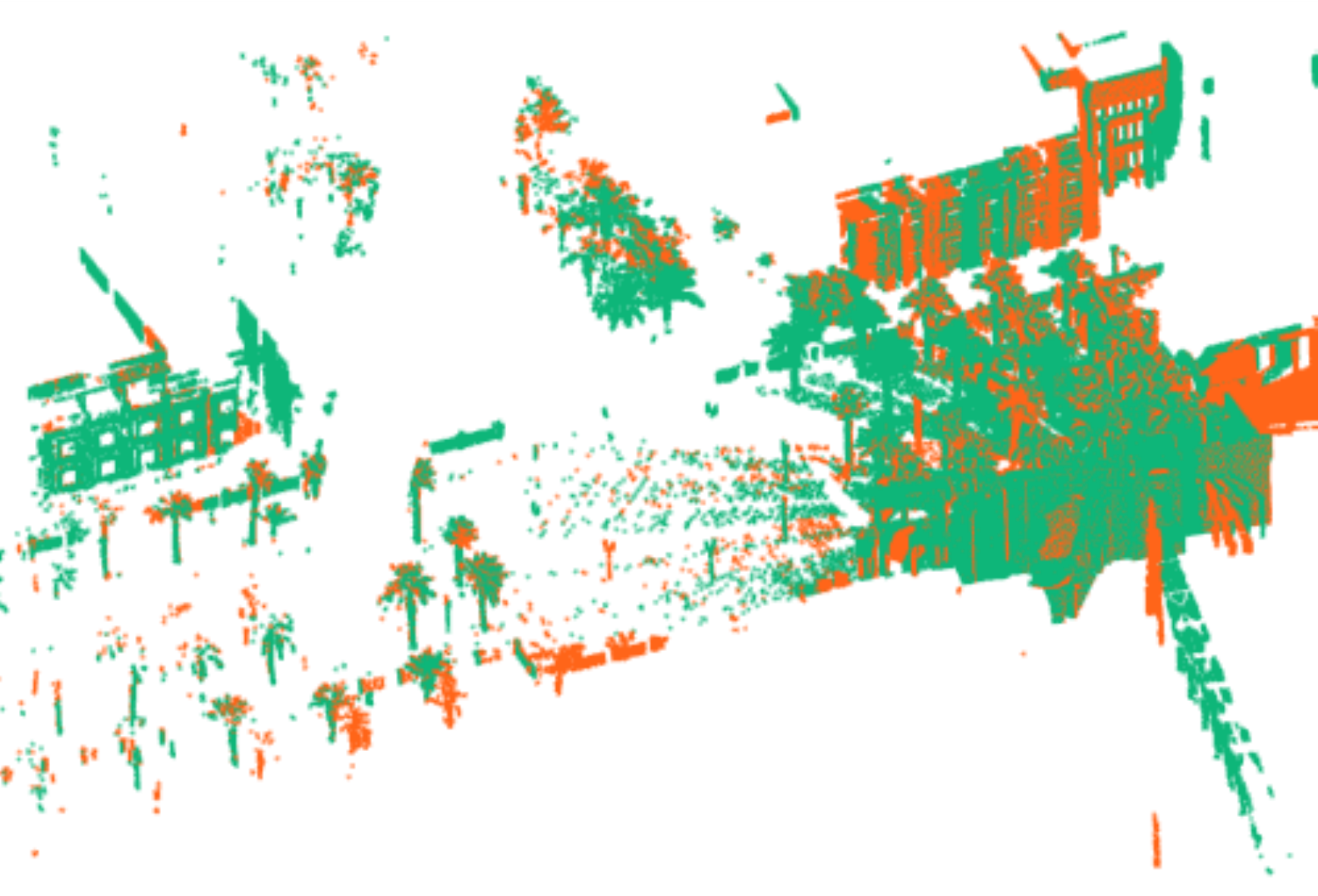}}
  \end{minipage} 
  & \begin{minipage}[b]{1\linewidth}
    \centering
    \raisebox{-.5\height}{\includegraphics[width=\linewidth]{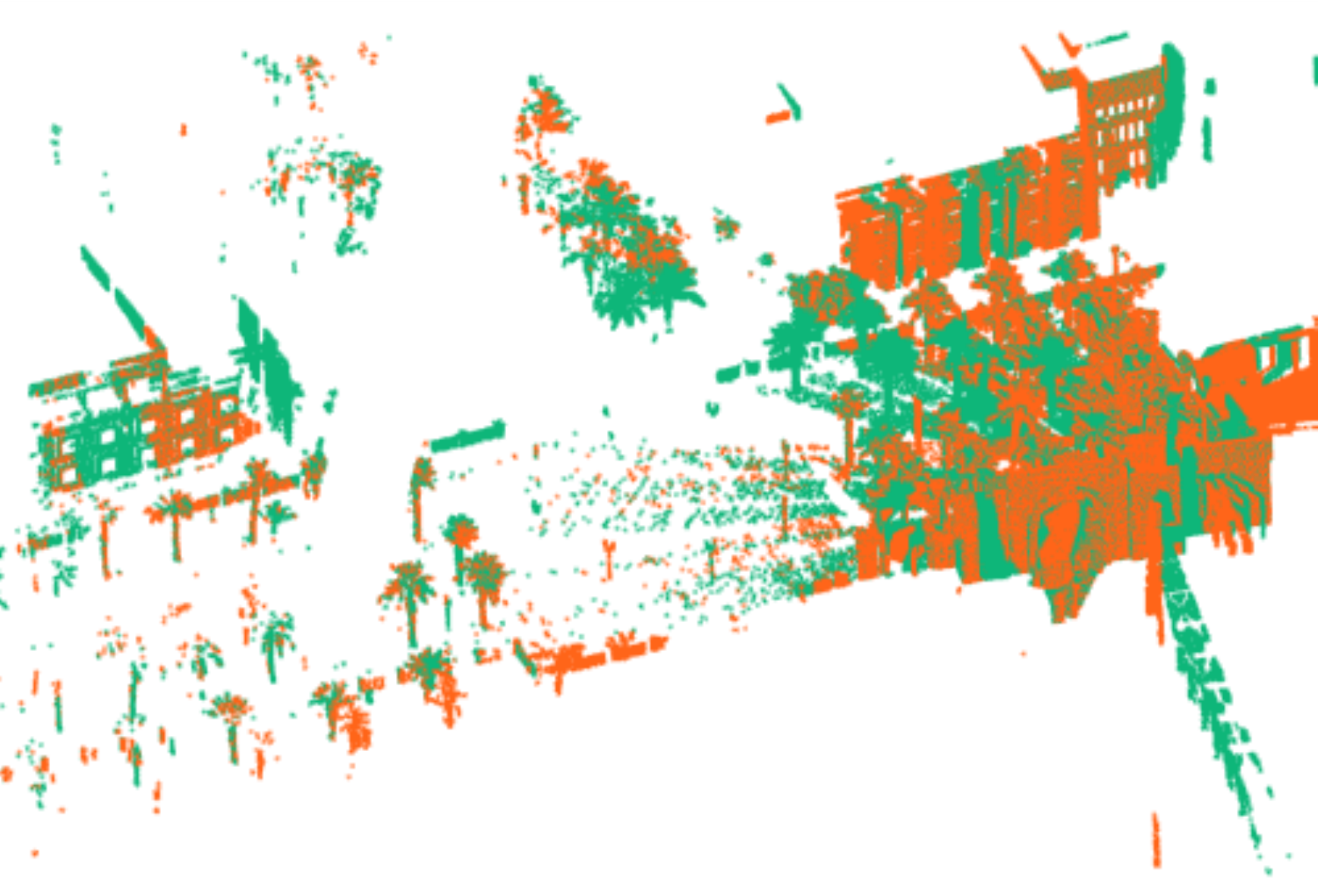}}
  \end{minipage} 
  &
  \textbf{7a}
  \\
    \begin{minipage}[b]{1\linewidth}
      \centering
      \raisebox{-.5\height}{\includegraphics[width=\linewidth]{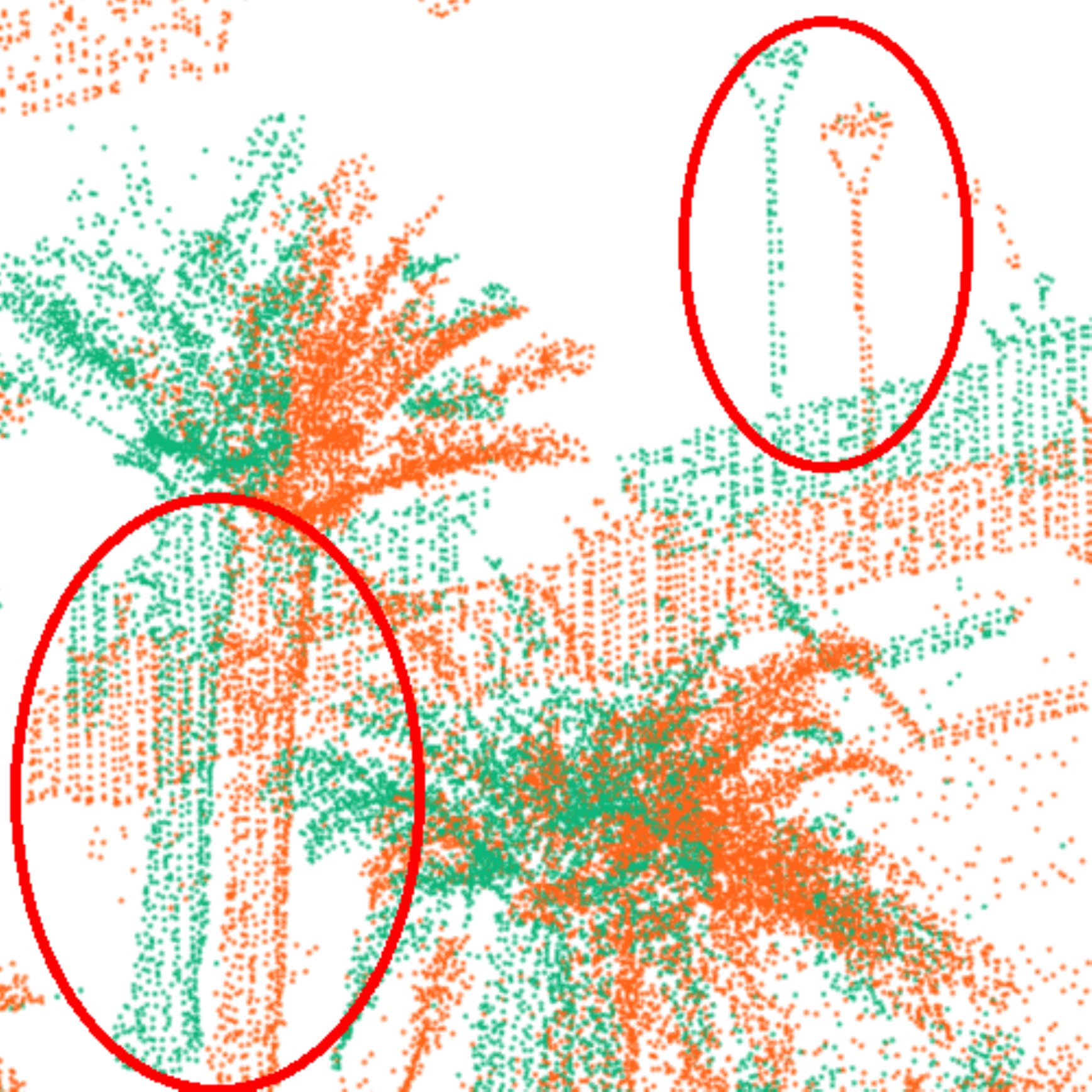}}
    \end{minipage}
  & 
    \begin{minipage}[b]{1\linewidth}
      \centering
      \raisebox{-.5\height}{\includegraphics[width=\linewidth]{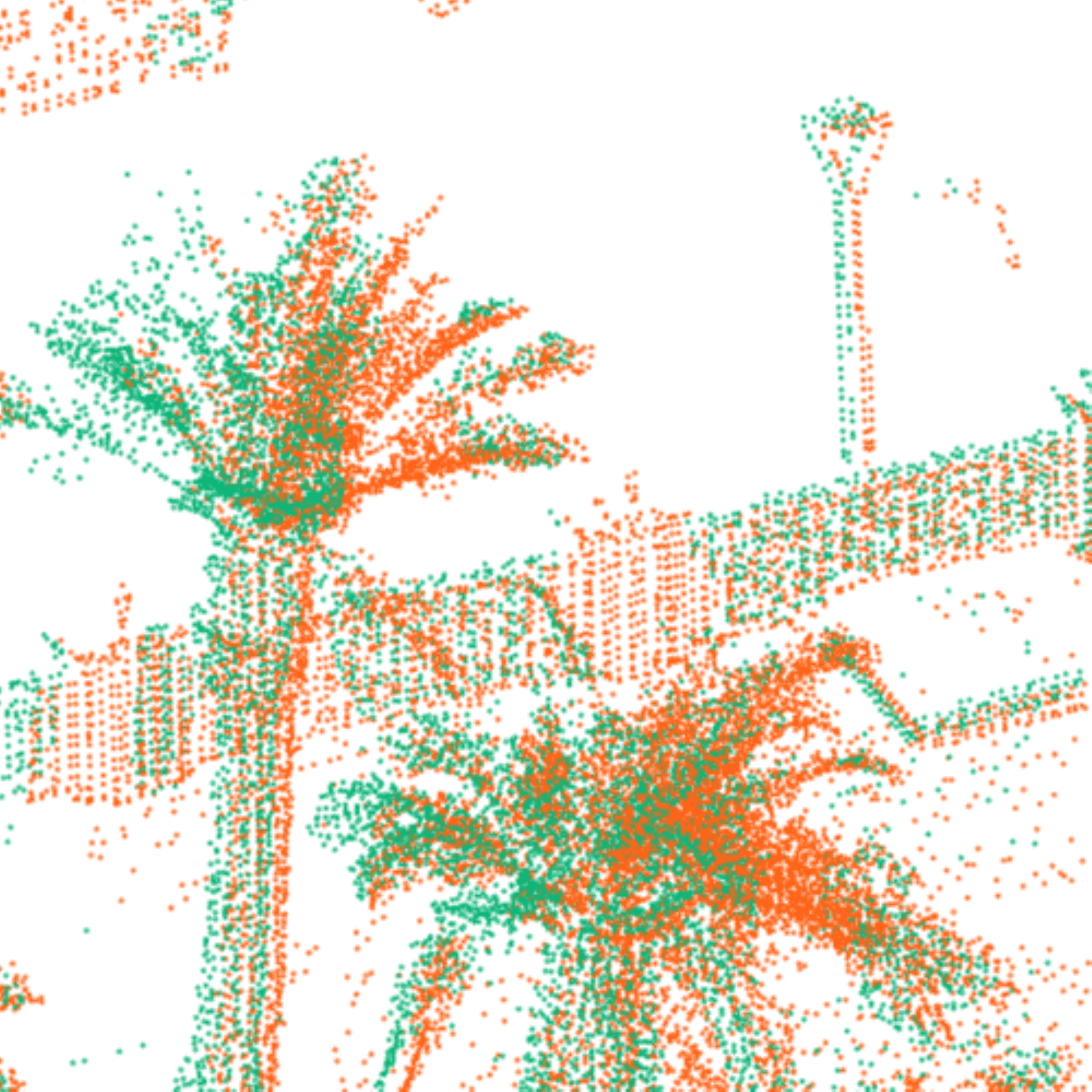}}
    \end{minipage}
  & 
    \begin{minipage}[b]{1\linewidth}
      \centering
      \raisebox{-.5\height}{\includegraphics[width=\linewidth]{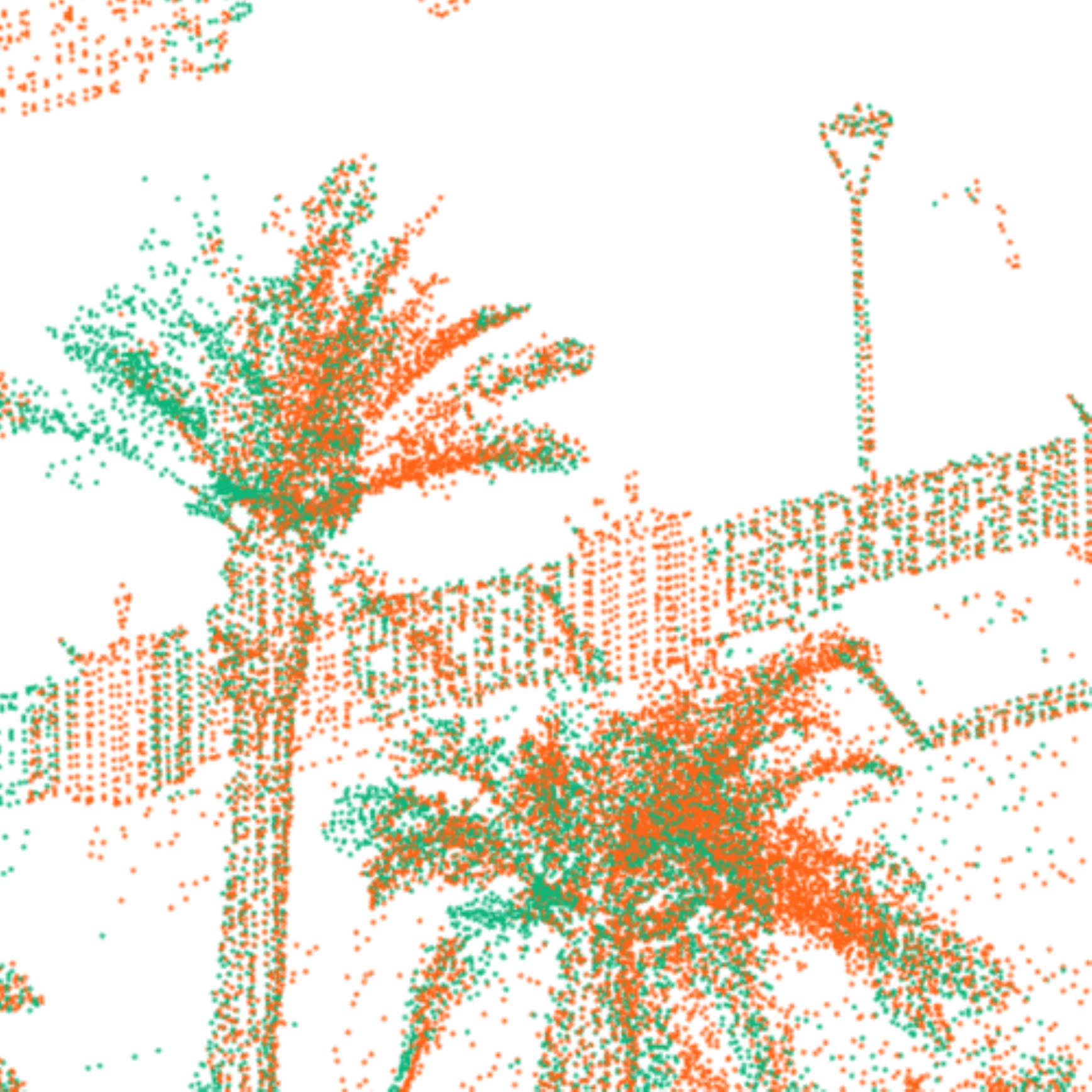}}
    \end{minipage}
    &
  \\
  \\
  \hdashline
  \\

  \begin{minipage}[b]{1\linewidth}
    \centering
    \raisebox{-.5\height}{\includegraphics[width=\linewidth]{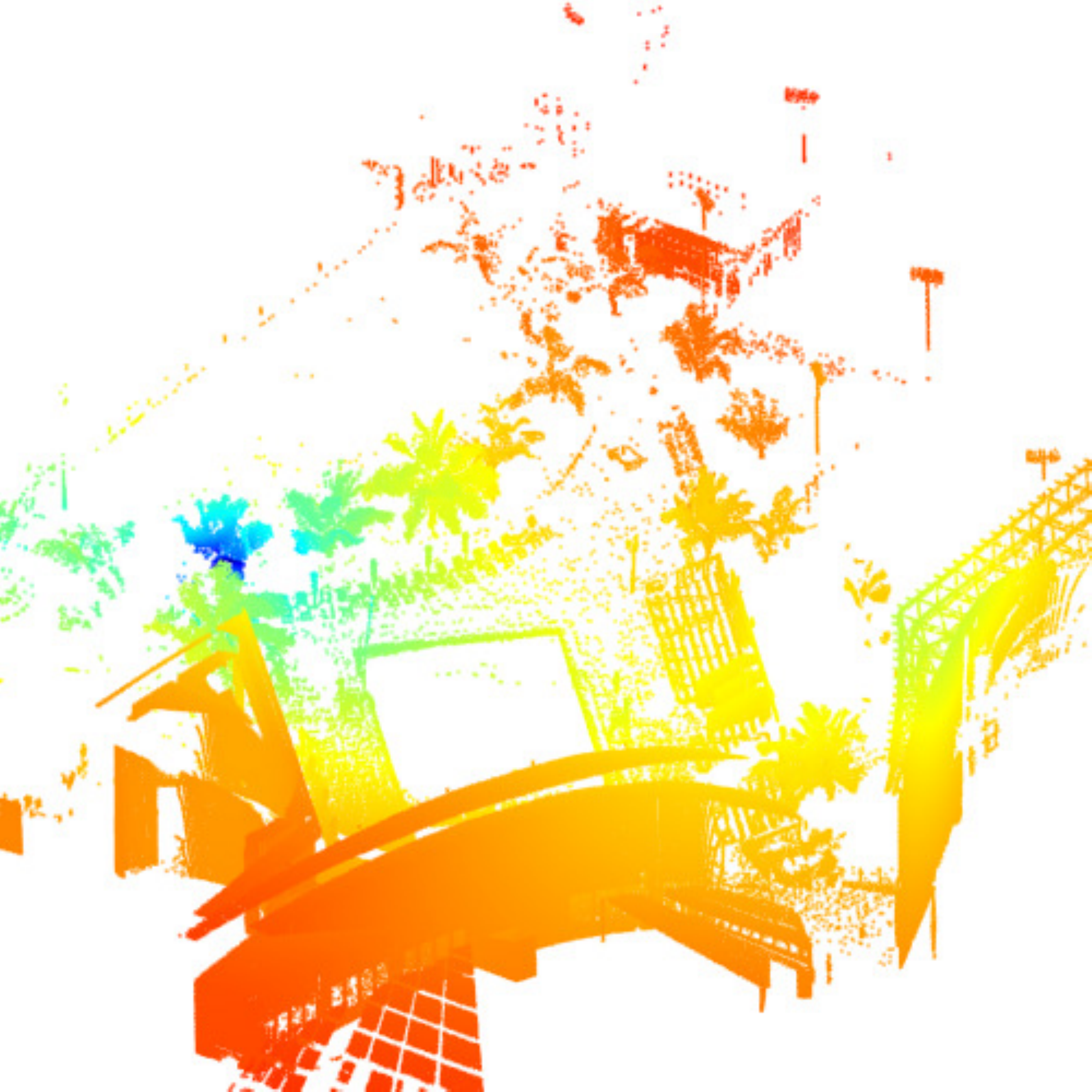}}
  \end{minipage}
  & \begin{minipage}[b]{1\linewidth}
    \centering
    \raisebox{-.5\height}{\includegraphics[width=\linewidth]{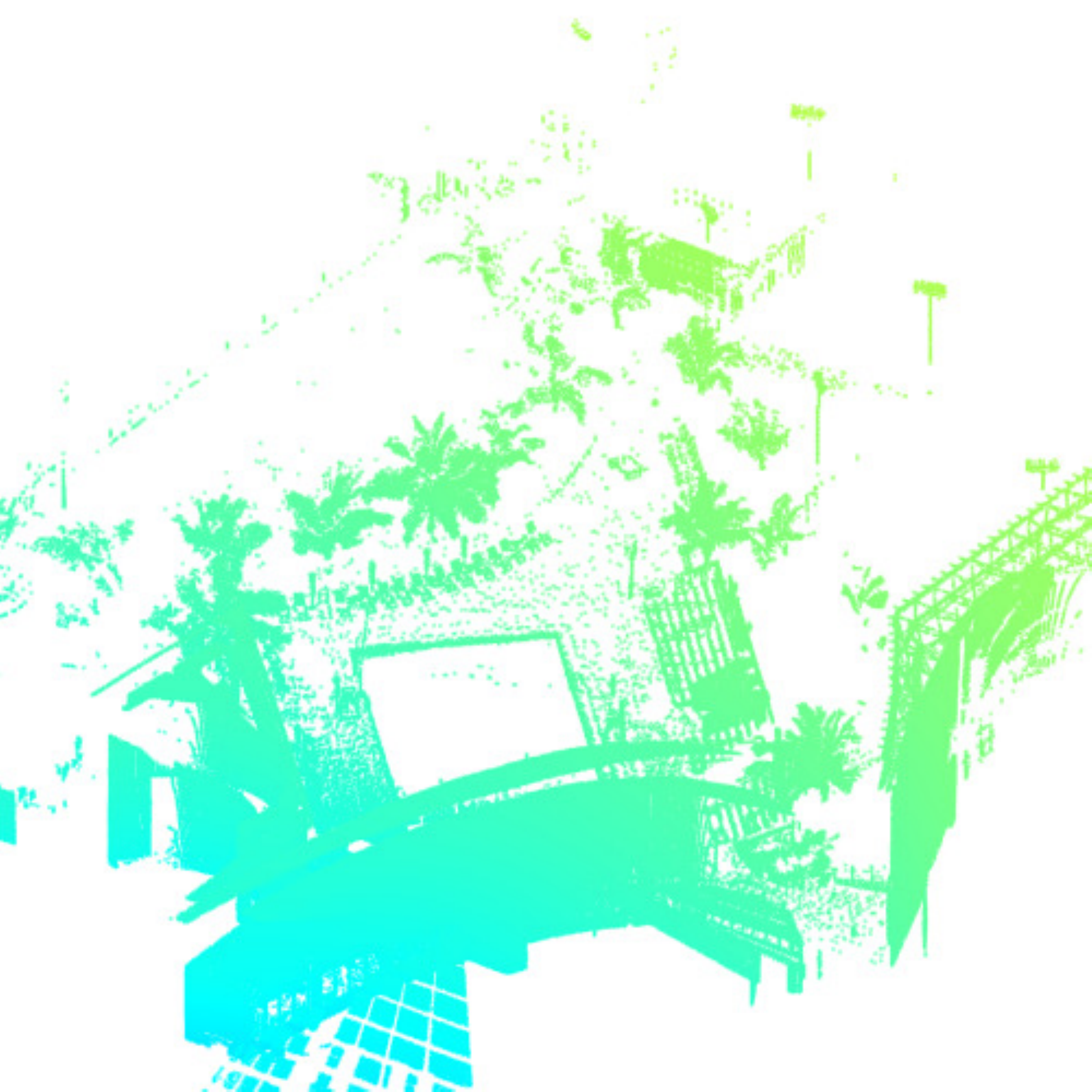}}
  \end{minipage} 
  & \begin{minipage}[b]{1\linewidth}
    \centering
    \raisebox{-.5\height}{\includegraphics[width=\linewidth]{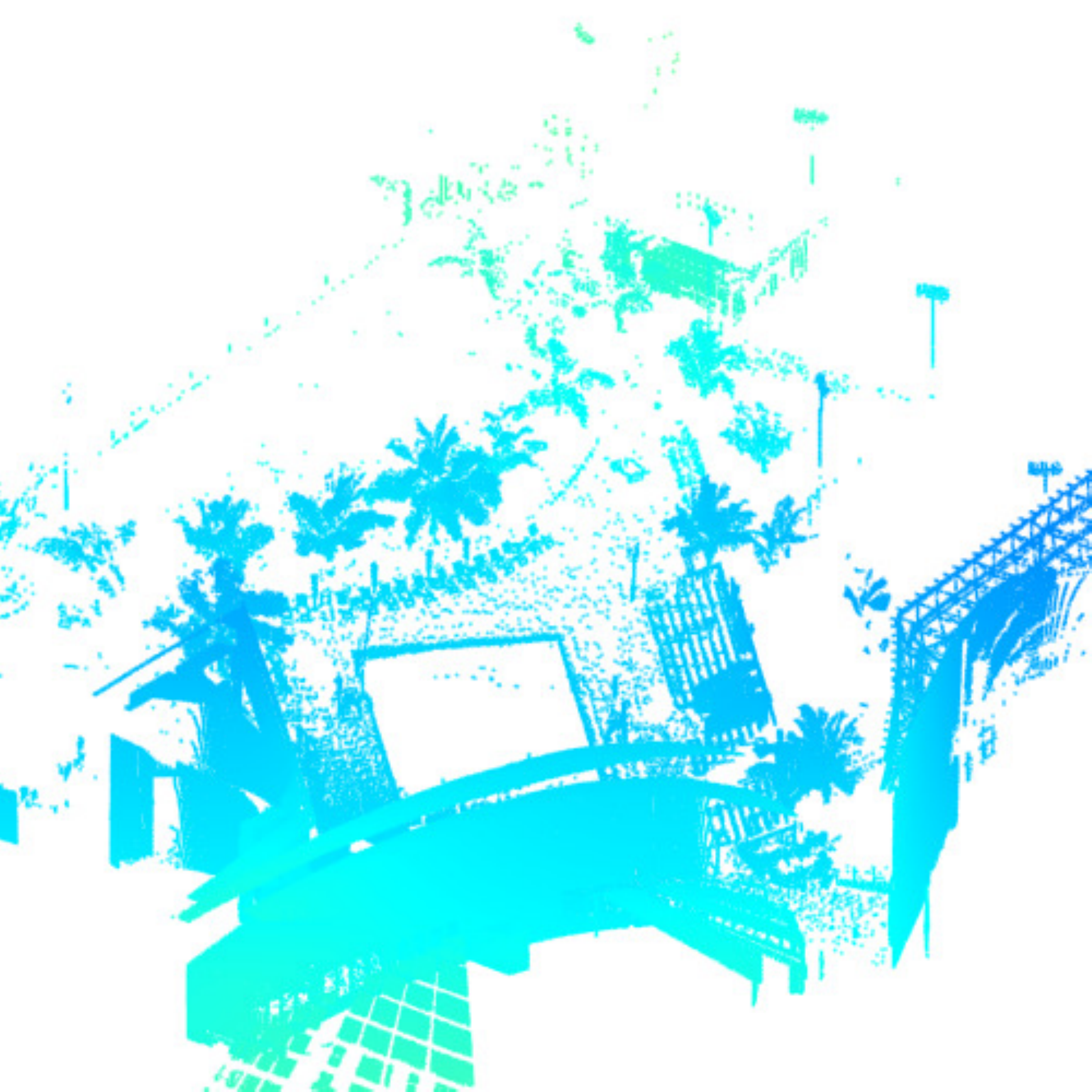}}
  \end{minipage} 
  & \begin{minipage}[b]{0.3\linewidth}
    \centering
    \raisebox{-.5\height}{\includegraphics[width=\linewidth]{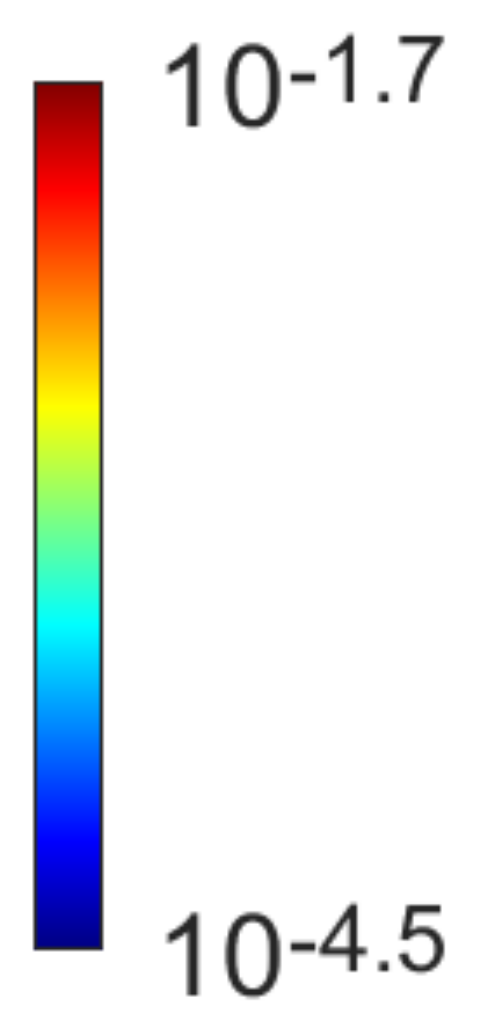}}
  \end{minipage} 
  \\
  \begin{minipage}[b]{1\linewidth}
    \centering
    \raisebox{-.5\height}{\includegraphics[width=\linewidth]{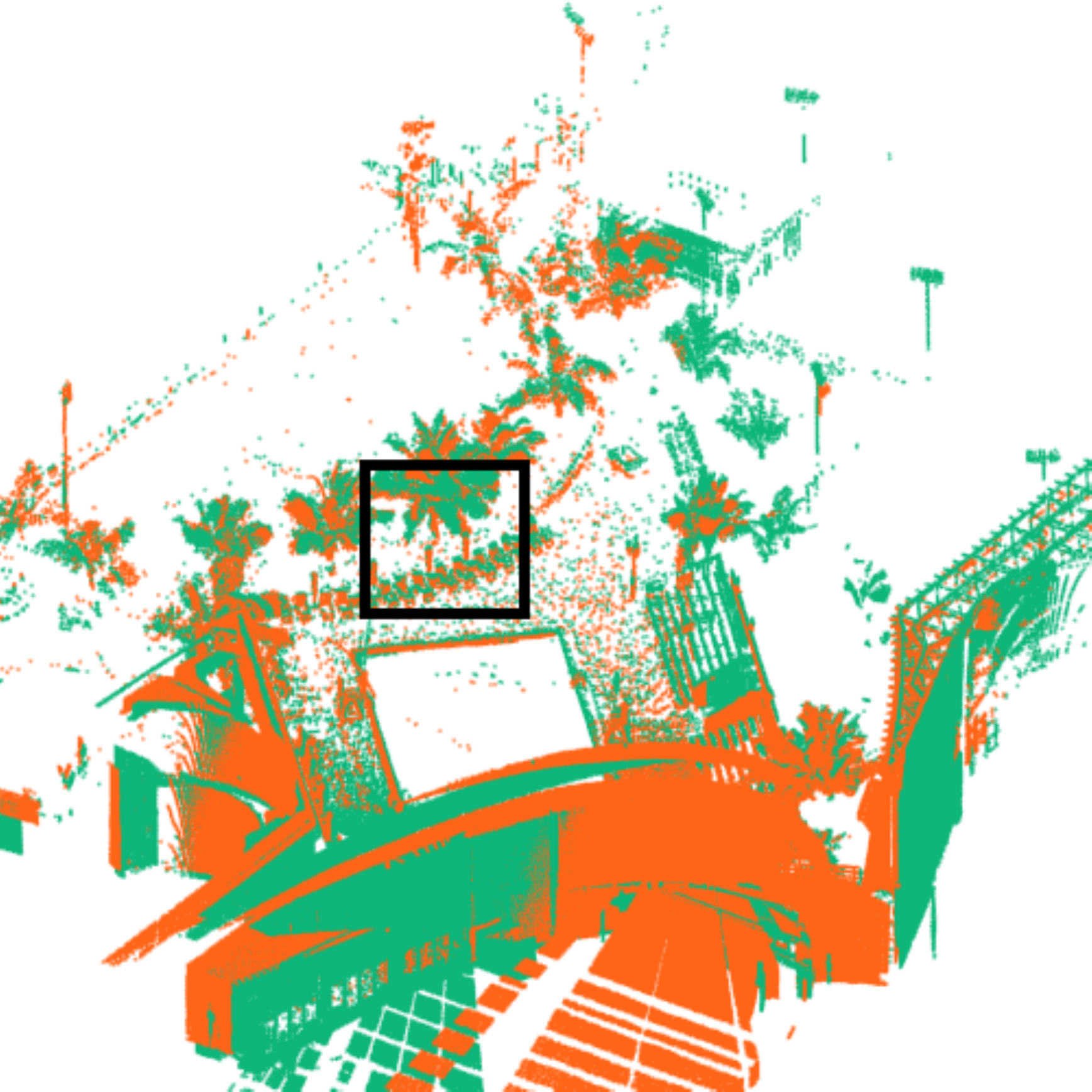}}
  \end{minipage} 
  & \begin{minipage}[b]{1\linewidth}
    \centering
    \raisebox{-.5\height}{\includegraphics[width=\linewidth]{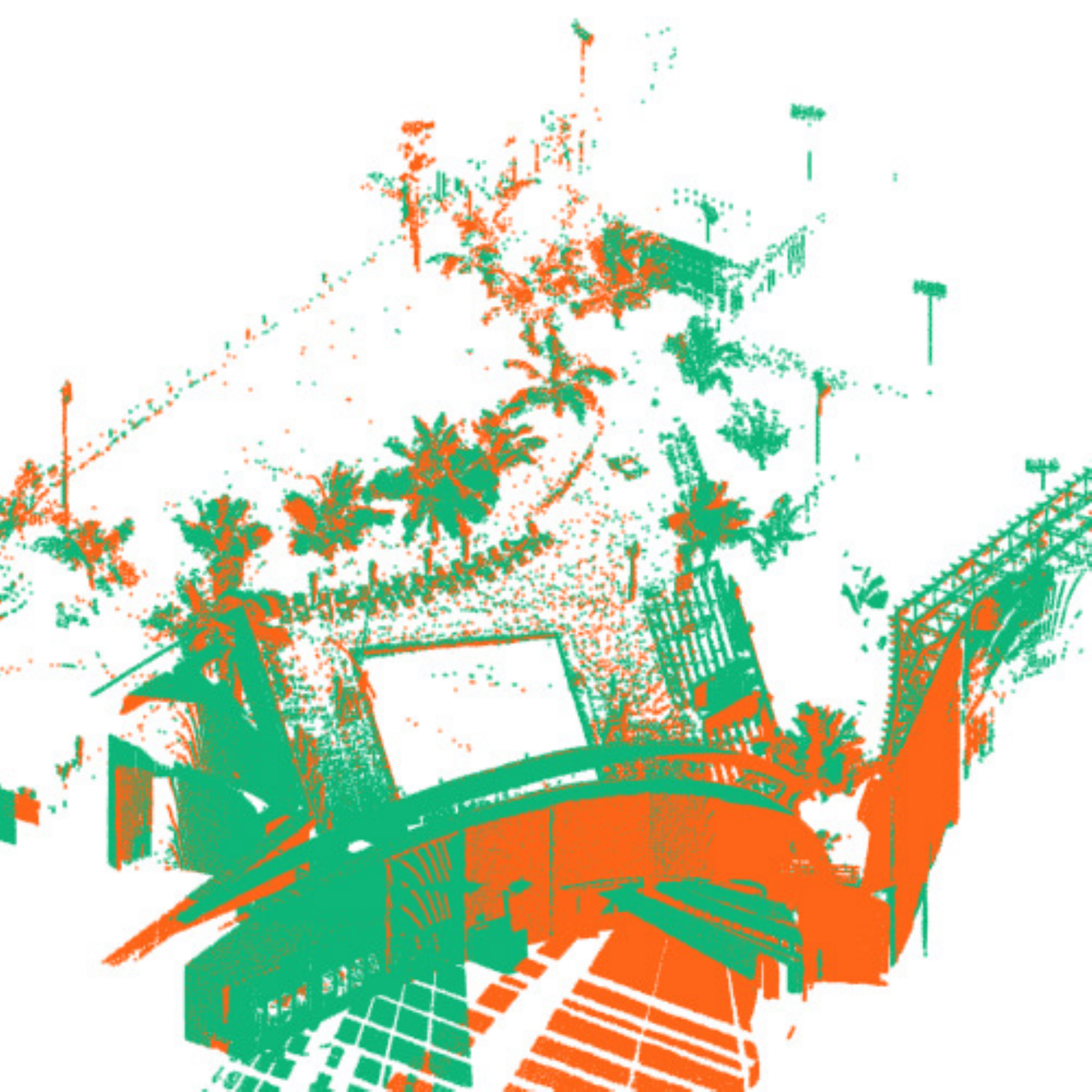}}
  \end{minipage} 
  & \begin{minipage}[b]{1\linewidth}
    \centering
    \raisebox{-.5\height}{\includegraphics[width=\linewidth]{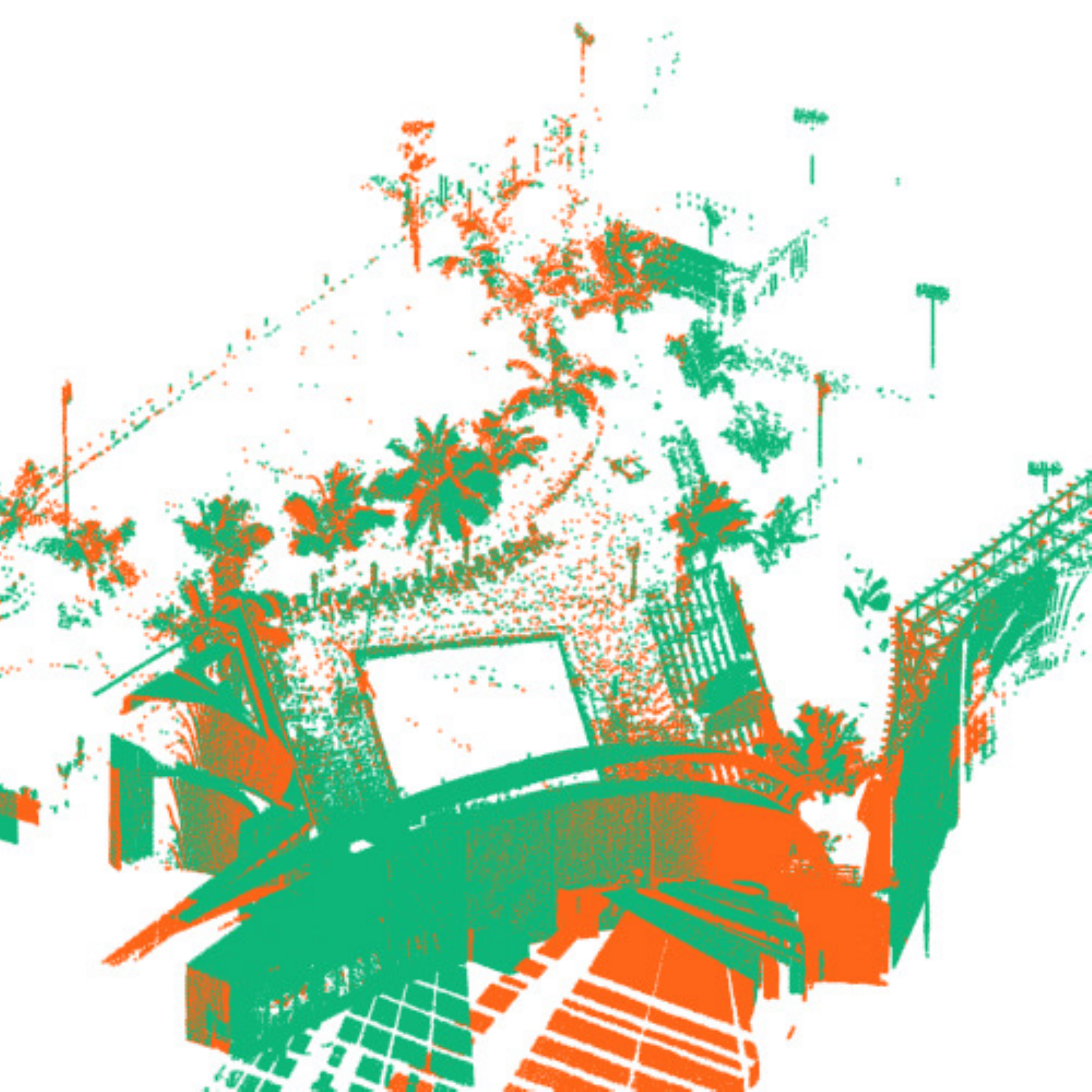}}
  \end{minipage} 
  &
  \textbf{7d}
  \\
    \begin{minipage}[b]{1\linewidth}
      \centering
      \raisebox{-.5\height}{\includegraphics[width=\linewidth]{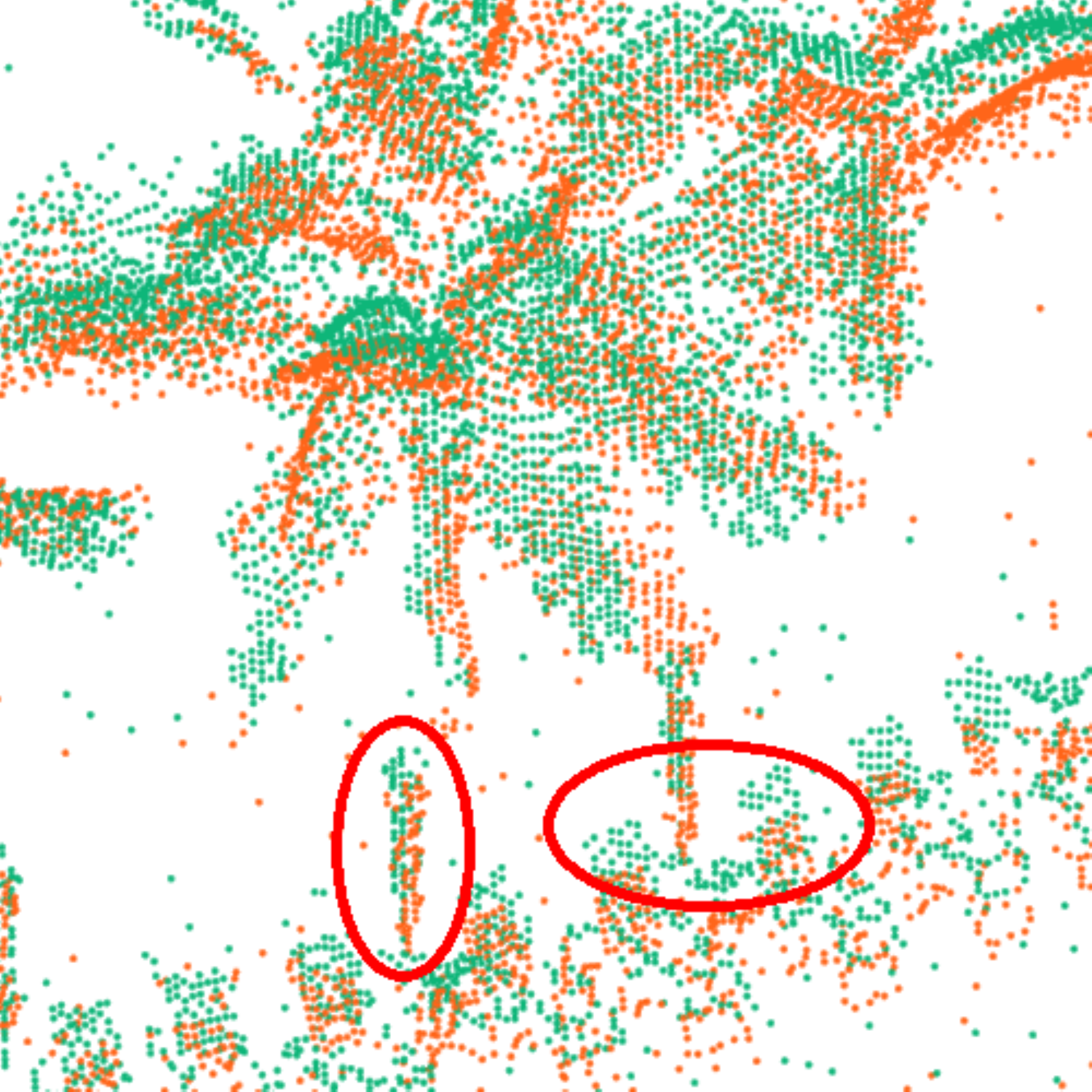}}
    \end{minipage}
  & 
    \begin{minipage}[b]{1\linewidth}
      \centering
      \raisebox{-.5\height}{\includegraphics[width=\linewidth]{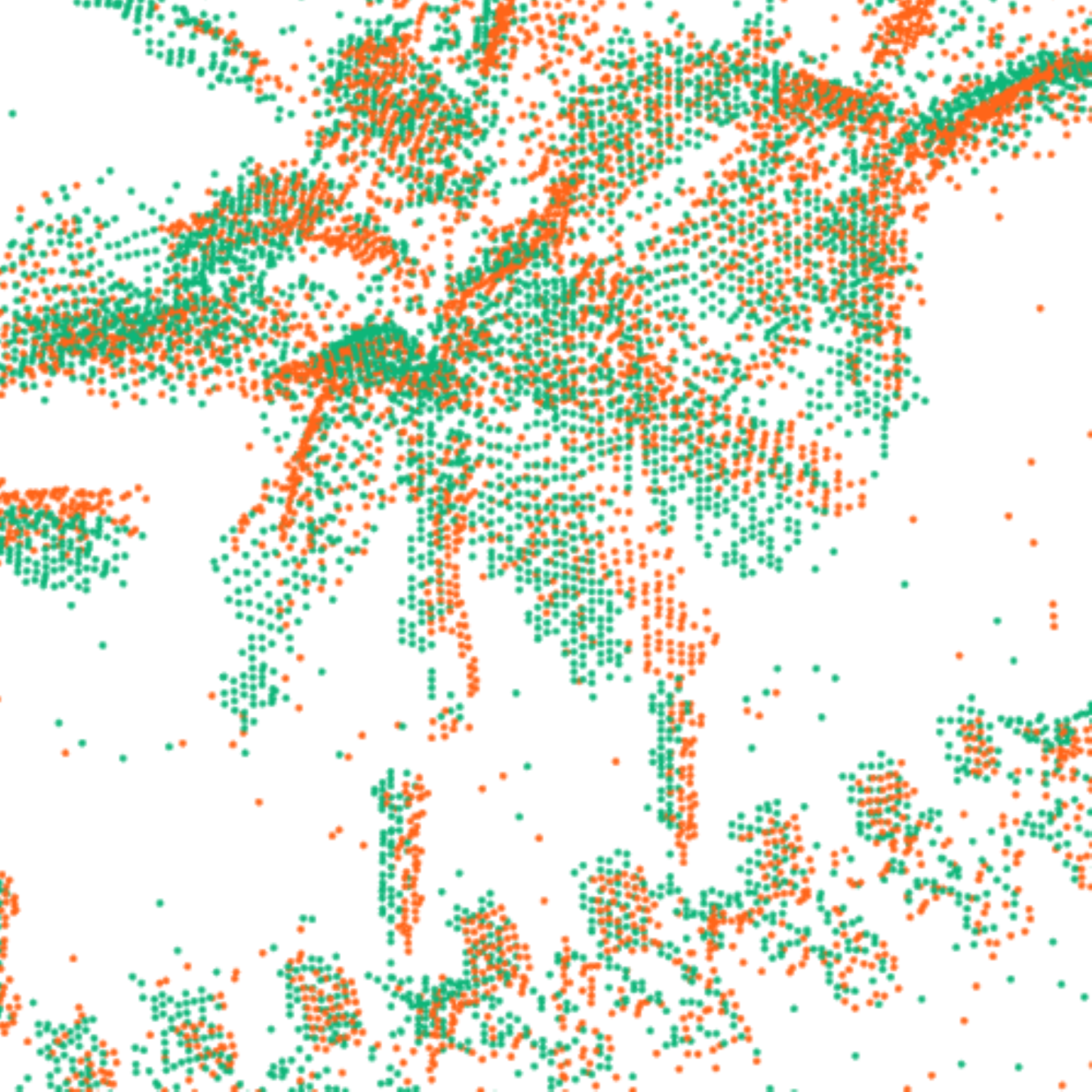}}
    \end{minipage}
  & 
    \begin{minipage}[b]{1\linewidth}
      \centering
      \raisebox{-.5\height}{\includegraphics[width=\linewidth]{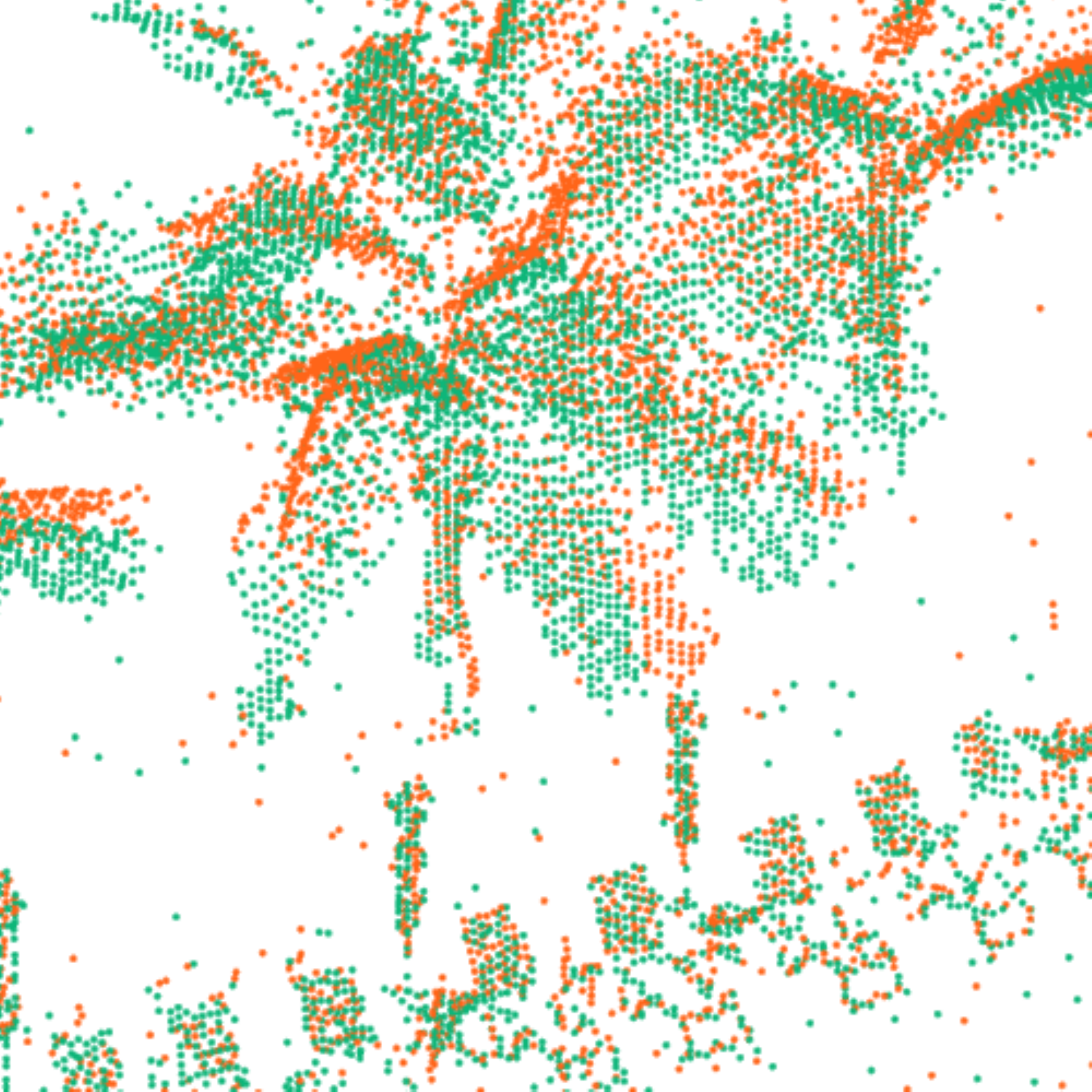}}
    \end{minipage}
    &
  \\
  \\
  \hdashline
  \\
    
  \begin{minipage}[b]{1\linewidth}
    \centering
    \raisebox{-.5\height}{\includegraphics[width=\linewidth]{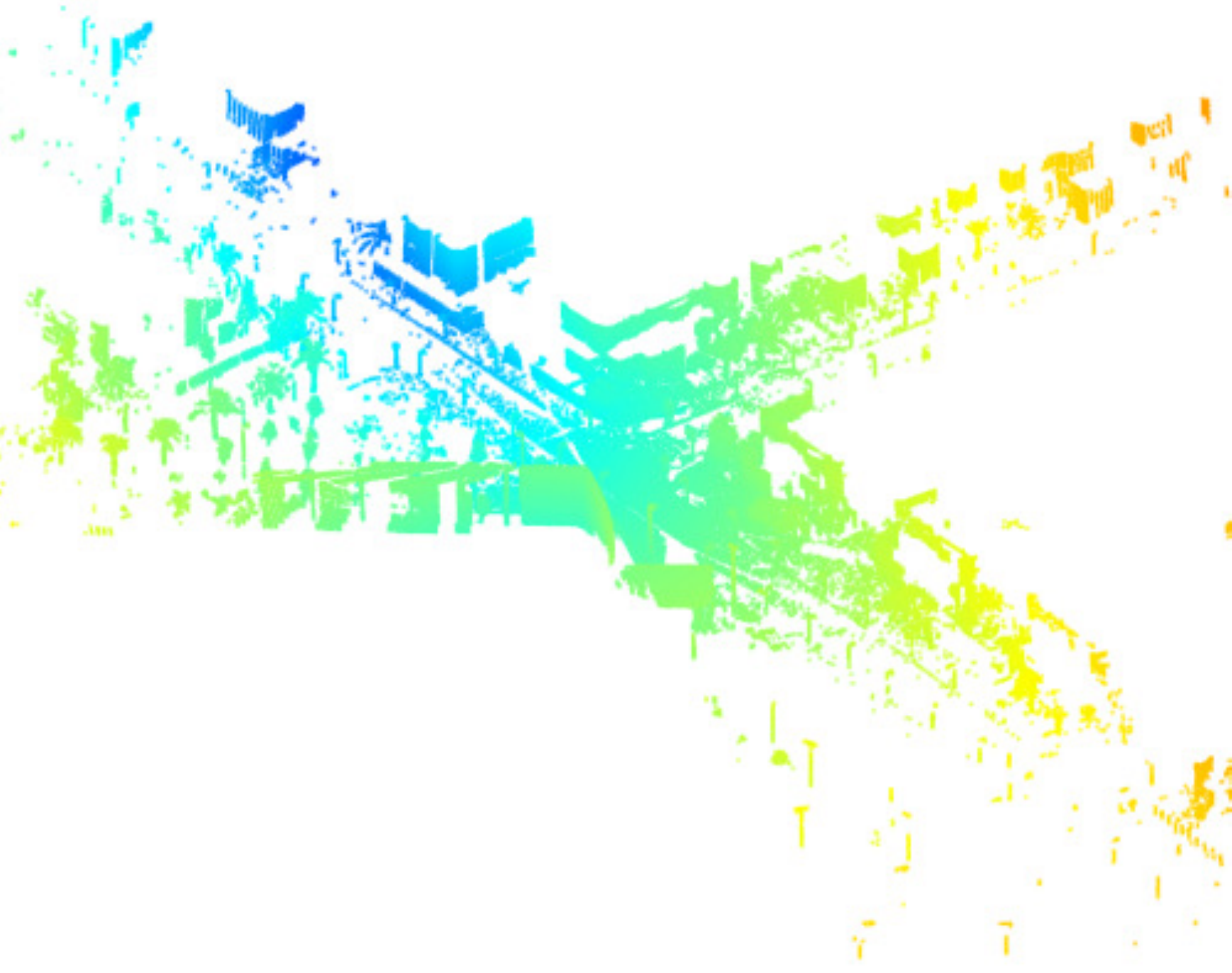}}
  \end{minipage}
  & \begin{minipage}[b]{1\linewidth}
    \centering
    \raisebox{-.5\height}{\includegraphics[width=\linewidth]{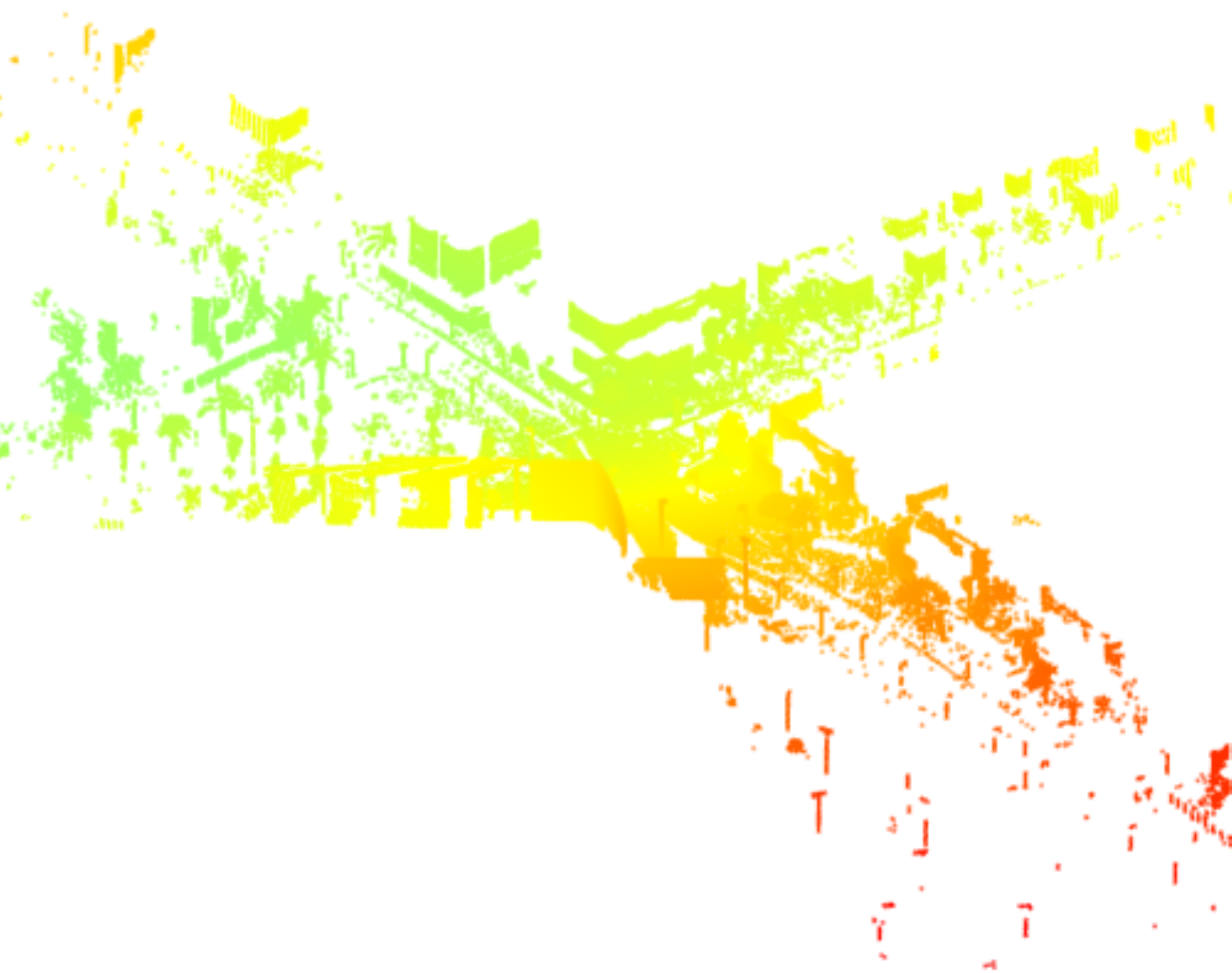}}
  \end{minipage} 
  & \begin{minipage}[b]{1\linewidth}
    \centering
    \raisebox{-.5\height}{\includegraphics[width=\linewidth]{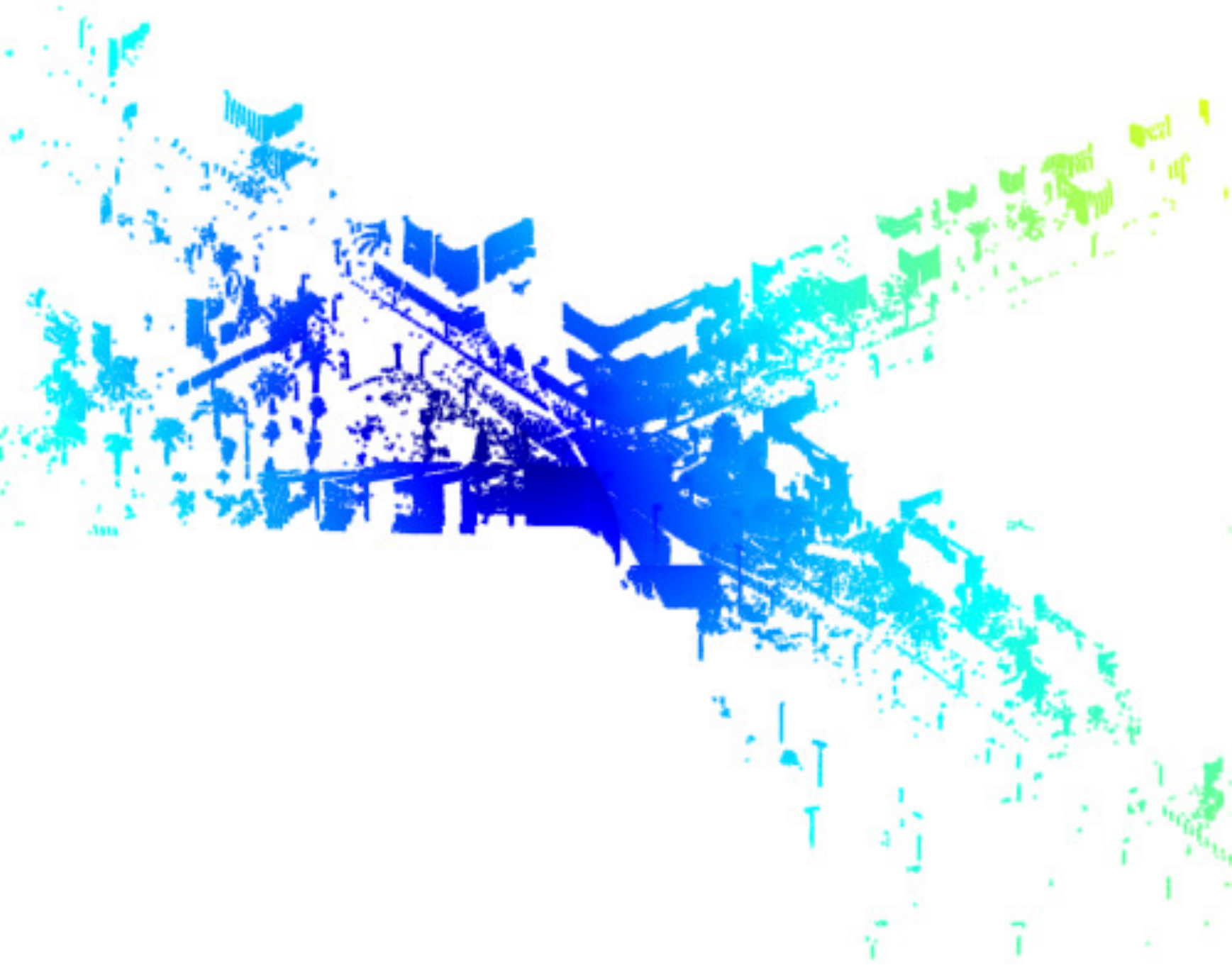}}
  \end{minipage} 
  & \begin{minipage}[b]{0.3\linewidth}
    \centering
    \raisebox{-.5\height}{\includegraphics[width=\linewidth]{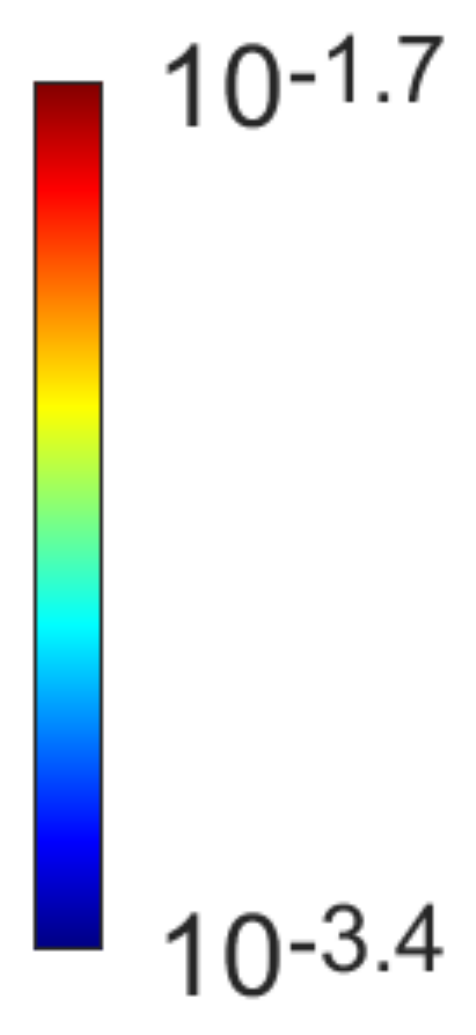}}
  \end{minipage} 
  \\
  \begin{minipage}[b]{1\linewidth}
    \centering
    \raisebox{-.5\height}{\includegraphics[width=\linewidth]{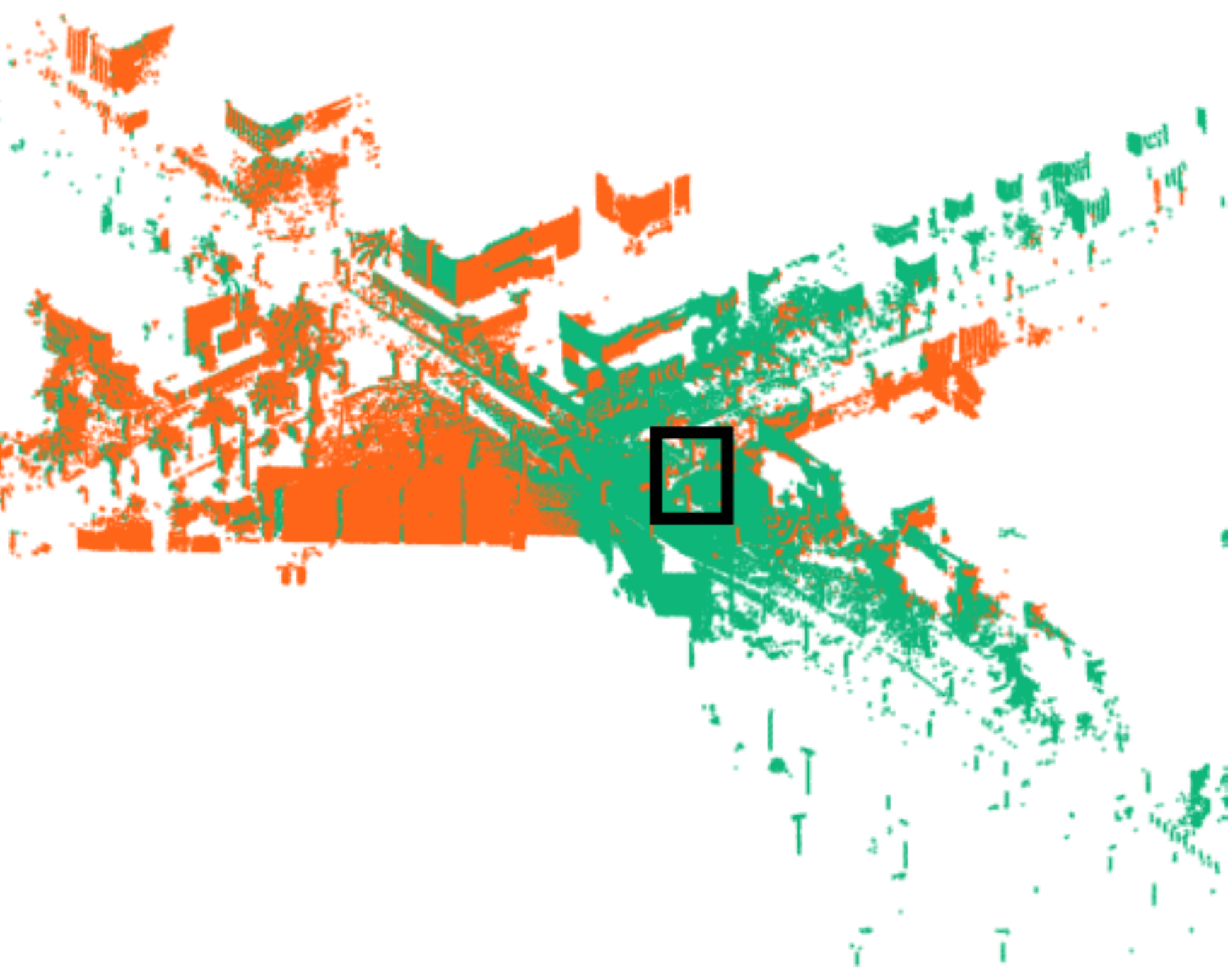}}
  \end{minipage} 
  & \begin{minipage}[b]{1\linewidth}
    \centering
    \raisebox{-.5\height}{\includegraphics[width=\linewidth]{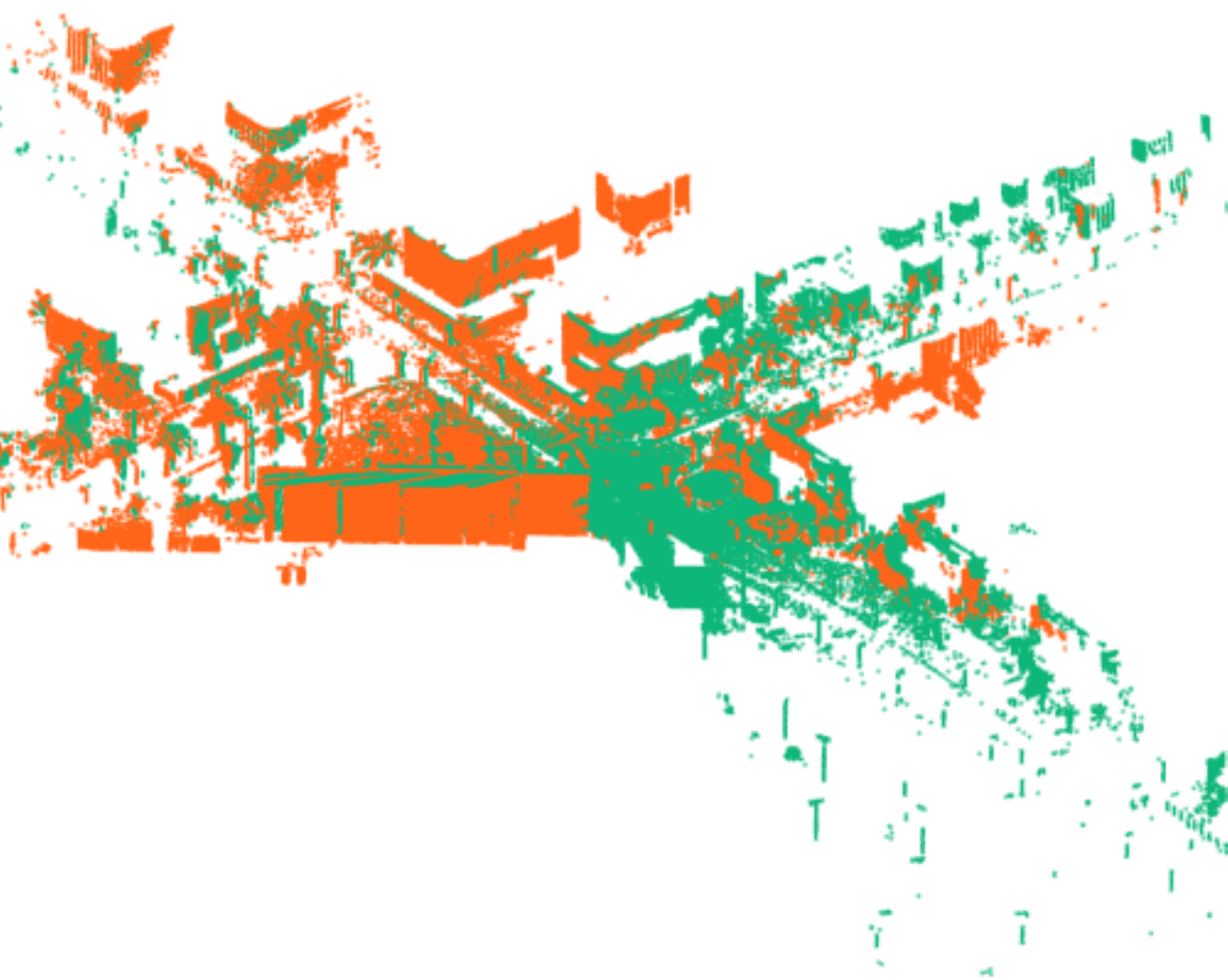}}
  \end{minipage} 
  & \begin{minipage}[b]{1\linewidth}
    \centering
    \raisebox{-.5\height}{\includegraphics[width=\linewidth]{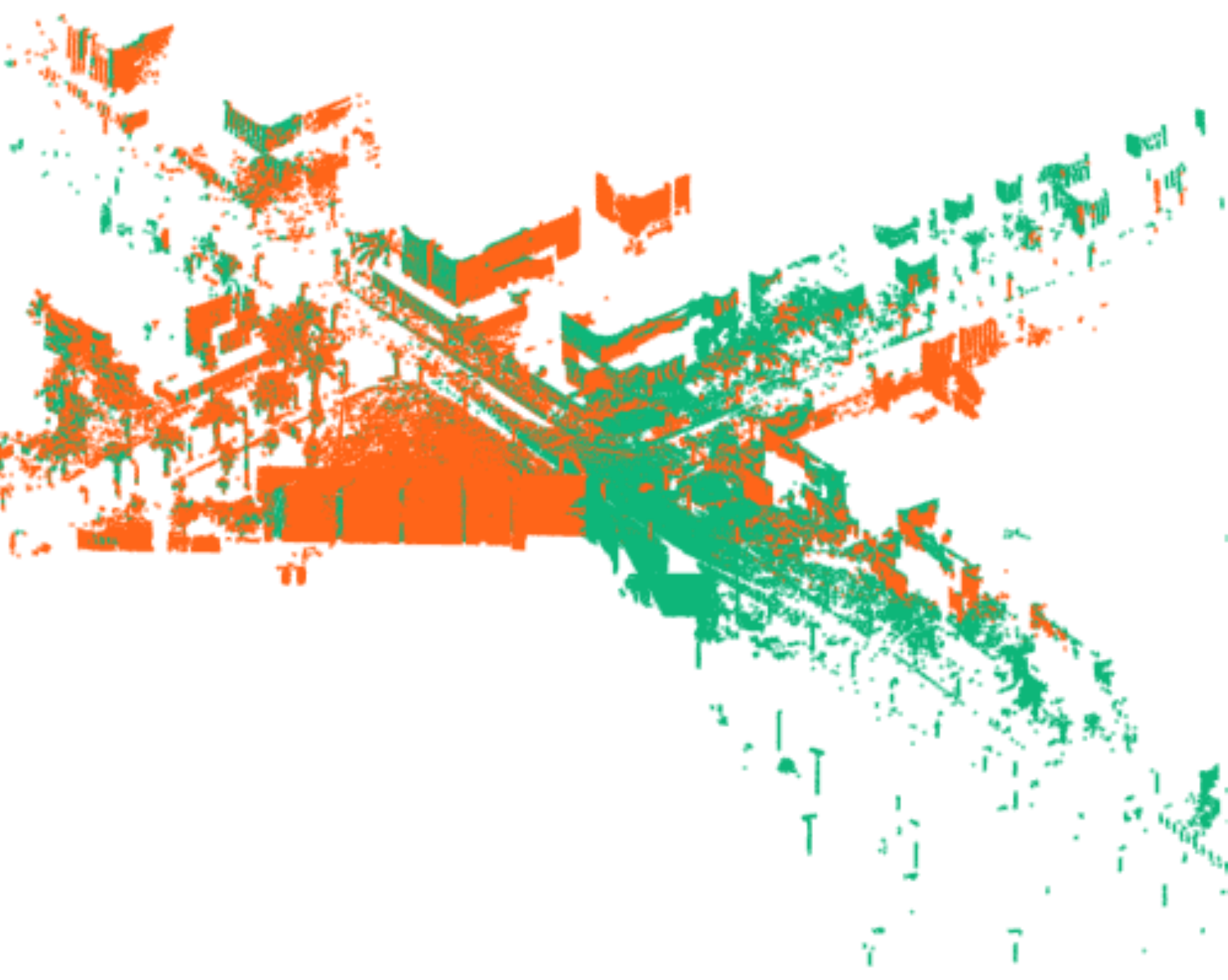}}
  \end{minipage} 
  &
  \textbf{7e}
  \\
    \begin{minipage}[b]{1\linewidth}
      \centering
      \raisebox{-.5\height}{\includegraphics[width=\linewidth]{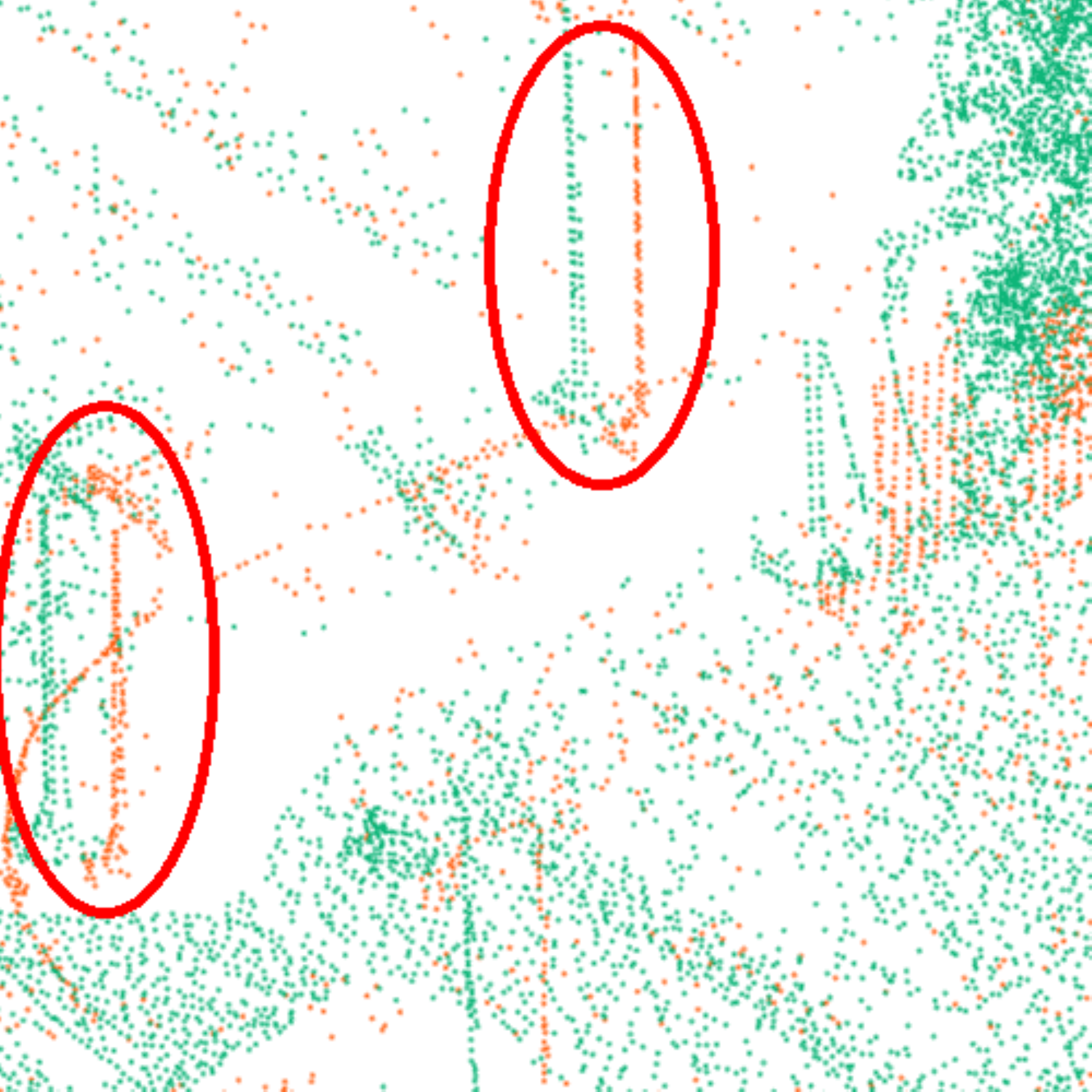}}
    \end{minipage}
  & 
    \begin{minipage}[b]{1\linewidth}
      \centering
      \raisebox{-.5\height}{\includegraphics[width=\linewidth]{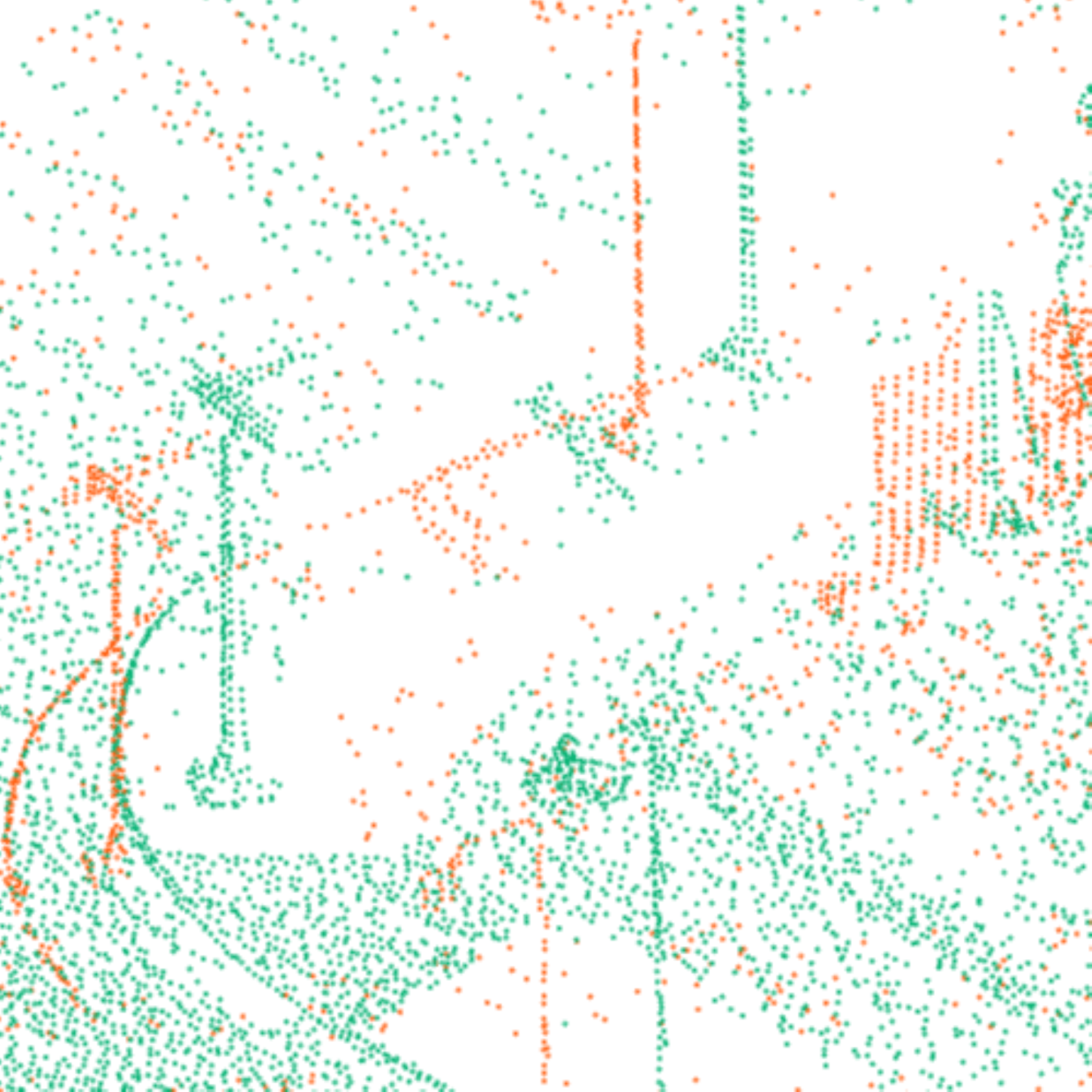}}
    \end{minipage}
  & 
    \begin{minipage}[b]{1\linewidth}
      \centering
      \raisebox{-.5\height}{\includegraphics[width=\linewidth]{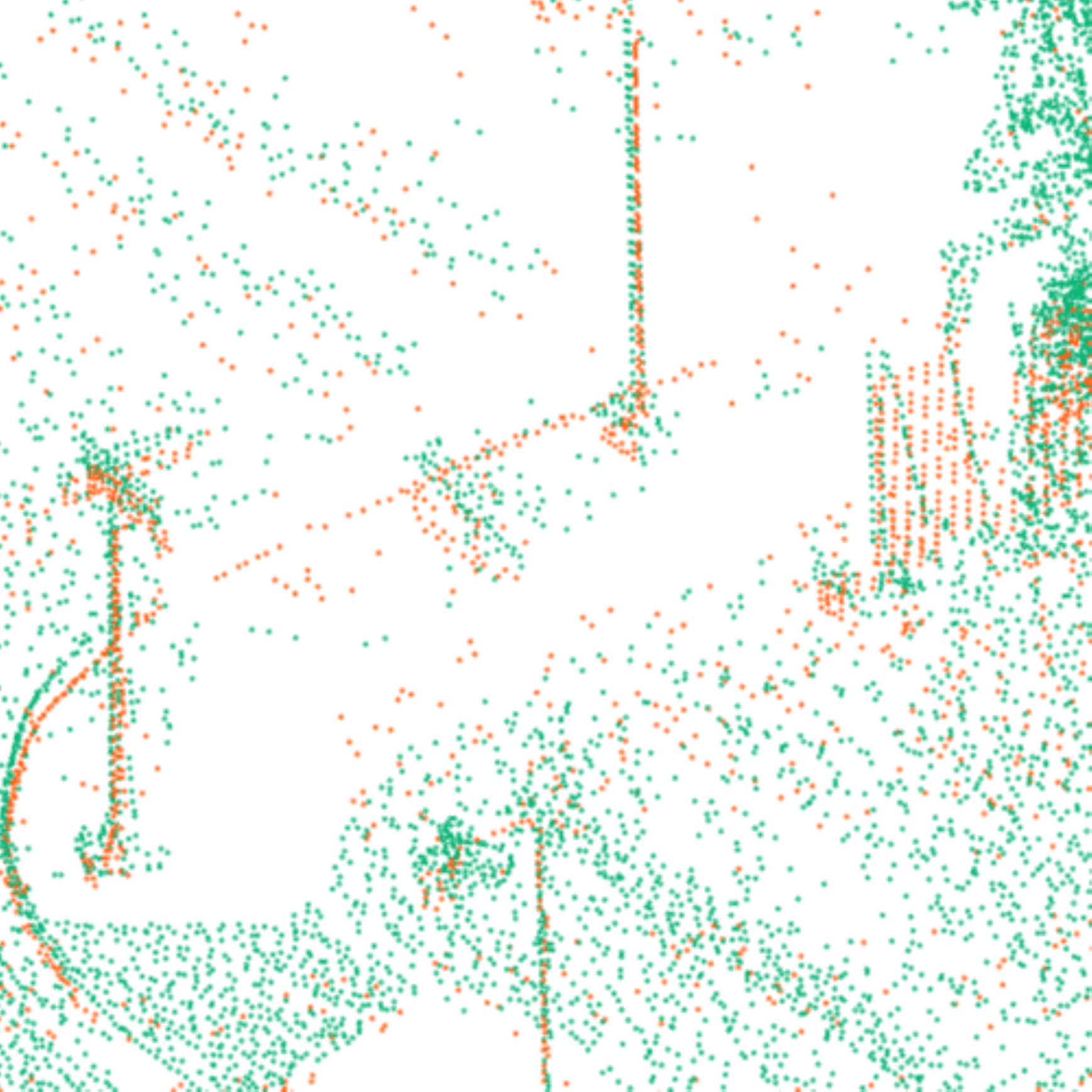}}
    \end{minipage}
    &
  \\
\end{tabular}}
\caption{Visualized registration results on scene-scale example instances, 
which contain overall views(middle), their error maps(top) and partial views(bottom).
\label{fig:vslz_scene}}
\end{figure}

Table \ref{tab:aver_SR} records the average success ratio obtained by these eight approaches in the 15 data.
As can be seen from Table \ref{tab:aver_SR}, the proposed EMTR-SSC obtains 54.7\% improvement over the best single-task peer competitors, i.e., JADE.
Compared with four state-of-the-art EMTO approaches, i.e., MFEA-II, MTEA-AD, EMFPCR and DEORA, EMTR-SSC achieves 52.0\%, 56.4\%, 48\% and 32.7\% improvements in terms of average success ratio, respectively. 
The success ratio of MFEA is the worst among all algorithms due to much negative transfer caused by the inappropriate knowledge transfer mechanism.
Furthermore, the average error calculation strategy makes approaches with high success ratios more likely to have large average errors as they contain more difficult data when calculating average error \cite{chen2022sc2}.
Therefore, the precision comparison of EMTR-SSC and evolving approaches, which is unfair under this calculation strategy, is not conducted.

\subsection{Comparisons with Traditional Approaches}

\subsubsection{Quantization Accuracy}
Table \ref{tab:table_accu_trad} compares EMTR-SSC with the other four traditional approaches in terms of registration precision in the 15 data, where '-' denotes that all twenty independent runs failed. 
In this table, an approach with $^+$ means a coarse registration by RANSAC is conducted before performing this approach.
As can be seen from Table \ref{tab:table_accu_trad}, the proposed EMTR-SSC achieves registration with the lowest average translation error in all 15 data among these five approaches.
Regarding the average rotation error, EMTR-SSC is not optimal in only one data, i.e., \emph{Dragon} data.
The error obtained by EMTR-SSC is either optimal or sub-optimal, indicating that high-quality registration can be achieved by optimizing the designed fitness function.
Furthermore, the proposed approach also performs well when tackling middle overlapping point clouds.
This is because of the use of M-estimator in fitness function, which can effectively identify outliers and reduce their interference.

\subsubsection{Visualization of Object-Scale Data}
Fig. \ref{fig:vslz_object} shows straightforward comparisons of visualized results in object-scale data.
Since most of the registration errors are small, magnification is required to observe the difference among the visualized results of different algorithms. 
However, due to the limited space, the error map method is used for comparison rather than magnification. 
This method reflects the difference between the estimated position and the ground truth one of the source point cloud via pointwise RMSE and log-scale color-coding. 
In brief, for the color of each point, dark blue represents a small RMSE, while dark red represents a large one.
In other words, the wider blue area in visualized point cloud means better registration, while the wider red area means worse registration. 
As can be seen from Fig. \ref{fig:vslz_object}, the selected input point clouds are presented in the first column, which includes many challenging cases with large transformations, such as \emph{Dragon}, \emph{Angel}, \emph{Bimba} and \emph{Dancing children} data. 
In these cases, it is difficult to achieve successful registration without coarse registration by RANSAC for the three local algorithms, i.e., ICP, NDT and TrICP.
Columns two to six are the registration results of five algorithms. 
The seventh column shows the results registered by the ground truth from the dataset. 
The last column contains the colorbars used for mapping between RMSE and color. 
Similar to the conclusions derived from Table \ref{tab:table_accu_trad}, both ICP and NDT yield only coarse registrations in the 15 data, and TrICP achieves great registrations only in a specific overlap rate interval.
FGR can obtain great registrations for most of the data.
The proposed EMTR-SSC yields registrations with the highest accuracy among these five algorithms.

\subsubsection{Visualization of Scene-Scale Data}
Fig. \ref{fig:vslz_scene} shows comprehensive comparisons of visualized results in scene-scale data.
In this figure, only well-performing algorithms, such as FGR and TrICP, are presented.
For each data, the first row shows error maps of three algorithms, and the usage of the error map is the same as before.
The second row presents overall views, where source and target point clouds are colored in atrovirens and orange, respectively.
The final row of each data displays partial views, which are obtained by enlarging the parts marked with \emph{black box} in overall views.
In addition, the details that can reflect the difference in accuracy are marked with \emph{red ellipses}.
Although the quantization errors of the three algorithms in Table \ref{tab:table_accu_trad} are close, the obvious offset can be observed in partial views of Fig. \ref{fig:vslz_scene}.
These prove that the proposed EMTR-SSC also has superior performance in handling scene-scale data.

\begin{figure*}[!t]
  \centering
  \begin{tabular}{cccc:cccc}
  \multicolumn{4}{c:}{Clean} & \multicolumn{4}{c}{Noisy}
  \\
  \begin{minipage}[b]{0.2\columnwidth}
    \centering
    \raisebox{-.5\height}{\includegraphics[width=\linewidth]{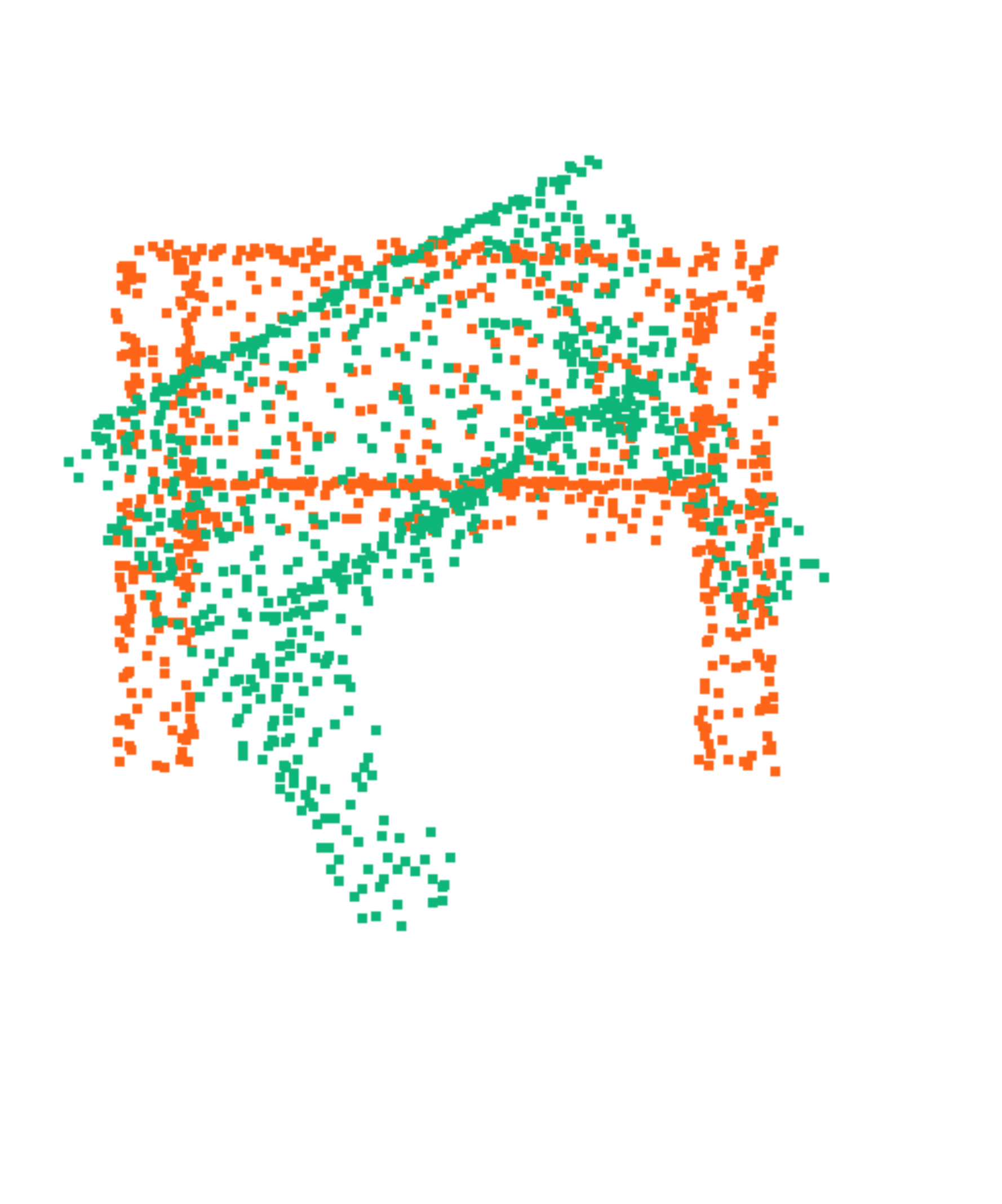}}
  \end{minipage} 
  & \begin{minipage}[b]{0.2\columnwidth}
    \centering
    \raisebox{-.5\height}{\includegraphics[width=\linewidth]{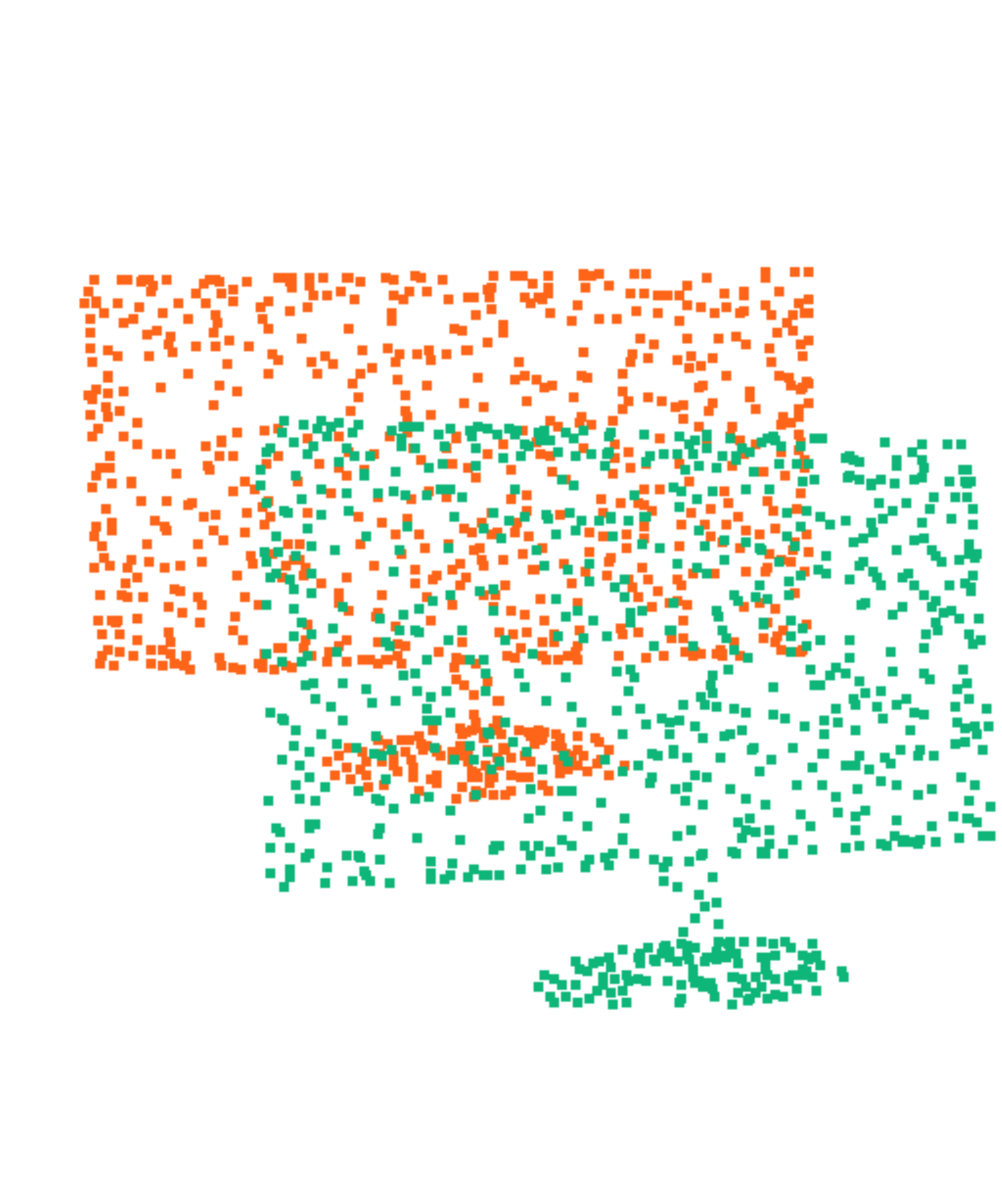}}
  \end{minipage} 
  & \begin{minipage}[b]{0.2\columnwidth}
    \centering
    \raisebox{-.5\height}{\includegraphics[width=\linewidth]{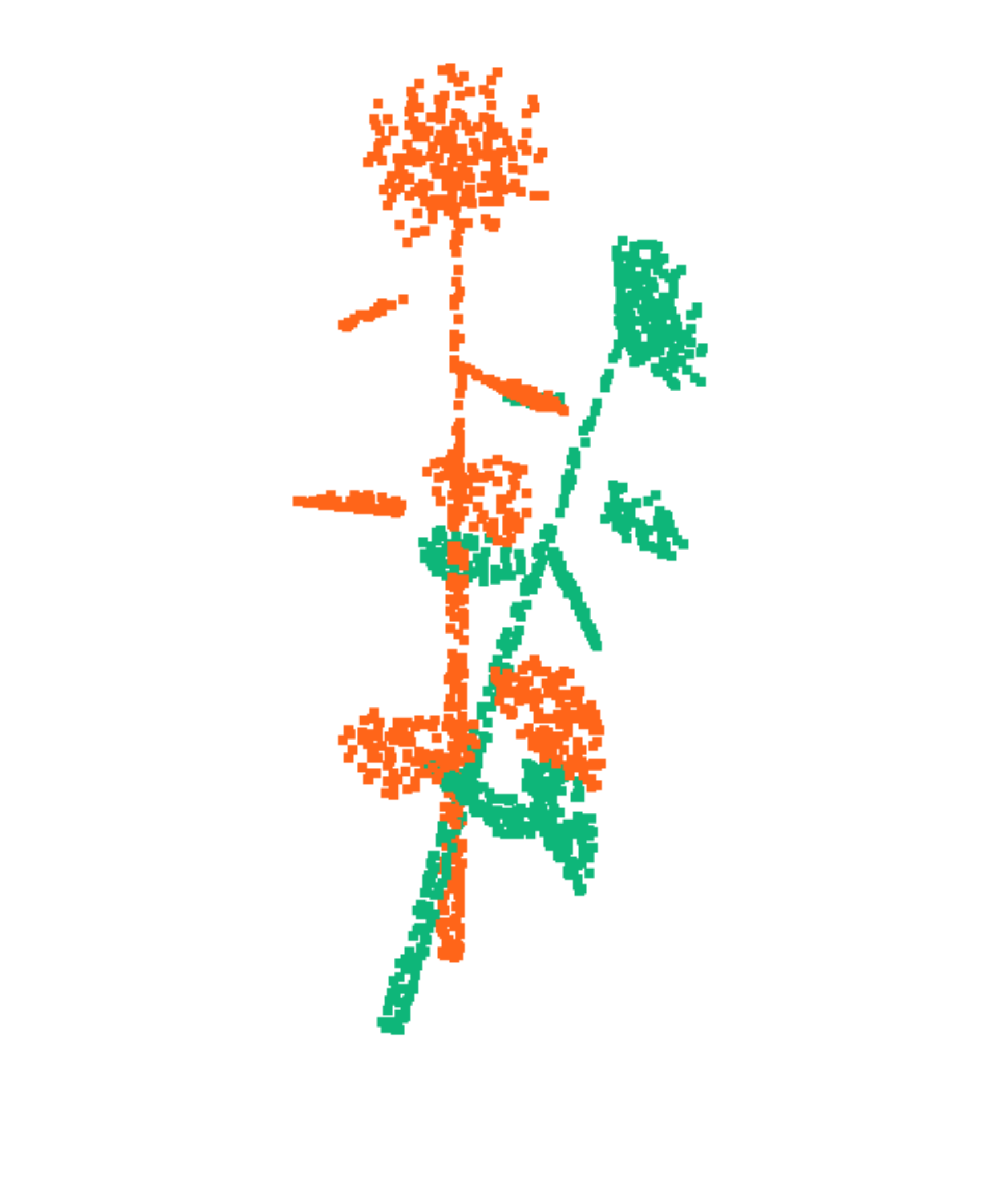}}
  \end{minipage}
  & \begin{minipage}[b]{0.2\columnwidth}
    \centering
    \raisebox{-.5\height}{\includegraphics[width=\linewidth]{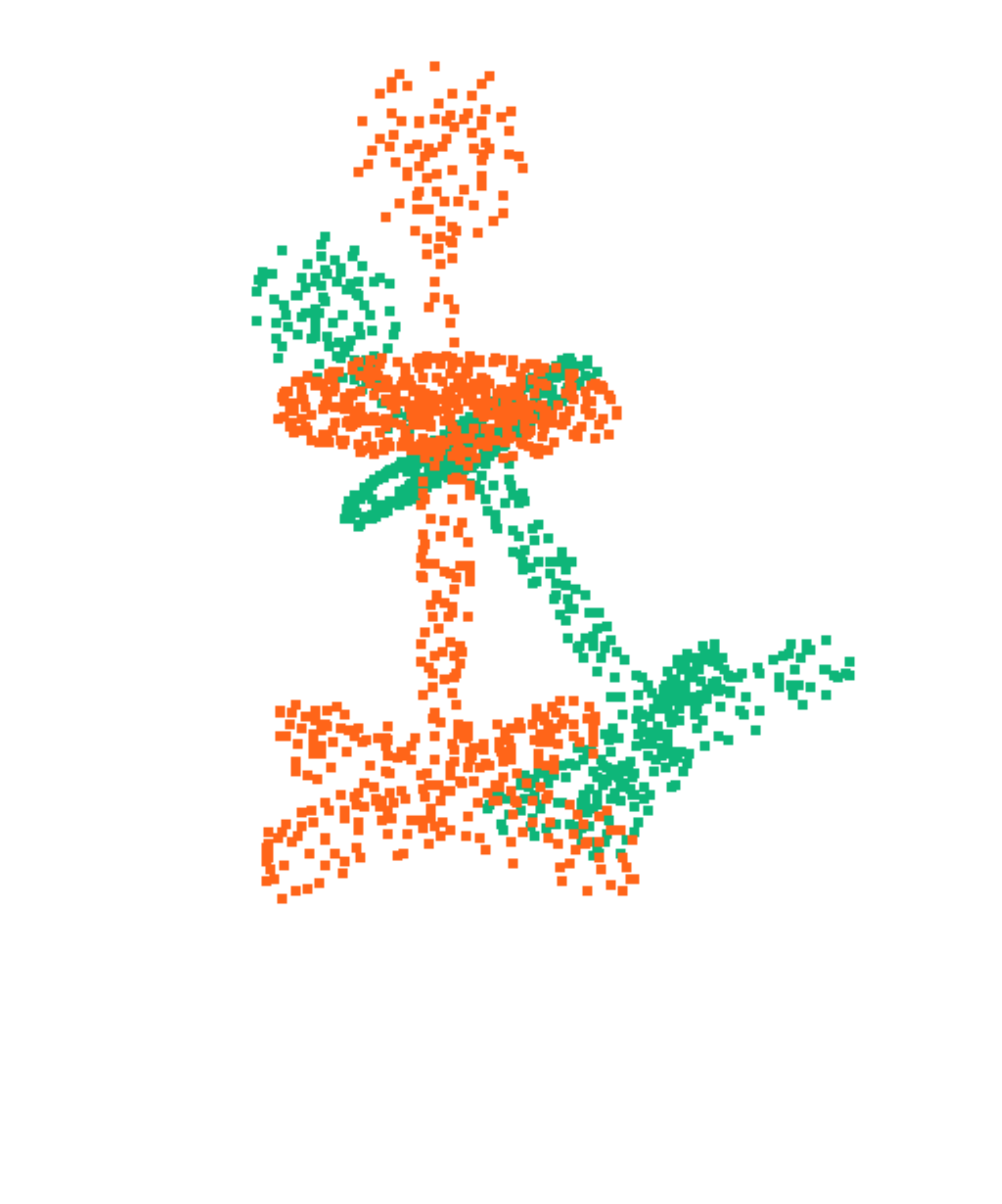}}
  \end{minipage} 
  & \begin{minipage}[b]{0.2\columnwidth}
    \centering
    \raisebox{-.5\height}{\includegraphics[width=\linewidth]{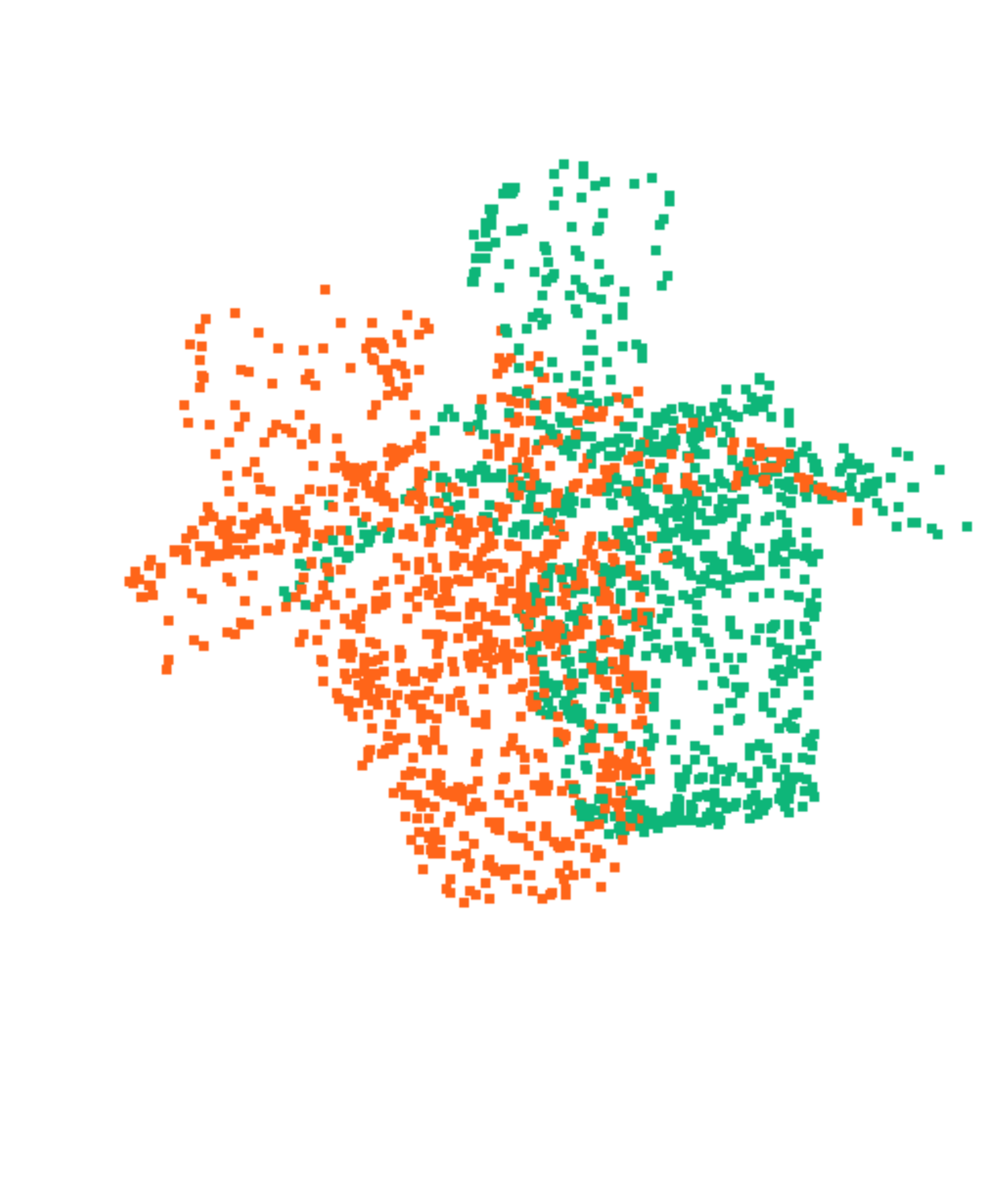}}
  \end{minipage} 
  & \begin{minipage}[b]{0.2\columnwidth}
    \centering
    \raisebox{-.5\height}{\includegraphics[width=\linewidth]{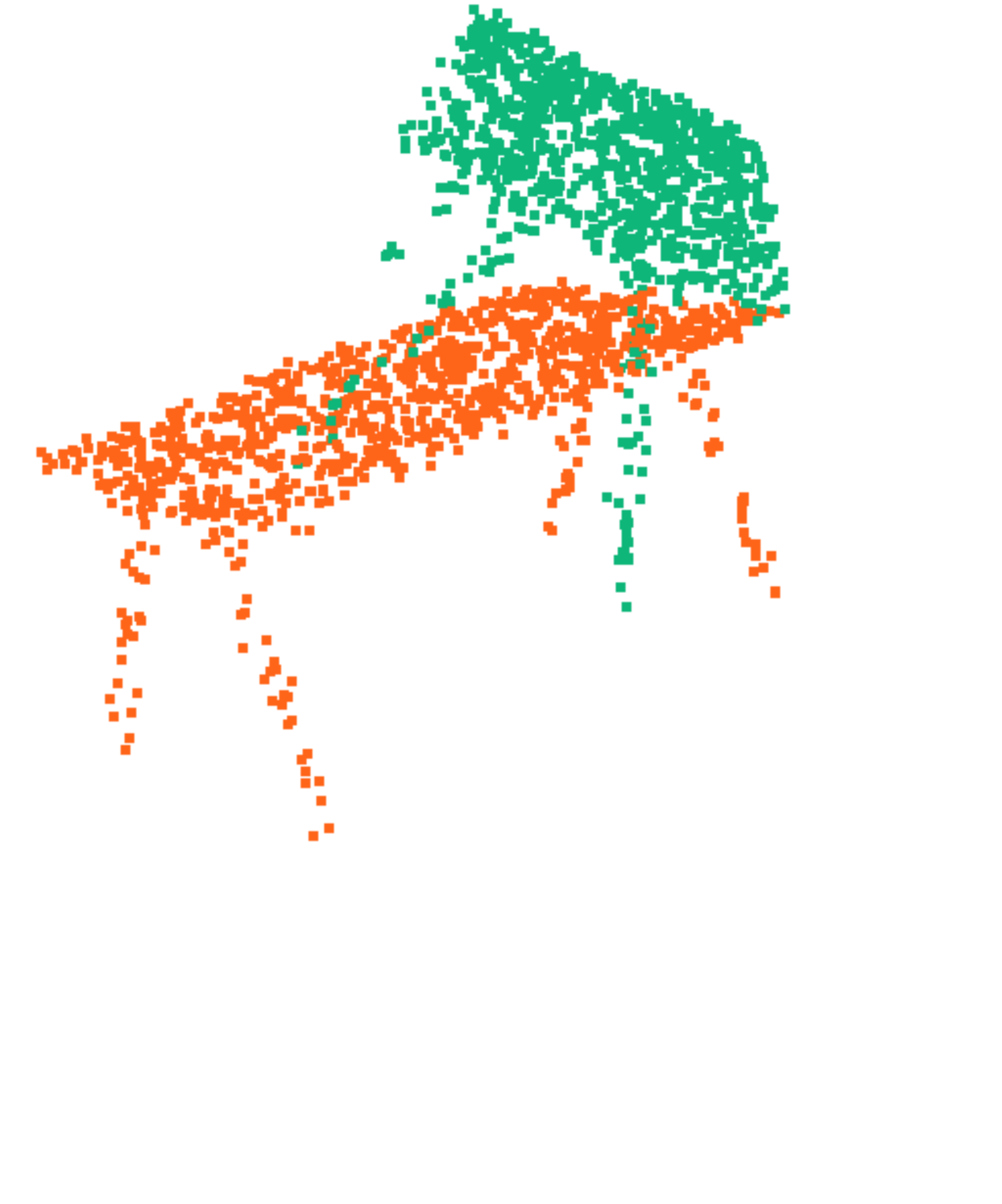}}
  \end{minipage} 
  & \begin{minipage}[b]{0.2\columnwidth}
    \centering
    \raisebox{-.5\height}{\includegraphics[width=\linewidth]{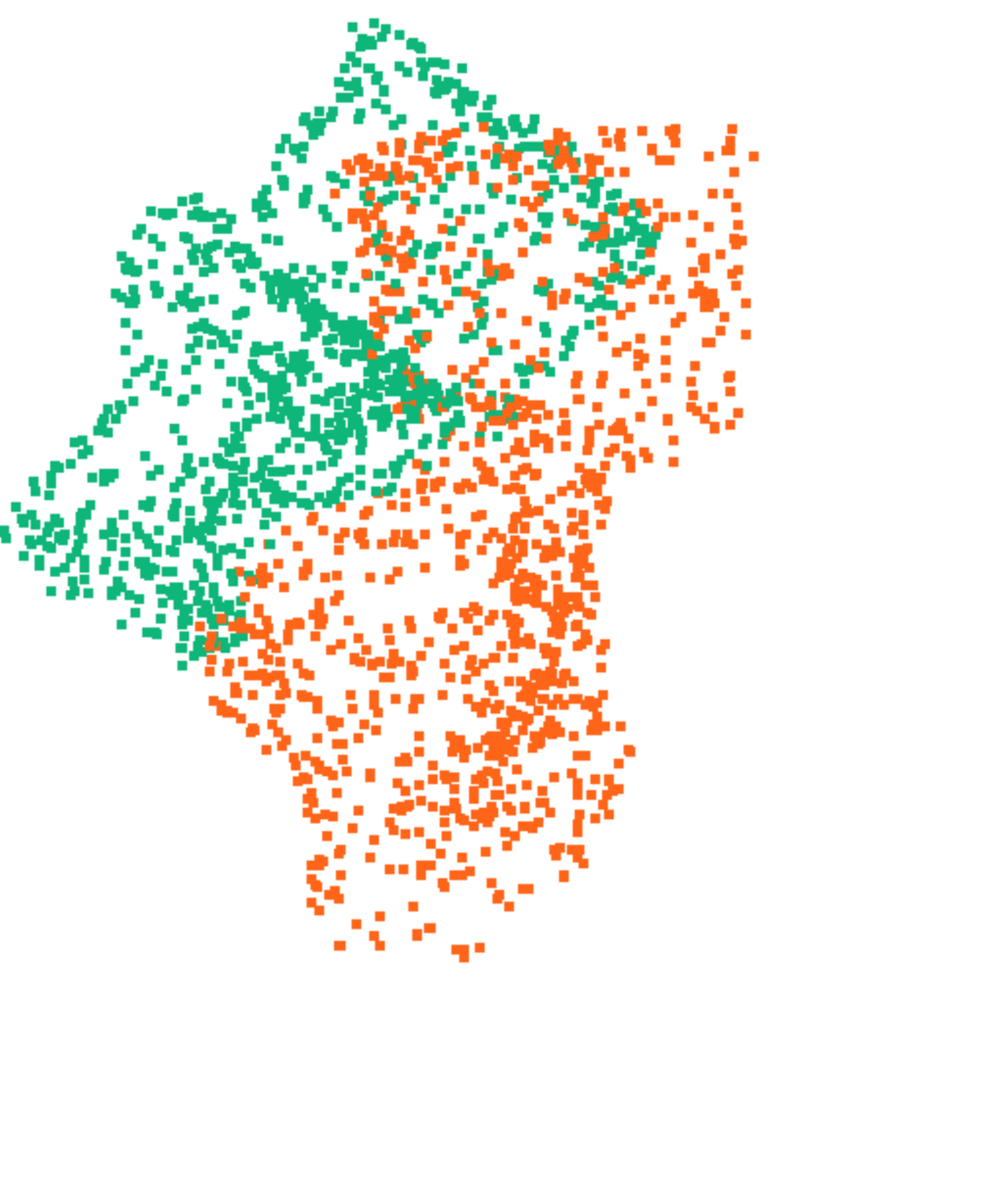}}
  \end{minipage} 
  & \begin{minipage}[b]{0.2\columnwidth}
    \centering
    \raisebox{-.5\height}{\includegraphics[width=\linewidth]{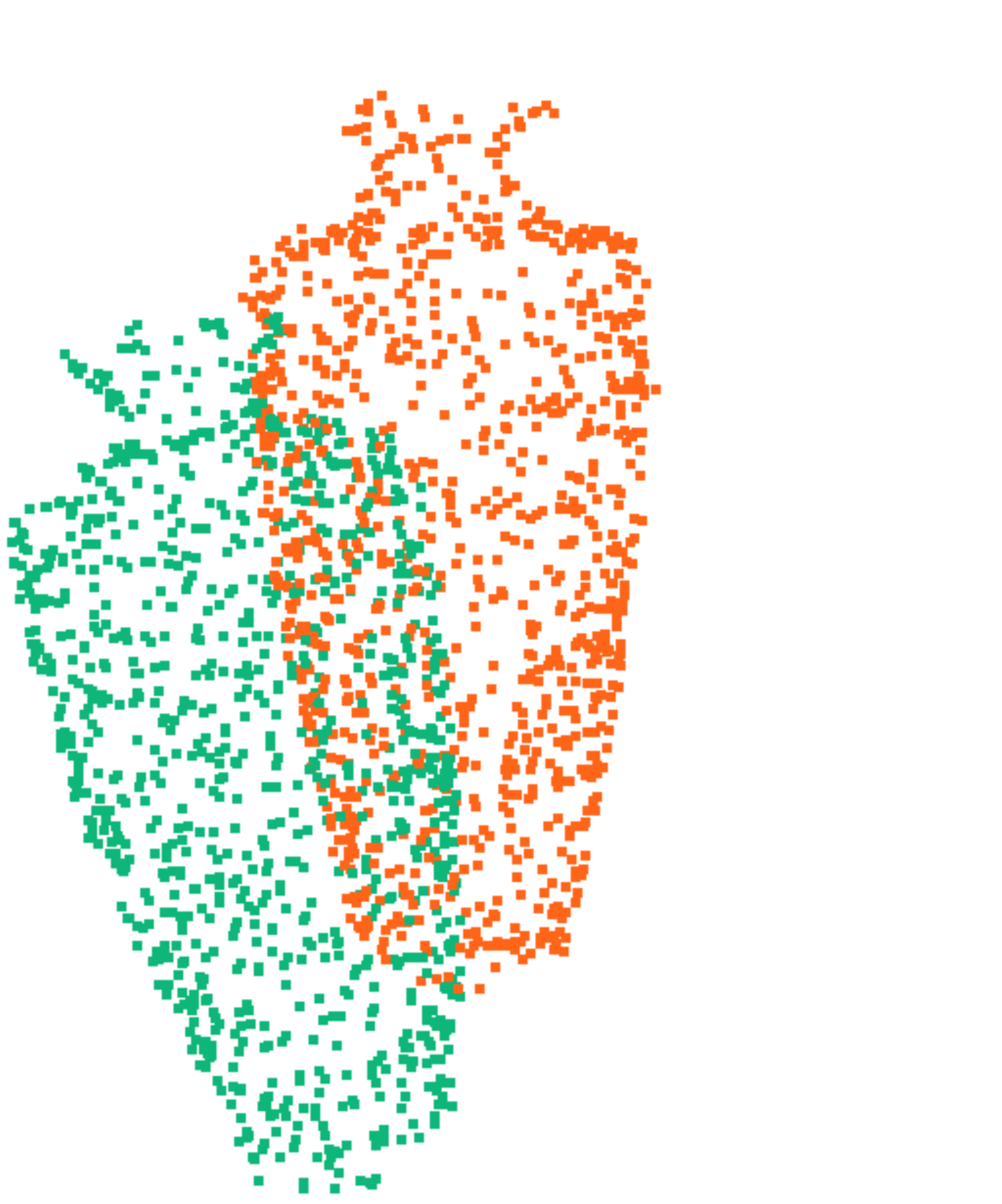}}
  \end{minipage} 
  \\

  \begin{minipage}[b]{0.2\columnwidth}
    \centering
    \raisebox{-.5\height}{\includegraphics[width=\linewidth]{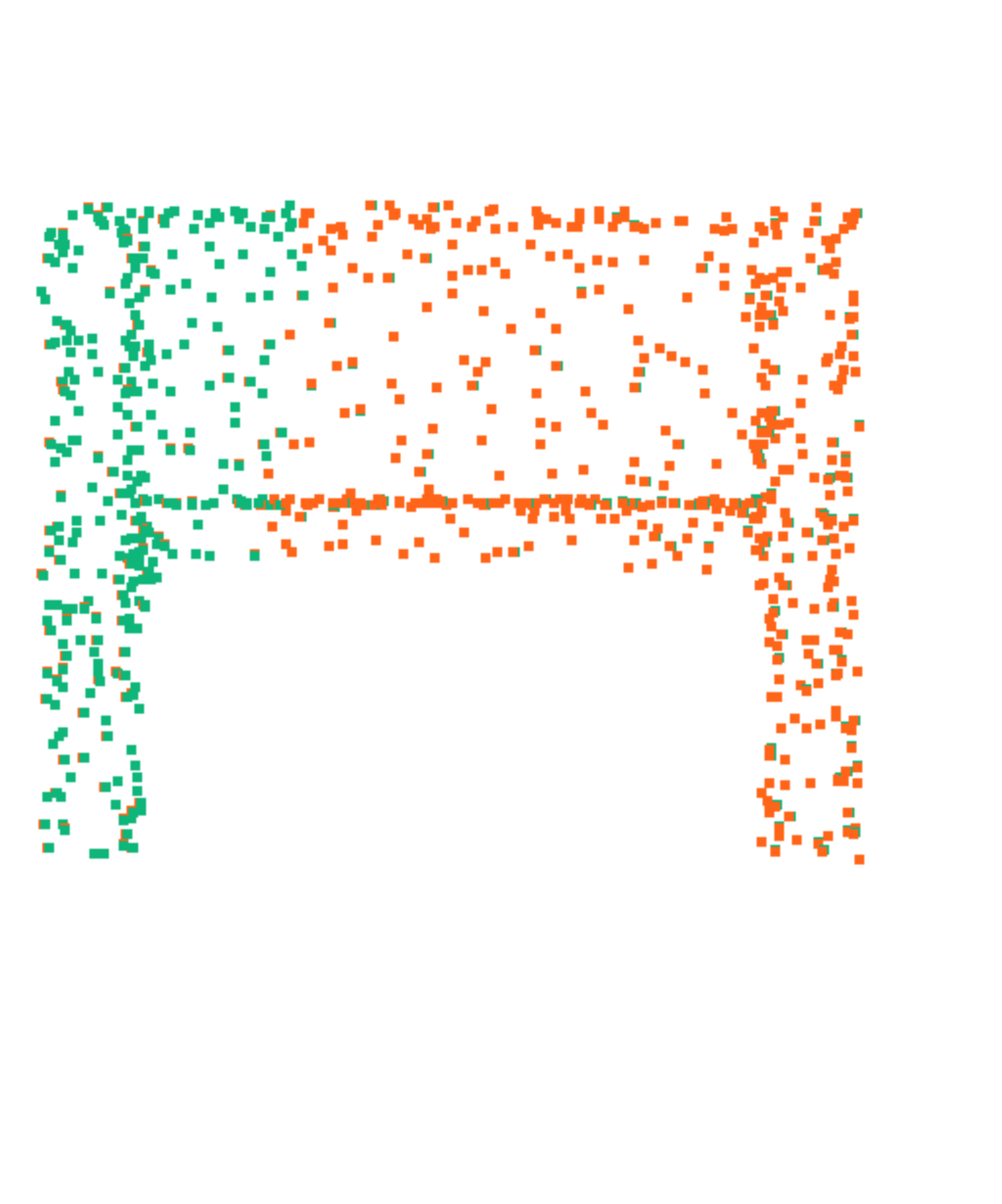}}
  \end{minipage} 
  & \begin{minipage}[b]{0.2\columnwidth}
    \centering
    \raisebox{-.5\height}{\includegraphics[width=\linewidth]{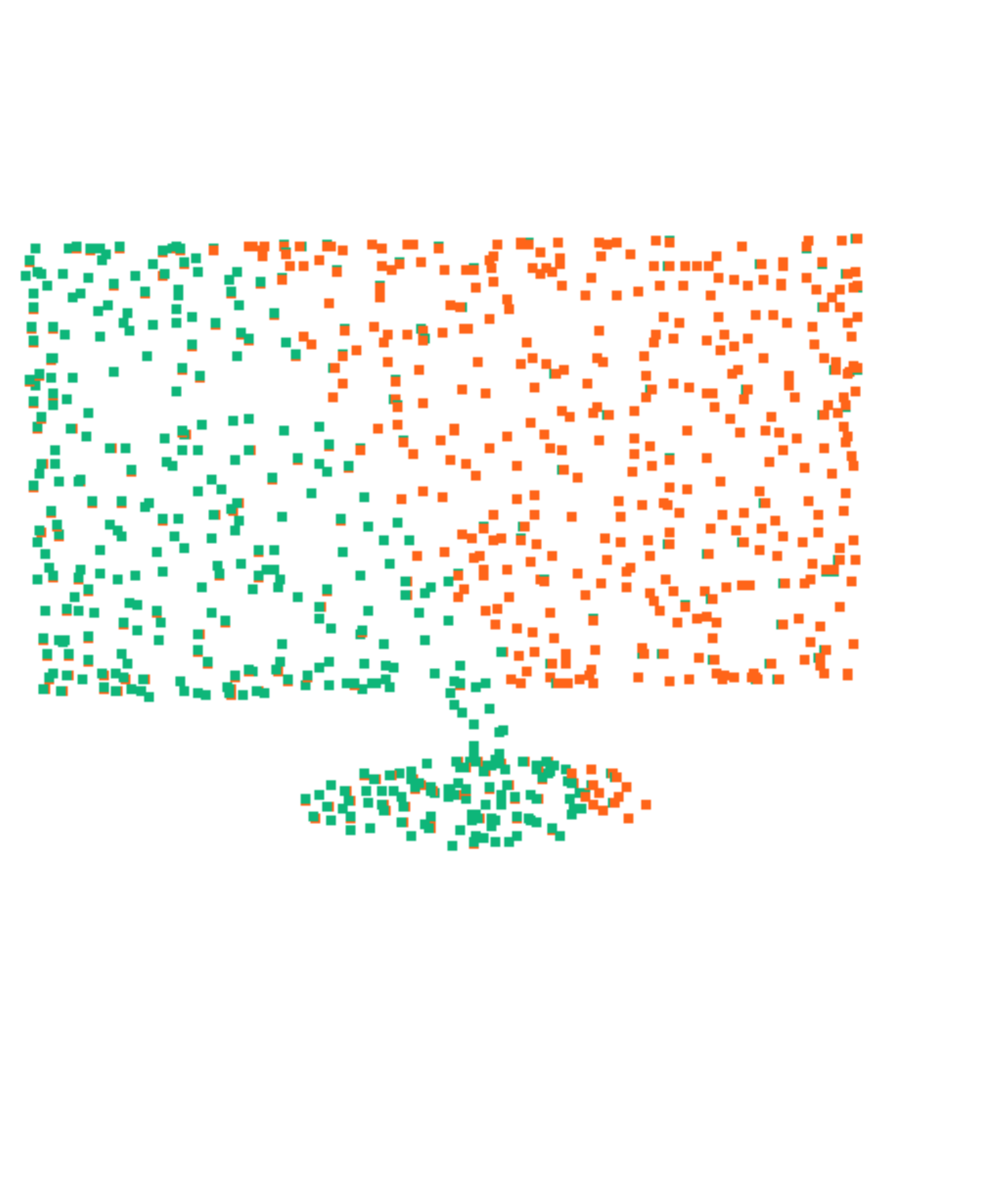}}
  \end{minipage} 
  & \begin{minipage}[b]{0.2\columnwidth}
    \centering
    \raisebox{-.5\height}{\includegraphics[width=\linewidth]{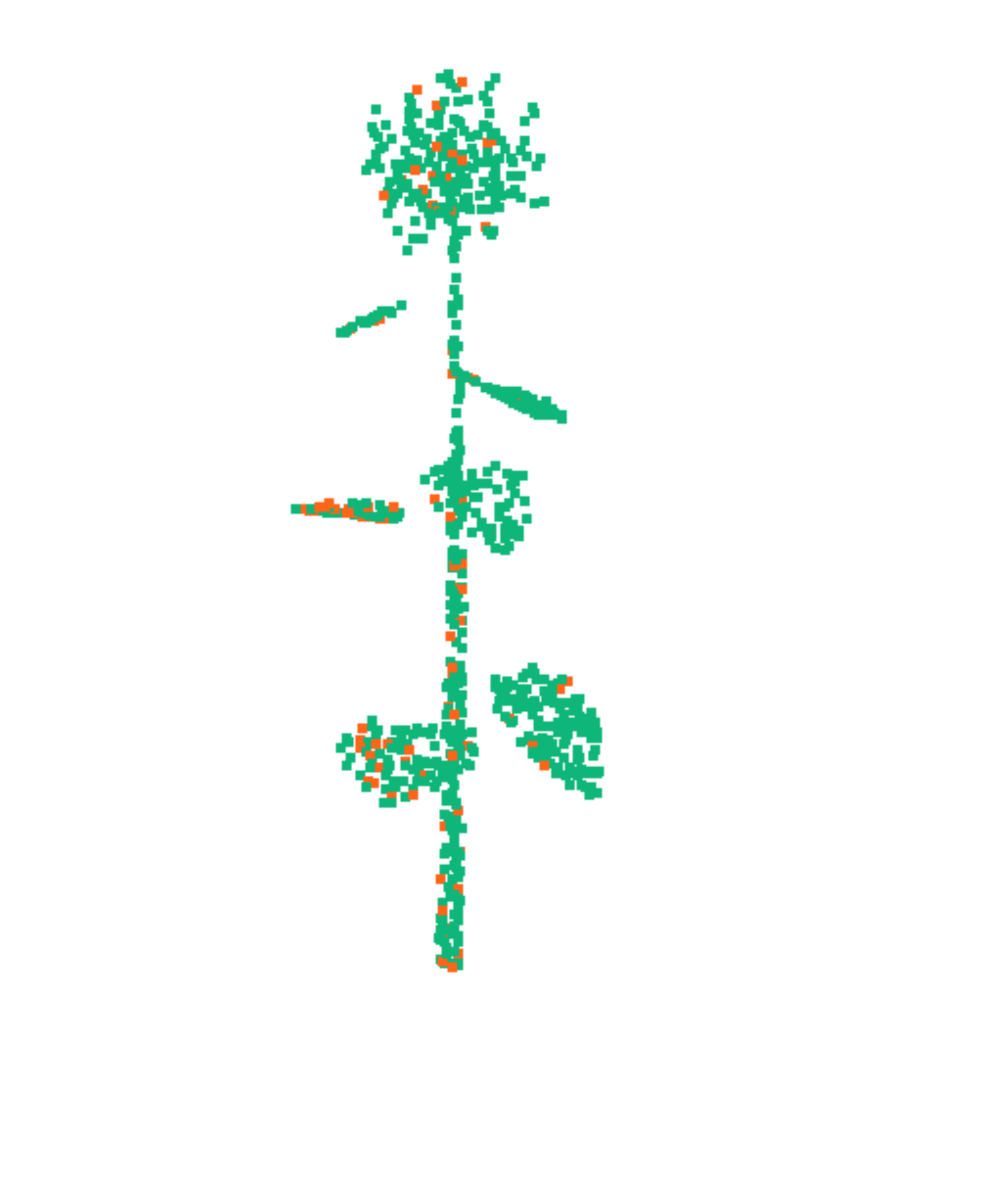}}
  \end{minipage}
  & \begin{minipage}[b]{0.2\columnwidth}
    \centering
    \raisebox{-.5\height}{\includegraphics[width=\linewidth]{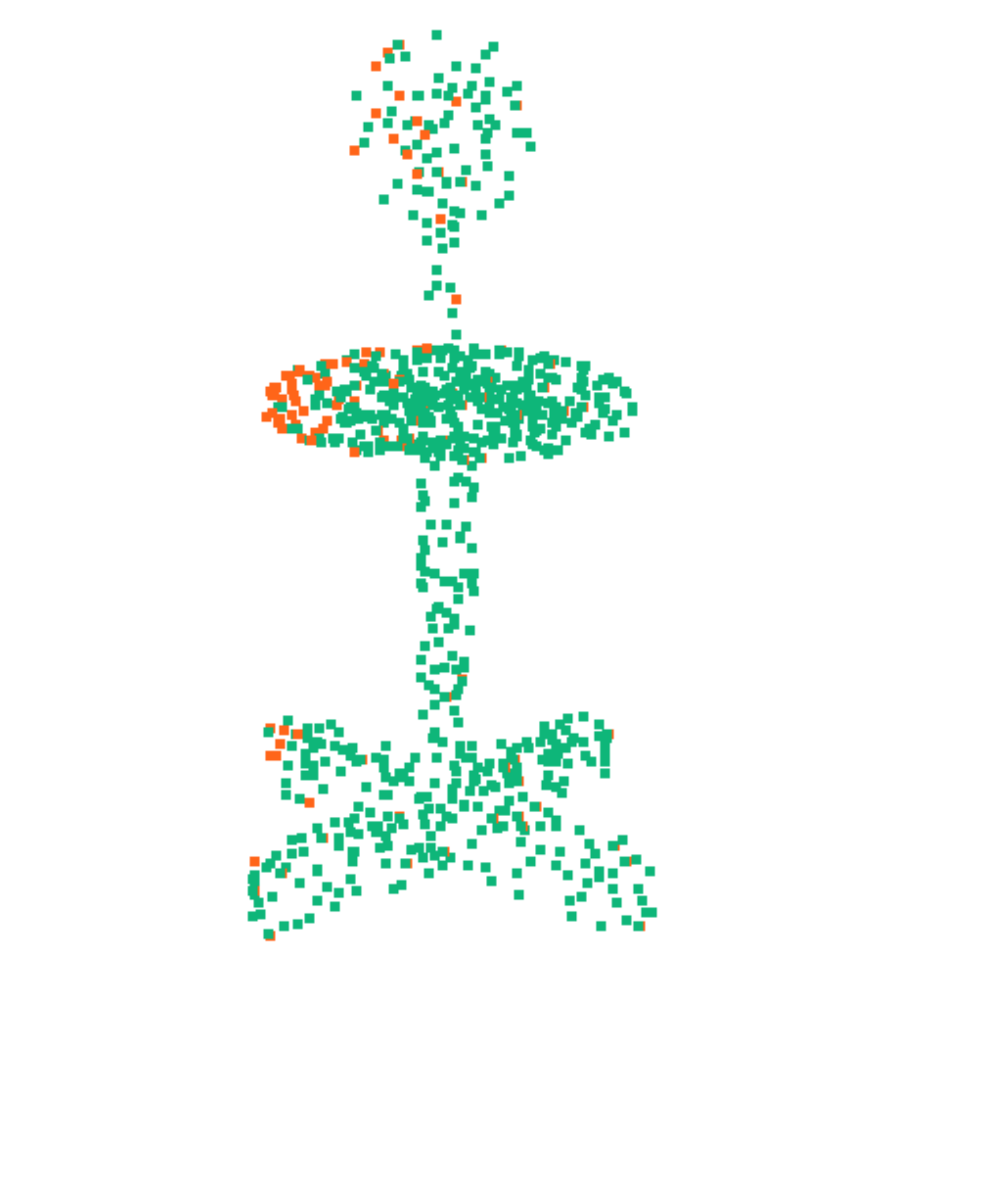}}
  \end{minipage} 
  & \begin{minipage}[b]{0.2\columnwidth}
    \centering
    \raisebox{-.5\height}{\includegraphics[width=\linewidth]{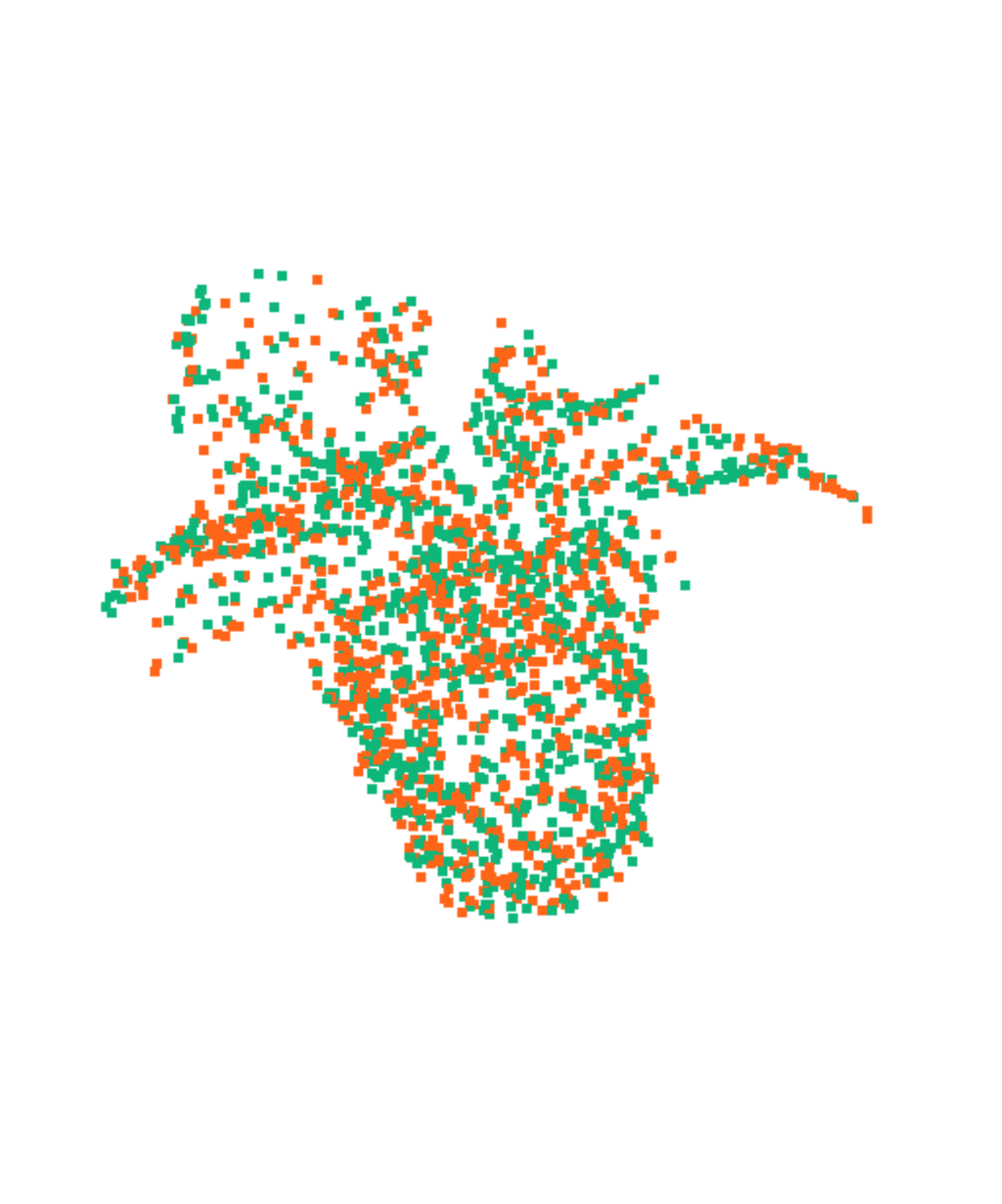}}
  \end{minipage} 
  & \begin{minipage}[b]{0.2\columnwidth}
    \centering
    \raisebox{-.5\height}{\includegraphics[width=\linewidth]{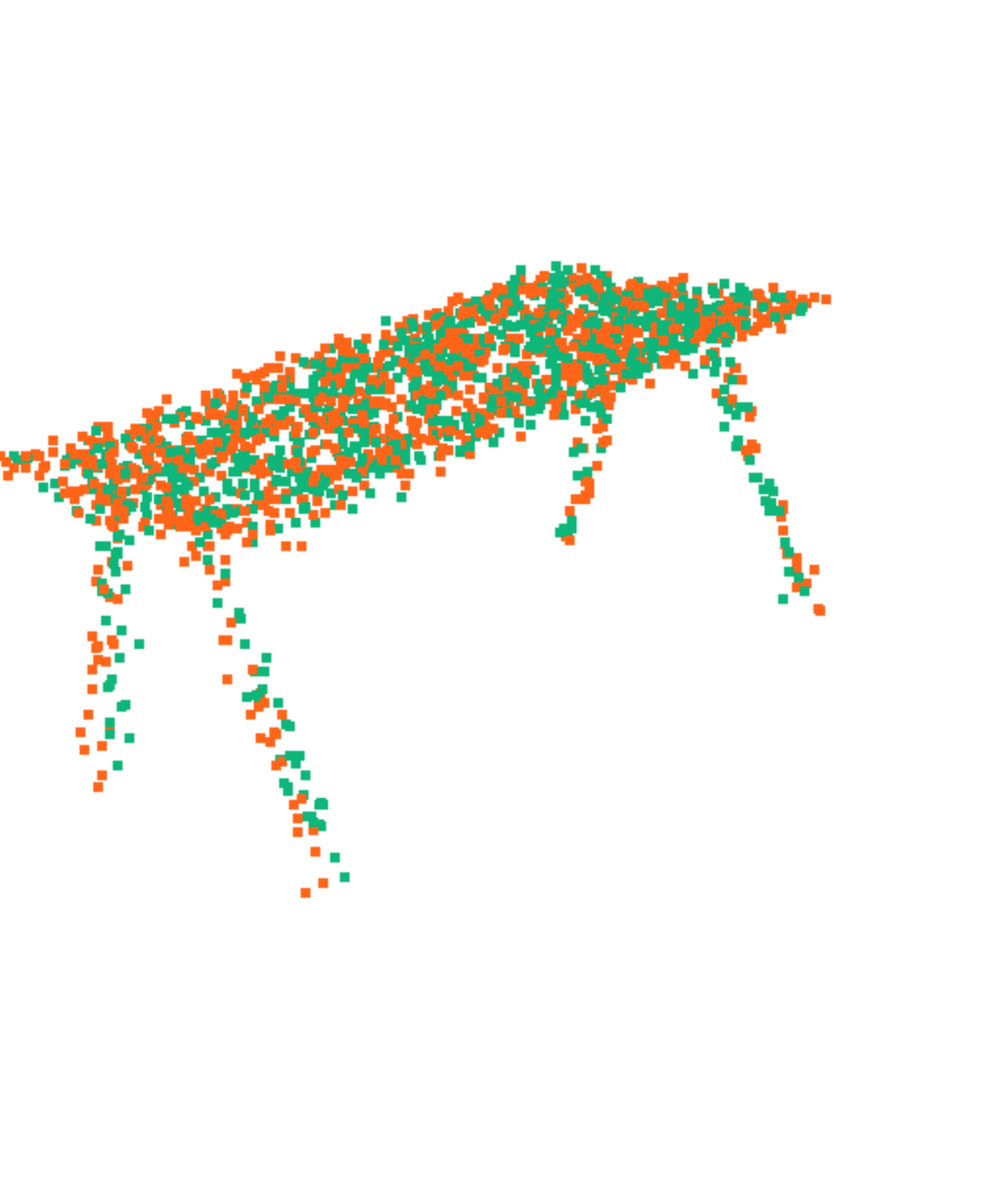}}
  \end{minipage} 
  & \begin{minipage}[b]{0.2\columnwidth}
    \centering
    \raisebox{-.5\height}{\includegraphics[width=\linewidth]{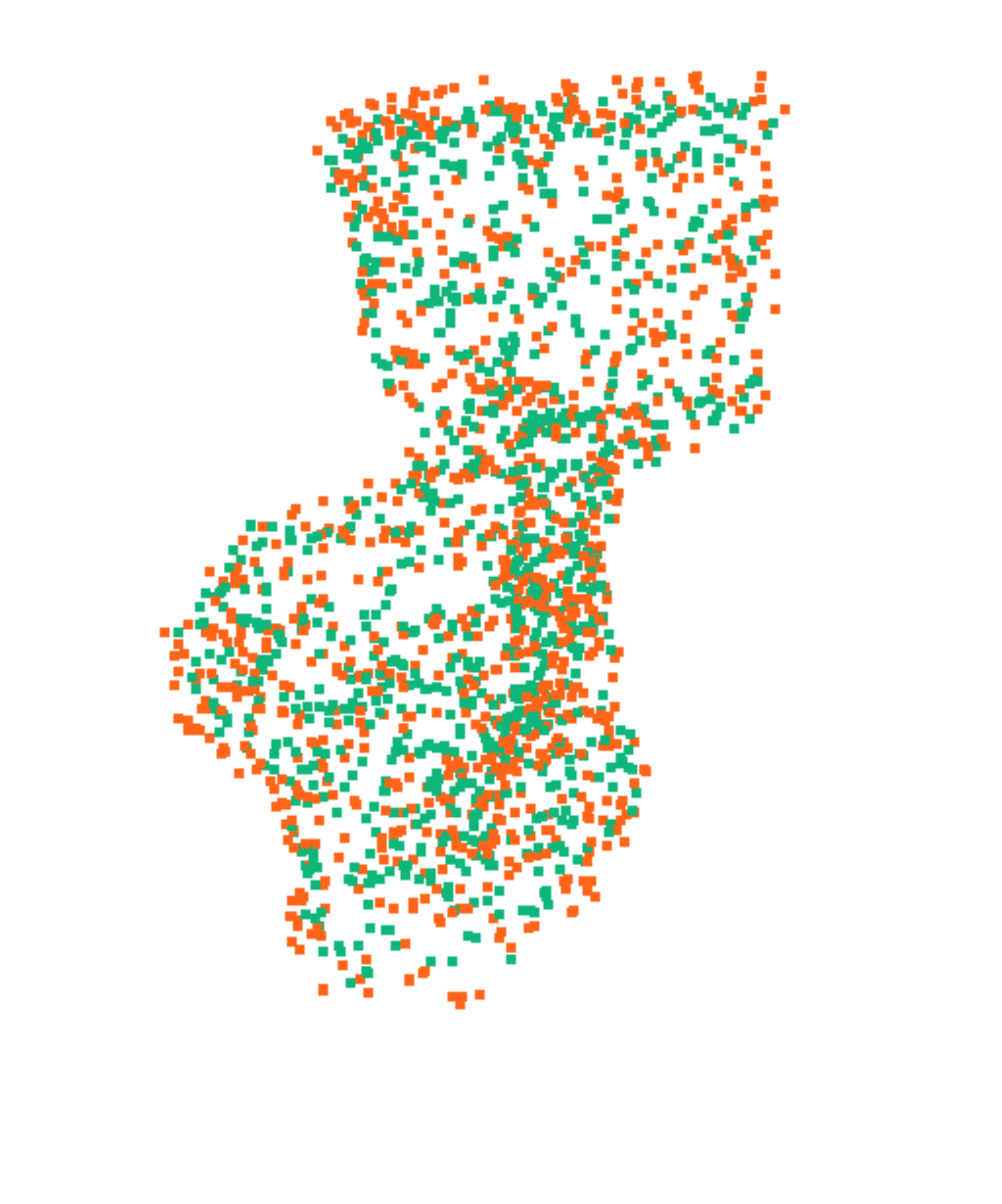}}
  \end{minipage} 
  & \begin{minipage}[b]{0.2\columnwidth}
    \centering
    \raisebox{-.5\height}{\includegraphics[width=\linewidth]{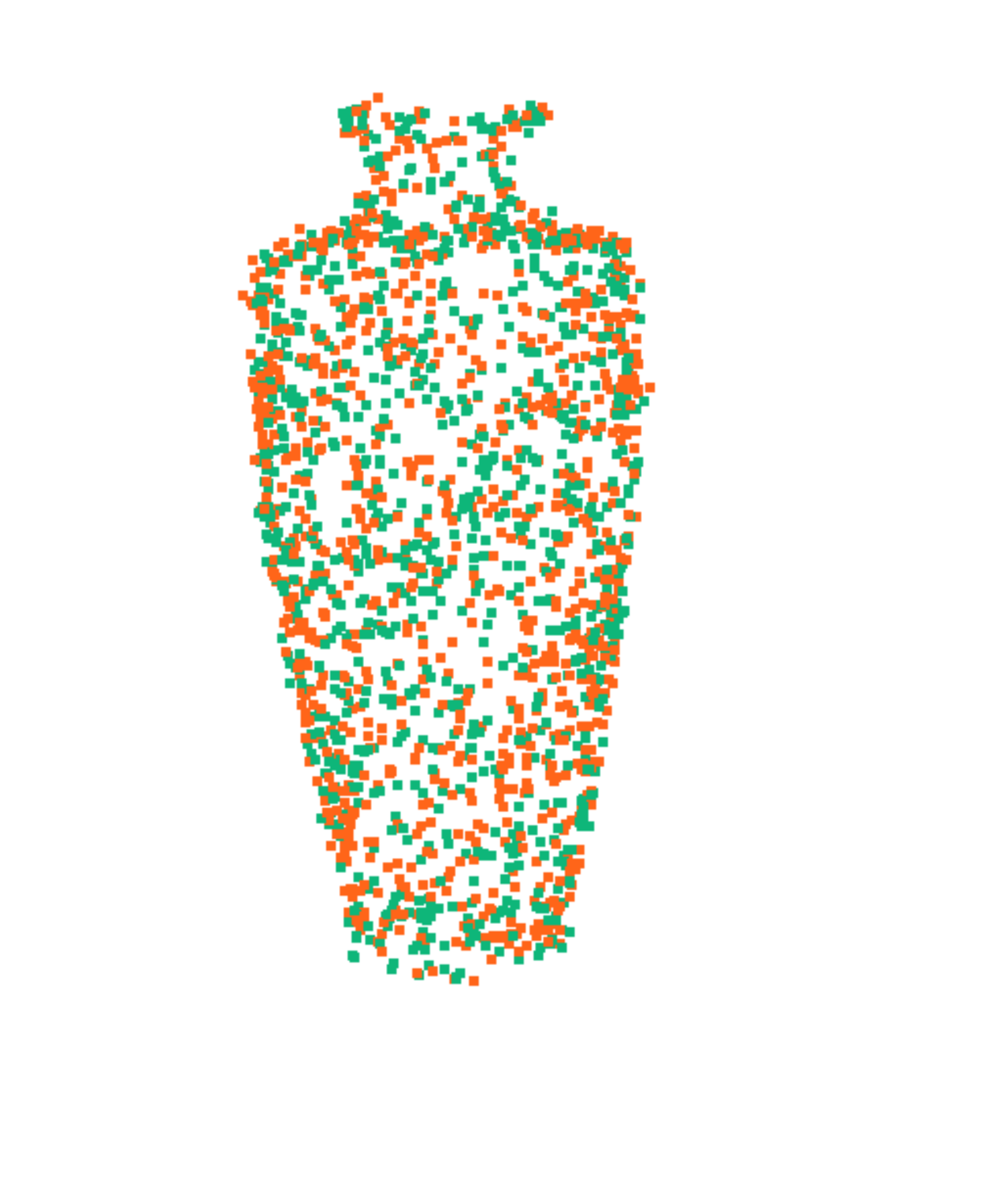}}
  \end{minipage} 
  \\

  \end{tabular}
  \caption{Visualization of ModelNet40. 
  The first row shows the input point clouds and the second row displays the point clouds aligned by the proposed method.
  \label{fig:vlz_DL}}
\end{figure*}

\begin{table}[!t]
  \caption{
  Rotation error and translation error for each deep learning method on the clean and noisy ModelNet40 Datasets. 
  The superior performance is highlighted in {\bf BOLD}.
  The second best one is typed in \underline{underline} type.
  \label{tab:DL}}
  \renewcommand\arraystretch{1.2}
  \resizebox{1\columnwidth}{!}{
  \begin{tabular}{c|cc|cc}
  \hline
              & \multicolumn{2}{c|}{Clean}                            & \multicolumn{2}{c}{Noisy}                             \\
              & $E_{\boldsymbol{R}}$ & $E_{\boldsymbol{t}}$ & $E_{\boldsymbol{R}}$ & $E_{\boldsymbol{t}}$ \\ \hline
  PointNetLK  & 0.847                & 0.00540              & 2.926                & 0.0262               \\
  DCP-v2      & 3.992                & 0.02920              & 8.922                & 0.0707               \\
  RPM-Net     & \underline{0.056}    & 0.00030              & \underline{0.664}    & \underline{0.0062}   \\ \hline
  \begin{tabular}[c]{@{}c@{}}EMTR-SSC(all)\\ ($SR_C=99.76\%$, $SR_N=83.65\%$)\end{tabular} & 0.338                & \underline{0.00028}  & 18.955               & 0.0143               \\
  EMTR-SSC(successful) & \textbf{0.007}       & \textbf{0.00006}     & \textbf{0.541}       & \textbf{0.0053}      \\ \hline
  \end{tabular}}
  \end{table}

\subsection{Comparisons with Deep Learning Approaches}
\subsubsection{Quantization Accuracy}
Table \ref{tab:DL} compares EMTR-SSC with three state-of-the-art deep learning methods in terms of registration precision in clean and noisy datasets, where '-' means there is no available result reported from the provider. 
EMTR-SSC(all) and EMTR-SSC(successful) represent the error calculation using all registration results and only successful ones, respectively.
$SR_C$ and $SR_N$ denote the success ratio of EMTR-SSC on clean and noisy datasets, respectively.
The superior performance is highlighted in bold and the second best one is typed in underline type.
The results of the compared methods are extracted from the original publication \cite{yew2020rpm}. 
Due to the simplicity of data and the small range of random transformations, our method can achieve a success ratio close to 100\%. 
On the other hand, this method can yield registration results with very low errors due to one-to-one correspondences.
The presence of outliers and noise makes the registration more intractable, resulting in a 16.11\% decrease in success ratio.
However, the proposed method can still achieve satisfactory precision as the fitness function of subpopulation $\alpha$ is robust to the outlier and noise.

On the whole, deep learning methods are highly competitive. 
However, it is worth emphasizing that these methods are supervised, while our method is unsupervised. 
In addition, deep learning methods need to be trained on clean and noisy datasets respectively. 
However, our method can directly handle both two datasets due to its strong generalization ability.

\subsubsection{Visualization of ModelNet40}
Fig. \ref{fig:vlz_DL} shows visualized result comparisons of ModelNet40, where the source and target point clouds are colored in atrovirens and orange, respectively.
As can be seen from Fig. \ref{fig:vlz_DL}, these data, consisting of CAD models, are simpler than scanned data in terms of geometric structure and surface texture.
In the clean dataset, many point pairs are perfectly aligned as if part of the point cloud disappears.

\subsection{Analysis on the Two Strategies}

\subsubsection{The Strategy of Knowledge Complement}
\begin{figure}[!t]
  \centering
  \subfloat[Success ratio]{\includegraphics[width=.9\columnwidth]{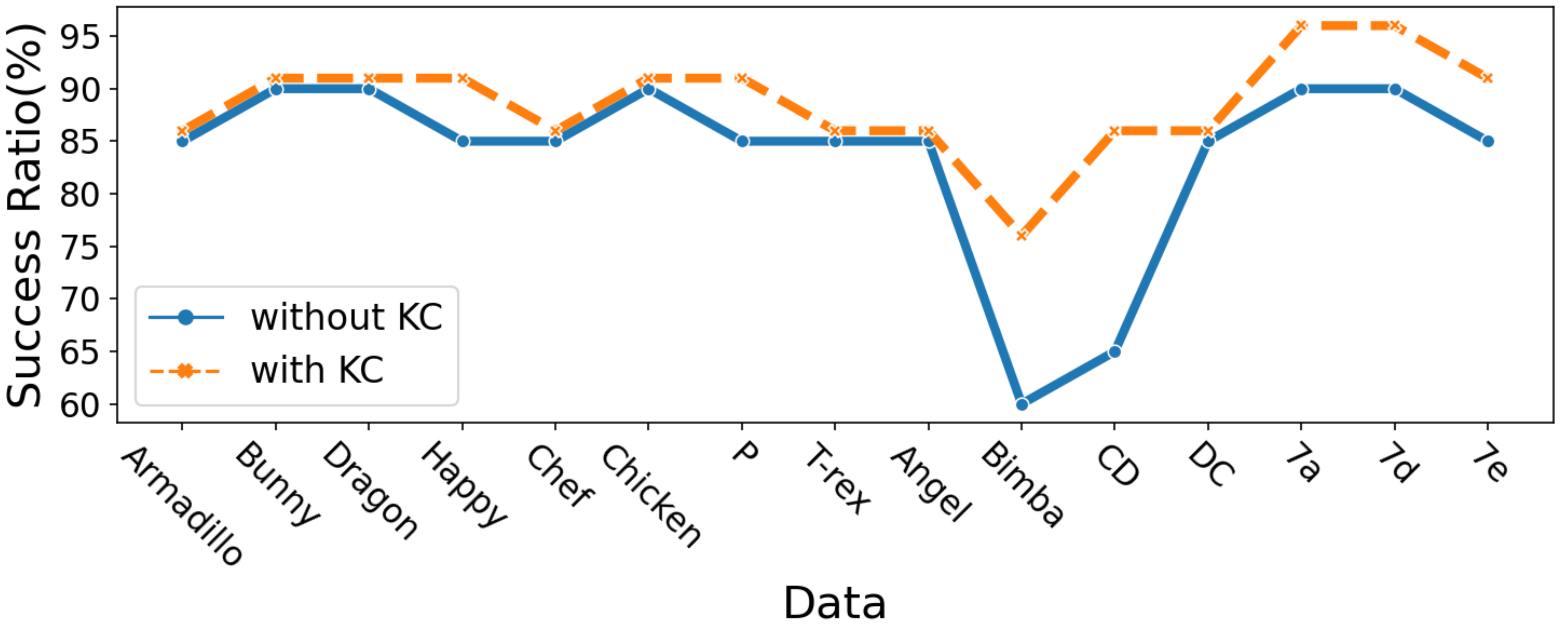}\label{fig:KC_SR}}
  \hspace{5pt}
  \\
  \subfloat[Rotation error]{\includegraphics[width=.45\columnwidth]{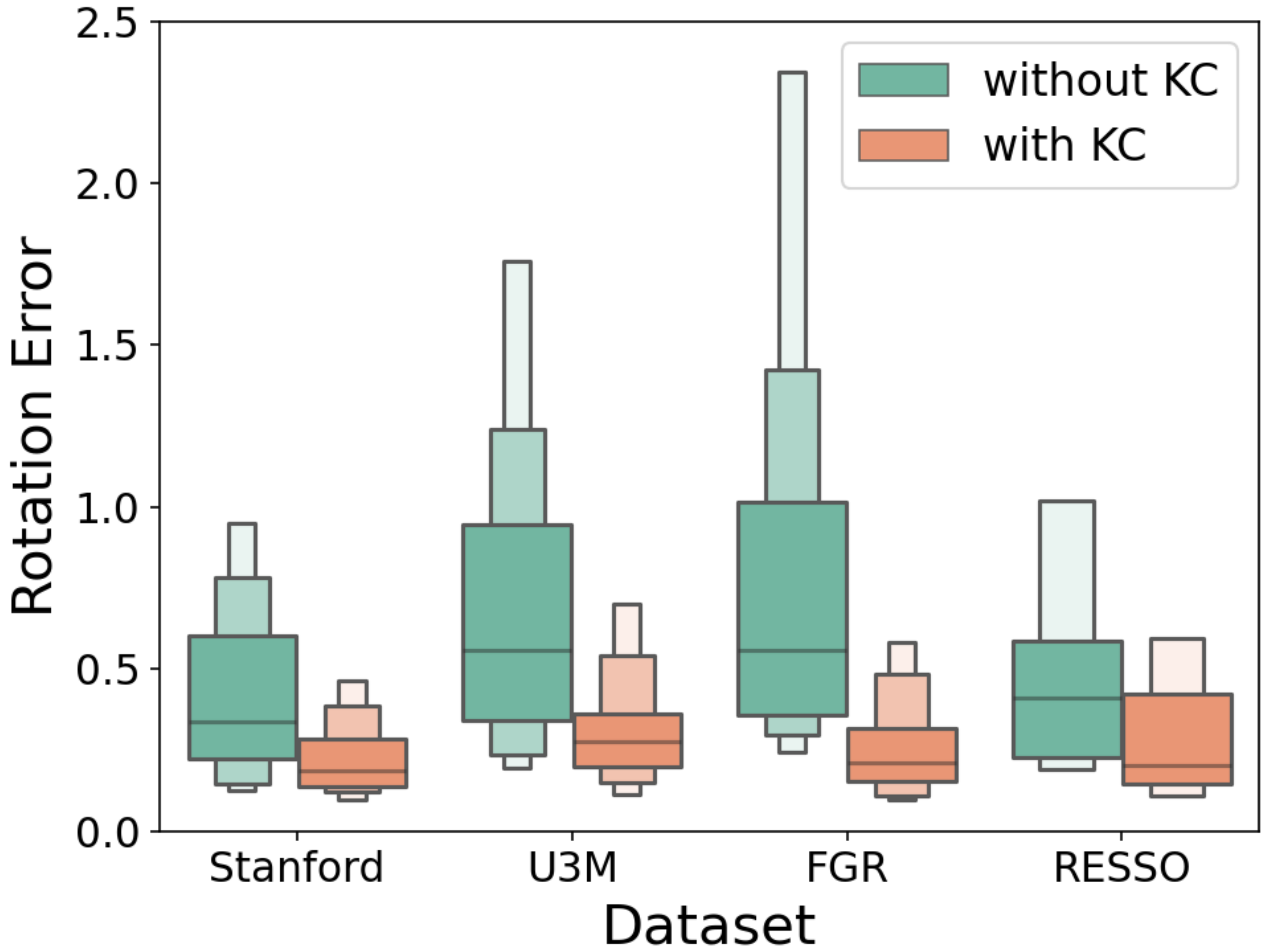}\label{fig:KC_Er}}
  \hspace{5pt}
  \subfloat[Translation error]{\includegraphics[width=.45\columnwidth]{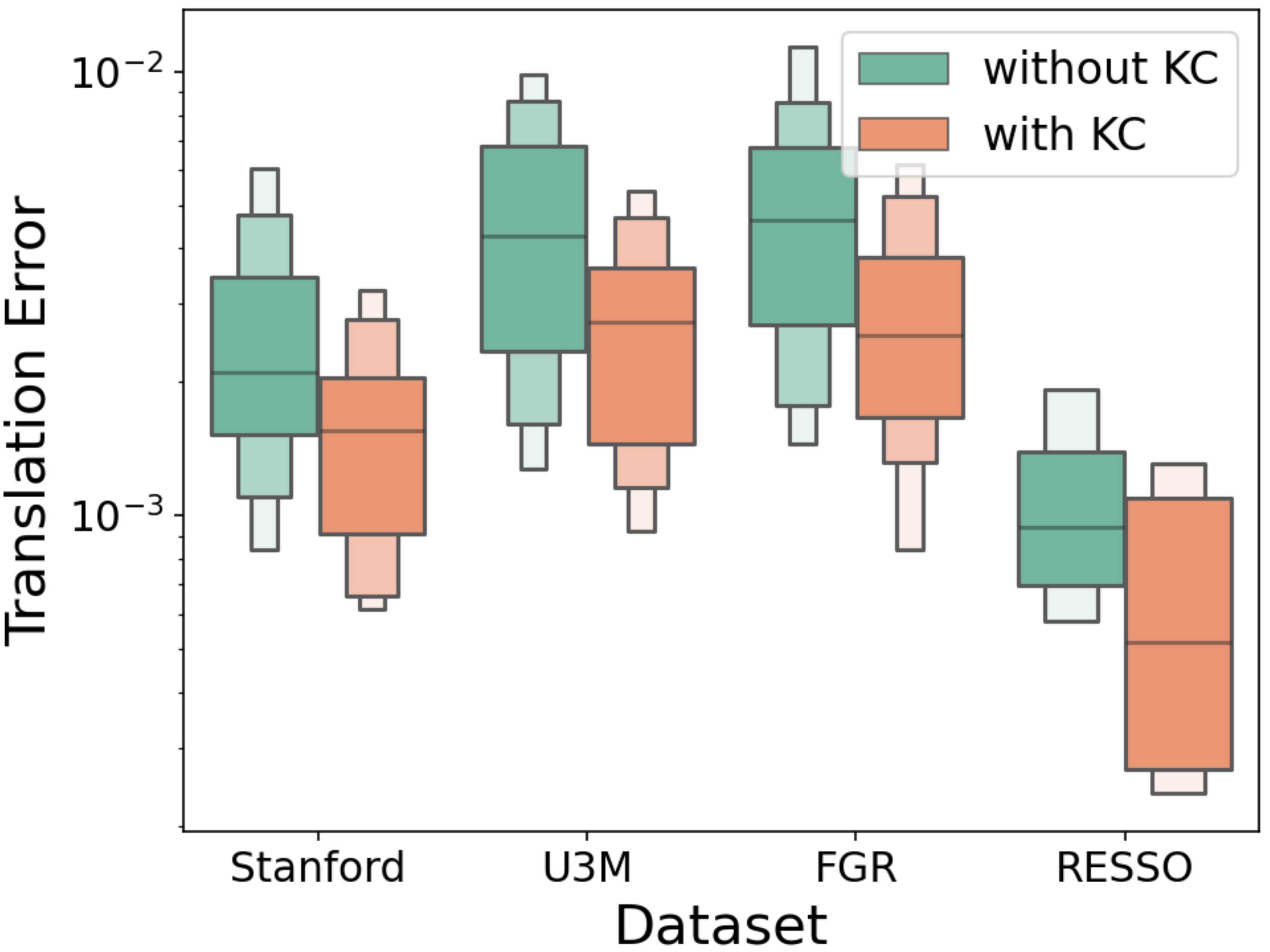}\label{fig:KC_Et}}
  \caption{Line plot of success ratio on 15 data of the methods with and without knowledge complement.
  Enhanced box plot of rotation error and translation error on 4 datasets of the methods with and without knowledge complement. 
  }
  \label{fig:KC}
\end{figure}

Fig. \ref{fig:KC} records the success ratio, rotation error and translation error of two methods based on 20 independent runs, where KC is the shorthand of knowledge complement.
As can be seen from Fig. \ref{fig:KC_SR}, except for the \emph{Bimba} and \emph{CD} data, the method with the knowledge complement strategy only achieves a tiny improvement regarding the success ratio.
Fig. \ref{fig:KC_Er} shows the mean and variance of rotation error are much smaller by using this strategy for all datasets. 
In Fig. \ref{fig:KC_Et}, for object-scale datasets, the optimal value, mean and variance of translation error are all much lower compared with the method without this strategy.
However, regarding the scene-scale dataset, only the variance of the method with this strategy prevails in the comparison.
Hence, the strategy for estimating translation parameters is effective to improve the precision and success ratio of registration. 

\begin{table}[!t]
  \caption{The results of methods with and without sparse-to-dense strategy on the 15 data.
  The superior performance is highlighted in {\bf BOLD}.
  \label{tab:table_S2D_cmp}}
  \centering
  \renewcommand\arraystretch{1.1}
  \resizebox{1\columnwidth}{!}{
    \begin{tabular}{cccclccclccc}
      \hline
      & \multicolumn{3}{c}{$E_{\boldsymbol{R}}$}                       &  & \multicolumn{3}{c}{$E_{\boldsymbol{t}}$}                       &  & \multicolumn{3}{c}{\textbf{NN}}                                             \\ \cline{2-4} \cline{6-8} \cline{10-12} 
      & \textbf{w/. S2D}            & \textbf{w/o. S2D}            & ${\bf{\Delta}}$ &  & \textbf{w/. S2D}            & \textbf{w/o. S2D}            & $\bf{\Delta}$ &  & \textbf{w/. S2D}  & \textbf{w/o. S2D} & $\frac{\text{w/. S2D}}{\text{w/o. S2D}}$ \\
\hline
\textit{(i)object-scale} \\
\textbf{Armadillo} & 0.161$\pm$0.073          & \textbf{0.159$\pm$0.064} & 0.001    &  & 0.084$\pm$0.034          & \textbf{0.078$\pm$0.029} & 0.006    &  & \textbf{9.86}  & 20.04                             & 49.2\%                 \\
\textbf{Bunny}     & 0.233$\pm$0.120          & \textbf{0.230$\pm$0.118} & 0.003    &  & 0.175$\pm$0.085          & \textbf{0.167$\pm$0.082} & 0.007    &  & \textbf{8.42}  & 17.29                             & 48.7\%                 \\
\textbf{Dragon}    & \textbf{0.190$\pm$0.074} & 0.194$\pm$0.084          & -0.004   &  & 0.168$\pm$0.074          & \textbf{0.163$\pm$0.067} & 0.005    &  & \textbf{8.75}  & 17.64                             & 49.6\%                 \\
\textbf{Happy}     & 0.369$\pm$0.195          & \textbf{0.353$\pm$0.167} & 0.015    &  & 0.220$\pm$0.087          & \textbf{0.199$\pm$0.062} & 0.021    &  & \textbf{8.00}  & 16.32                             & 49.0\%                 \\
\textbf{Chef}      & 0.379$\pm$0.208          & \textbf{0.377$\pm$0.198} & 0.001    &  & 0.197$\pm$0.137          & \textbf{0.180$\pm$0.111} & 0.016    &  & \textbf{8.70}  & 17.90                             & 48.6\%                 \\
\textbf{Chicken}   & 0.262$\pm$0.181          & \textbf{0.258$\pm$0.218} & 0.004    &  & \textbf{0.266$\pm$0.120} & 0.276$\pm$0.148          & -0.010   &  & \textbf{7.83}  & 16.18                             & 48.4\%                 \\
\textbf{P}         & 0.335$\pm$0.186          & \textbf{0.313$\pm$0.198} & 0.022    &  & 0.299$\pm$0.148          & \textbf{0.286$\pm$0.074} & 0.013    &  & \textbf{7.22}  & 14.82                             & 48.7\%                 \\
\textbf{T-rex}     & 0.296$\pm$0.165          & \textbf{0.268$\pm$0.300} & 0.029    &  & 0.366$\pm$0.176          & \textbf{0.356$\pm$0.258} & 0.010    &  & \textbf{9.06}  & 18.26                             & 49.6\%                 \\
\textbf{Angel}     & 0.178$\pm$0.076          & \textbf{0.173$\pm$0.089} & 0.005    &  & 0.277$\pm$0.182          & \textbf{0.257$\pm$0.189} & 0.020    &  & \textbf{11.41} & 23.52                             & 48.5\%                 \\
\textbf{Bimba}     & 0.417$\pm$0.237          & \textbf{0.395$\pm$0.257} & 0.022    &  & 0.419$\pm$0.146          & \textbf{0.398$\pm$0.125} & 0.021    &  & \textbf{9.57}  & 19.51                             & 49.0\%                 \\
\textbf{CD}        & 0.258$\pm$0.123          & \textbf{0.203$\pm$0.073} & 0.056    &  & 0.251$\pm$0.149          & \textbf{0.235$\pm$0.129} & 0.015    &  & \textbf{8.41}  & 17.36                             & 48.4\%                 \\
\textbf{DC}        & \textbf{0.199$\pm$0.108} & 0.202$\pm$0.089          & -0.003   &  & 0.265$\pm$0.267          & \textbf{0.205$\pm$0.146} & 0.060    &  & \textbf{8.26}  & 16.69                             & 49.5\%                 \\
\hline
\textit{(ii)scene-scale} \\
\textbf{7a}        & 0.147$\pm$0.063          & \textbf{0.141$\pm$0.073} & 0.006    &  & 0.051$\pm$0.029          & \textbf{0.049$\pm$0.024} & 0.002    &  & \textbf{11.30} & 22.11                             & 51.1\%                 \\
\textbf{7d}        & 0.192$\pm$0.075          & \textbf{0.183$\pm$0.059} & 0.009    &  & 0.035$\pm$0.020          & \textbf{0.032$\pm$0.010} & 0.003    &  & \textbf{12.15} & 24.23                             & 50.1\%                 \\
\textbf{7e}        & 0.549$\pm$0.220          & \textbf{0.480$\pm$0.199} & 0.069    &  & 0.135$\pm$0.064          & \textbf{0.117$\pm$0.057} & 0.018    &  & \textbf{12.40} & 24.07                             & 51.5\%                 \\
\hline
\end{tabular}}
  \end{table}

\subsubsection{The Strategy of Sparse-to-Dense}
Table \ref{tab:table_S2D_cmp} compares methods with and without the sparse-to-dense strategy in terms of accuracy and computational cost.
In this table, w/. S2D and w/o. S2D represent EMTR-SSC with and without this strategy, respectively. 
$\Delta$ denotes the difference between the errors of these two methods.
$\frac{\text{w/. S2D}}{\text{w/o. S2D}}$ indicates the ratio of the two methods with respect to the computational cost.
As can be seen from Table \ref{tab:table_S2D_cmp}, the method without this strategy achieves a greater than 10\% improvement of error in only 3 out of 15 data, i.e., \emph{CD}, \emph{DC} and \emph{7e} data.
However, the method without this strategy requires twice the computational cost on most of the data according to the number of calls to the NNS algorithm. 
Obviously, the designed strategy, sparse-to-dense, significantly reduces the computational cost and it is able to minimize the loss of precision.

Table \ref{tab:aver_time} compared the proposed method with other EMTO methods in terms of run time, where EMTR-SSC(-) represents the proposed EMTR-SSC without Sparse-to-Dense strategy. 
As can be seen from Table \ref{tab:aver_time}, the designed knowledge complement mechanism would increase the computational cost to some extent. 
Since each algorithm has an equal number of NNS algorithm, they are close in average run time. 
With the introduction of Sparse-to-Dense mechanism, the computational cost can be reduced substantially.

On the other hand, although EMTR-SSC(-) and EMTR-SSC have the same number of evaluations, their run times are much different.
Therefore, it is not applicable for PCR problems using the number of evaluations to measure the computational cost.
For the two methods, the ratio of run time, 47.4\%, is close to the ratio of computational cost, which reflects that the proposed metric can accurately measure the computational cost for PCR problems.

\section{Conclusion}
\label{sec:Conclusion}

The objective of this paper is to develop an effective PCR approach for the low success ratio problem of evolving registration. 
The goal has been successfully accomplished by injecting the idea of EMTO and solution space cutting into evolving registration approaches. 
Furthermore, we propose the knowledge complement strategy for enhancing the quality of transferred information, the sparse-to-dense strategy for saving unnecessary computational cost and a novel fitness function MCD for higher precision and handling partial overlap. 
Notably, we also propose a novel metric for the computational cost calculation of evolving registration. 
From an experimental perspective, we compare 15 competitive approaches, containing 4 state-of-the-art EMTO approaches, with our approach on 5 datasets widely used in PCR. 
The experimental results show that this approach is capable of significantly enhancing the success ratio of evolving registration approach significantly.  
The proposed approach also can achieve great registration precision on both object-scale and scene-scale datasets.
In addition, EMTR-SSC reduces unnecessary computational expenses to a large extent.

\begin{table}[!t]
  \caption{The average run time for each EMTO method.\label{tab:aver_time}}
  \renewcommand\arraystretch{1.5}
  \resizebox{1\columnwidth}{!}{
    \begin{tabular}{c|ccccccc}
      \hline
                & \textbf{MFEA} & \textbf{MFEA-II} & \textbf{MTEA-AD} & \textbf{EMFPCR} & \textbf{DEORA} & \textbf{EMTR-SSC(-)} & \textbf{EMTR-SSC} \\ \hline
    Avg time(s) & 14.89         & 15.01            & 14.56            & 14.18           & 15.29          & 15.58                & \textbf{7.39}     \\ \hline    
    \end{tabular}}
  \end{table}

In this study, several parameters need to be determined experimentally or manually, such as {\it rmp} and switch parameter $\delta$. 
This limits the adaptability of this approach to various kinds of data. 
In the future, we will investigate how to convert these constant parameters to adaptive ones, like MFEA-II \cite{bali2019multifactorial}.

\bibliographystyle{IEEEtran}
\bibliography{IEEEabrv,myrefs}

\begin{IEEEbiography}[{\includegraphics[width=1in,height=1.25in,clip,keepaspectratio]{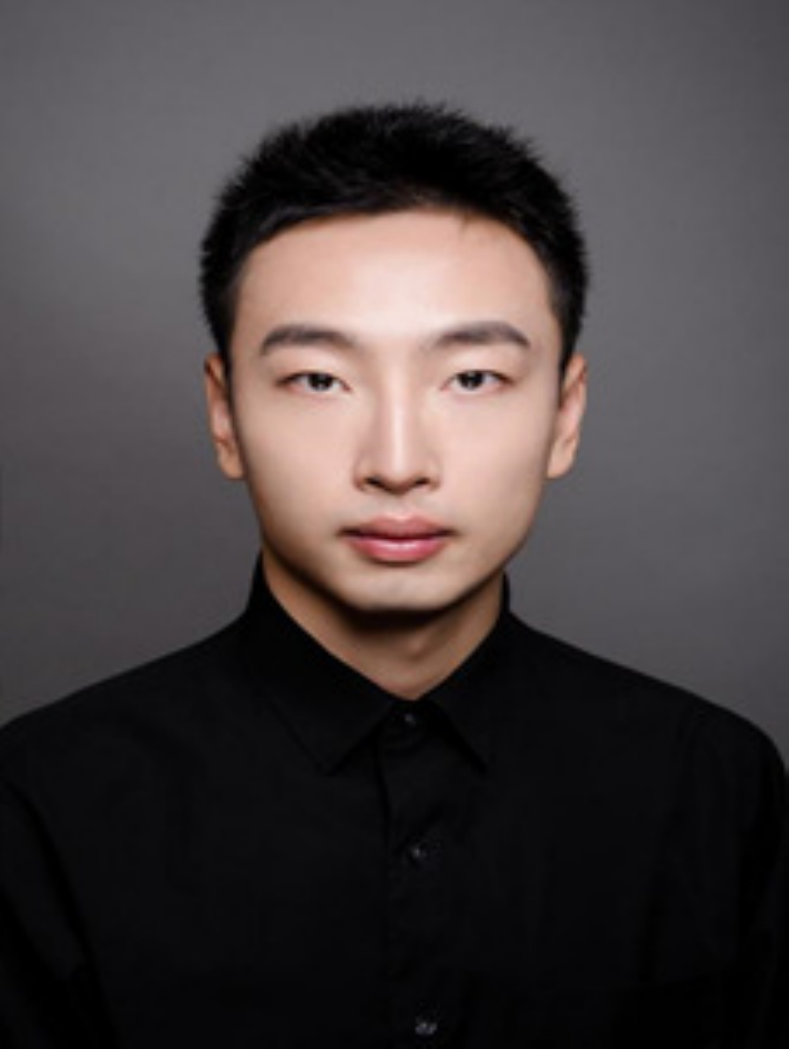}}]{Yue Wu}
	(Member, IEEE) received the B.Eng. and Ph.D. degrees from Xidian University, Xi'an, China, in 2011 and 2016, respectively. 
	
	Since 2016, he has been a Teacher with Xidian University. He is currently an Associate Professor with Xidian University. He has authored or co-authored more than 70 papers in refereed journals and proceedings. 
	
	His research interests include computational intelligence and its Applications. He is the Secretary General of Chinese Association for Artificial Intelligence-Youth Branch, Chair of CCF YOCSEF Xi'an, Senior Member of Chinese Computer Federation. He is Editorial Board Member for over five journals, including Remote Sensing, Applied Sciences, Electronics, Mathematics.
\end{IEEEbiography}

\begin{IEEEbiography}[{\includegraphics[width=1in,height=1.25in,clip,keepaspectratio]{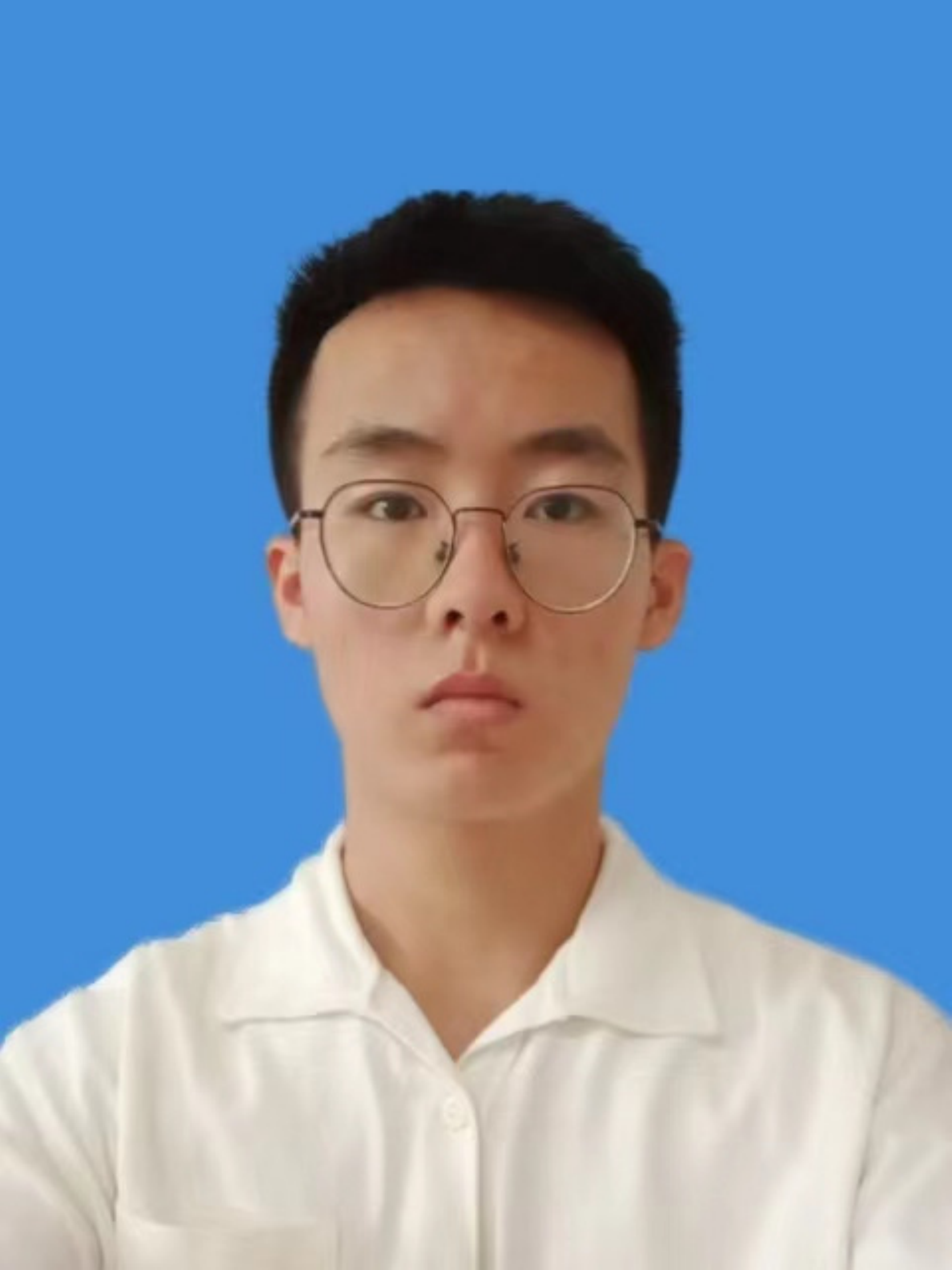}}]{Peiran Gong}
  received the B.Eng. degree from Nanjing Tech University, Nanjing, China, in 2021. He is currently pursuing the master's degree from the School of Computer Science and Technology, Xidian University, Xi'an, China.
  
  His research interests include evolutionary computation and 3-D registration.
\end{IEEEbiography}

\begin{IEEEbiography}[{\includegraphics[width=1in,height=1.25in,clip,keepaspectratio]{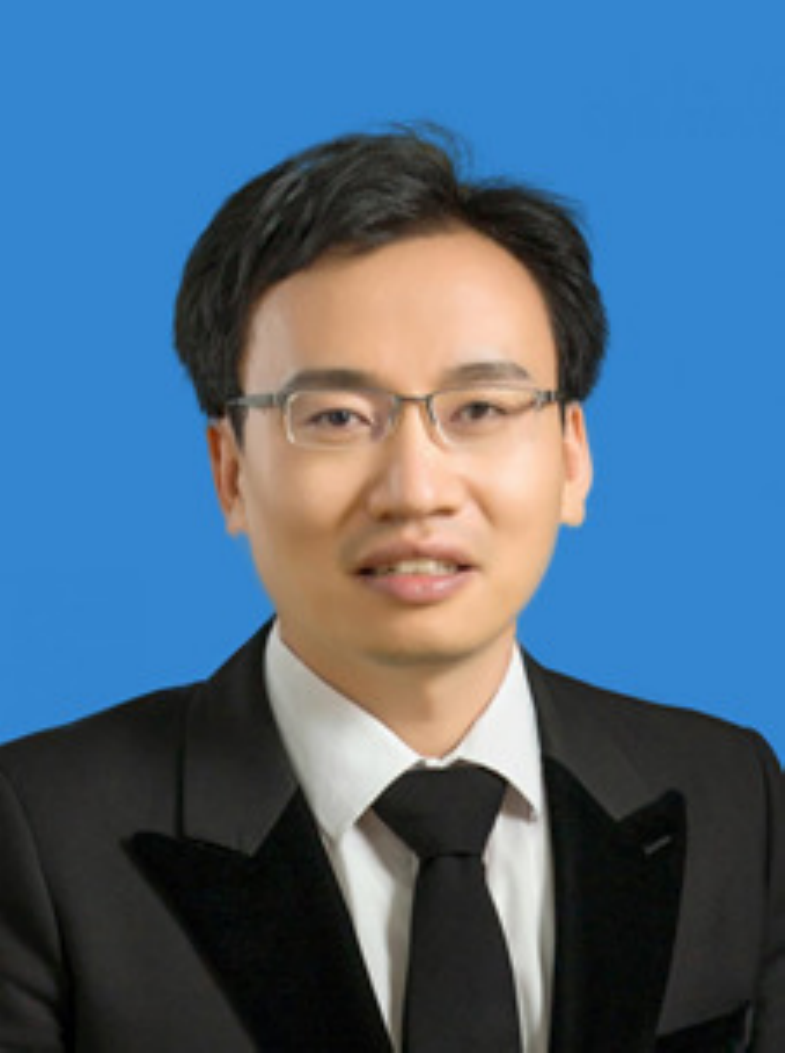}}]{Maoguo Gong}
	(Senior Member, IEEE) received the B.Eng. and Ph.D. degrees from Xidian University, Xi'an, China, in 2003 and 2009, respectively. 
	
	Since 2006, he has been a Teacher with Xidian University. He was promoted to an Associate Professor and a Full Professor, in 2008 and 2010, respectively, with exceptive admission. He has authored or coauthored over 100 articles in journals and conferences. He holds over 20 granted patents as the first inventor. He is leading or has completed over twenty projects as the Principal Investigator, funded by the National Natural Science Foundation of China, the National Key Research and Development Program of China, and others. 
	
	His research interests are broadly in the area of computational intelligence, with applications to optimization, learning, data mining, and image understanding. Prof. Gong is the Executive Committee Member of Chinese Association for Artificial Intelligence and a Senior Member of Chinese Computer Federation. He was the recipient of the prestigious National Program for Support of the Leading Innovative Talents from the Central Organization Department of China, the Leading Innovative Talent in the Science and Technology from the Ministry of Science and Technology of China, the Excellent Young Scientist Foundation from the National Natural Science Foundation of China, the New Century Excellent Talent from the Ministry of Education of China, and the National Natural Science Award of China. He is an Associate Editor or an Editorial Board Member for over five journals including the \textsc{IEEE Transactions on Evolutionary Computation} and the \textsc{IEEE Transactions on Neural Networks and Learning Systems}.
\end{IEEEbiography}

\begin{IEEEbiography}[{\includegraphics[width=1in,height=1.25in,clip,keepaspectratio]{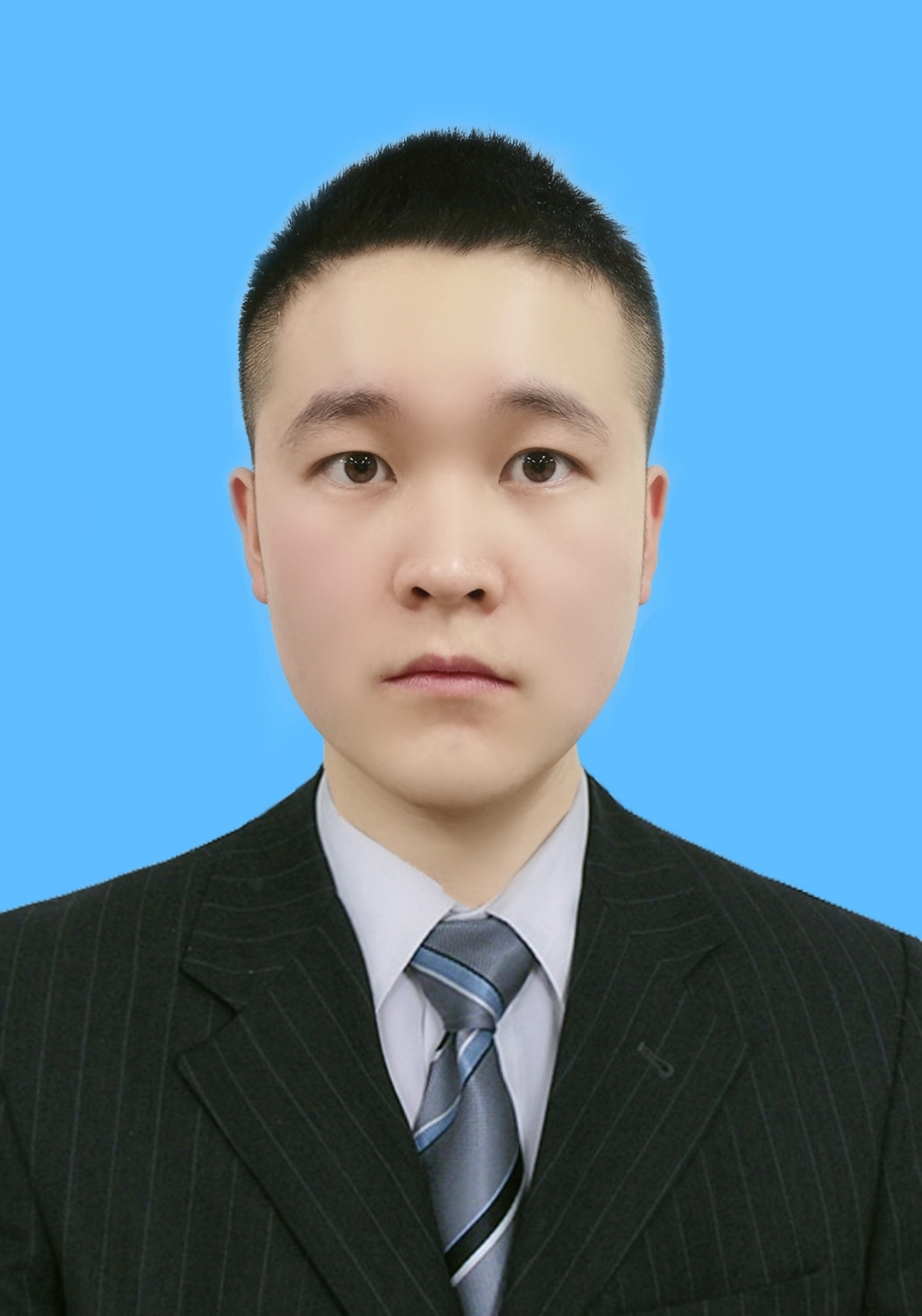}}]{Hangqi Ding}
	received the B.Eng. degree from Shandong University of Technology, Zibo, China, in 2021. He is currently pursuing the master's degree with the Key Laboratory of Big Data and Intelligent Vision, School of Computer Science and Technology, Xidian University, Xi'an, China. 
	
	His research interests include computational intelligence and point cloud registration.
\end{IEEEbiography}

\begin{IEEEbiography}[{\includegraphics[width=1in,height=1.25in,clip,keepaspectratio]{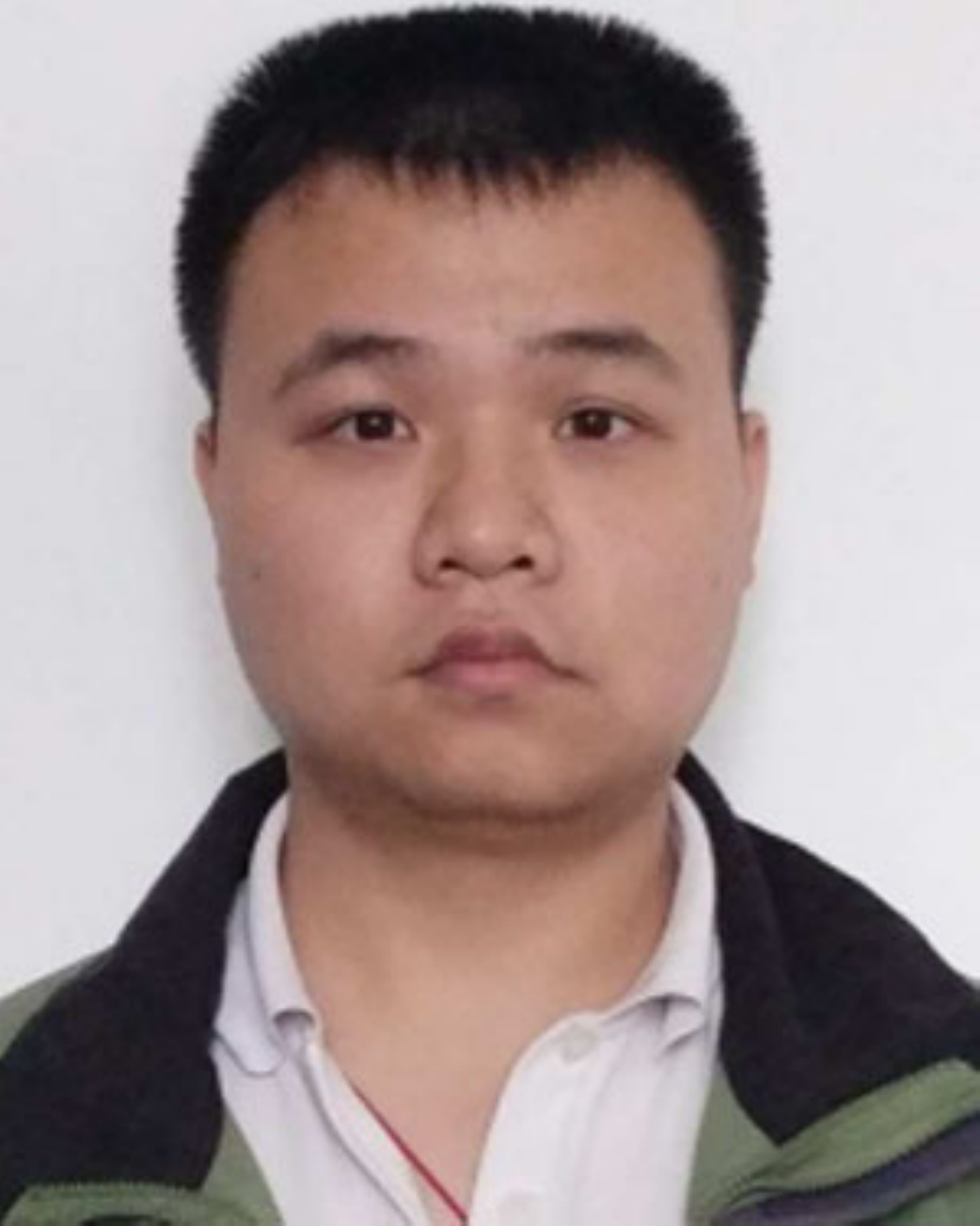}}]{Zedong Tang}
  received the B.Eng. and Ph.D. degrees from Xidian University, Xi'an, China, in 2014 and 2020, respectively. Since 2020, he has been a Lecturer with the Academy of Advanced Interdisciplinary Research, Xidian University. 

  His current research interests include computational intelligence and machine learning.
\end{IEEEbiography}

\begin{IEEEbiography}[{\includegraphics[width=1in,height=1.25in,clip,keepaspectratio]{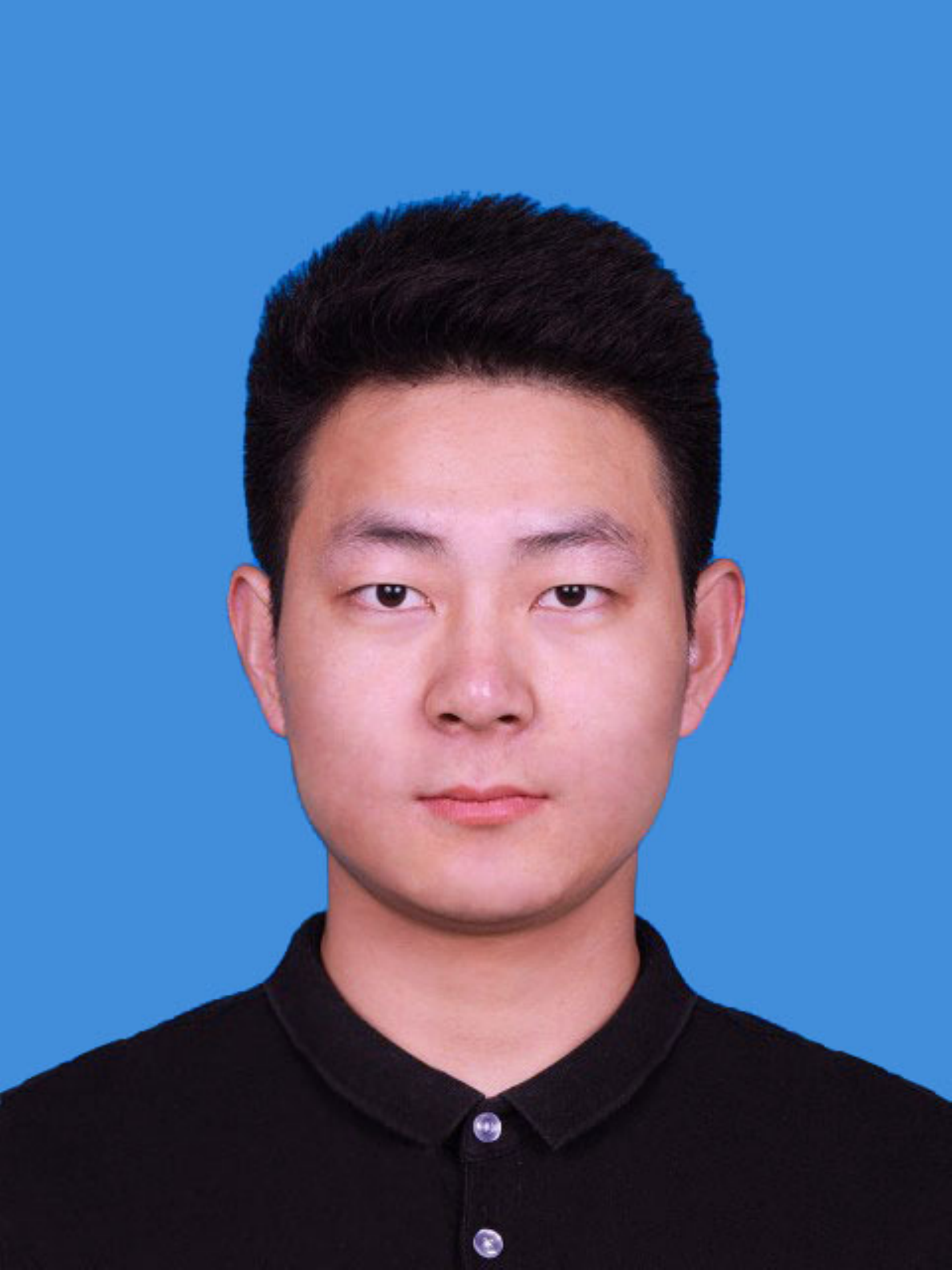}}]{Yibo Liu} 
  received the B.Eng. degree in computer science and technology from North China Electric Power University, Beijing, China, in 2020. He is currently pursuing the master's degree with the Key Laboratory of Big Data and Intelligent Vision, School of Computer Science and Technology, Xidian University, Xi'an, China.
	
  His research interests include point cloud registration and computational intelligence.
\end{IEEEbiography}
\begin{IEEEbiography}[{\includegraphics[width=1in,height=1.25in,clip,keepaspectratio]{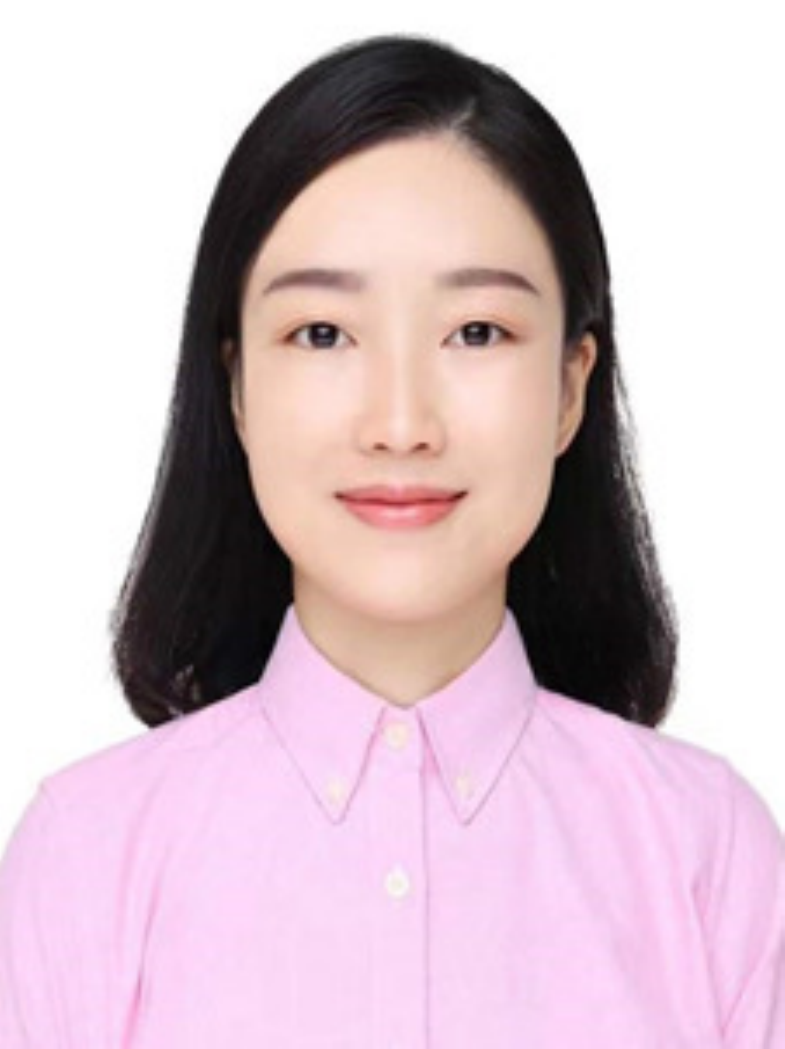}}]{Wenping Ma}
	(Senior Member, IEEE) received the B.S. degree in computer science and technology and the Ph.D. degree in pattern recognition and intelligent systems from Xidian University, Xi'an, China, in 2003 and 2008, respectively.
	
  She has published more than 30 SCI papers in international academic journals, including the \textsc{IEEE Transactions on Evolutionary Computation} and the \textsc{IEEE  Transactions on Image Processing}.
  Her research interests include natural computing and intelligent image processing.
	
\end{IEEEbiography}

\begin{IEEEbiography}[{\includegraphics[width=1in,height=1.25in,clip,keepaspectratio]{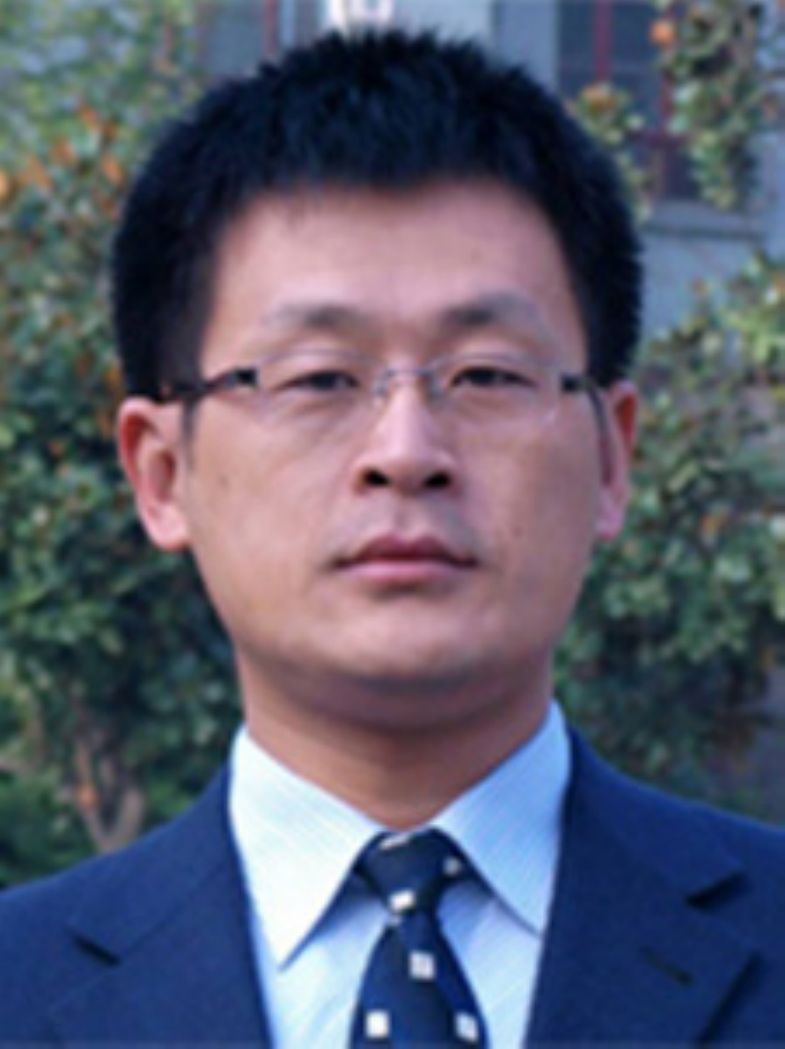}}]{Qiguang Miao}
	(Senior Member, IEEE) received the M.Eng. and Ph.D. degrees from Xidian University, Xi'an, China, respectively.

	He is currently a Professor with the School of Computer Science and Technology, Xidian University. His research interests include intelligent image processing and multiscale geometric representations for images.
\end{IEEEbiography}

\end{document}